\RequirePackage{fix-cm}
\documentclass[twocolumn]{svjour3}          
\smartqed  
\usepackage{graphicx}
\usepackage{hyperref}
\usepackage{amsmath,amssymb,amsfonts}%
\usepackage{subcaption}
\usepackage{booktabs} 
\usepackage{tabularx} 
\usepackage{multirow} 
\usepackage{wrapfig}
\usepackage{xcolor}
\journalname{Multimedia Systems}
\begin{document}
\sloppy

\title{ALEN: A Dual-Approach for Uniform and Non-Uniform Low-Light Image Enhancement}


\author{Ezequiel Perez-Zarate  \and Oscar Ramos-Soto \and Chunxiao Liu \and Diego Oliva \and Marco Perez-Cisneros}


\institute{
    Ezequiel Perez-Zarate \at
    School of Computer Science and Technology, Zhejiang Gongshang University, Hangzhou, 310018, Zhejiang, China\\
    \email{isaias.perez@alumnos.udg.mx} %
    \and
    Oscar Ramos-Soto \at
    Depto. de Ingeniería Electro-fotónica, Universidad de Guadalajara, Guadalajara, 44430, Jalisco, México\\
    \email{oscar.ramos9279@alumnos.udg.mx}
    \and
    Chunxiao Liu \at
    School of Computer Science and Technology, Zhejiang Gongshang University, Hangzhou, 310018, Zhejiang, China\\
    \email{cxliu@zjgsu.edu.cn}
    \and
    Diego Oliva \at
    Depto. de Ingeniería Electro-fotónica, Universidad de Guadalajara, Guadalajara, 44430, Jalisco, México \\
    \email{diego.oliva@cucei.udg.mx}
    \and
    Marco Perez-Cisneros \at
    Depto. de Ingeniería Electro-fotónica, Universidad de Guadalajara, Guadalajara, 44430, Jalisco, México \\
    \email{marco.perez@cucei.udg.mx}\\
    \emph{Corresponding Authors:} Diego Oliva, Marco Perez-Cisneros
}

\date{Received: date / Accepted: date}

\maketitle

\begin{abstract}
Low-light image enhancement is an important task in computer vision, essential for improving the visibility and quality of images captured in non-optimal lighting conditions. Inadequate illumination can lead to significant information loss and poor image quality, impacting various applications such as surveillance, photography, or even autonomous driving. In this regard, automated methods have been developed to automatically adjust illumination in the image for a better visual perception. Current enhancement techniques often use specific datasets to enhance low-light images, but still present challenges when adapting to diverse real-world conditions, where illumination degradation may be localized to specific regions. To address this challenge, the Adaptive Light Enhancement Network (ALEN) is introduced, whose main approach is using a classification mechanism to determine whether local or global illumination enhancement is required. Subsequently, estimator networks adjust illumination based on this classification and simultaneously enhance color fidelity. ALEN integrates the Swin Light-Classification Transformer (SLCformer) for illuminance categorization, complemented by the Single-Channel Network (SCNet), and Multi-Channel Network (MCNet) for precise estimation of illumination and color, respectively. Extensive experiments on publicly available datasets for low-light conditions were carried out to underscore ALEN's robust generalization capabilities, demonstrating superior performance in both quantitative metrics and qualitative assessments when compared to recent state-of-the-art methods. The ALEN not only enhances image quality in terms of visual perception but also represents an advancement in high-level vision tasks, such as semantic segmentation, as presented in this work. The code of this method is available at \href{https://github.com/xingyumex/ALEN}{https://github.com/xingyumex/ALEN}.

\end{abstract}

\section{Introduction}
\label{IT}

Images captured under low-light conditions, such as indoor settings or dark outdoor environments during the day or night, pose significant challenges for accurate visual perception due to limited light reflected by objects. These conditions result in reduced image quality, often characterized by color distortions and noticeable noise, which impact critical fields such as transportation surveillance~\cite{qu2024double}, professional photography~\cite{cao2023physics}, and autonomous driving~\cite{li2021deep}. Furthermore, these limitations extend to high-level vision tasks like object detection~\cite{cui2021multitask} and semantic segmentation~\cite{xia2023cmda}, where image quality plays a pivotal role.

Numerous image enhancement strategies have been proposed to address these challenges. Traditional techniques, including histogram equalization~\cite{pizer1987adaptive,kaur2011survey,singh2015enhancement,wang2007fast,dale1993study,khan2014segment}, have been widely adopted to adjust contrast and improve image visibility. However, these methods often fail to effectively capture local lighting variations, leading to suboptimal textural and chromatic details enhancement. Alternatively, the Retinex theory \cite{land1971lightness}, through approaches like Single-scale Retinex (SSR) \cite{choi2008color}, Multiscale Retinex (MSR) \cite{rahman1996multi}, and Multiscale Retinex with Color Restoration (MSRCR) \cite{jobson1997multiscale,rahman2004retinex}, offers a framework to decompose a low-light image into reflectance and illumination components. Despite these advancements, Retinex-based methods are inherently limited when applied in the RGB space, where learning brightness features poses significant challenges.

In recent years, approaches based on convolutional neural networks (CNNs)\cite{khan2020survey}, generative adversarial networks (GANs) \cite{pan2019recent}, and transformers \cite{han2022survey,khan2022transformers} have shown significant progress in enhancing low-light images. Despite its advancements and achieving the best results in the visual enhancement of low-light images, these methods still face challenges in generalization across different datasets, which can lead to a tendency to amplify noise and produce unnatural or oversaturated colors in real-world image datasets \cite{lee2013contrast,guo2016lime,wang2013naturalness,ma2015perceptual,VV_TM-DIED}

\vspace{-0.5cm}
\subsection{Motivation}

\begin{figure}[ht]
	\centering
	\begin{subfigure}{0.24\linewidth}
		\centering
		\includegraphics[width=\linewidth]{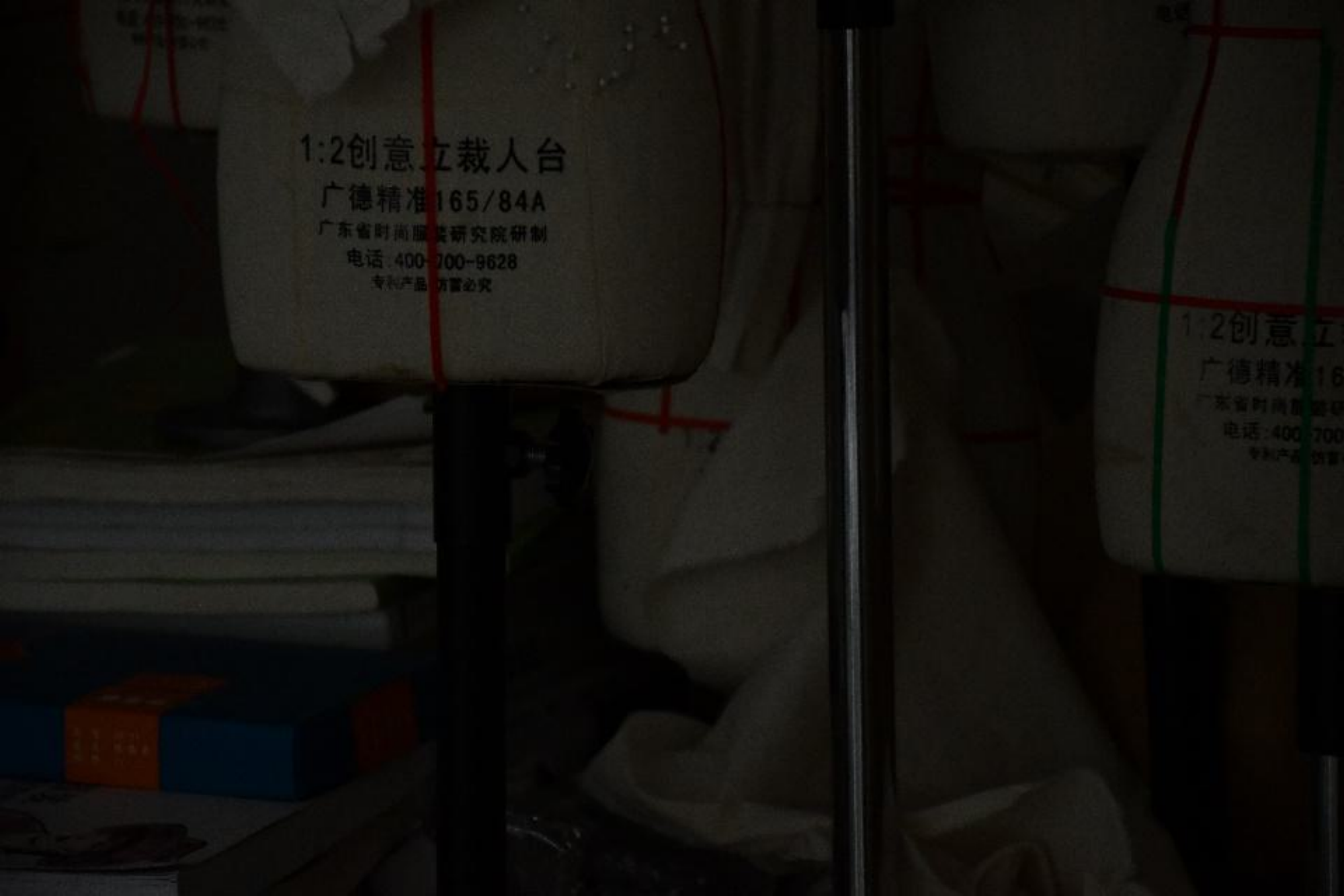} 
	\end{subfigure}
	\begin{subfigure}{0.24\linewidth}
		\centering
		\includegraphics[width=\linewidth]{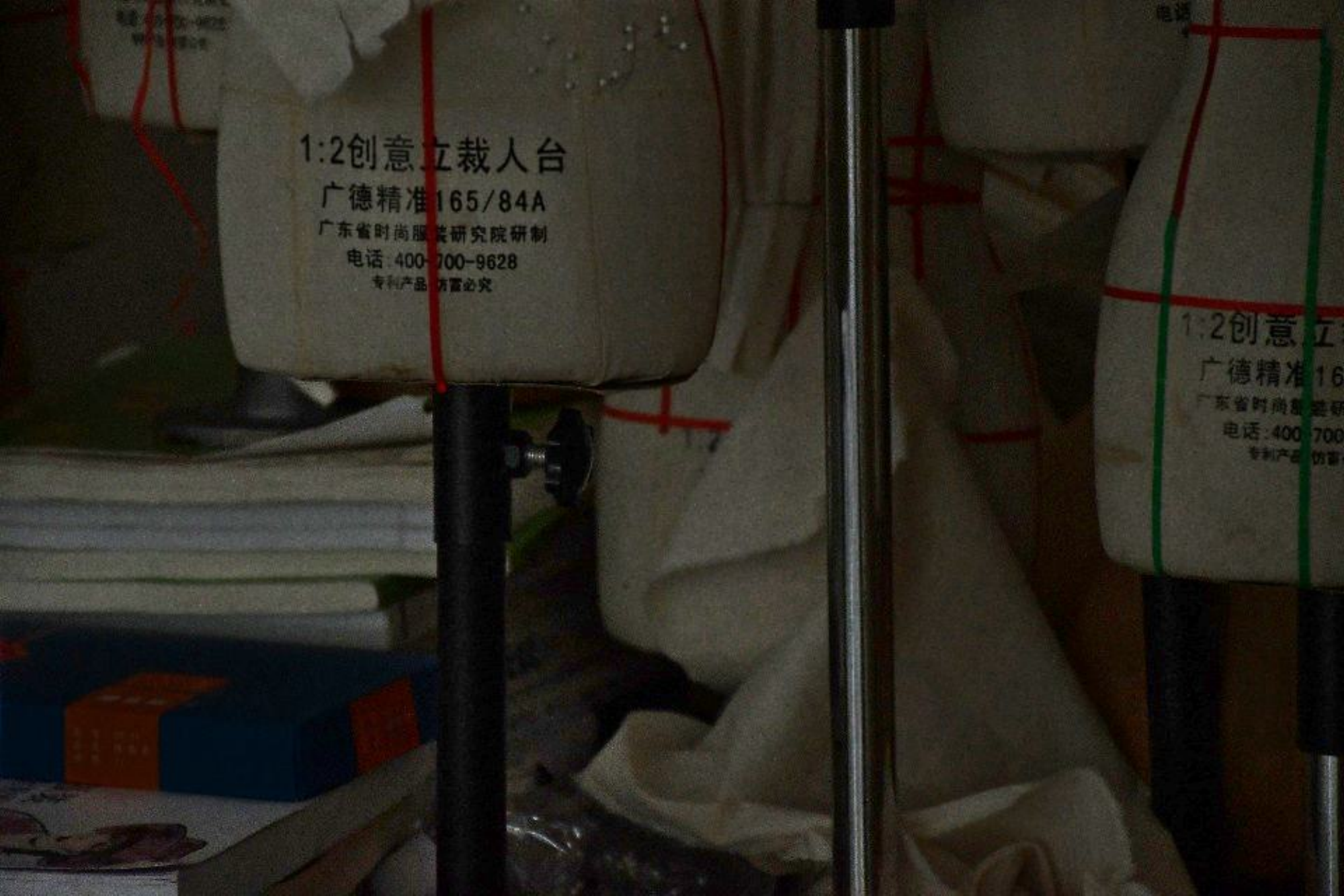} 
	\end{subfigure}
	\begin{subfigure}{0.24\linewidth}
		\centering
		\includegraphics[width=\linewidth]{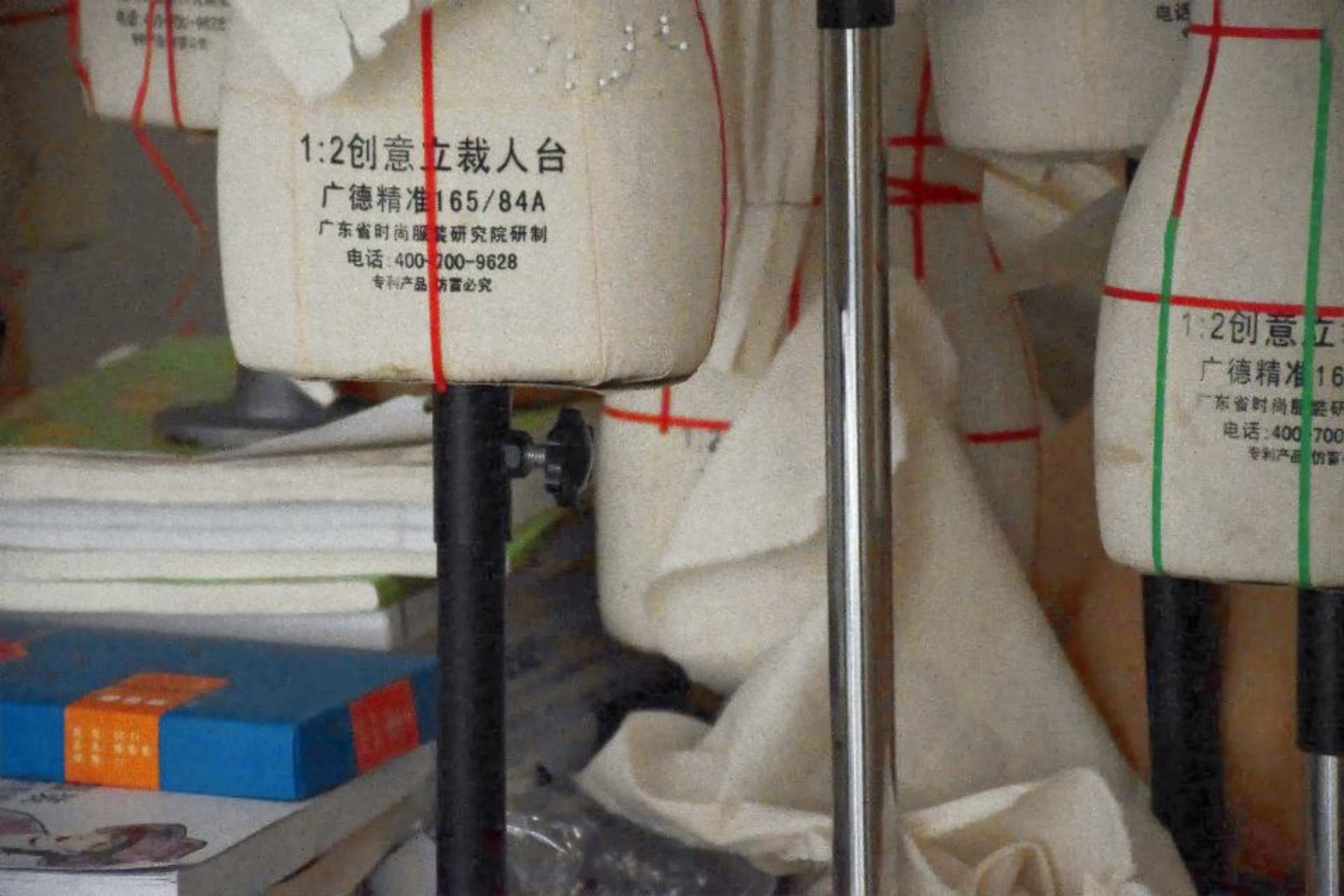}
	\end{subfigure}
    \begin{subfigure}{0.24\linewidth}
		\centering
		\includegraphics[width=\linewidth]{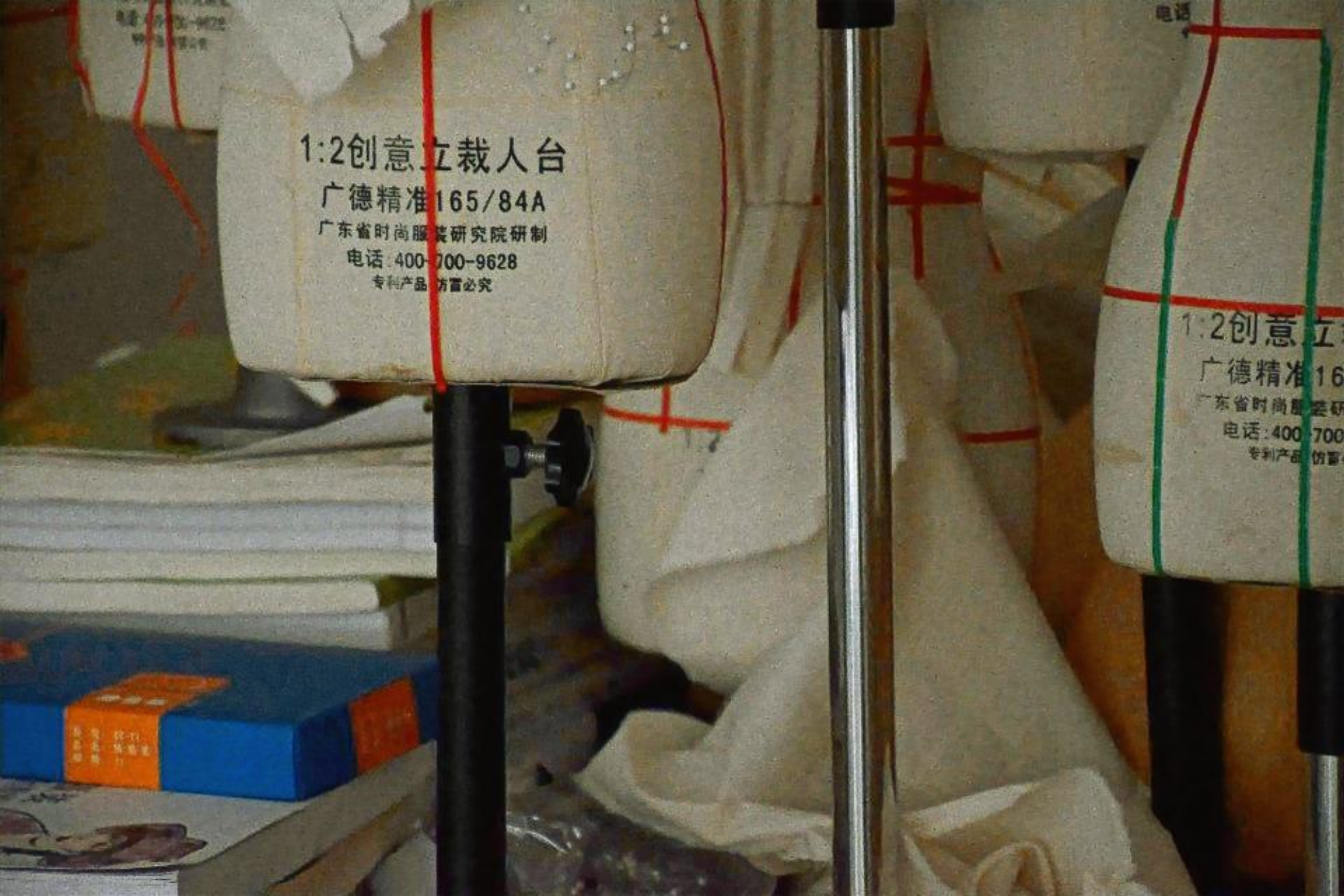}
	\end{subfigure}

	\begin{subfigure}{0.24\linewidth}
		\centering
		\includegraphics[width=\linewidth]{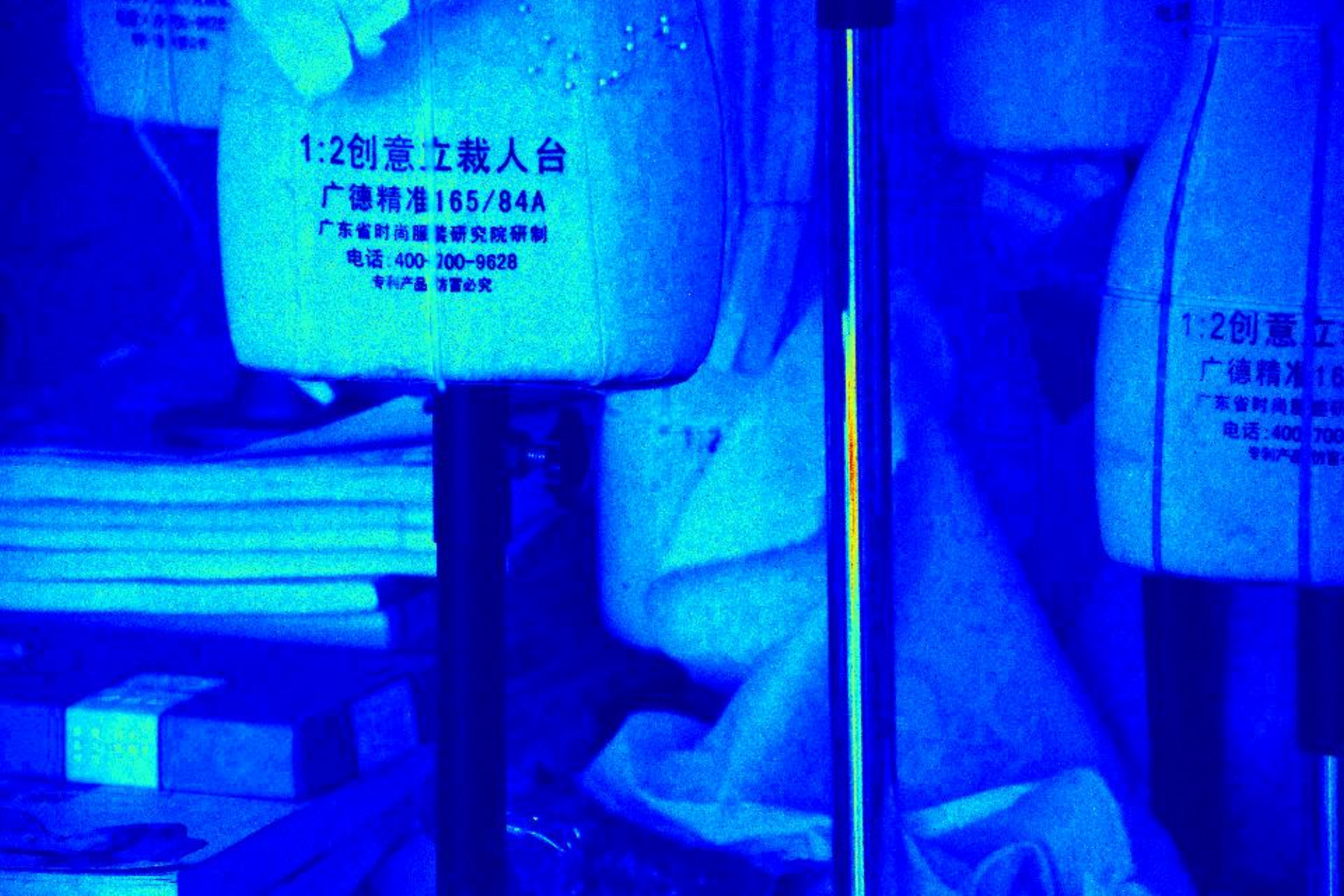} 
	\end{subfigure}
	\begin{subfigure}{0.24\linewidth}
		\centering
		\includegraphics[width=\linewidth]{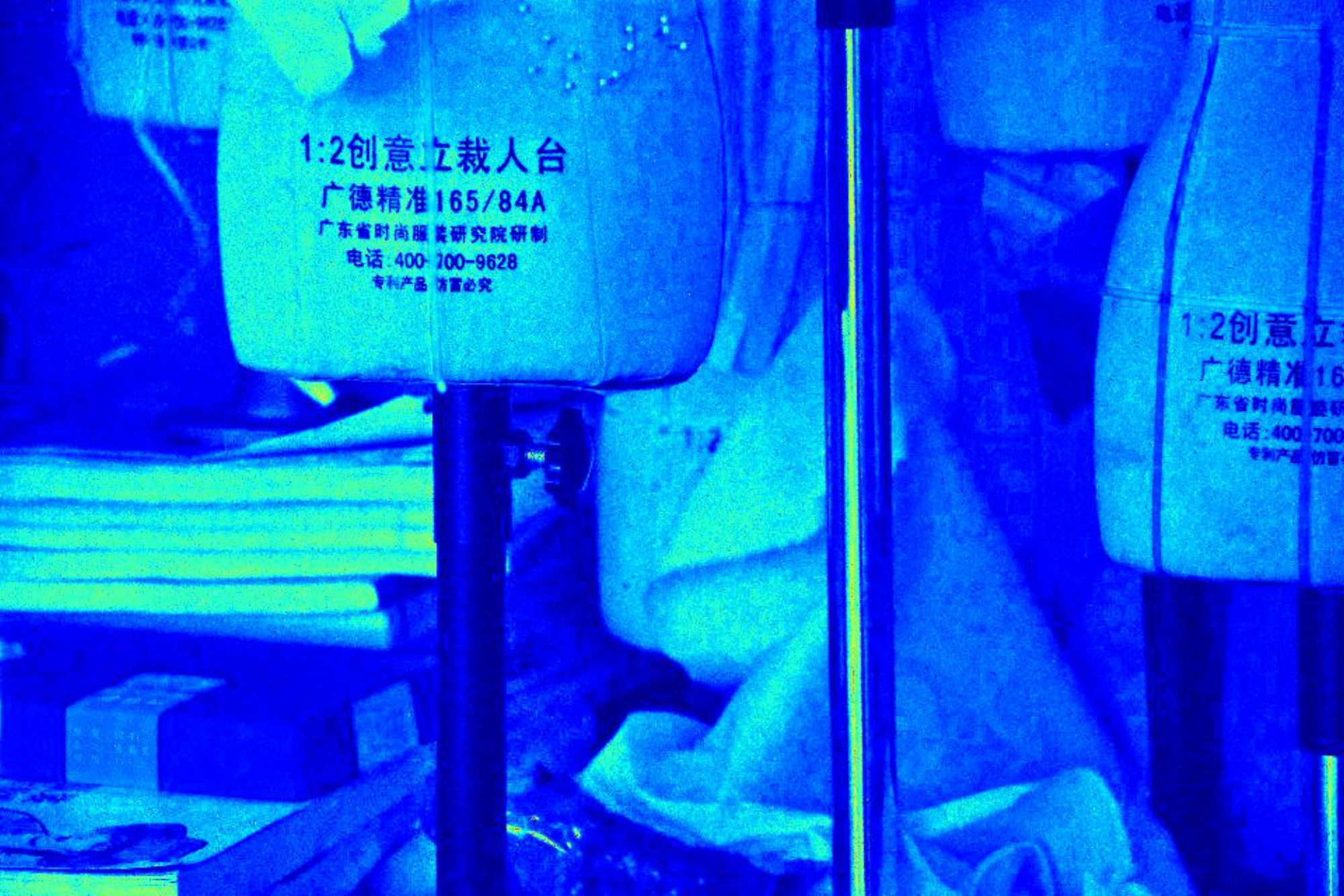} 
	\end{subfigure}
	\begin{subfigure}{0.24\linewidth}
		\centering
		\includegraphics[width=\linewidth]{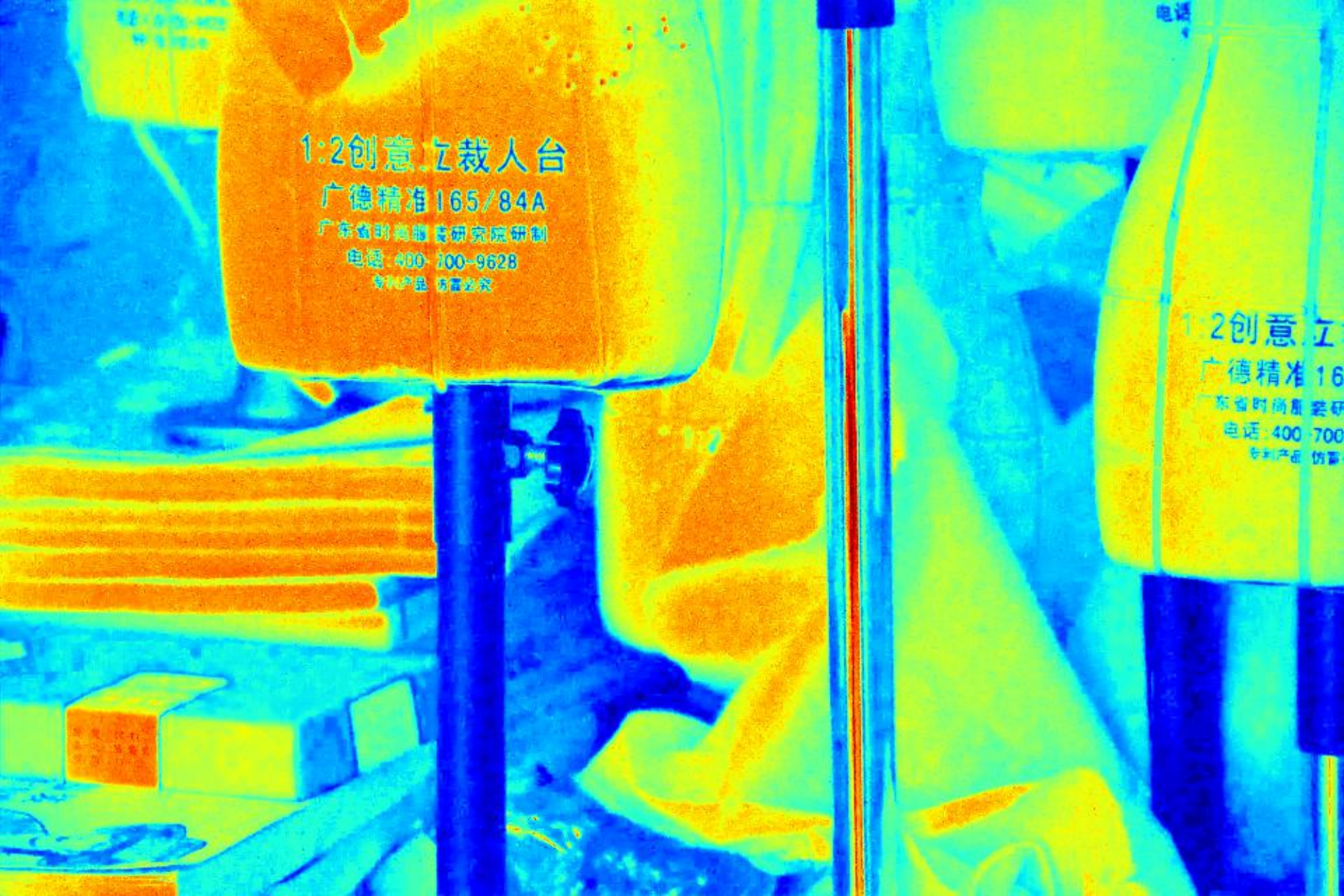}
	\end{subfigure}
    \begin{subfigure}{0.24\linewidth}
		\centering
		\includegraphics[width=\linewidth]{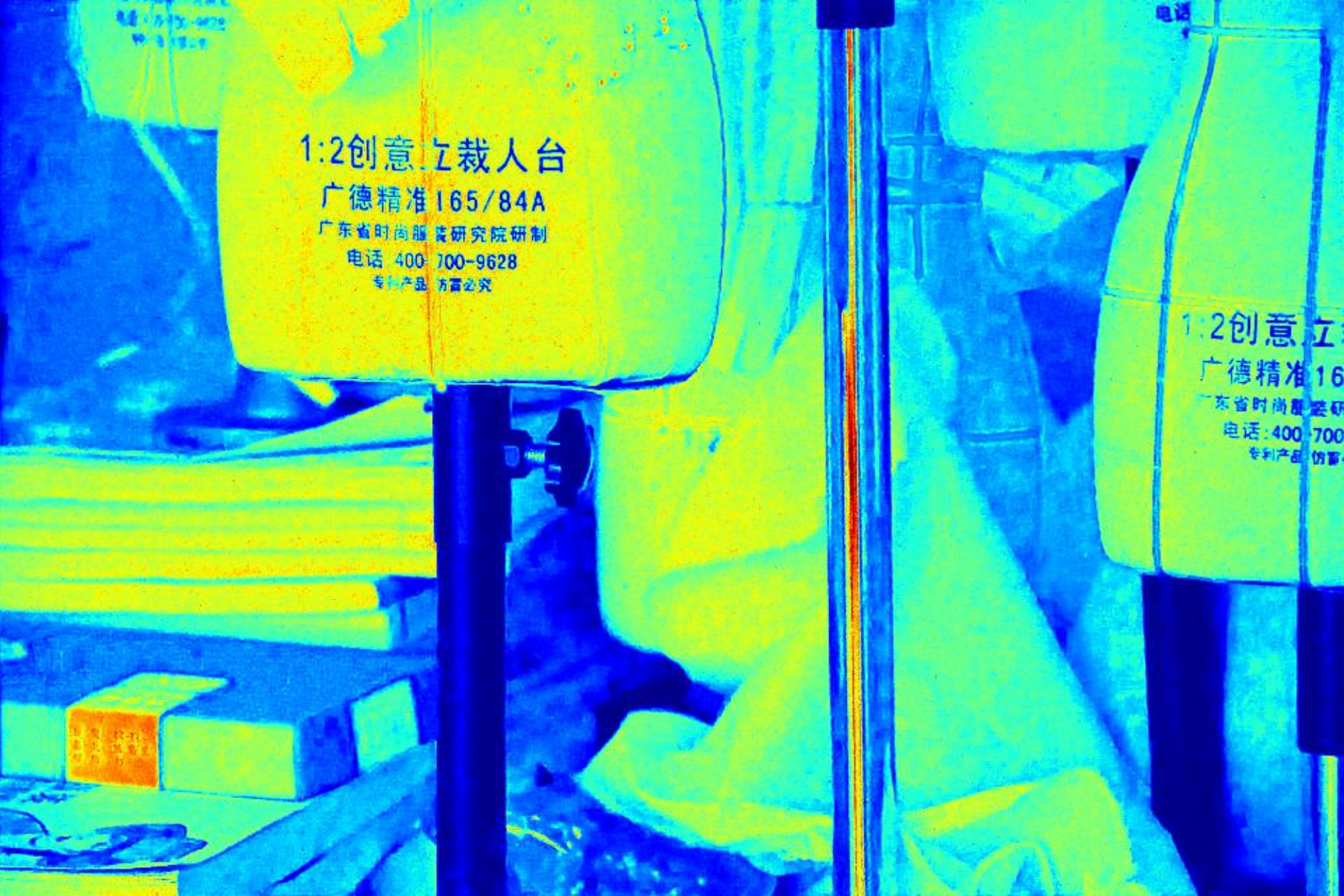}
	\end{subfigure}

    \begin{subfigure}{0.24\linewidth}
		\centering
		\includegraphics[width=\linewidth]{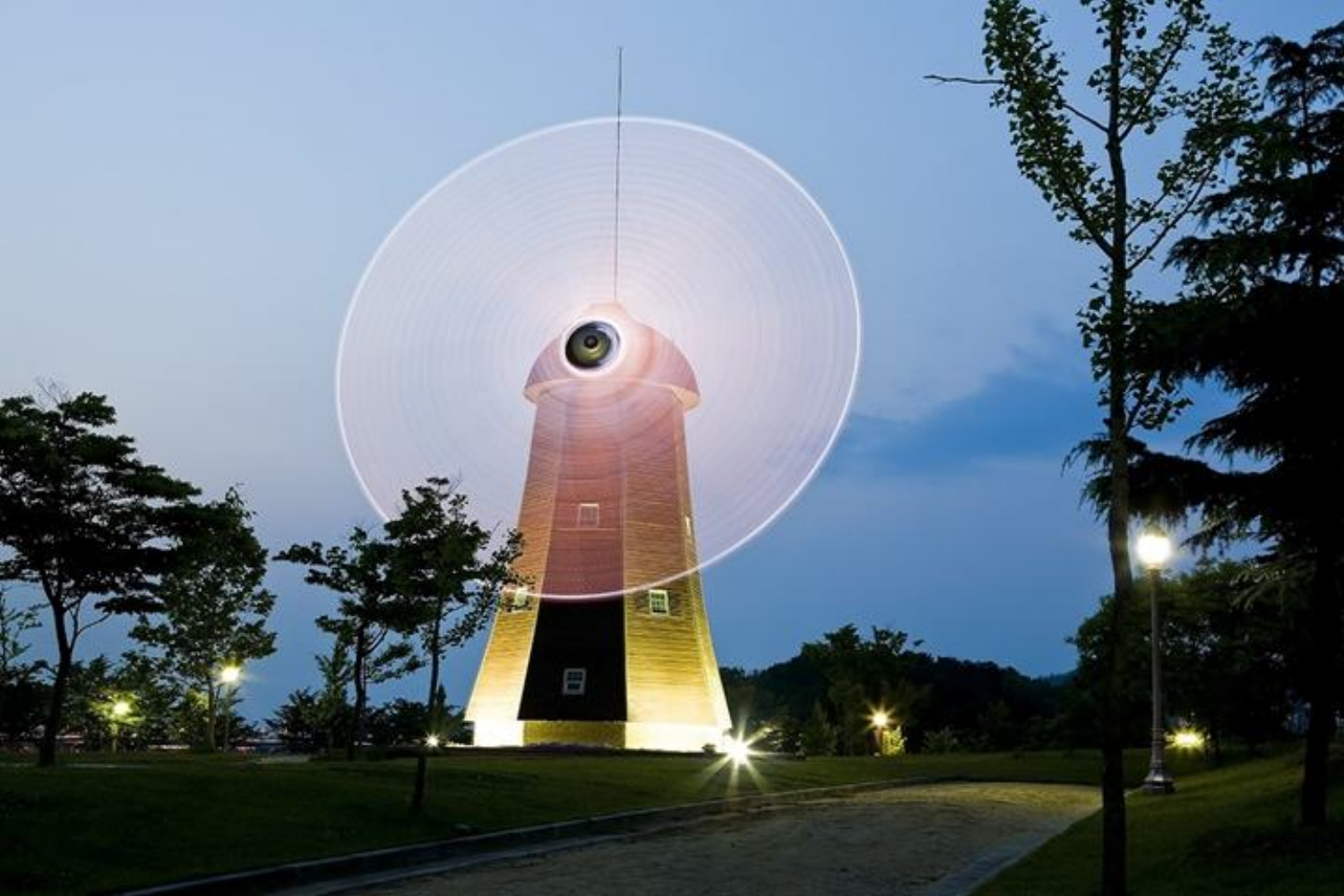} 
	\end{subfigure}
	\begin{subfigure}{0.24\linewidth}
		\centering
		\includegraphics[width=\linewidth]{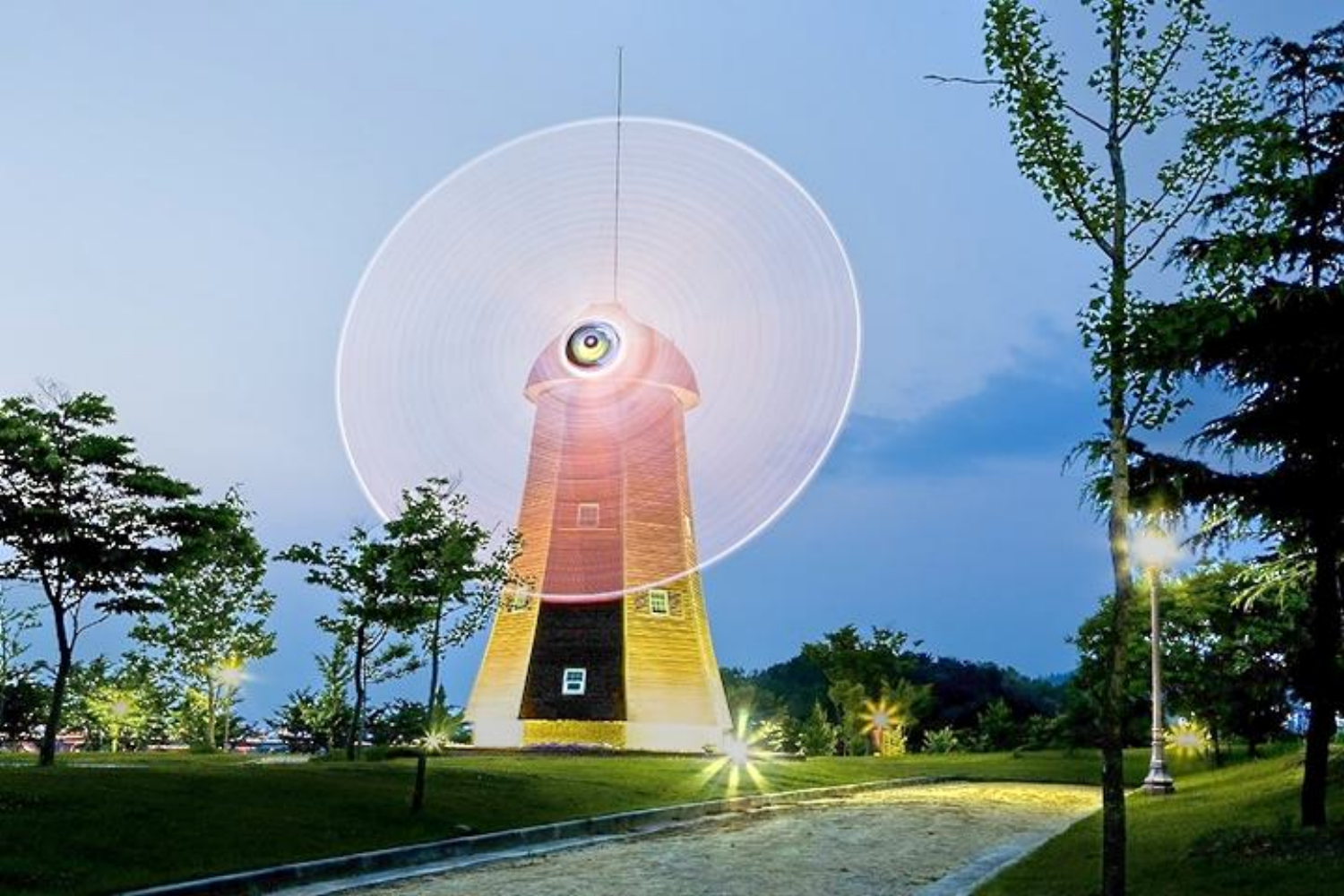} 
	\end{subfigure}
	\begin{subfigure}{0.24\linewidth}
		\centering
		\includegraphics[width=\linewidth]{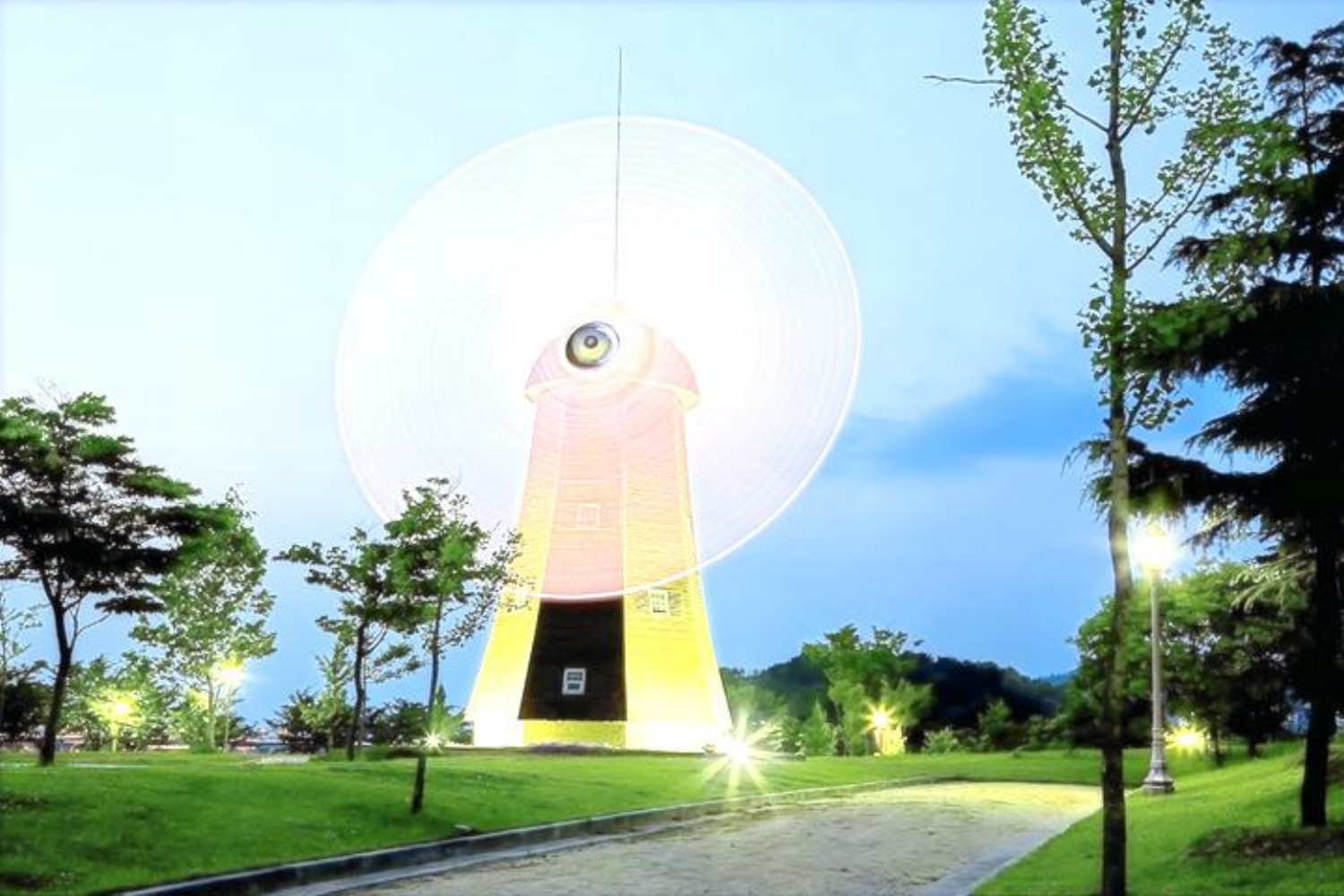}
	\end{subfigure}
    \begin{subfigure}{0.24\linewidth}
		\centering
		\includegraphics[width=\linewidth]{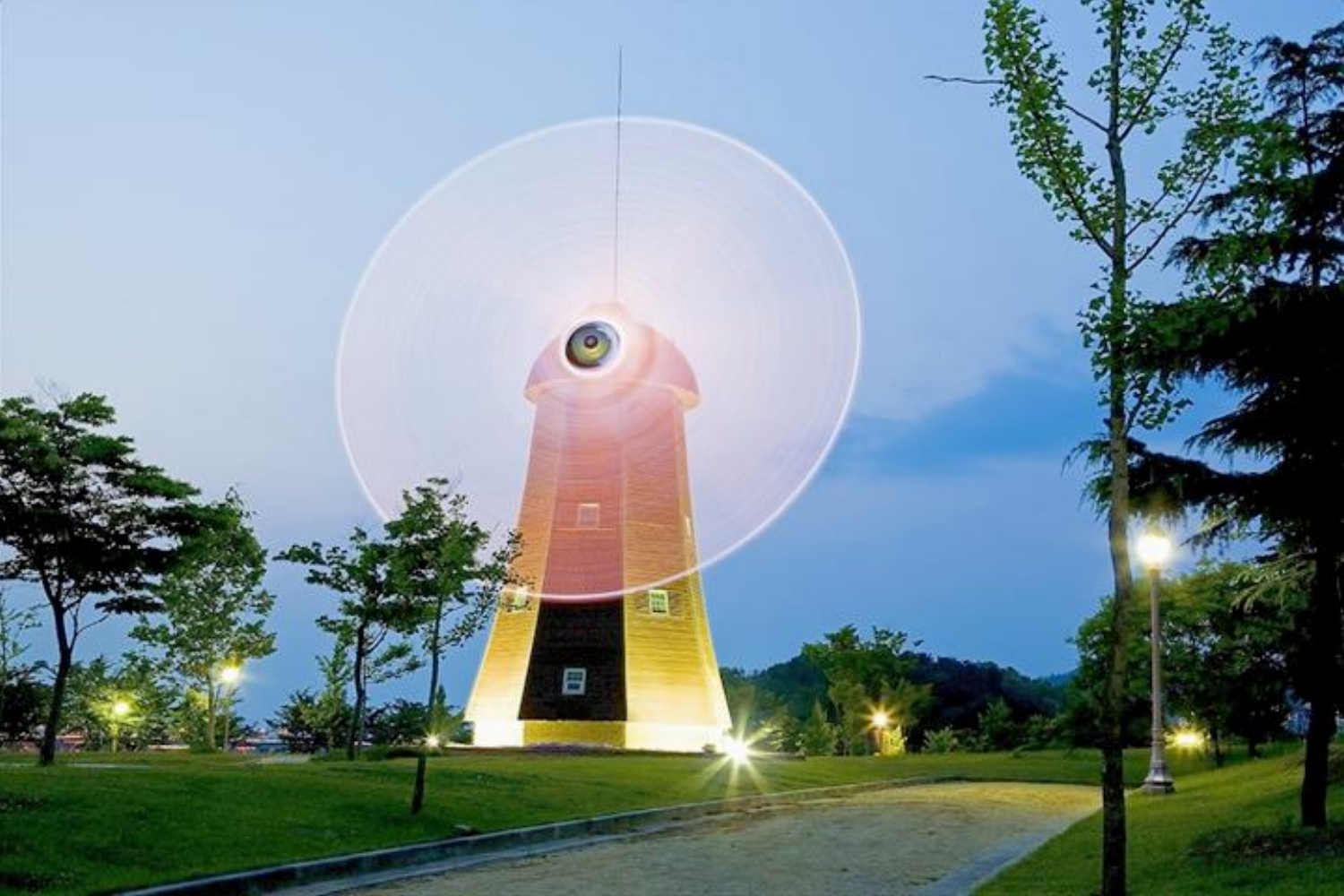}
	\end{subfigure}

    \begin{subfigure}{0.24\linewidth}
		\centering
		\includegraphics[width=\linewidth]{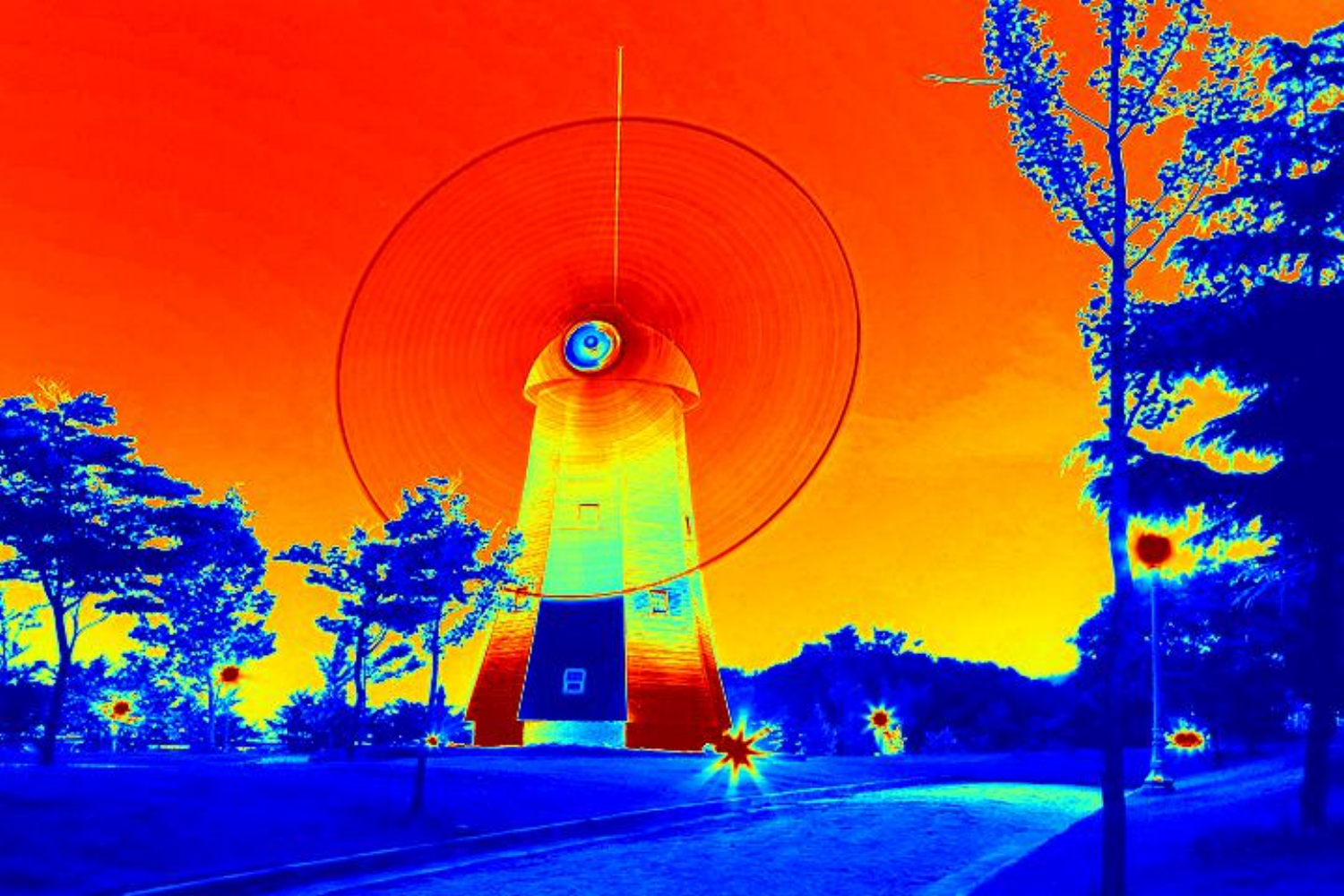} 
		\caption{Low-light}
		\label{Input_a}
	\end{subfigure}
	\begin{subfigure}{0.24\linewidth}
		\centering
		\includegraphics[width=\linewidth]{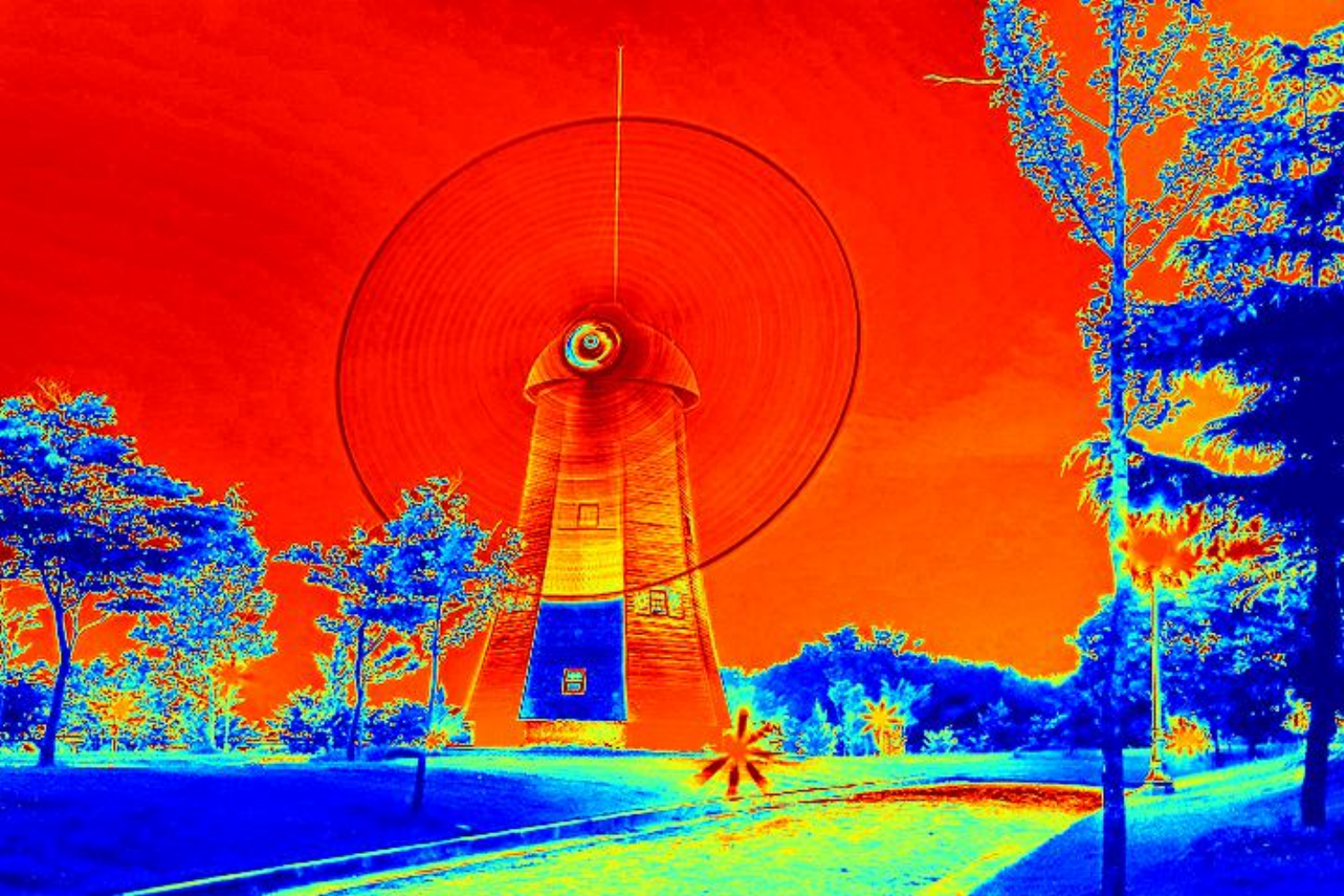} 
		\caption{ITRE}
		\label{ITRE_b}
	\end{subfigure}
	\begin{subfigure}{0.24\linewidth}
		\centering
		\includegraphics[width=\linewidth]{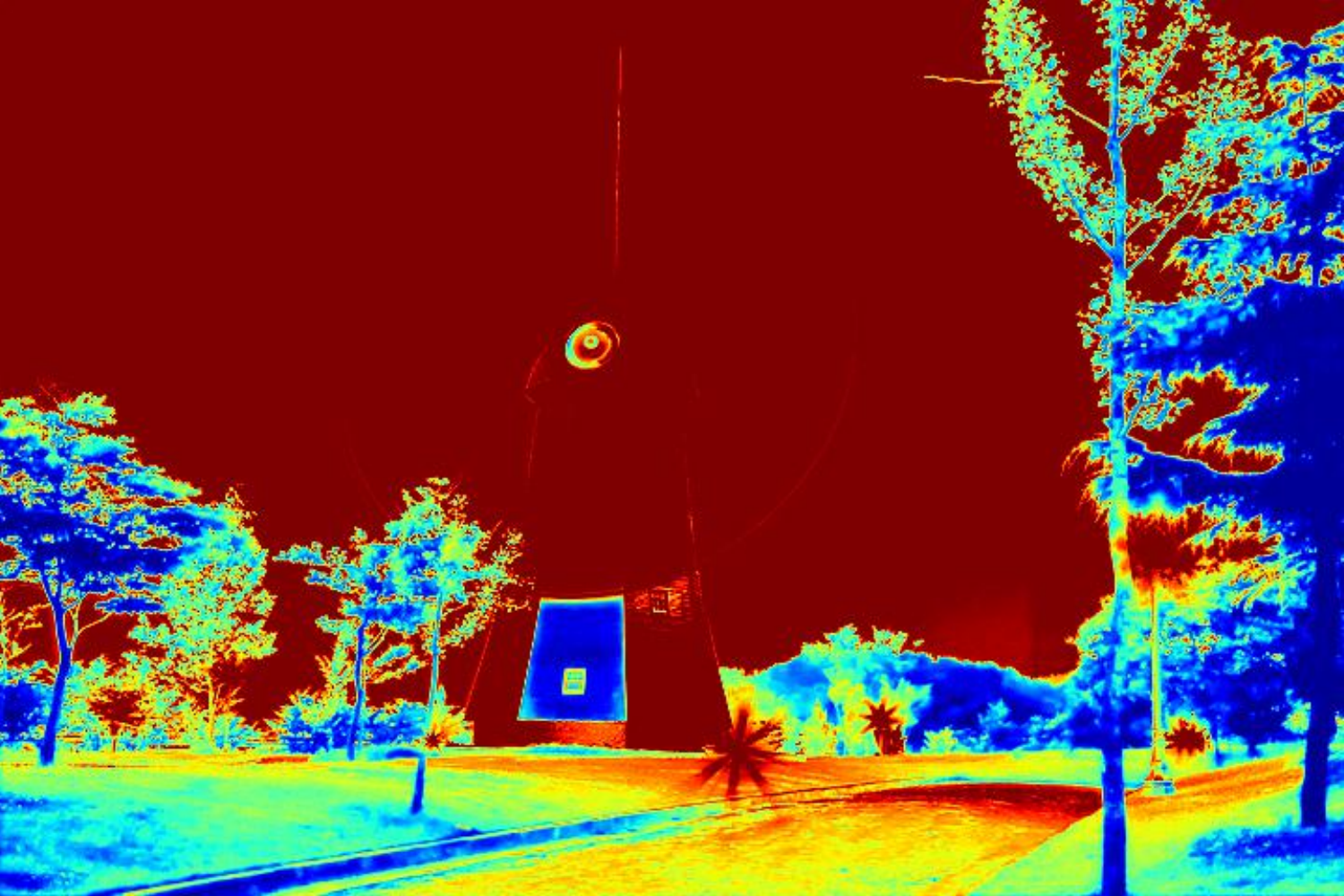}
		\caption{PPformer}
		\label{PPformer_c}
	\end{subfigure}
    \begin{subfigure}{0.24\linewidth}
		\centering
		\includegraphics[width=\linewidth]{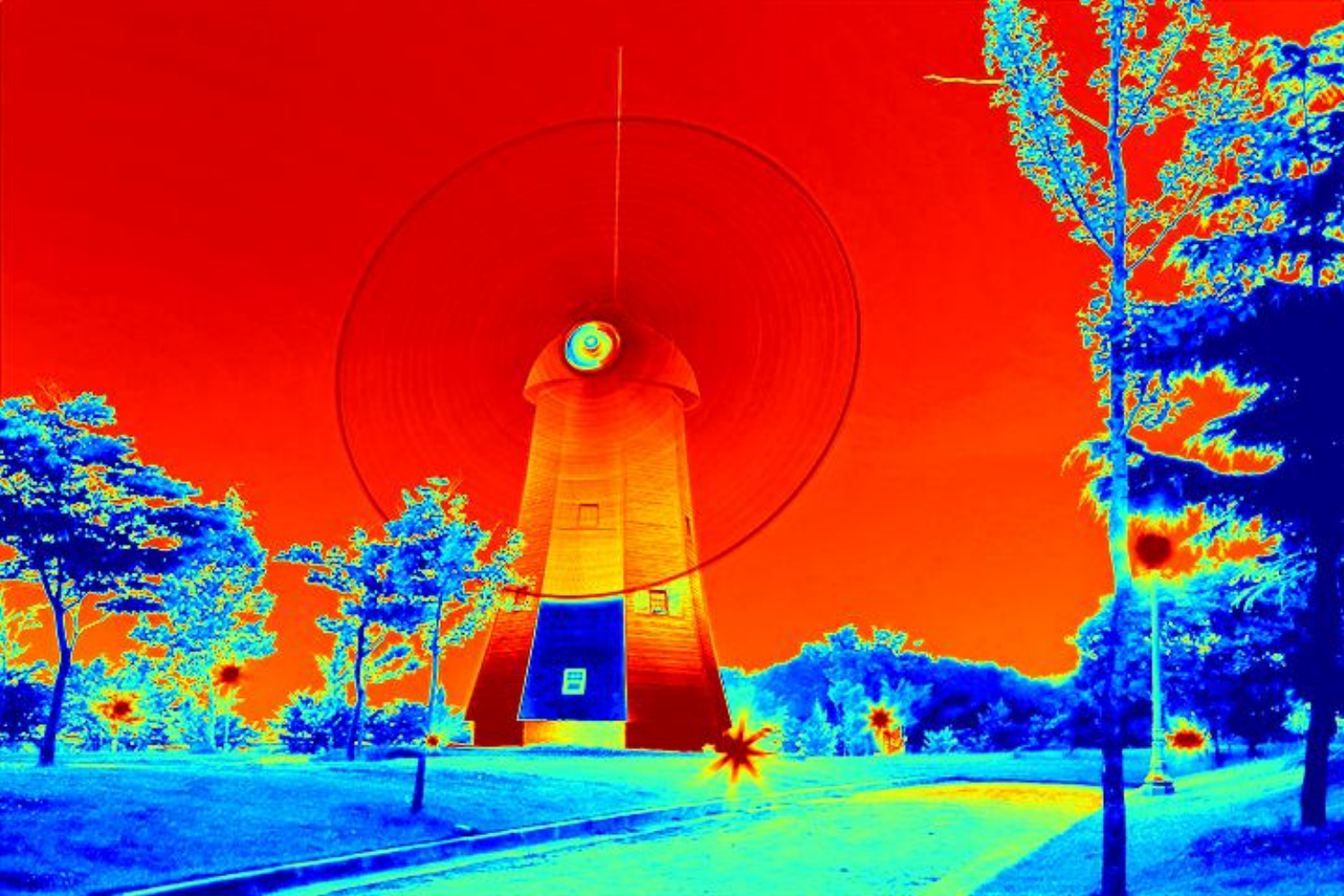}
		\caption{Proposed}
		\label{ALEN_d}
	\end{subfigure}
	
	\caption{Low-light images and their enhancements by ITRE, PPformer, and the proposed method.}
	\label{Mot_images}
\end{figure}

The primary challenge in low-light image enhancement lies in the difficulty of simultaneously addressing underexposed and overexposed regions. Figure \ref{Mot_images} presents two examples: one uniformly dark and another with uneven illumination, alongside their corresponding illumination distribution maps.

In column \ref{Input_a}, the first image suffers from uniform darkness, while the second exhibits uneven lighting. Column \ref{ITRE_b} presents results obtained using the ITRE method~\cite{wang2024itre}, which is based on Retinex theory. While ITRE improves the second image, it fails to address the dark regions in the first image due to its reliance on local enhancement. In contrast, column \ref{PPformer_c} shows results from the PPformer method~\cite{dang2024ppformer}, which utilizes a deep learning Transformer architecture. Although PPformer significantly enhances the illumination of the first image, overexposure in the second image results in detail loss. This phenomenon underscores a common limitation of global enhancement methods and their inherent difficulty in adapting to local illumination variations.

\begin{figure*}[ht]
	\centering
	\includegraphics[width=0.9\textwidth]{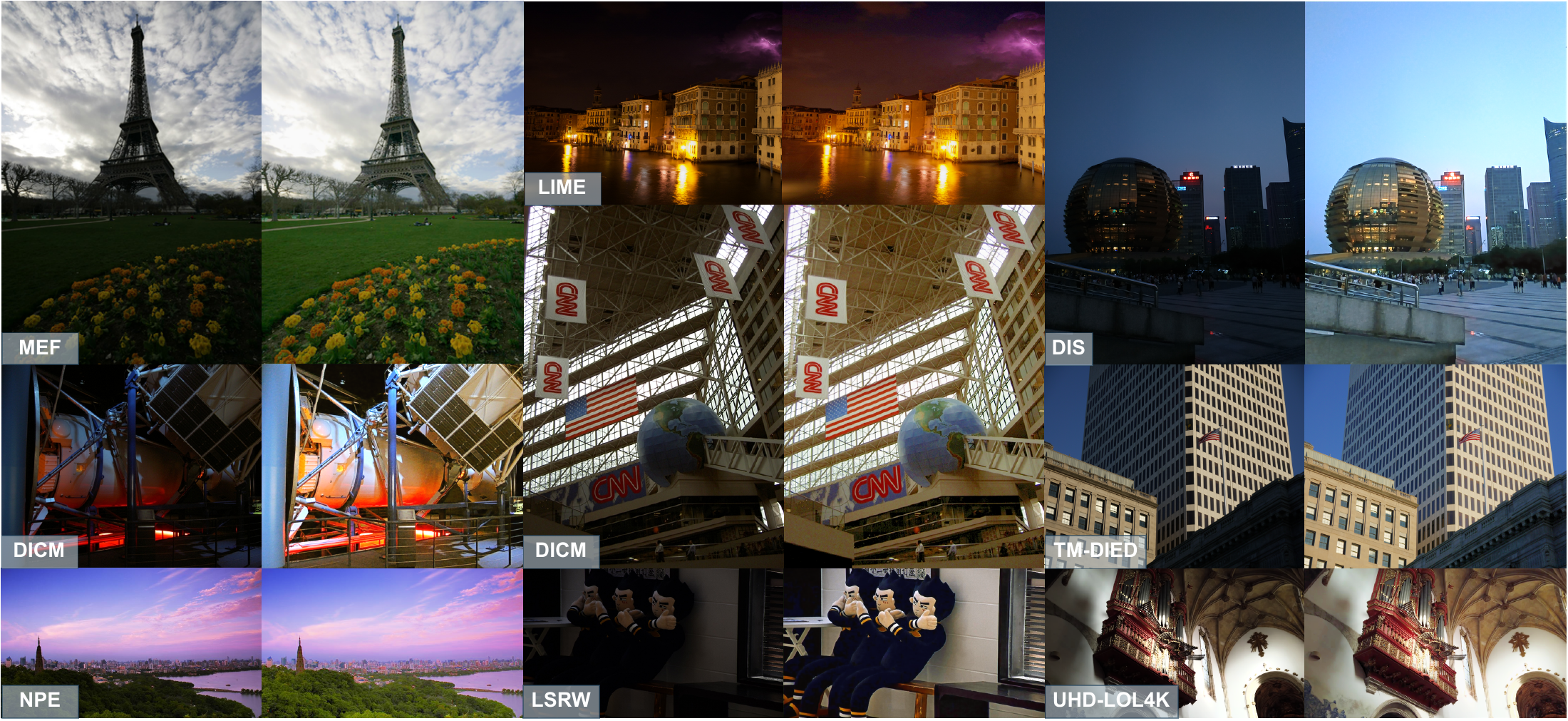}
	\caption{Example of enhancing images in low-light environments using the ALEN method across different scenarios.}
	\label{example_enhancement}
\end{figure*}

To overcome these challenges, a novel method called the Adaptive Light Enhancement Network (ALEN) is presented, inspired by the Low-Light Image Enhancement Algorithm (LoLi-IEA)~\cite{perez2023loli}. ALEN incorporates a dynamic classifier that determines whether an image requires local or global enhancement, applying the most suitable technique to each region. This adaptive approach effectively mitigates both underexposure and overexposure issues, ensuring balanced and uniform illumination across the entire image. The effectiveness of ALEN is demonstrated in various scenarios, as shown in Figure \ref{example_enhancement}.

\subsection{Contributions}

This paper introduces ALEN for low-light image enhancement in real-world environments. The proposed method addresses challenges in diverse lighting conditions while maintaining computational efficiency. The main contributions are:

\begin{itemize}
    \item \textbf{A novel approach to enhance both uniform and non-uniform image light conditions:} The ALEN technique is introduced, which is constructed from three distinct sub-networks: Swin Light-Classification Transformer (SLCformer) for illumination classification, Single-Channel Network (SCNet) for single-channel estimation, and Multi-Channel Network (MCNet) for multi-channel estimation.
    \item \textbf{An illuminance classification dataset based on modified existing data:} The Global-Local Illuminance (GLI) dataset is introduced, which is obtained from a modification of an existing set of images, based on human perception and histogram analysis to categorize illuminance enhancement as either global or local.
    \item \textbf{A novel dataset for evaluating real-world low-light images:} The Diverse Illumination Scene (DIS) dataset is presented, encompassing real-world low-light images in various indoor and outdoor scenes, spanning both day-time and night-time periods.  
\end{itemize}

The remainder of the paper is structured as follows: Section \ref{RW} presents the most relevant related work, while Section \ref{MM} introduces the proposed method. Section \ref{RD} details the experiments and results obtained, and finally, Section \ref{CF} presents the conclusions and future work.

\section{Related Work}
\label{RW}
Over the past decades, various methods have been explored to enhance low-light images, from traditional approaches to advanced deep-learning techniques. This section categorizes related works according to the different processing techniques used.

\subsection{Histogram equalization-based methods}

Classical methods, such as histogram adjustment and equalization \cite{pizer1987adaptive,kaur2011survey}, have been fundamental in enhancing the visibility of images under low-light conditions. One of the first approaches was developed in 2007, Ibrahim et al.~\cite{ibrahim2007brightness} proposed the Brightness Preserving Dynamic Histogram Equalization (BPDHE), a method that uses a Gaussian filter to smooth the input histogram and then applies equalization over segments defined by local maxima, preserving the mean brightness of the original image. That same year, Abdullah-Al-Wadud et al.~\cite{abdullah2007dynamic} introduced Dynamic Histogram Equalization (DHE), a histogram equalization-based technique that enhances low-light images by dividing the histogram into segments based on local minima, assigning them specific ranges of grayscale levels, and verifying their redistribution to prevent dominance.

Later in 2013, Hitam et al.~\cite{hitam2013mixture} introduced the so-called Contrast Limited Adaptive Histogram Equalization (CLAHE), a method specifically designed to enhance underwater images. This approach applies CLAHE to RGB and HSV color models, combining the results using the Euclidean norm. More recently, in 2022, Thepade et al.~\cite{thepade2022contrast} introduced an innovative fusion combining the CLAHE with BPDHE along with a linearly weighted color restoration technique. The next year, in 2023, Han et al.~\cite{han2023low} proposed the Histogram Equalization–Multiscale Retinex Combination (HE-MSR-COM) method, which consists of three main modules: MSR enhancement for edges, HE enhancement for illumination, and frequency domain fusion. This approach adapts edge and illumination information using derived weights for different scenarios. Finally, in 2024, Jeon et al.~\cite{jeon2024low} presented a fast method for low-light image enhancement based on an atmospheric scattering model applied to an inverted image.

\subsection{Retinex-based methods}

Retinex theory~\cite{land1971lightness} enhances visual perception by adjusting the color space through techniques that optimize dynamic range and color fidelity, especially under varying lighting conditions.~\cite{reflectance2011retinex,hussein2019retinex}

In 2016, Guo et al.~\cite{guo2016lime} introduced the Low-light Image Enhancement via Illumination Map Estimation (LIME) method, which individually estimates the illumination of each pixel in a low-light image, refines this illumination map and applies it to enhance the image quality. A few years later, in 2018, Wei et al.~\cite{wei2018deep} proposed Retinex-Net, which consists of two sub-networks: Decom-Net, for the decomposition of the image into reflectance and illumination, and Enhance-Net, to adjust the illumination and improve the image quality. In 2022, Wu et al.~\cite{wu2022uretinex} presented the URetinex method, which includes three key modules: initialization, unfolded optimization, and illumination adjustment. These modules allow the image to be decomposed into reflectance and illumination, adapting the results to suppress noise and preserve details.

In 2023, Yi et al.~\cite{yi2023diff} introduced the Diff-Retinex method, which proposes a generative diffusion model for low-light image enhancement. It uses a Retinex Transformer Decomposition Network (TDN) to decompose the image into illumination and reflectance maps. That same year, Cai et al.~\cite{cai2023retinexformer} presented Retinexformer for enhancing low-light images and restoring corruptions using Retinex theory. They also developed an Illumination-Guided Transformer (IGT) for modeling non-local interactions under varying lighting conditions. In 2024, Bai et al.~\cite{bai2024retinexmamba} proposed RetinexMamba, combining traditional Retinex methods' intuition with advanced deep learning techniques from Retinexformer. It improves processing speed with State Space Models (SSMs), introduces new illumination estimators and restoration mechanisms, and enhances model interpretability with a Fused Attention mechanism.

\subsection{Unsupervised and self-calibrated approaches}

Unsupervised approaches, such as self-calibrated methods, enhance low-light images by automatically adjusting parameters to optimize their visual quality.

In 2019, Zhang et al.~\cite{zhang2019dual} introduced a method that improves images through dual illumination estimation. It separately corrects underexposed and overexposed parts and then adaptively merges visually better-exposed parts from multiple images to create a well-exposed single image. In 2021, Zhang et al.\cite{zhang2021unsupervised} proposed a method that uses a Histogram Equalization Prior (HEP) to guide the process. This unsupervised learning process is divided into the Light Up Module (LUM) and Noise Disentanglement Module (NDM).

Later in 2022, Jin et al.~\cite{jin2022unsupervised} presented an unsupervised method for enhancing night images using layer decomposition and light-effects suppression. In the same year, Ma et al.~\cite{ma2022toward} introduced SCI, employing cascaded learning with weight sharing to reduce computational cost and adapt to various scenes through unsupervised training. In 2023, Liang et al.~\cite{liang2023iterative} proposed the Contrastive Language-Image Pre-Training (CLIP-LIT), a method that utilizes CLIP to distinguish between backlit and well-lit images. It improves the network through a prompt-based learning framework that optimizes text-image similarity. 

More recently, in 2024, Xu et al.~\cite{xu2024degraded} introduced Degraded Structure and Hue Guided Auxiliary Learning (SHAL-Net), an unsupervised method that utilizes auxiliary learning guided by the degraded structure and tone of the image. On the other hand, Zhao et al.\cite{zhao2024non} proposed a self-supervised low-light enhancement method that simplifies illumination estimation without regularization. Wang et al.\cite{wang2024itre} followed with ITRE, addressing noise, artifacts, and over-exposure, supported by a Robust-Guard module. Liang et al.~\cite{liang2024pie} presented PIE, a physics-inspired approach that leverages contrastive learning with unpaired data and unsupervised segmentation. Finally, Shi et al. introduced MaCo~\cite{shi2024maco}, a low-light image enhancement method that uses magnitude control for pixel-wise enhancement and avoids overexposure. It employs LCE-Net for low-resolution feature estimation and HIE Module for upsampling.

\subsection{Deep learning-based methods}

Deep Learning methods yield favorable generalization results based on the quantity and quality of input images. In 2020, Guo et al.~\cite{guo2020zero} proposed the Zero-reference Deep Curve Estimation (Zero-DCE) method, which enhances lighting using the deep network DCE-Net. This network adjusts specific image curves and employs non-reference loss functions to guide learning.

The following year, in 2021, Liu et al.~\cite{Liu21} proposed the Retinex-Inspired Unrolling With Cooperative Prior Architecture Search architecture, which models the intrinsic structure of underexposed images and develops optimization processes to construct a comprehensive propagation structure. On the other hand, in the same year, Zheng et al.~\cite{zheng2021adaptive} introduced the Adaptive Unfolding Total Variation Network (UTVNet) architecture, which estimates noise level and learns a corresponding map through total variation regularization. Guided by this map, UTVNet effectively recovers fine details and suppresses noise. 

Later in 2022, Fan et al~\cite{fan2022half} introduced HWMNet, an image enhancement network based on the enhanced hierarchical model M-Net+. This network uses a half-wavelet attention block to enrich wavelet domain features. Within the same year, Wang et al.~\cite{wang2022lowlight} proposed a method using a normalizing flow model to enhance low-light images toward normal exposure. This approach employs an invertible network that adjusts low-light images to map them to a Gaussian distribution of normally exposed images. On the other hand, Cui et al.~\cite{cui2022you} presented a lightweight and fast Illumination Adaptive Transformer (IAT) to restore normal sRGB images from low-light, underexposed, or overexposed conditions.

More recently, in 2023, Ma et al.~\cite{ma2023bilevel} introduced a method to enhance images in low-light scenes using a learning approach that explores relationships between different scenes. They used a bi-level learning framework to optimize hyperparameters and improve the encoder's adaptability to diverse scenes. In the same year, Ye et al.\cite{ye2023glow} introduced EMNet, a network that leverages external memory to enhance images by capturing sample-specific properties. In addition, Fu et al.~\cite{fu2023learning} proposed PairLIE, an approach that learns adaptive prior values from pairs of low-light images. Finally, in 2024, Dang et al.~\cite{dang2024ppformer} introduced PPformer, a lightweight CNN-transformer hybrid designed for low-light image enhancement through pixel- and patch-wise cross-attention. Similarly, Fang et al. proposed MSATr~\cite{fang2024non}, a multi-scale attention Transformer for low-light image enhancement. It adapts to varying illumination by refining pixel-level features and improving brightness perception.

\section{Materials and Methods}
\label{MM}
This section presents the datasets used and generated, describes the ALEN design criterion and its modules, and details the experimental details. 

\subsection{Datasets}
Four different datasets are used in this proposal: the custom-developed Global-Local Illumination (GLI) dataset is designed for illumination classification. For the estimation part, two public datasets are employed for illumination estimation, and one specific dataset is used for color estimation. These datasets are described in detail below.

\subsubsection{Dataset for illuminance classification}
\label{GLI_D}

A key challenge in the proposed framework is developing a dataset to distinguish between images needing global or local enhancement. Due to the lack of a suitable dataset, a dataset called GLI was constructed from multiple public databases focused on low-light image enhancement. Each dataset offers unique features for estimating image enhancement under these conditions. The primary goal is identifying illumination variations and common features and creating a classification dataset. 

\begin{figure*}[ht]
	\centering
    \begin{subfigure}{0.35\textwidth}
    	\includegraphics[width=\textwidth]{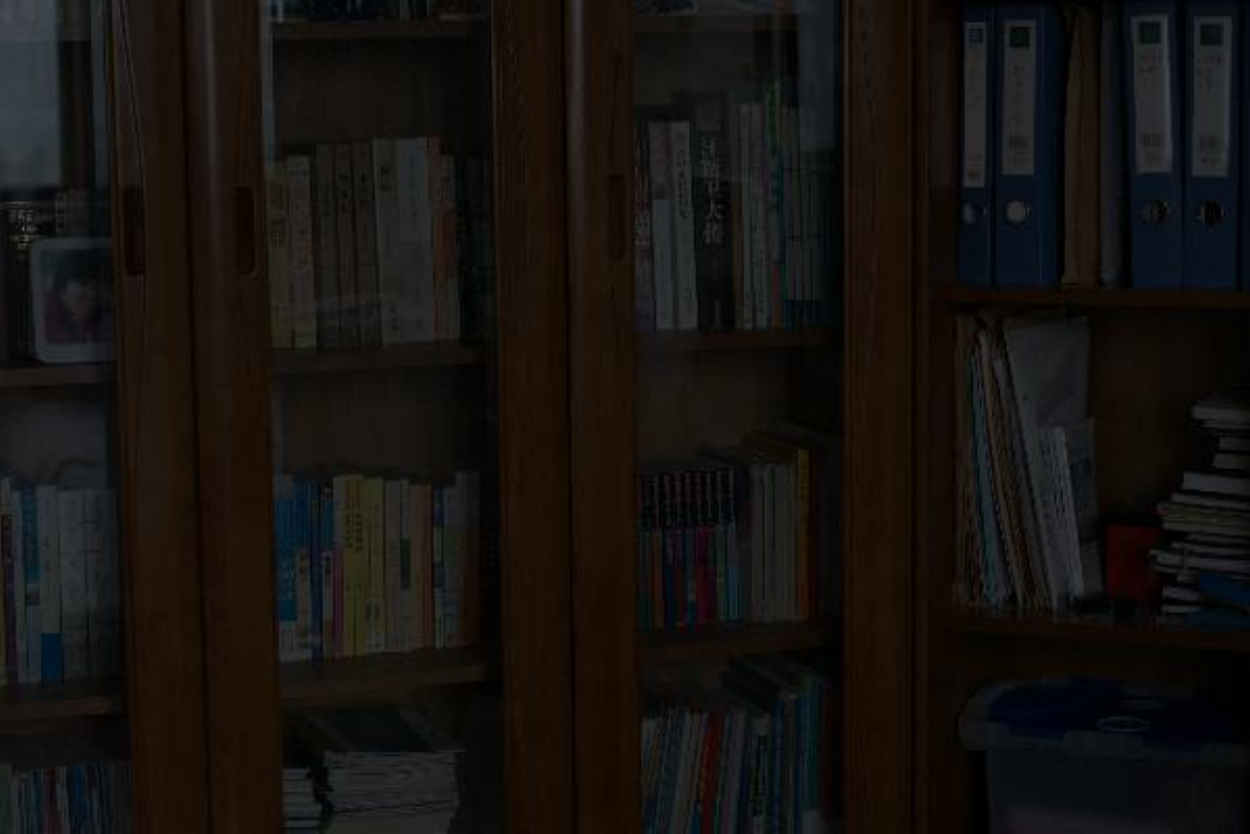}
  	\end{subfigure}
    \hspace{1.4cm}
    \begin{subfigure}{0.35\textwidth}
    	\includegraphics[width=\textwidth]{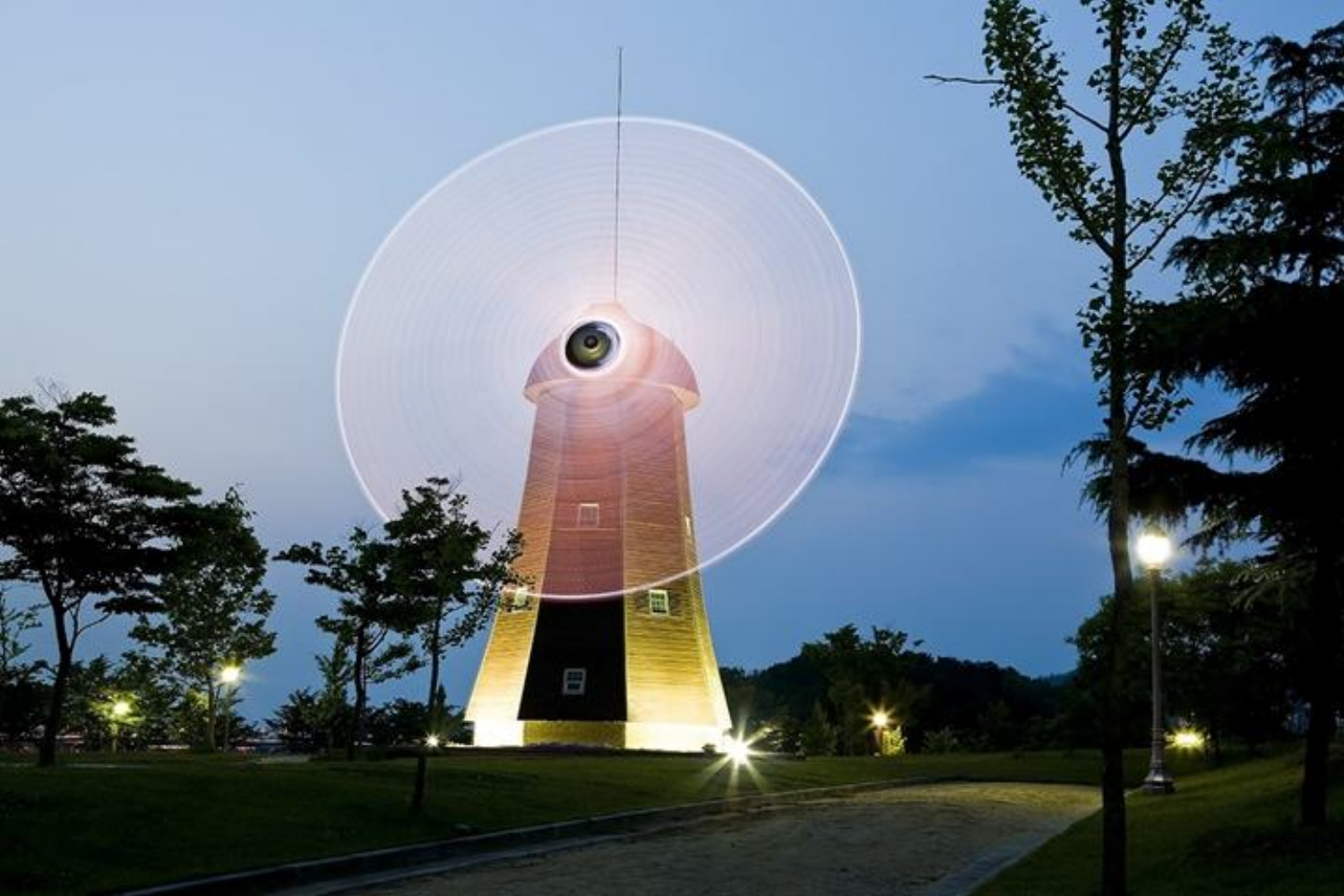}
	\end{subfigure} 
	\hfill
    \begin{subfigure}{0.45\textwidth}
		\includegraphics[width=\textwidth]{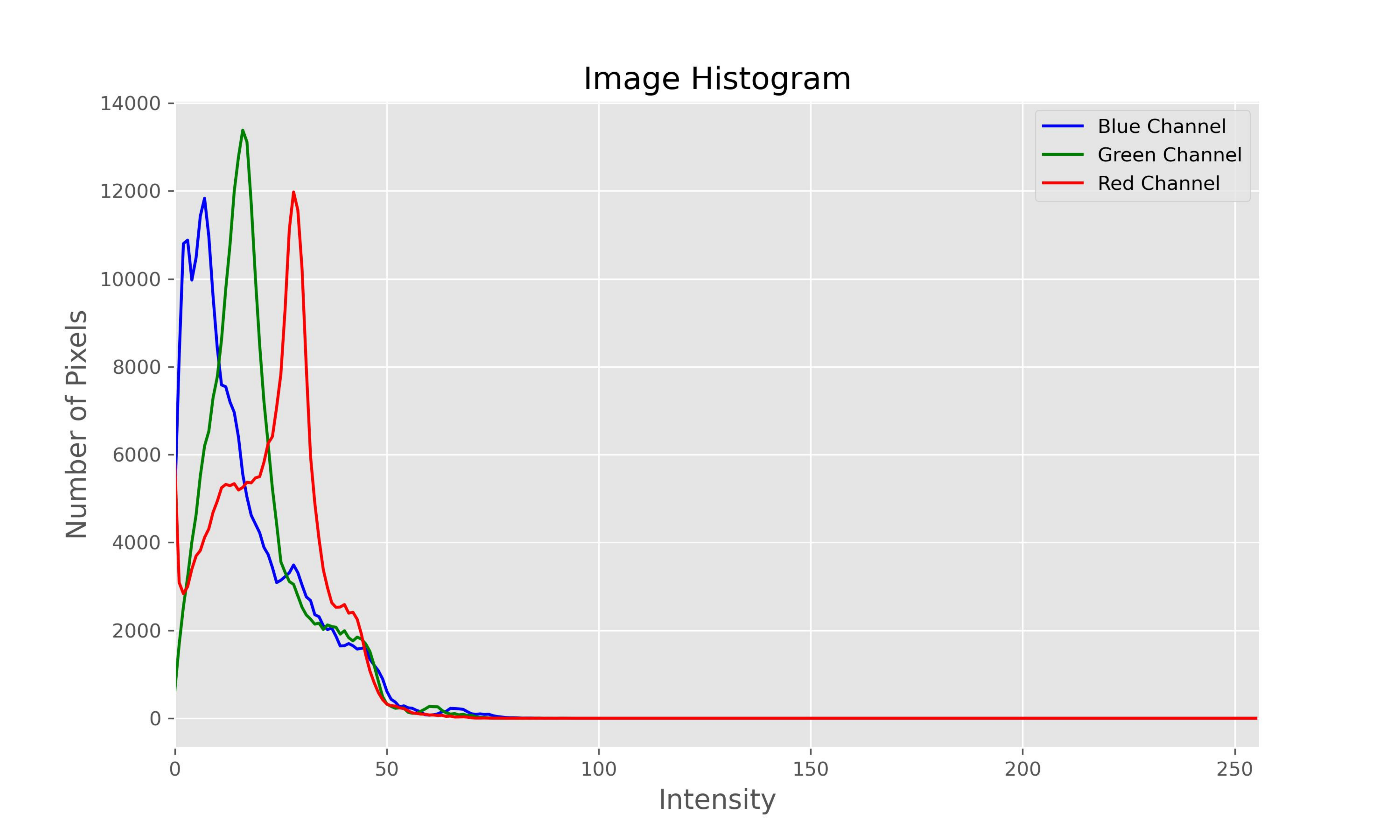}
		\caption{Global illumination features}
		\label{fig:GH}
	\end{subfigure} 
	\begin{subfigure}{0.45\textwidth}
		\includegraphics[width=\textwidth]{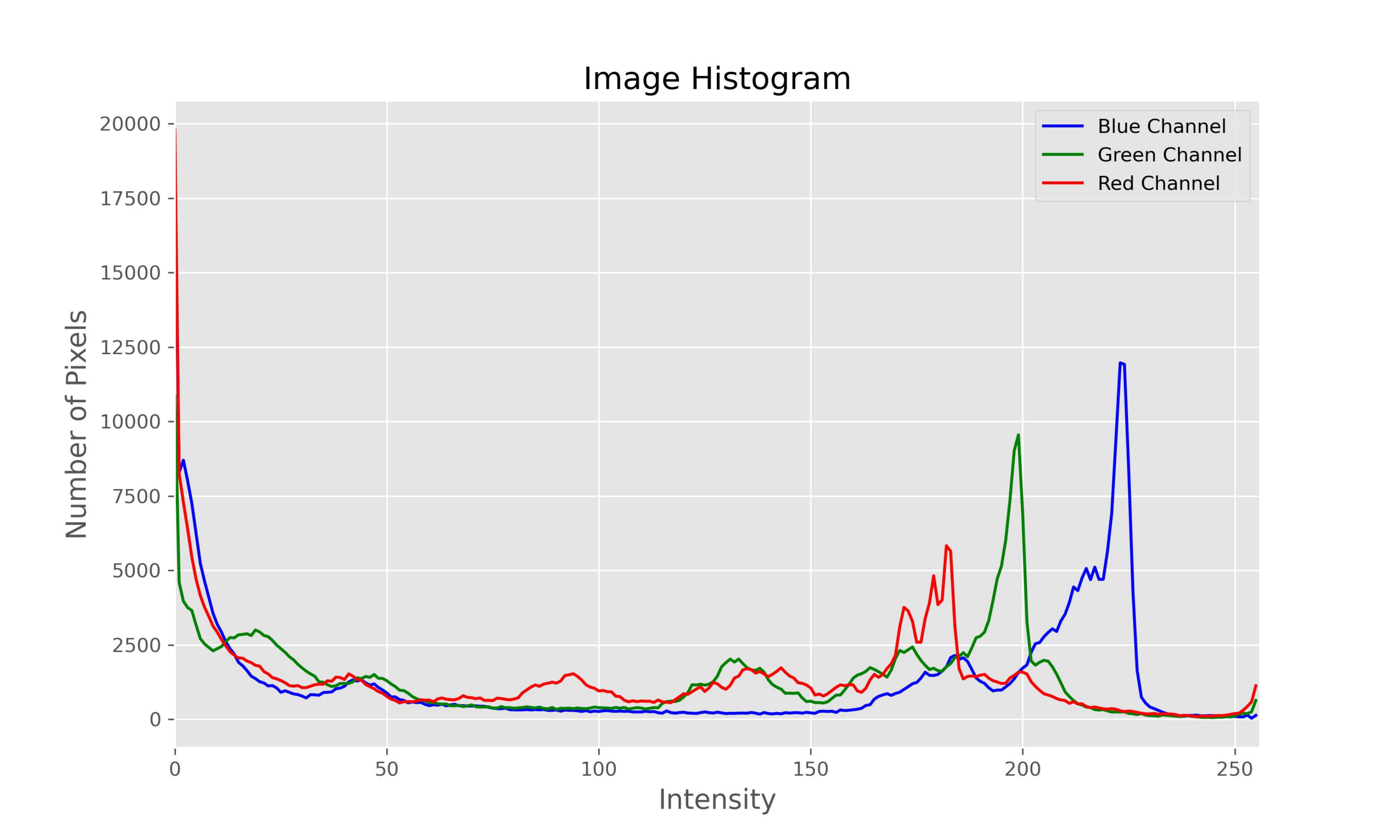}
		\caption{Local illumination features}
		\label{fig:LH}
	\end{subfigure}
	\caption{Comparison of global and local illumination features.}
	\label{fig_illumination_features}
\end{figure*}

During the initial stages of creating the GLI dataset, each image undergoes visual inspection and classification from a human perspective to determine whether global or local enhancement is required. To aid in this process, histograms of various image types are analyzed to evaluate illumination uniformity. Analyzing histograms is essential for understanding how pixel intensities are distributed in low-light images. Pixel intensity is assessed individually for each channel in RGB-format images, ranging from 0 to 255. Based on the histogram analysis shown in Fig. \ref{fig_illumination_features}, a detailed representation of intensity value distributions is provided. Figure \ref{fig:GH} shows an example of an image requiring global enhancement from the LOL~\cite{wei2018deep} dataset, while Fig. \ref{fig:LH} presents the histogram of an image requiring local enhancement taken from the DICM~\cite{lee2013contrast} dataset.

Through this analysis, it can be inferred that images requiring global enhancements predominantly exhibit pixel intensities on the left side of the histogram, resulting in decreased values on the right side. Conversely, images needing local enhancements display intensity values covering the entire histogram range, focusing mainly on the extremes near the road and tree zones. 

\begin{figure}[ht]
	\centering
	\begin{subfigure}{0.45\textwidth}
		\includegraphics[width=\textwidth]{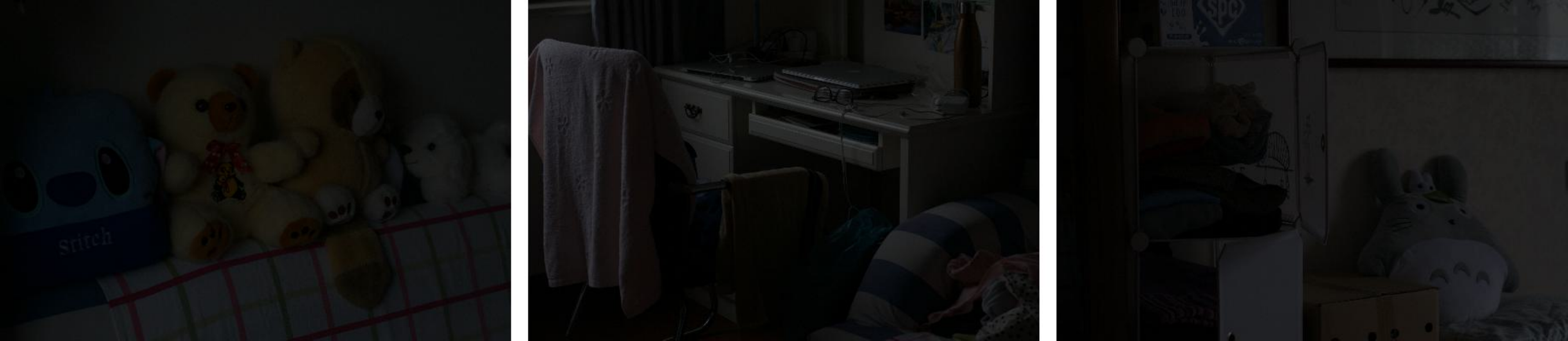}
		\caption{Globally labeled dataset}
		\label{global_ds}
	\end{subfigure} 
 
	\begin{subfigure}{0.45\textwidth}
		\includegraphics[width=\textwidth]{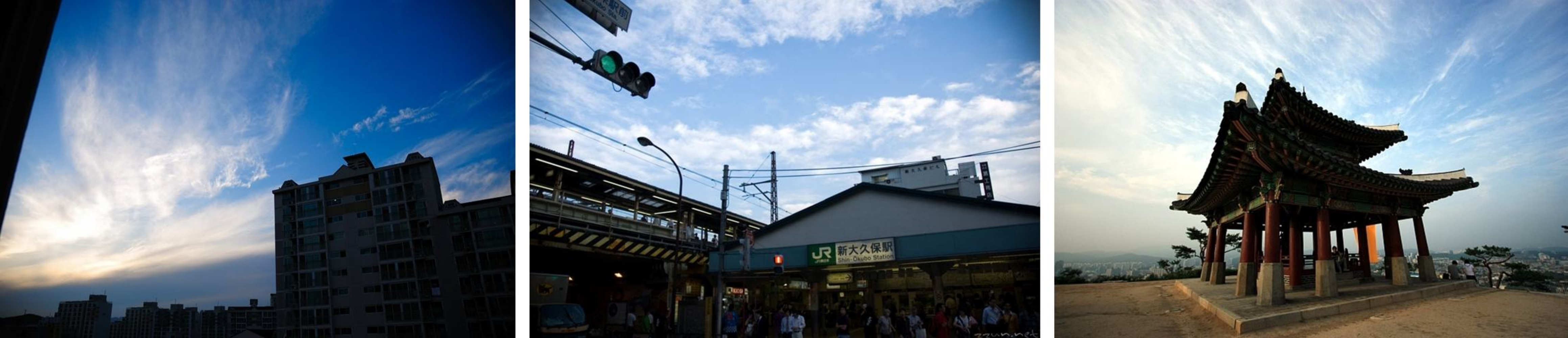}
		\caption{Locally labeled dataset}
		\label{local_ds}
	\end{subfigure}
	\caption{Example images from the training set for classification.}
	\label{fig_local_global_classification}
\end{figure}

In the second stage of creating the GLI dataset, images are converted to grayscale after the human visual inspection, marking the beginning of parameter-based classification to determine whether an image belongs to the local or global category. For any grayscale image $I$, the histogram $Hist(i)   $ is computed by quantifying a pixel according to its respective intensity levels. This can be mathematically expressed as:

\begin{equation}
    \label{eq:delta}
	Hist(i) = \sum_{i=0}^{255} \delta(I(p), i)
\end{equation}

\noindent where $\delta$ is a function derived from the Kronecker delta function. Its output is $1$ if its two parameters are equal and $0$ otherwise. Therefore, if the intensity of pixel $p$ is equal to $i$, $\delta(I(p), i)$ returns $1$; otherwise, it returns $0$. By summing these values for all pixels in the image, the count of pixels displaying intensity $i$ is obtained. The threshold defining the image category labels is mathematically set as follows: 

\begin{equation}
	I_{\text{thr}} = \max \{ i \ | \ Hist(i) \geq T \}
    \label{eq:thres}
\end{equation}
\noindent where $Hist(i)$ denotes the count of pixels with intensity $i$, while $T$ represents the predefined threshold for image classification. According to Eq. \ref{eq:thres}, $I_{\text{thr}}$ represents the sought threshold intensity, crucial for distinguishing whether an image is classified as global or local. 
A total of 2000 images were used, divided into two main categories: 1000 images labeled as ``global" and another 1000 as ``local". Figure \ref{global_ds} shows an example of ``global" labeled images from the LOL dataset, while Fig. \ref{local_ds} presents an example of ``local" labeled images from the DICM dataset.

\subsubsection{Datasets for illuminance and color estimation}

The illumination estimation sections of the proposed method use two specific datasets: HDR+ \cite{hasinoff2016burst} for global illumination enhancement and Synthetic Low-Light (SLL) \cite{lv2021attention} for local illumination enhancement. While the color estimation section employs the MIT-Adobe FiveK \cite{bychkovsky2011learning}. HDR+ dataset \cite{hasinoff2016burst} used for global illumination enhancement consists of 3640 scenes, each with 2 to 10 photos in their original format. Specifically, the version selected by Zeng et al. \cite{zeng2020learning} was used, which includes 922 paired images of low and normal illumination. 

Synthetic Low-Light (SLL) dataset \cite{lv2021attention} is generated from original images of public datasets such as COCO \cite{lin2014microsoft} or ImageNet \cite{deng2009imagenet}. This dataset employs strategies including gamma adjustments, reduced dynamic range to simulate loss of detail in dark areas, and the addition of various types of noise to replicate low-light conditions. Additionally, SLL provides multiple exposure levels (original, 1/4x, 1/8x, 1/16x, and 1/32x) to enhance generalization. This dataset, comprising 22,656 paired training images and 965 paired test images, is specifically designed to improve local illumination.

\begin{figure}[ht]
	\centering
	\begin{subfigure}{0.32\linewidth}
		\centering
		\includegraphics[width=\linewidth]{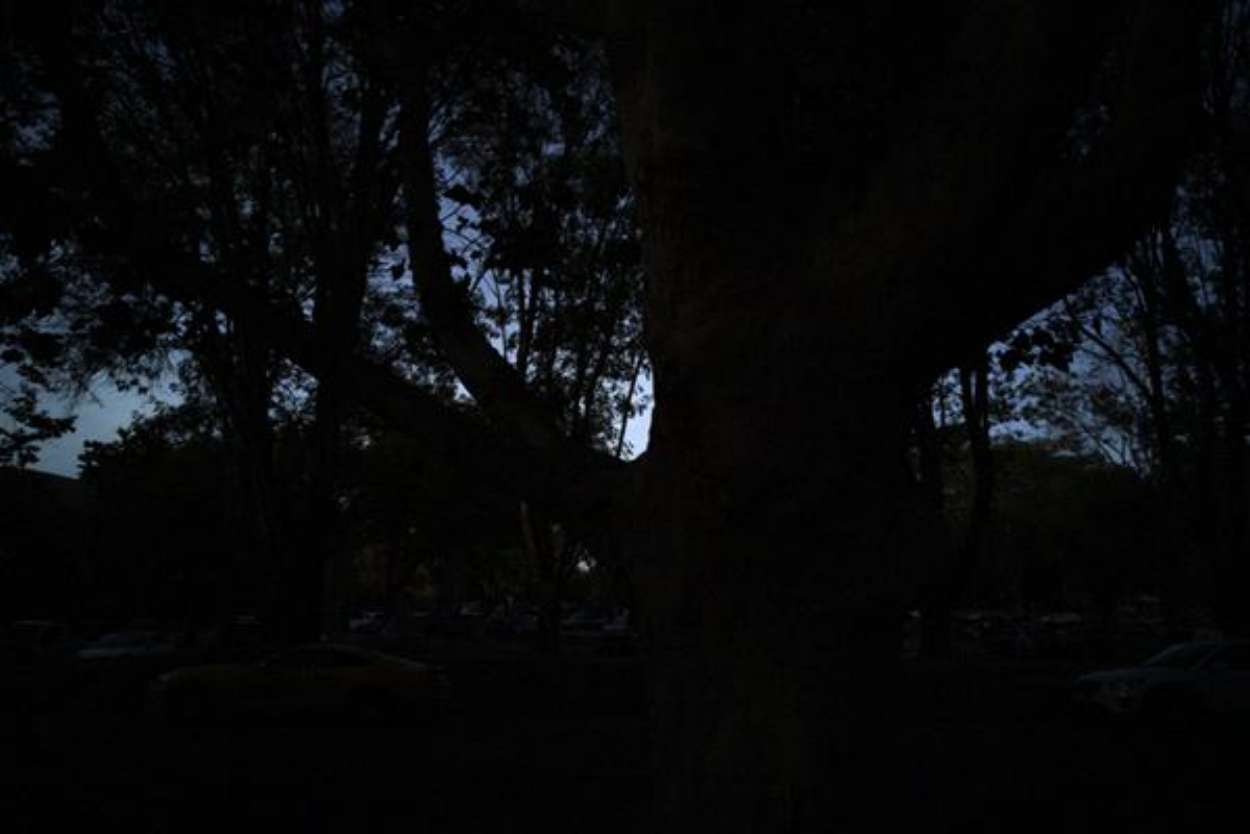} 
	\end{subfigure}
	\begin{subfigure}{0.32\linewidth}
		\centering
		\includegraphics[width=\linewidth]{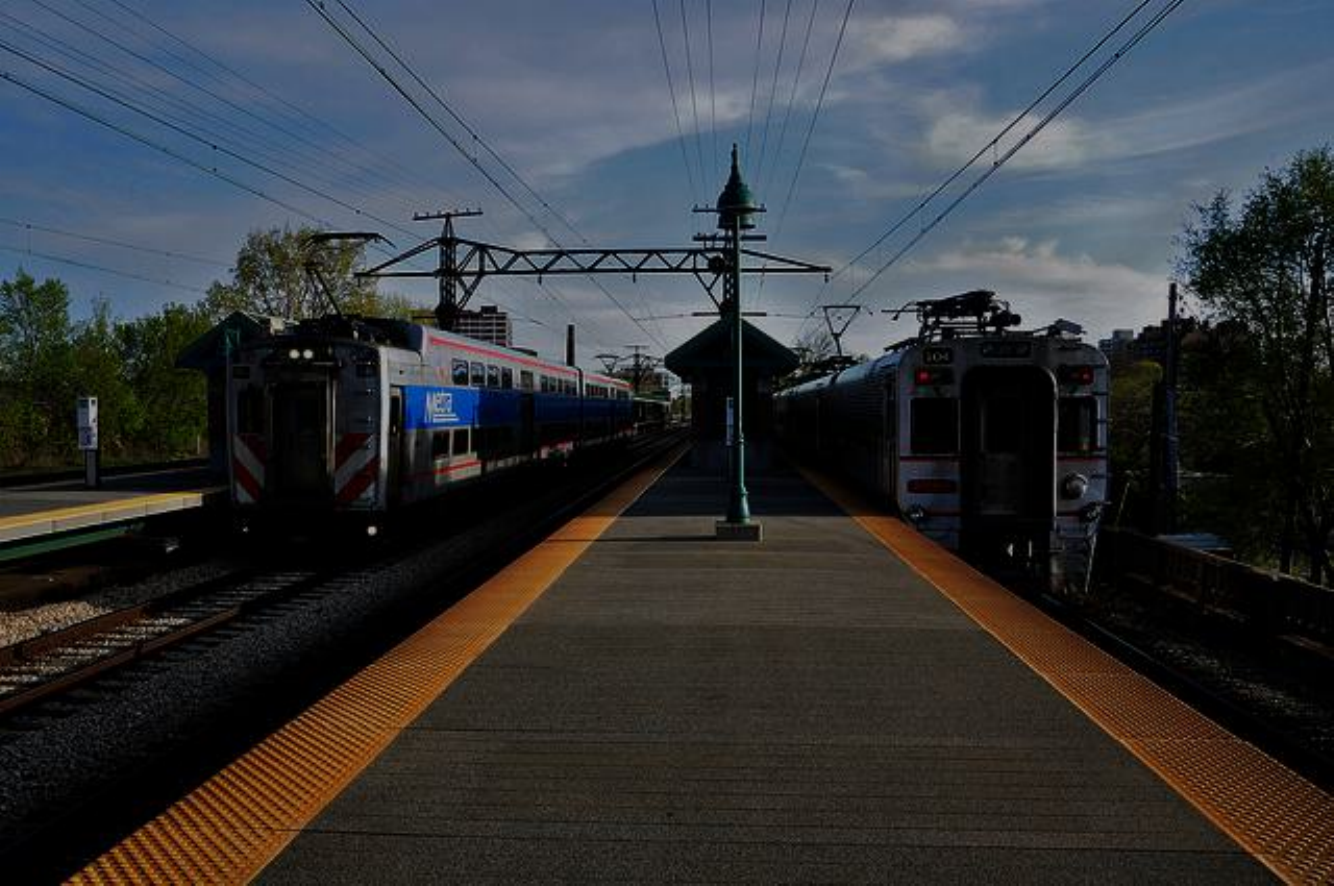} 
	\end{subfigure}
	\begin{subfigure}{0.32\linewidth}
		\centering
		\includegraphics[width=\linewidth]{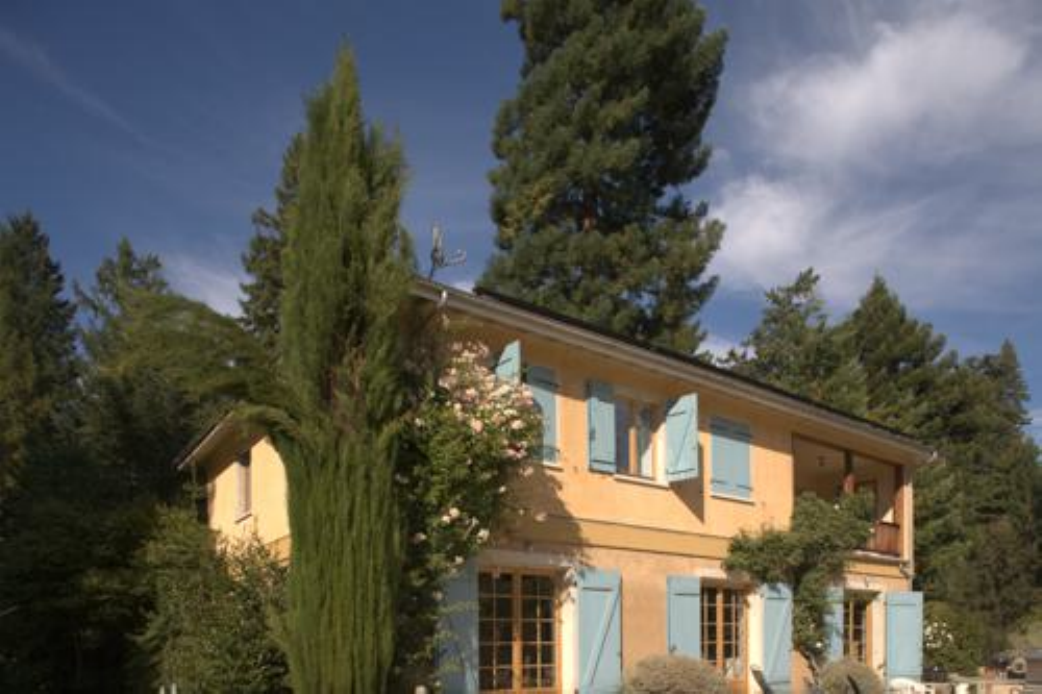}
	\end{subfigure}

	\begin{subfigure}{0.32\linewidth}
		\centering
		\includegraphics[width=\linewidth]{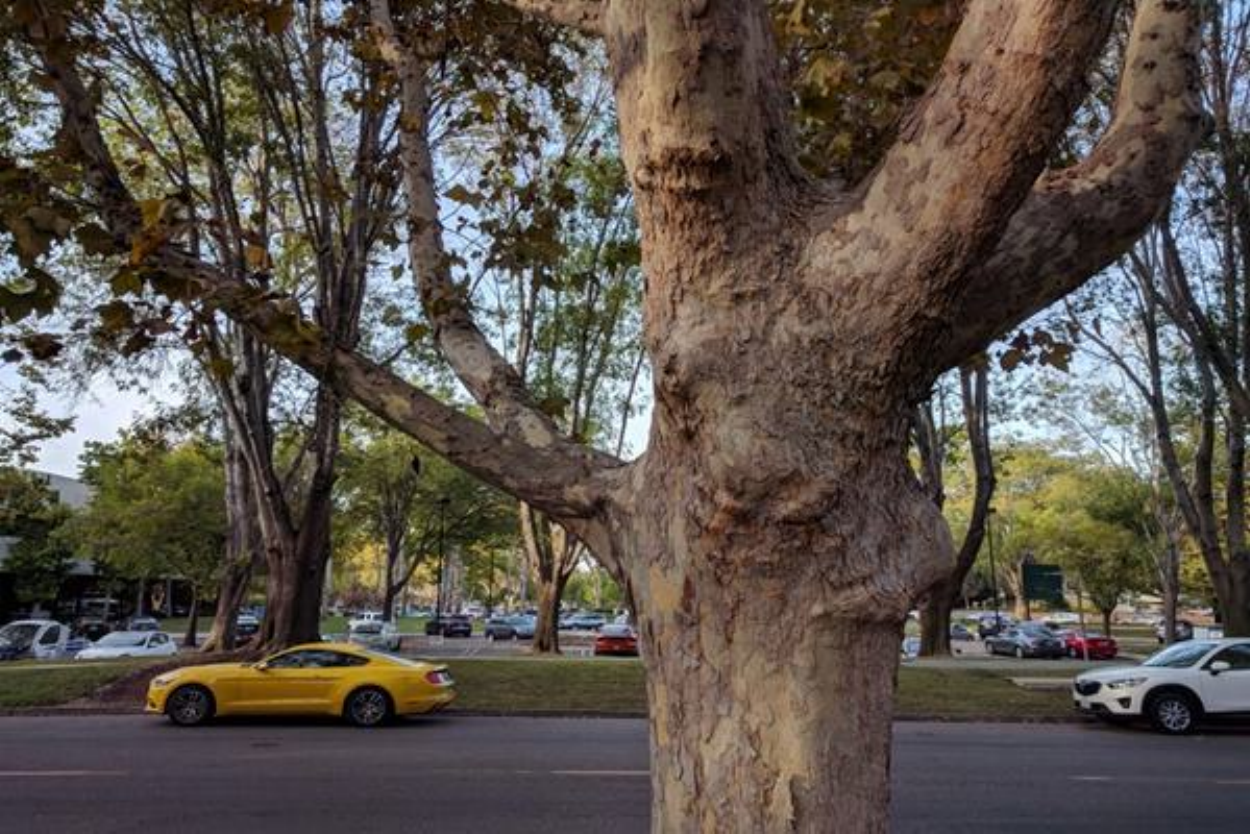} 
		\caption{HDR+}
		\label{fig:HDR}
	\end{subfigure}
	\begin{subfigure}{0.32\linewidth}
		\centering
		\includegraphics[width=\linewidth]{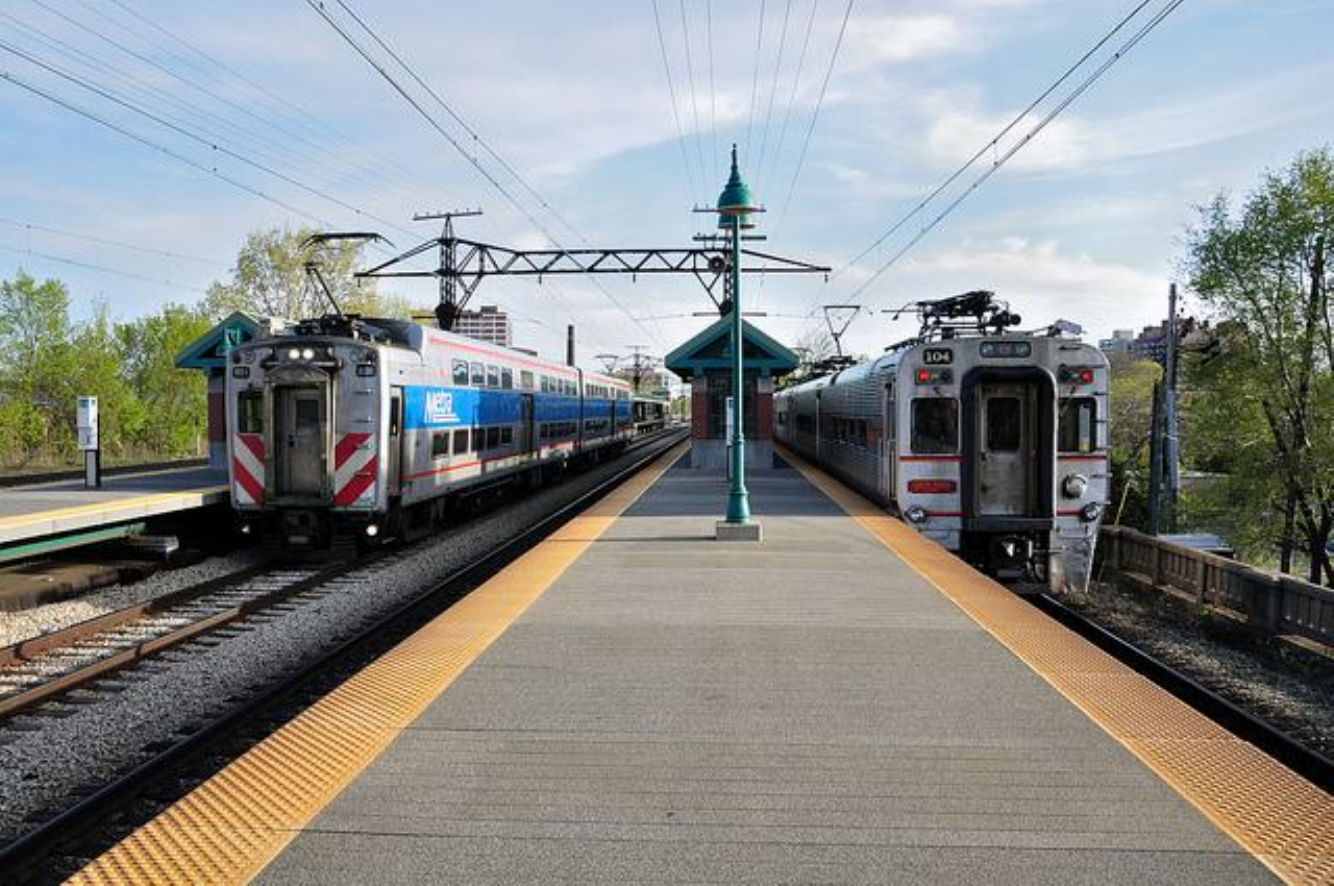} 
		\caption{SLL}
		\label{fig:SLL}
	\end{subfigure}
	\begin{subfigure}{0.32\linewidth}
		\centering
		\includegraphics[width=\linewidth]{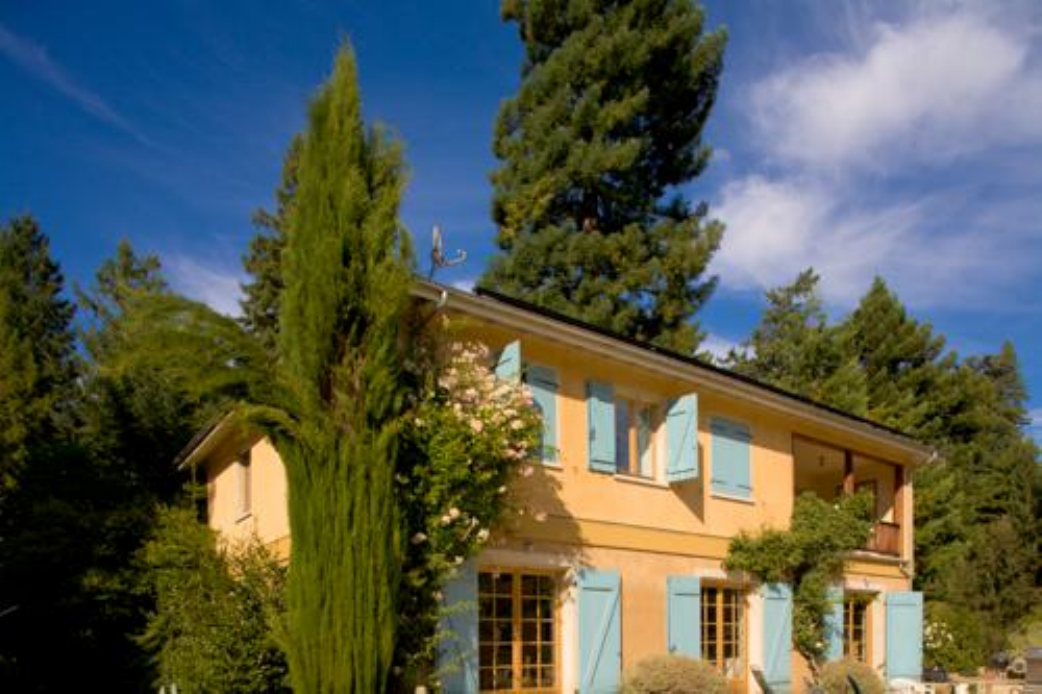}
		\caption{MIT}
		\label{fig:MIT}
	\end{subfigure}
	
	\caption{Example images from the training set for estimation.}
	\label{fig:trainingdatasets}
\end{figure}

Finally, the MIT-Adobe FiveK dataset \cite{bychkovsky2011learning} consists of 5,000 paired images, each enhanced by five different experts. This dataset balances global and local illumination improvements, achieving a more vivid and realistic color restoration. Figure \ref{fig:trainingdatasets} presents the three aforementioned datasets for training the lighting and color estimation sections. The first row displays the original images to be enhanced, while the second row showcases reference images representing the ideal versions sought through the enhancement process.

\vspace{-1cm}
\subsection{ALEN Design}
\begin{figure*}[ht]
	\centering
	\includegraphics[width=0.90\textwidth]{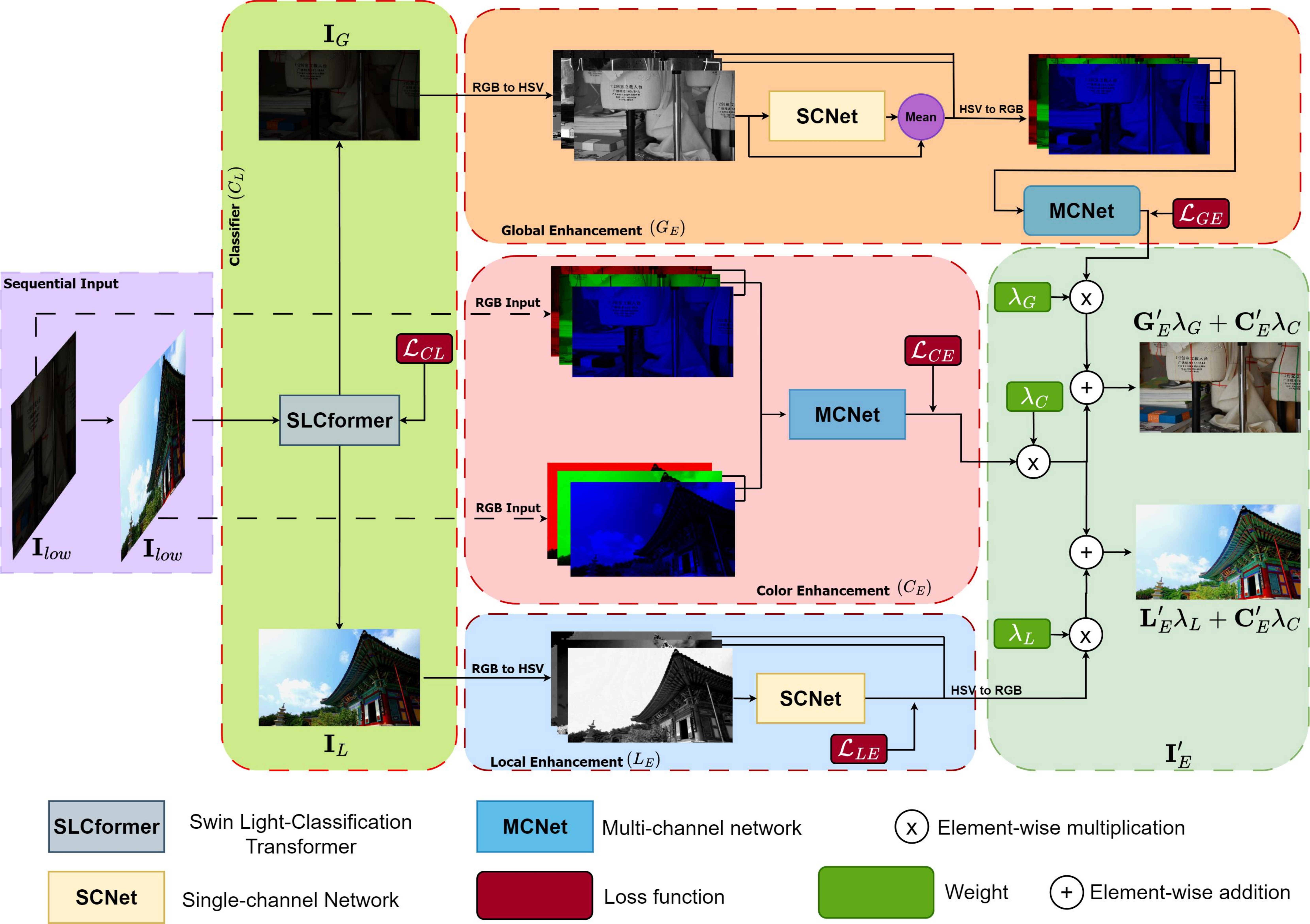}
	\caption{The architecture of the proposed method.}
	\label{fig:diagram}
\end{figure*}

Recent studies have employed various methods to enhance low-light images, including CNNs, GANs, Transformer-based architectures, and statistical methods. The performance of each technique varies depending on the image capture conditions and the training dataset, especially for neural network-based approaches. However, the adaptability of these algorithms does not always guarantee consistent improvement across different datasets. Therefore, the proposed method, ALEN, addresses this diversity by combining multiple neural network designs. Figure \ref{fig:diagram} illustrates the overall structure of the proposed ALEN architecture.

Based on the diagram shown in Fig. \ref{fig:diagram}, various sections that encompass different aspects of the comprehensive architecture aimed at enhancing low-light images can be identified. The main components of the diagram include the classifier ($C_L$), Global Enhancement ($G_E$), Local Enhancement ($L_E$), and Color Enhancement ($C_E$). 


Considering $H$ and $W$ as the height and width of the image in pixels, respectively, the $C_L$ component relies on the Swin Light-Classification Transformer (SLCformer), a lightweight classification sub-network that analyzes the images $\mathbf{I}_{low} \in \mathbb{R}^{H \times W \times 3}$ sequentially, without any prior preprocessing. The only operation applied to $\mathbf{I}_{low}$ is labeling, as outlined in Section \ref{GLI_D}. Based on the features extracted by SLCformer, the $C_L$ component dynamically selects between global enhancement, represented by $\mathbf{I}_{G} \in \mathbb{R}^{H \times W \times 3}$, and local enhancement, represented by $\mathbf{I}_{L} \in \mathbb{R}^{H \times W \times 3}$, according to the specific characteristics of the image.

For the estimation of $G_E$, $L_E$, and $C_E$, the Single-Channel Network (SCNet) and Multi-Channel Network (MCNet) sub-networks are used. SCNet employs a single channel of information for illumination enhancement, specifically the V channel in the HSV color space. On the other hand, MCNet allows for the independent extraction of information from each channel in the RGB color space. For $G_E$, a global input image $\mathbf{I}_{G}$ is considered. The process begins with converting from RGB to HSV and extracting the V channel from $\mathbf{I}_{G}$. Subsequently, this V channel extraction is processed by the SCNet sub-network. This procedure can be mathematically expressed in Eqs. \ref{v_channel_global} and \ref{v_process_SCNet_global}, respectively.

\begin{equation}
    \label{v_channel_global}
    \mathbf{V}_{G} = \mathcal{T}_{\text{HSV}}(\mathbf{I}_{G}[R], \mathbf{I}_{G}[G], \mathbf{I}_{G}[B])[V]
\end{equation}

\begin{equation}
    \label{v_process_SCNet_global}
    \mathbf{V}_{G}^{\prime} = \text{SCNet}(\mathbf{V}_{G})
\end{equation}

\noindent where $\mathbf{V}_{G} \in \mathbb{R}^{H \times W}$ represents the output of the channel extraction, $\mathcal{T}_{\text{HSV}}$ denotes the RGB to HSV transformation function, $[ \cdot ]$ indicates channel extraction, and $V$ is the specific channel extracted from this conversion. Following that, in Eq. \ref{v_process_SCNet_global},  $\mathbf{V}_{G}^{\prime} \in \mathbb{R}^{H \times W}$ represents the output of the channel processed by the SCNet sub-network. Continuing with the process,  $\mathbf{V}_{G}$ and $\mathbf{V}_{G}^{\prime}$ are averaged to preserve the characteristics of the V channel. Then, the conversion from HSV to RGB is performed, which is expressed as follows:

\begin{equation}
\mathbf{I}_{G}^{\prime} = \mathcal{T}_{\text{RGB}}\left(\mathbf{I}_{G}[H], \mathbf{I}_{G}[S], \frac{\mathbf{V}_{G} + \mathbf{V}_{G}^{\prime}}{2}\right)
\end{equation}

\noindent where $\mathbf{I}_{G}^{\prime} \in \mathbb{R}^{H \times W \times 3}$ stands for the output of the conversion using the averages of $\mathbf{V}_{G}$ + $\mathbf{V}_{G}^{\prime}$ and $\mathcal{T}_{\text{RGB}}$ is the transformation function from HSV to RGB. Finally, to obtain the global illumination enhancement in each of its RGB channels, it is passed through MCNet, which allows independent enhancement of information in each channel of the RGB color space. This can be expressed mathematically as:

\begin{equation}
    \mathbf{G}_{E}^{\prime} = \text{MCNet}(\mathbf{I}_{G}^{\prime}[R], \mathbf{I}_{G}^{\prime}[G], \mathbf{I}_{G}^{\prime}[B])  
\end{equation}

\noindent where $\mathbf{G}_{E}^{\prime} \in \mathbb{R}^{H \times W \times 3}$ denotes the output obtained after passing $\mathbf{I}_{G}^{\prime}$ through MCNet. 

In the case of $L_E$, the input image is $\mathbf{I}_{L}$. For $\mathbf{I}_{L}$ to be processed by the SCNet sub-network, it must first be transformed from RGB to HSV. This transformation allows for the extraction and specific processing of the V channel. 

The complete process, from the transformation to the processing by SCNet, is detailed in Eqs. \ref{v_channel_local} and \ref{v_process_SCNet_local}.

\begin{equation}
    \label{v_channel_local}
    \mathbf{V}_{L} = \mathcal{T}_{\text{HSV}}(\mathbf{I}_{L}[R], \mathbf{I}_{L}[G], \mathbf{I}_{L}[B])[V]
\end{equation}

\begin{equation}
    \label{v_process_SCNet_local}
    \mathbf{V}_{L}^{\prime} = \text{SCNet}(\mathbf{V}_{L})
\end{equation}

\noindent where $\mathbf{V}_{L} \in \mathbb{R}^{H \times W}$  represents the extraction of the V channel from the HSV color space, and $\mathcal{T}_{\text{HSV}}$ is the transformation function from RGB to HSV. On the other hand, $\mathbf{V}_{L}^{\prime} \in \mathbb{R}^{H \times W}$ is the output obtained after passing $\mathbf{V}_{L}$ through the SCNet subnetwork. After this processing, $\mathbf{V}_{L}^{\prime}$  is converted back to the RGB color space to obtain the final output of the local illumination enhancement. 

This process is expressed as follows:

\begin{equation}
\mathbf{L}_{E}^{\prime} = \mathcal{T}_{\text{RGB}}(\mathbf{I}_{L}[H], \mathbf{I}_{L}[S], \mathbf{V}_{L}^{\prime})
\end{equation}

\noindent where $\mathbf{L}_{E}^{\prime} \in \mathbb{R}^{H \times W \times 3}$ represents the output of the local illumination enhancement and $\mathcal{T}_{\text{RGB}}$ is the transformation from HSV to RGB. For $C_E$, the images $\mathbf{I}_{low}$ are considered as input. In this process, $\mathbf{I}_{low}$ is fed directly into the MCNet sub-network without passing through the classifier, allowing for color enhancement based on the primary characteristics of the input images.

This can be expressed mathematically as follows:

\begin{equation}
\mathbf{C}_{E}^{\prime} = \text{MCNet}(\mathbf{I}_{low}[R], \mathbf{I}_{low}[G], \mathbf{I}_{low}[B])  
\end{equation}

\noindent where $\mathbf{C}_{E}^{\prime} \in \mathbb{R}^{H \times W \times 3}$ is the resulting image with color enhancement.

Finally, the combination of the local and global enhancement processes with the color enhancement of the image can be expressed as follows:

\begin{equation}
    \label{eq:EGH}
	\mathbf{I}'_{E} = 
	\begin{cases}
		 \mathbf{G}_{E}^{\prime}\lambda_{G}+\mathbf{C}_{E}^{\prime}\lambda_{C} & \text{Global enhancement}\\
		 \mathbf{L}_{E}^{\prime}\lambda_{L}+\mathbf{C}_{E}^{\prime}\lambda_{C} & \text{Local enhancement}
	\end{cases}
\end{equation}

\noindent where $\mathbf{I}_{E}^{\prime}\in \mathbb{R}^{H \times W \times 3}$ represents the enhanced image, whether locally or globally, $\mathbf{G}_{E}^{\prime} $ denotes the global illumination enhancement, $\mathbf{L}_{E}^{\prime}$ is the local illumination enhancement, and $\mathbf{C}_{E}^{\prime}$ stand for color enhancement. On the other hand $\lambda_{G}$, $\lambda_{L}$, and $\lambda_{C}$ are corresponding weights used to achieve the optimal fusion between the images.

\begin{figure*}[ht]
	\centering
	\includegraphics[width=1\textwidth]{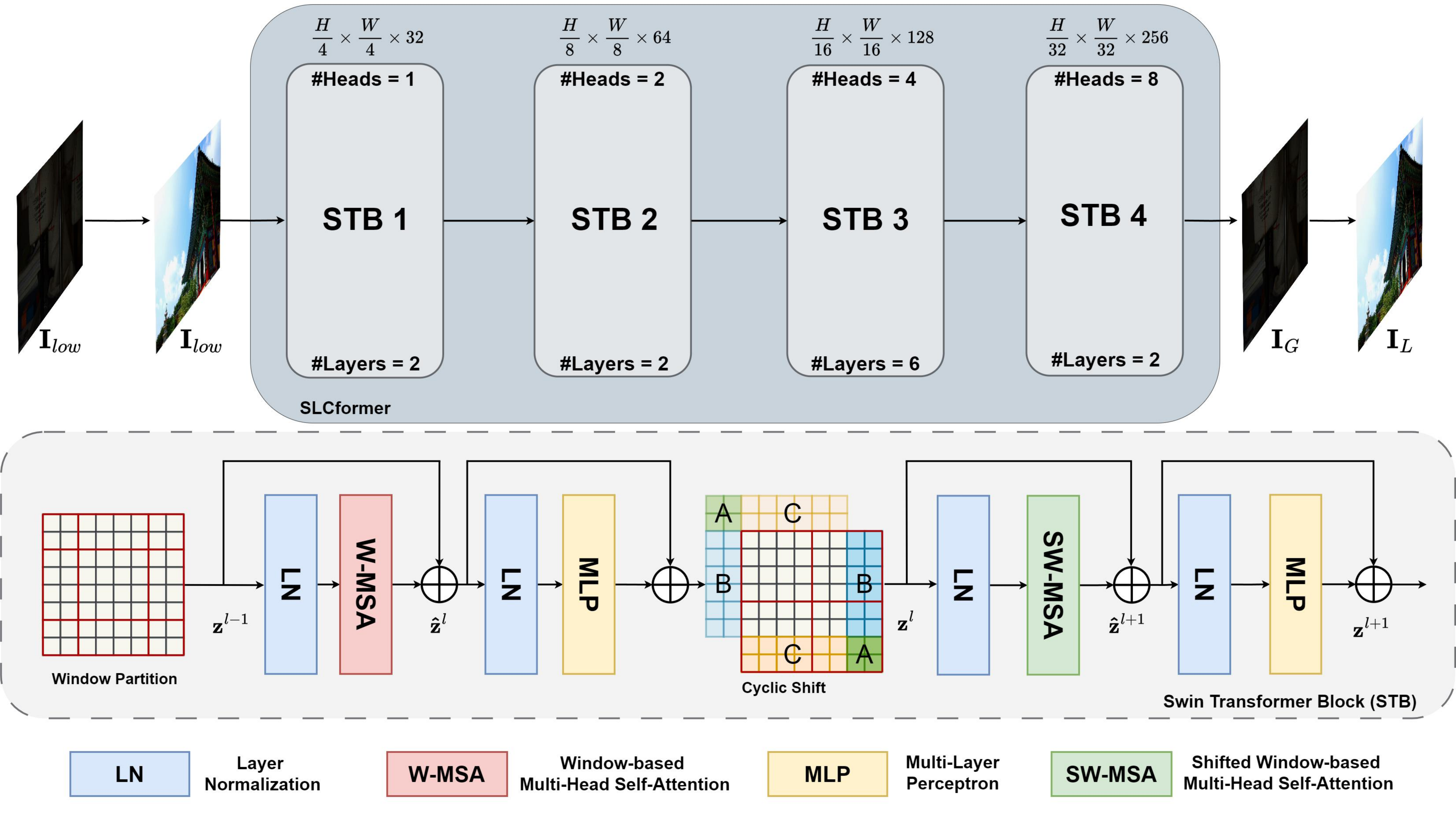}
	\caption{Classification architecture.}
	\label{SLCformer}
\end{figure*}

To provide a more precise description of the ALEN architecture, the SLCformer, SCNet, and MCNet sub-networks for light classification and estimation, as well as image color, are detailed first, covering Sections \ref{lighclas} and \ref{lightest}. These sections delve into the key elements of the required convolutional layers, including the core concepts of executing the proposed model. Subsequently, Section \ref{modeldes} thoroughly describes the loss functions used to achieve the desired model outcomes, which are employed in the models developed in Sections \ref{lighclas} and \ref{lightest}. This step-by-step analysis provides a deeper understanding of the neural network architectures required for light classification and estimation and image color training.

\subsubsection{Swin Transformer-based classification architecture}
\label{lighclas}

The SLCformer, based on the Swin Transformer\cite{liu2021swin}, extracts image features and uses a binary classifier to determine whether global or local enhancement is required. The image $\mathbf{I}_{low} \in \mathbb{R}^{H \times W \times 3}$ is first passed through multiple Swin Transformer Blocks (STB), as shown in the design of Figure \ref{SLCformer}.

SLCformer utilizes four STBs, where the first, second, and fourth blocks contain two layers each, while the third block consists of six layers. This design follows the Swin-T model architecture, balancing computational efficiency with enhanced feature extraction in the third block. Unlike other variants such as Swin-S, Swin-B, and Swin-L, where the third block typically includes 18 layers, this configuration improves the model’s ability to capture fine-grained details crucial for accurately classifying whether global or local enhancement is required.

\begin{figure}[ht]
	\centering
	\includegraphics[width=0.95\columnwidth]{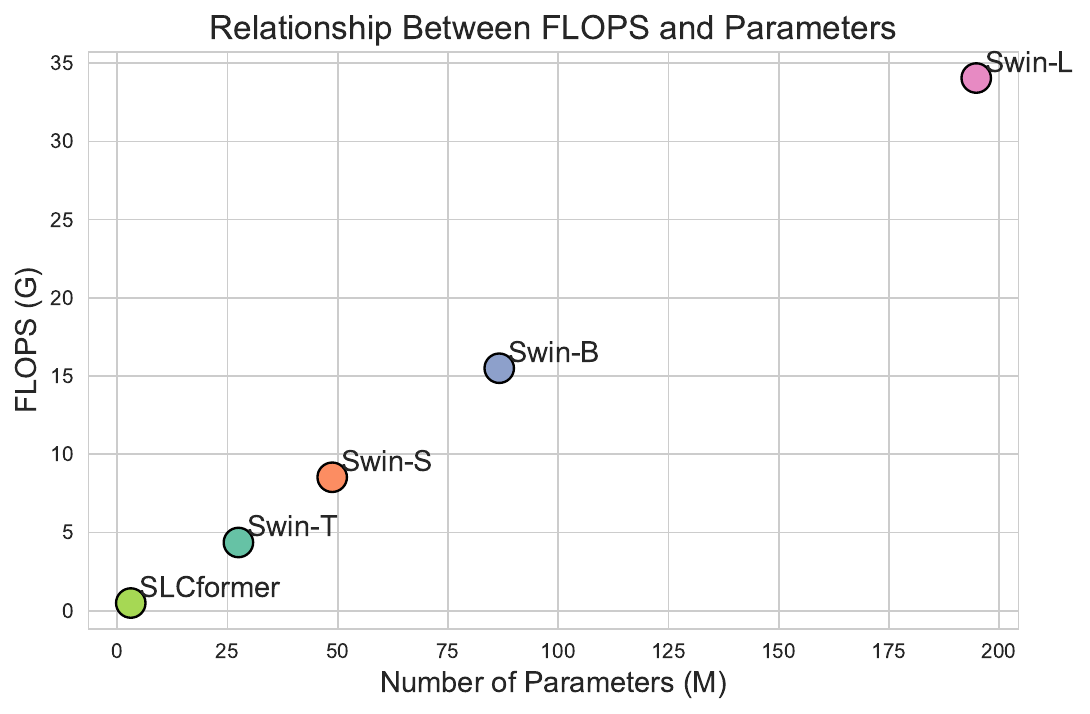}
	\caption{Computational complexity comparison between the different variants of the Swin Transformer and the proposed model.}
	\label{SLCformer_comparison}
\end{figure}

\begin{table*}[htbp]
    \centering
    \caption{Comparison of the number of parameters and FLOPS for different methods.}
    \label{C3Tab1}
    \begin{tabular}{lccc|cc} 
        \toprule
        Method & \#Channels & \#Heads & \#Layers & \#Parameters & FLOPs  \\ 
        \midrule
        Swin-T & \{96, 192, 384, 768\} & \{3, 6, 12, 24\} & \{2, 2, 6, 2\} & 27.47M & 4.36G\\
        Swin-S & \{96, 192, 384, 768\} & \{3, 6, 12, 24\} & \{2, 2, 18, 2\} & 48.74M & 8.53G\\
        Swin-B & \{128, 256, 512, 1024\} & \{4, 8, 16, 32\} & \{2, 2, 18, 2\} & 86.62M & 15.15G\\
        Swin-L & \{192, 384, 768, 1,536\} & \{6, 12, 24, 48\} & \{2, 2, 18, 2\} & 194.80M & 34.05G \\
        SLCformer & \{32, 64, 128, 256\} & \{1, 2, 4, 8\} & \{2, 2, 6, 2\} & 3.06M & 0.49G \\
        \bottomrule
    \end{tabular}
\end{table*}

To further reduce computational cost while maintaining effectiveness, SLCformer significantly decreases the number of channels in each stage from Swin-T’s \{96, 192, 384, 768\} to \{32, 64, 128, 256\}, and correspondingly reduces the number of attention heads from \{3, 6, 12, 24\} to \{1, 2, 4, 8\}. The layer configuration remains similar to Swin-T, ensuring a balance between efficiency and representational capacity.

Table \ref{C3Tab1} compares the parameters and FLOPS of different Swin Transformer variants and SLCformer, showing that SLCformer greatly reduces computational complexity while maintaining effective feature extraction. With only 3.06M parameters and 0.49G FLOPS, SLCformer is significantly lighter than Swin-T (27.47M parameters, 4.36G FLOPS), making it a suitable choice for resource-limited tasks. Figure \ref{SLCformer_comparison} further illustrates these differences, highlighting SLCformer’s efficiency in balancing performance and resource utilization.

The operations within each block are defined by Equations \eqref{STB}, where Self-Attention and Multi-Layer Perceptron (MLP) modules alternate to refine the extracted features.

\begin{subequations} \label{STB}
\begin{align}
    \mathbf{\hat{z}}^l &= \text{W-MSA}(\text{LN}(\mathbf{z}^{l-1})) + \mathbf{z}^{l-1}, \label{STB1} \\
    \mathbf{z}^l &= \text{MLP}(\text{LN}(\mathbf{\hat{z}}^l)) + \mathbf{\hat{z}}^l, \label{STB2} \\
    \mathbf{\hat{z}}^{l+1} &= \text{SW-MSA}(\text{LN}(\mathbf{z}^l)) + \mathbf{z}^l, \label{STB3} \\
    \mathbf{z}^{l+1} &= \text{MLP}(\text{LN}(\mathbf{\hat{z}}^{l+1})) + \mathbf{\hat{z}}^{l+1}. \label{STB4}
\end{align}
\end{subequations}

\noindent In the first stage, represented by Equations \eqref{STB1} and \eqref{STB2}, the normalized output from the previous layer ($\mathbf{z}^{l-1}$) undergoes Window-based Multi-head Self-Attention (W-MSA). The output of this operation is combined with the residual connection from $\mathbf{z}^{l-1}$, resulting in an intermediate representation, $\mathbf{\hat{z}}^l$. Next, this intermediate output is then subjected to Layer Normalization (LN) before being processed by the Multi-Layer Perceptron (MLP). Finally, a residual connection from $\mathbf{\hat{z}}^l$ is added, producing the final output, $\mathbf{z}^l$.

In the subsequent part, corresponding to Equations \eqref{STB3} and \eqref{STB4}, Shifted Window-based Multi-head Self-Attention (SW-MSA) is applied to the normalized representation of $\mathbf{z}^l$. The residual connection is then added to the result, allowing for the capture of broader contextual features and producing the output $\mathbf{\hat{z}}^{l+1}$. This output, $\mathbf{\hat{z}}^{l+1}$, passes through an LN and an MLP, producing the output $\mathbf{z}^{l+1}$. A residual connection from $\mathbf{\hat{z}}^{l+1}$ is also incorporated, completing the feature refinement process.

On the other hand, the attention mechanism for W-MSA and SW-MSA is defined by Equation \ref{SA}:

\begin{figure*}[ht]
	\centering
	\includegraphics[width=0.95\textwidth]{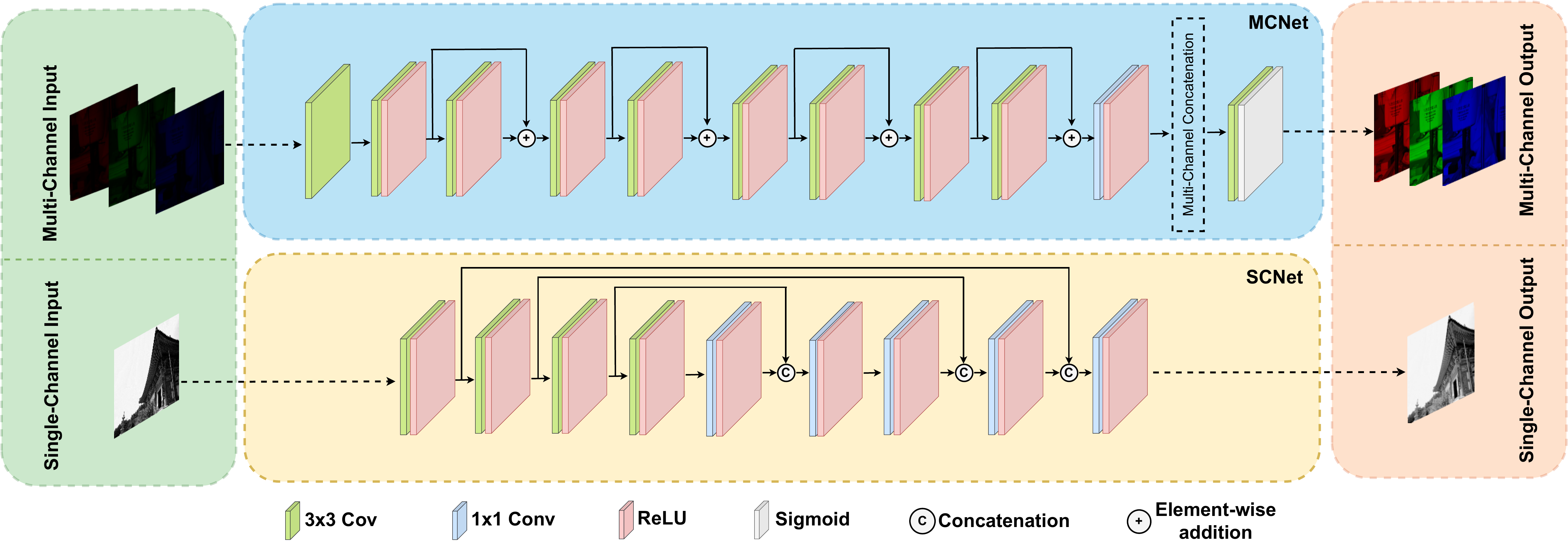}
	\caption{Estimation architectures.}
	\label{ESTINETS}
\end{figure*}

\begin{equation}
    \label{SA}
    \text{Attention}(\mathbf{Q}, \mathbf{K}, \mathbf{V}) = \text{SoftMax} \left( \frac{\mathbf{Q} \mathbf{K}^T}{\sqrt{d}} + \mathbf{B} \right) \mathbf{V}
\end{equation}

\noindent In Equation \ref{SA}, $ \mathbf{Q} $, $ \mathbf{K} $, and $ \mathbf{V} \in \mathbb{R}^{M^2 \times d}$ represent the query, key, and value matrices, respectively. The product $ \mathbf{Q} \mathbf{K}^T $ measures the similarity between the queries and keys. The variable $ d $ is the dimension of the queries and keys, used to normalize the result and prevent excessively large values. $ M^2 $ represents the number of patches per window, referring to the subregions analyzed within each window. Lastly, $ \mathbf{B} \in \mathbb{R}^{M^2 \times M^2} $ is the relative position bias, which adds information about the relative positions of elements within the attention windows.



\vspace{-0.5cm}
\subsubsection{Estimation architectures}
\label{lightest}
SCNet begins with three consecutive convolutional layers, each followed by ReLU activation, doubling the output channels at each layer up to 128. A fourth convolutional layer with ReLU activation increases the channels to 256. The fifth layer reduces the output to 128 channels, and its output is concatenated with the third layer's output to form a skip connection, preserving previous features. The sixth layer reduces the channels to 64 and is concatenated with the second layer's output, creating another skip connection. The seventh layer reduces the channels to 32, and its output is concatenated with the first layer's output, forming a third skip connection. Finally, an eighth layer reduces the output dimensions to the original input size. SCNet does not include pooling or fully connected layers; instead, it focuses on convolution and feature combination through skip connections to retain information from earlier layers. Table \ref{C3Tab2} details each convolutional layer and the convolutional kernels used.

\begin{table}[ht]
	\centering
	\caption{Characteristics of SCNet.}
	\label{C3Tab2}
	\begin{tabular*}{\columnwidth}{@{\extracolsep{-4pt}} lccccc@{}}
		\toprule
		Layer & Input & Output & Filter Kernel & Stride & Padding \\ 
        \midrule
		\multicolumn{6}{c}{Feature Extraction}\\
		Conv 1 & 1 & 32 & 3x3 & 1 & 1\\
		Conv 2 & 32 & 64 & 3x3 & 1 & 1\\
		Conv 3 & 64 & 128 & 3x3 & 1 & 1\\
		Conv 4 & 128 & 256 & 3x3 & 1 & 1\\
		\multicolumn{6}{c}{Filter Reduction}\\
		Conv 5 & 256 & 128 & 1x1 & 1 & 0\\
		Conv 6 & 256 & 64 & 1x1 & 1 & 0\\
		Conv 7 & 128 & 32 & 1x1 & 1 & 0\\
		Conv 8 & 64 & 1 & 1x1 & 1 & 0\\ 
        \bottomrule
	\end{tabular*}
\end{table}

MCNet begins with feature extraction and filter reduction. An initial 3x3 convolutional layer extracts basic features with a stride and padding of 1. Several subsequent 3x3 convolutional layers (Conv 1-6) deepen the features and reduce the number of filters. The final layer (Final) uses 1x1 filters to reduce the output to a single channel, compressing the information. MCNet also includes a channel adjustment phase, where Conv 1 and Conv 2 use 1x1 filters for channel dimension transformation. A fusion convolutional layer (Conv Fusion) with 3x3 filters merges multi-channel features. Most layers use ReLU activation, while the fusion layer uses the Sigmoid function to normalize the output to a range of 0 to 1. Table \ref{C3Tab3} details each convolutional layer and the convolutional kernels used, while Fig. \ref{ESTINETS} illustrates the structure of SCNet and MCNet, highlighting the distinctive features and differences in the design of each model.

\begin{table}[ht]
	\centering
	\caption{Characteristics of MCNet.}
	\label{C3Tab3}
	\begin{tabular*}{\columnwidth}{@{\extracolsep{-8pt}} lccccc@{}}
		\toprule
		Layer & Input & Output & Filter Kernel & Stride & Padding \\ 
        \midrule
		\multicolumn{6}{c}{Channel Adjustment}\\
		Conv 1  & 32   & 64  &  1x1 & 1 & 0\\
		Conv 2  & 64   & 128  &  1x1 & 1 & 0\\ 
		\multicolumn{6}{c}{Feature Extraction/Filter Reduction}\\
		Initial   & 1   & 32  &  3x3 & 1 & 1\\
		Conv 1 & 32  & 32  &  3x3 & 1 & 1\\
		Conv 2 & 32  & 64  &  3x3 & 1 & 1\\
		Conv 3 & 64  & 64  &  3x3 & 1 & 1\\
		Conv 4 & 64  & 128 &   3x3 & 1 & 1\\
		Conv 5 & 128 & 64  &   3x3 & 1 & 1\\ 
		Conv 6 & 32  & 32  &  3x3 & 1 & 1\\
		Final     & 32  & 1   &   1x1 & 1 & 0\\  
		Conv Fusion & 3 & 3 & 3x3 & 1 & 1\\
        \bottomrule
	\end{tabular*}
\end{table}

\begin{figure*}[ht]
	\centering
	\begin{subfigure}{0.16\linewidth}
		\centering
		\includegraphics[width=\linewidth]{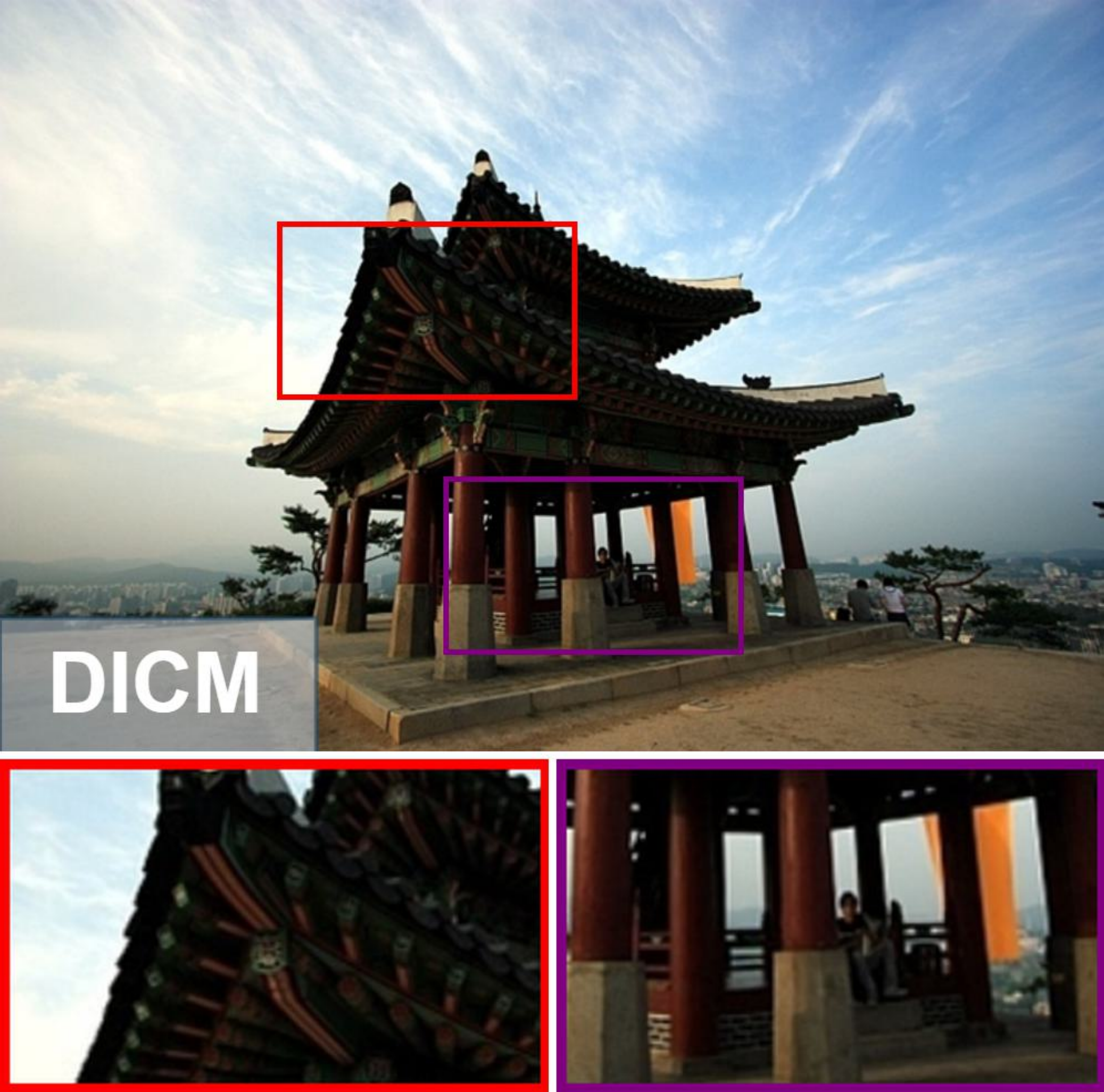} 
        \caption{\footnotesize Low-light} 
	\end{subfigure}
	\begin{subfigure}{0.16\linewidth}
		\centering
		\includegraphics[width=\linewidth]{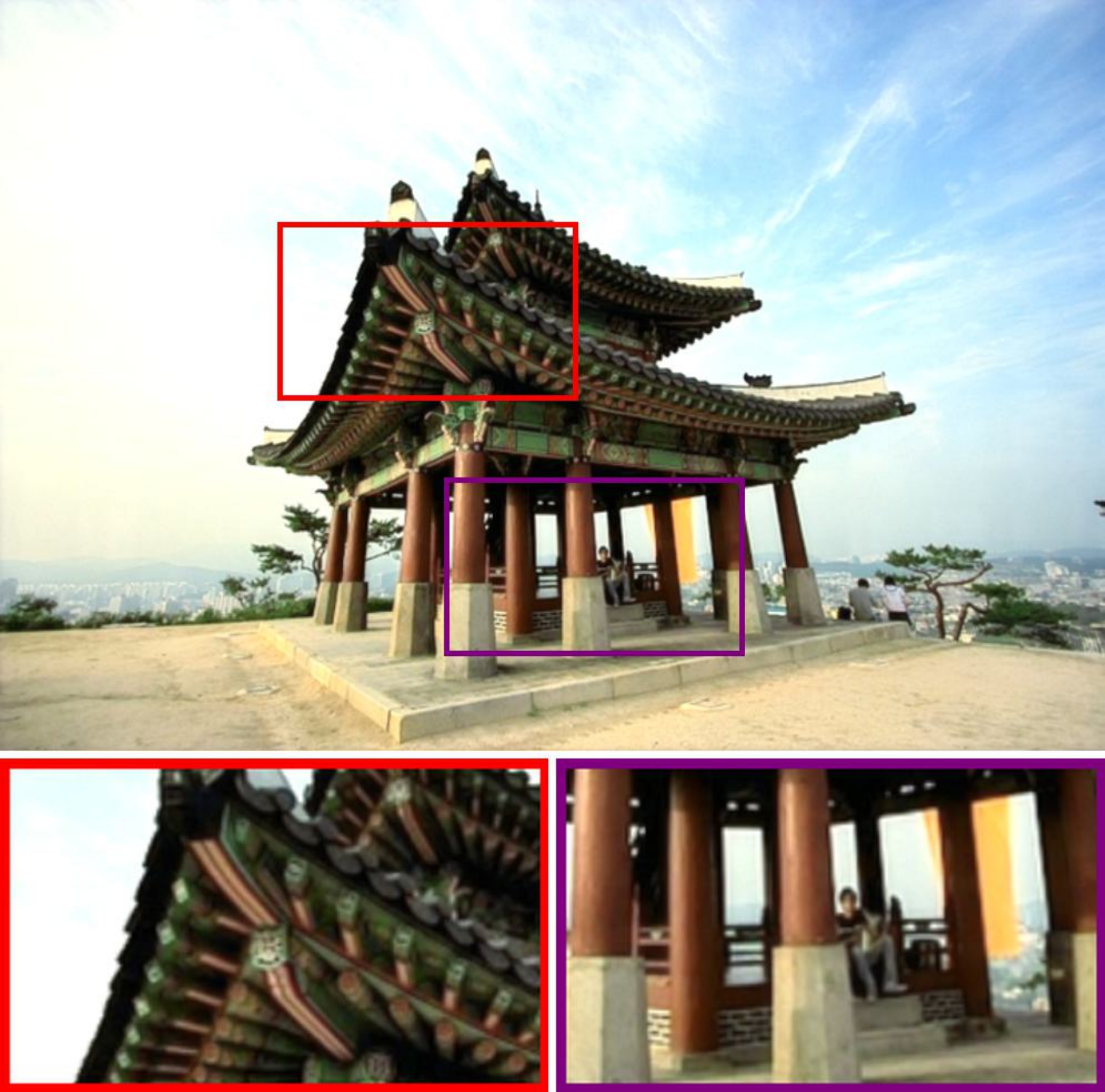} 
        \caption{\footnotesize PPformer}
	\end{subfigure}
	\begin{subfigure}{0.16\linewidth}
		\centering
		\includegraphics[width=\linewidth]{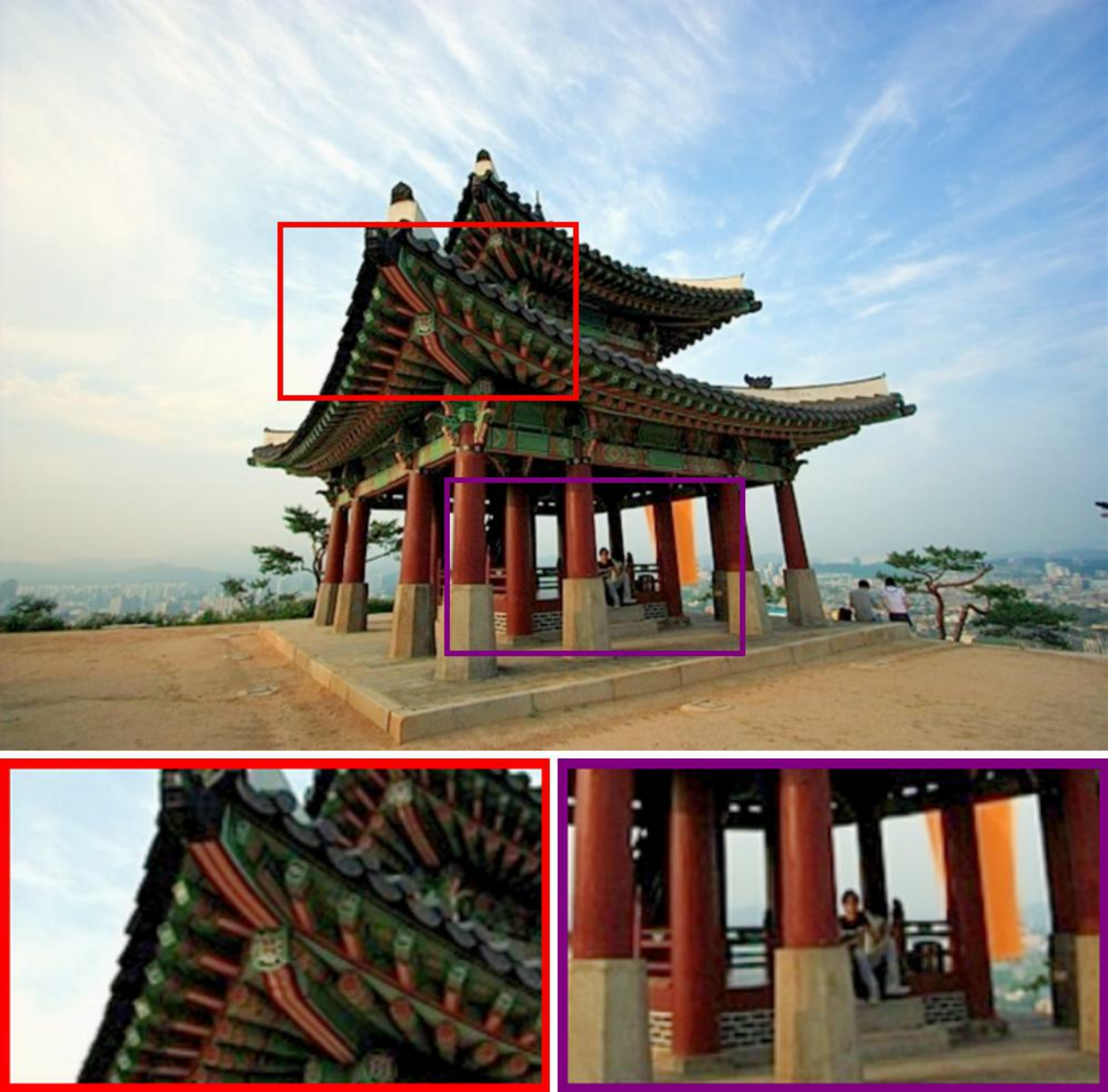}
        \caption{\footnotesize Proposed}
	\end{subfigure}
	\begin{subfigure}{0.16\linewidth}
		\centering
		\includegraphics[width=\linewidth]{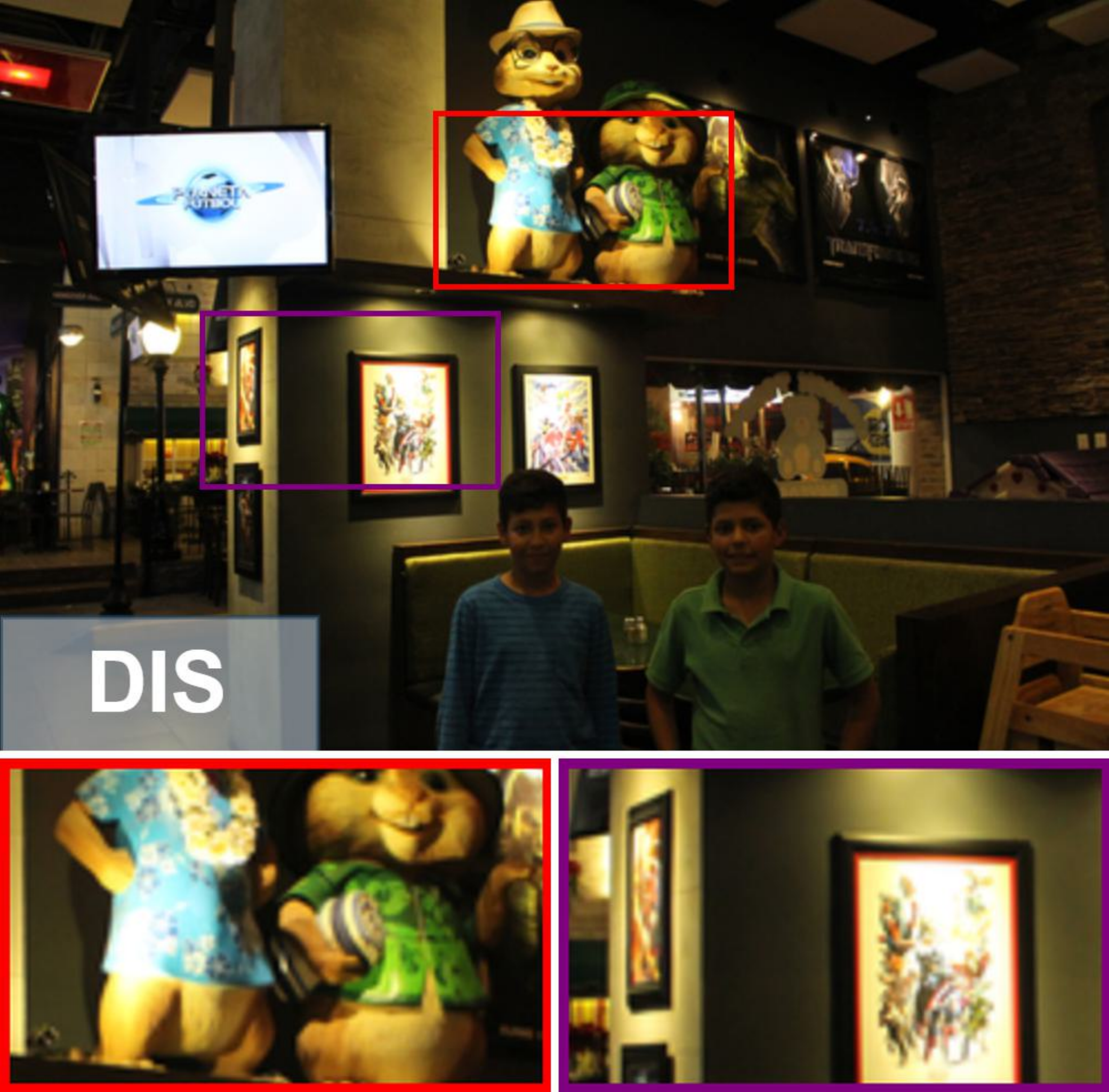} 
		\caption{\footnotesize Low-light}
	\end{subfigure}
	\begin{subfigure}{0.16\linewidth}
		\centering
		\includegraphics[width=\linewidth]{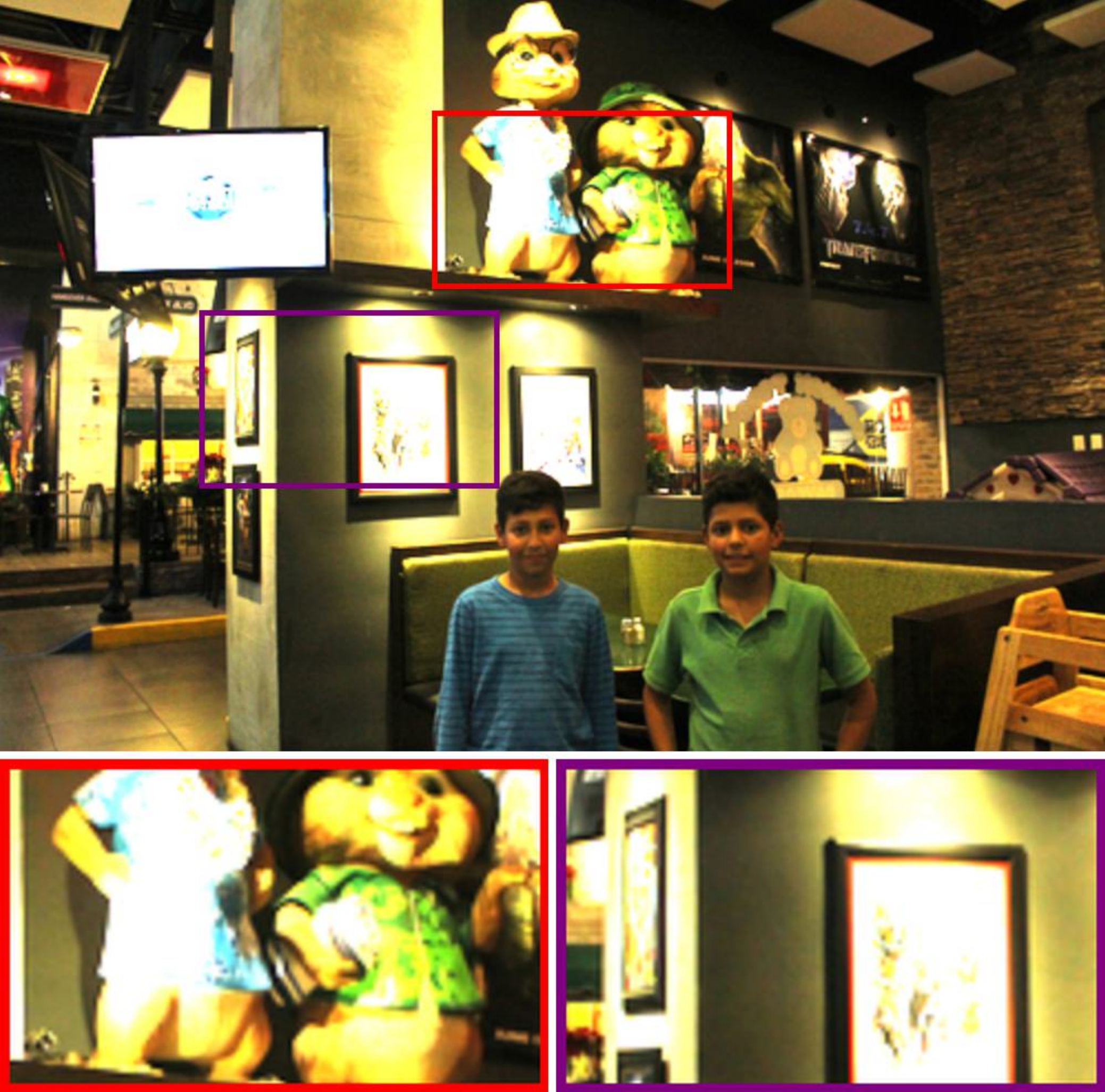} 
		\caption{\footnotesize BL}
	\end{subfigure}
	\begin{subfigure}{0.16\linewidth}
		\centering
		\includegraphics[width=\linewidth]{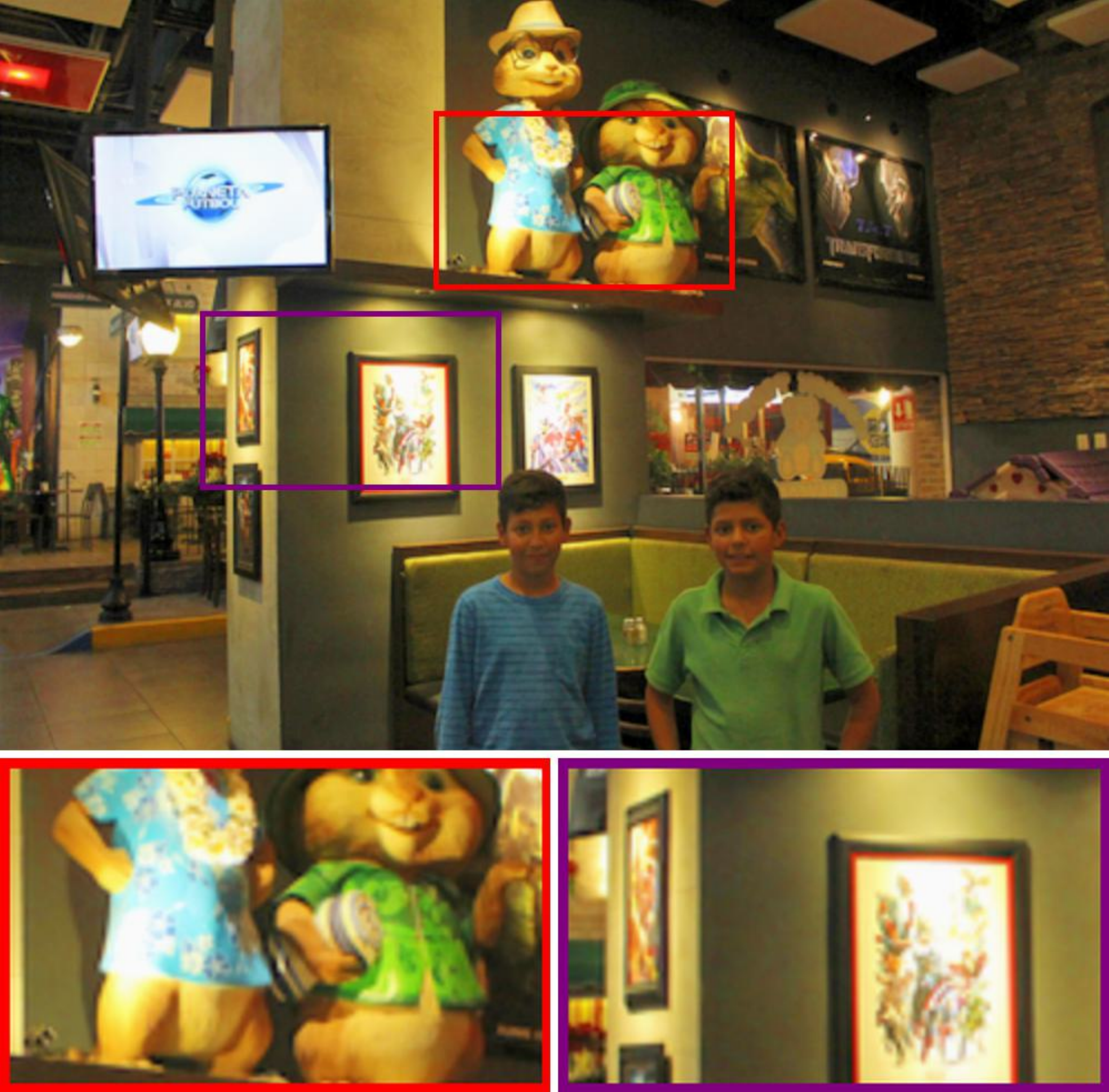}
		\caption{\footnotesize Proposed}
	\end{subfigure}

    \begin{subfigure}{0.16\linewidth}
		\centering
		\includegraphics[width=\linewidth]{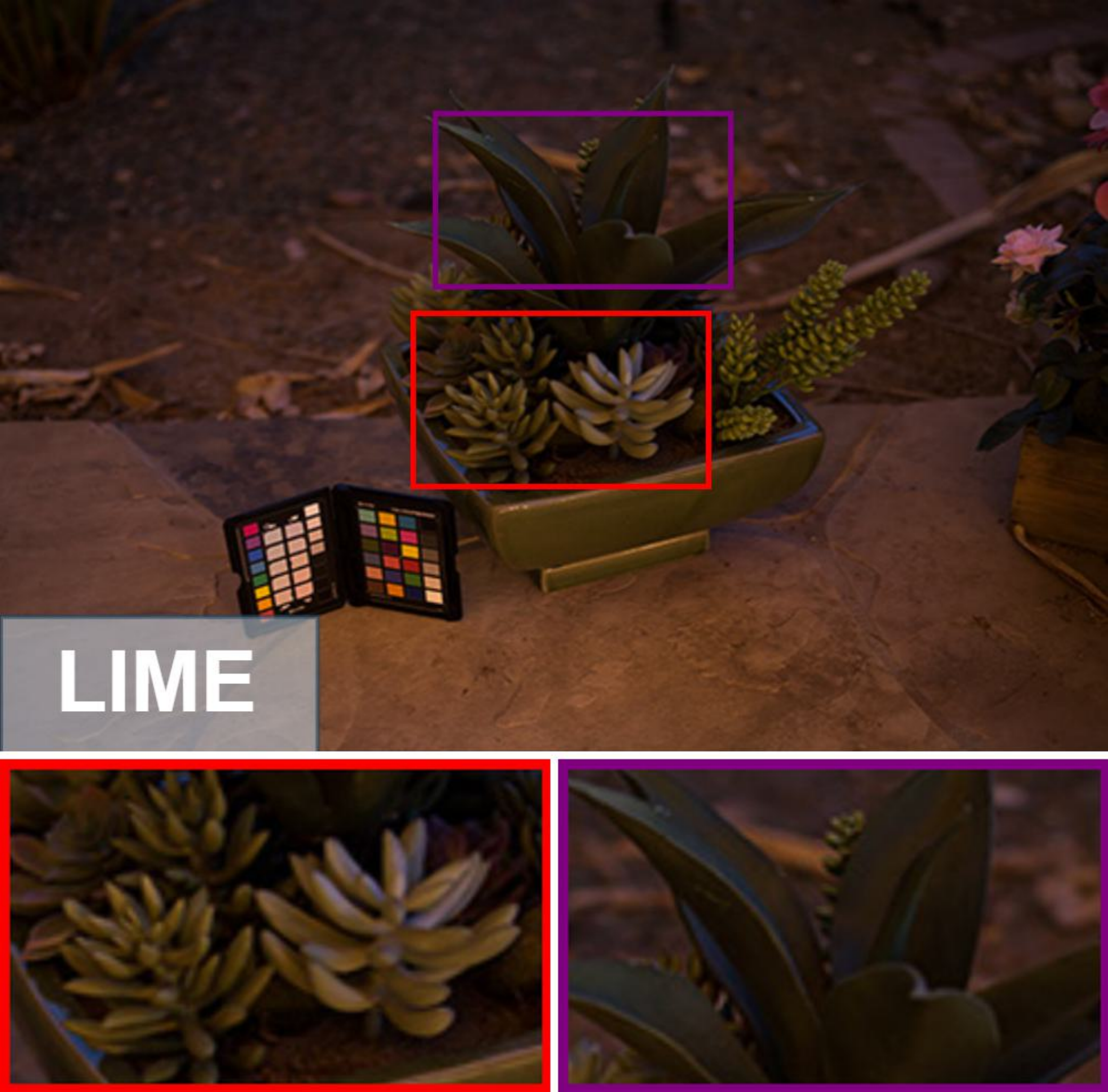} 
		\caption{\footnotesize Low-light}
	\end{subfigure}
	\begin{subfigure}{0.16\linewidth}
		\centering
		\includegraphics[width=\linewidth]{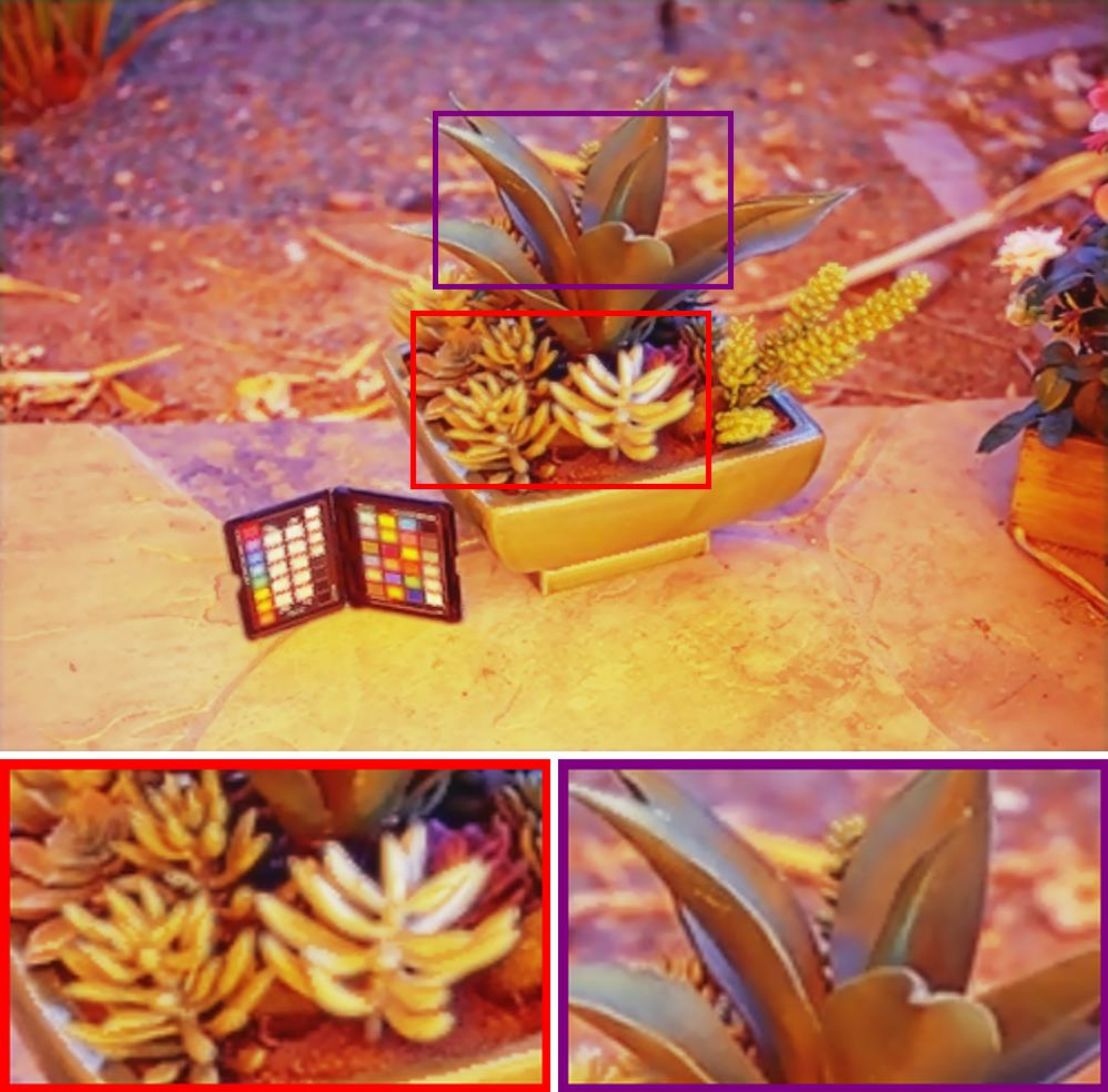} 
		\caption{\footnotesize SHAL-Net}
	\end{subfigure}
	\begin{subfigure}{0.16\linewidth}
		\centering
		\includegraphics[width=\linewidth]{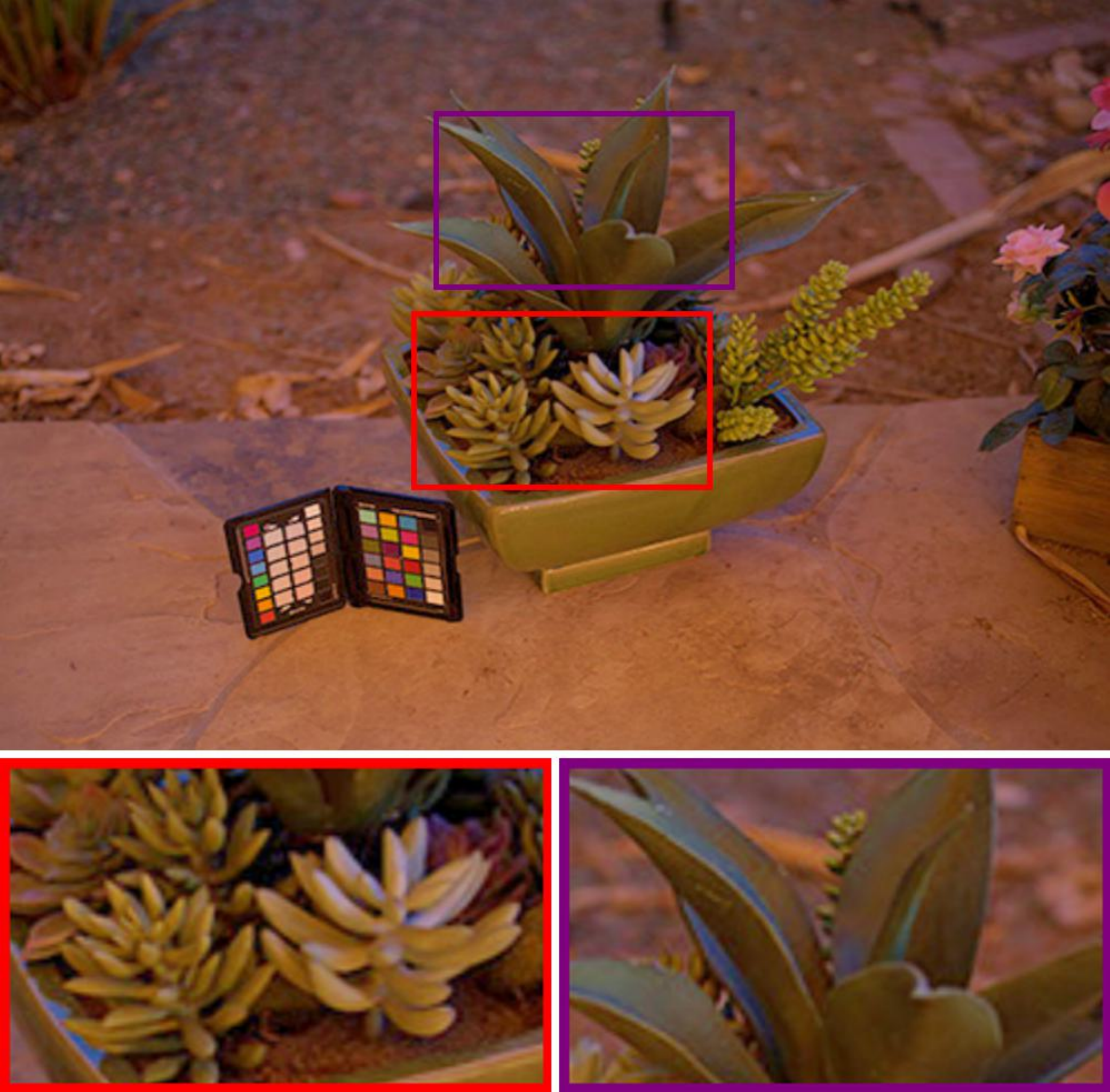}
		\caption{\footnotesize Proposed}
	\end{subfigure}
     \begin{subfigure}{0.16\linewidth}
		\centering
		\includegraphics[width=\linewidth]{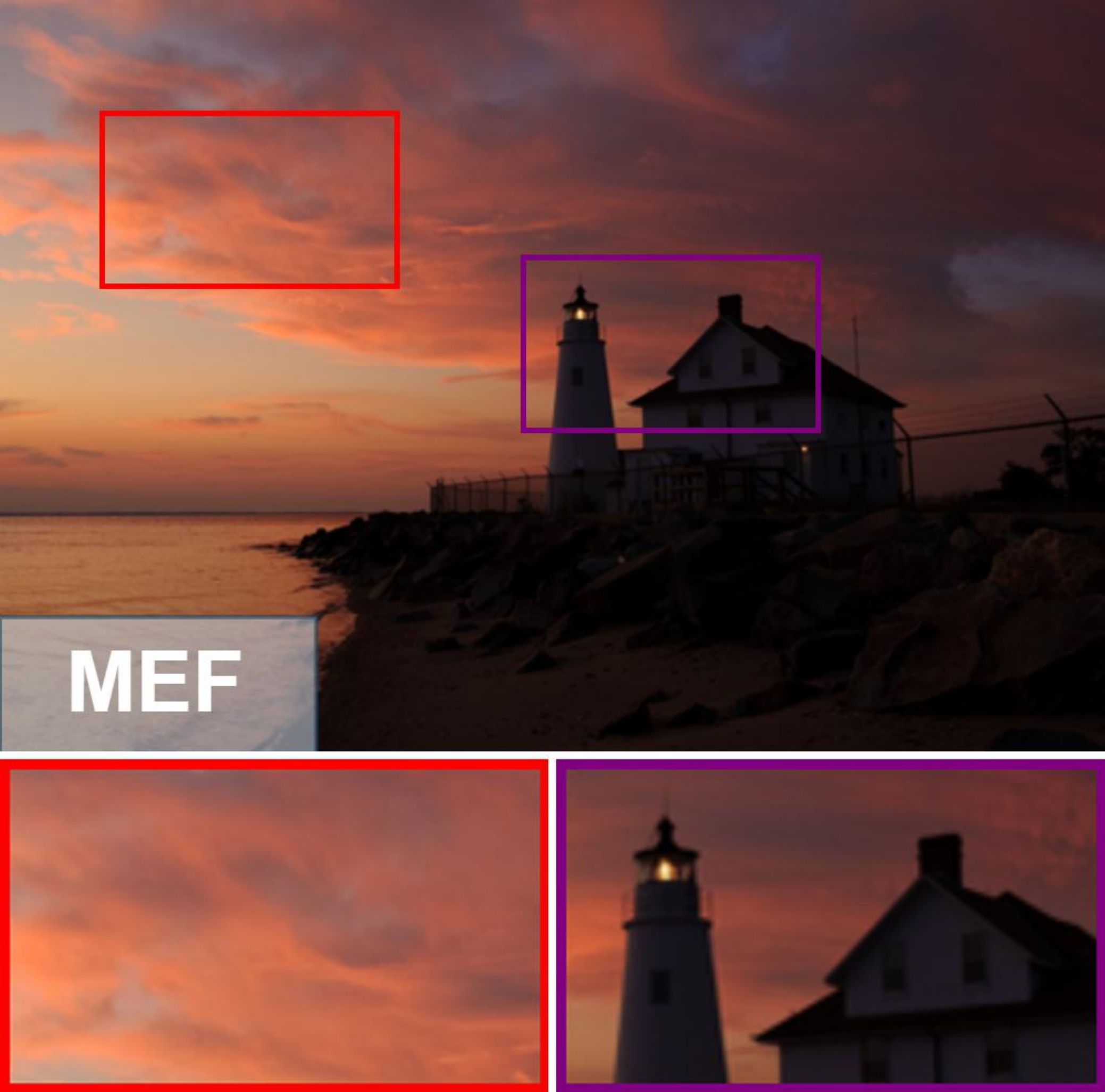} 
		\caption{\footnotesize Low-light}
	\end{subfigure}
	\begin{subfigure}{0.16\linewidth}
		\centering
		\includegraphics[width=\linewidth]{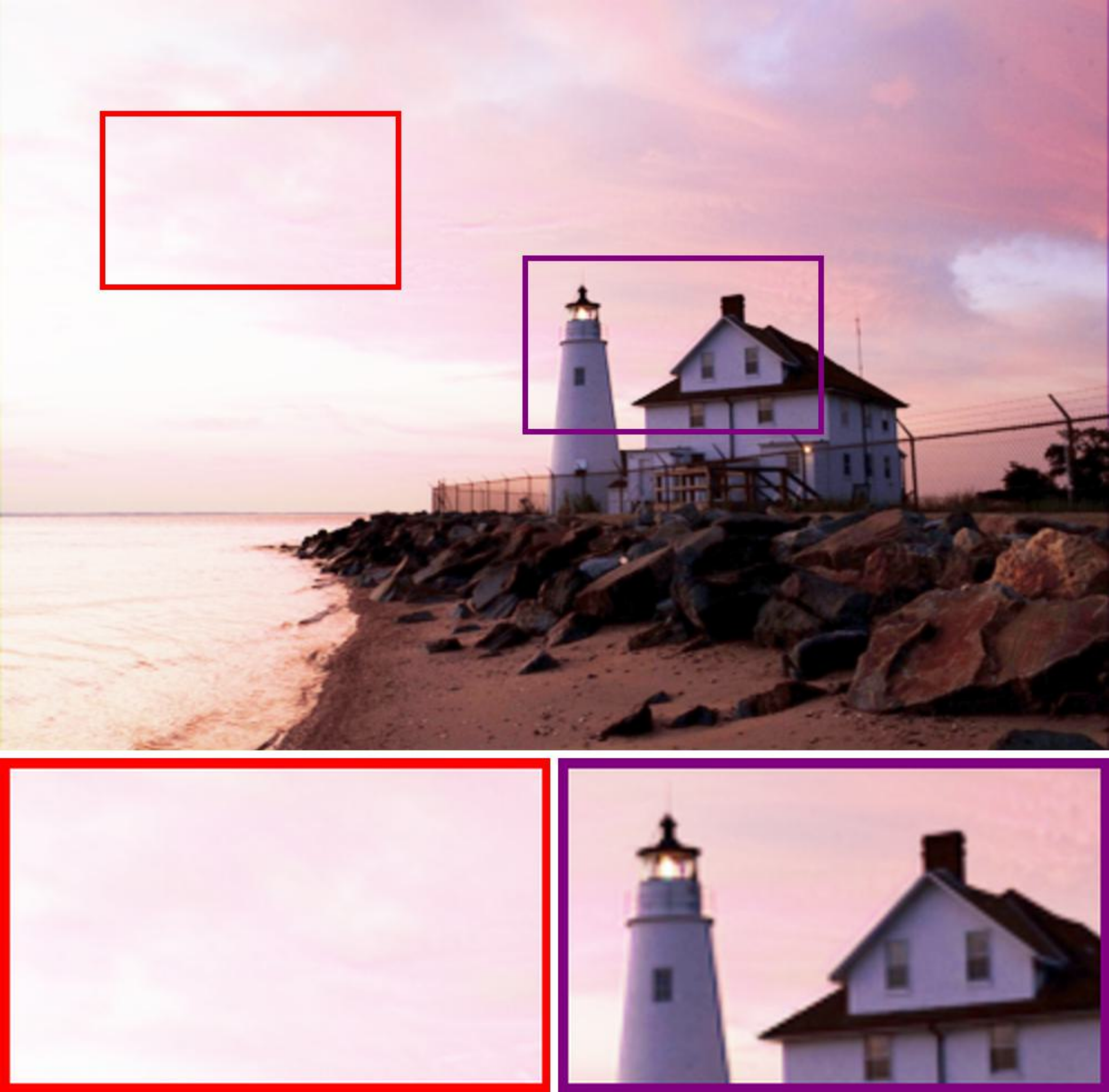} 
		\caption{\footnotesize SCI}
	\end{subfigure}
	\begin{subfigure}{0.16\linewidth}
		\centering
		\includegraphics[width=\linewidth]{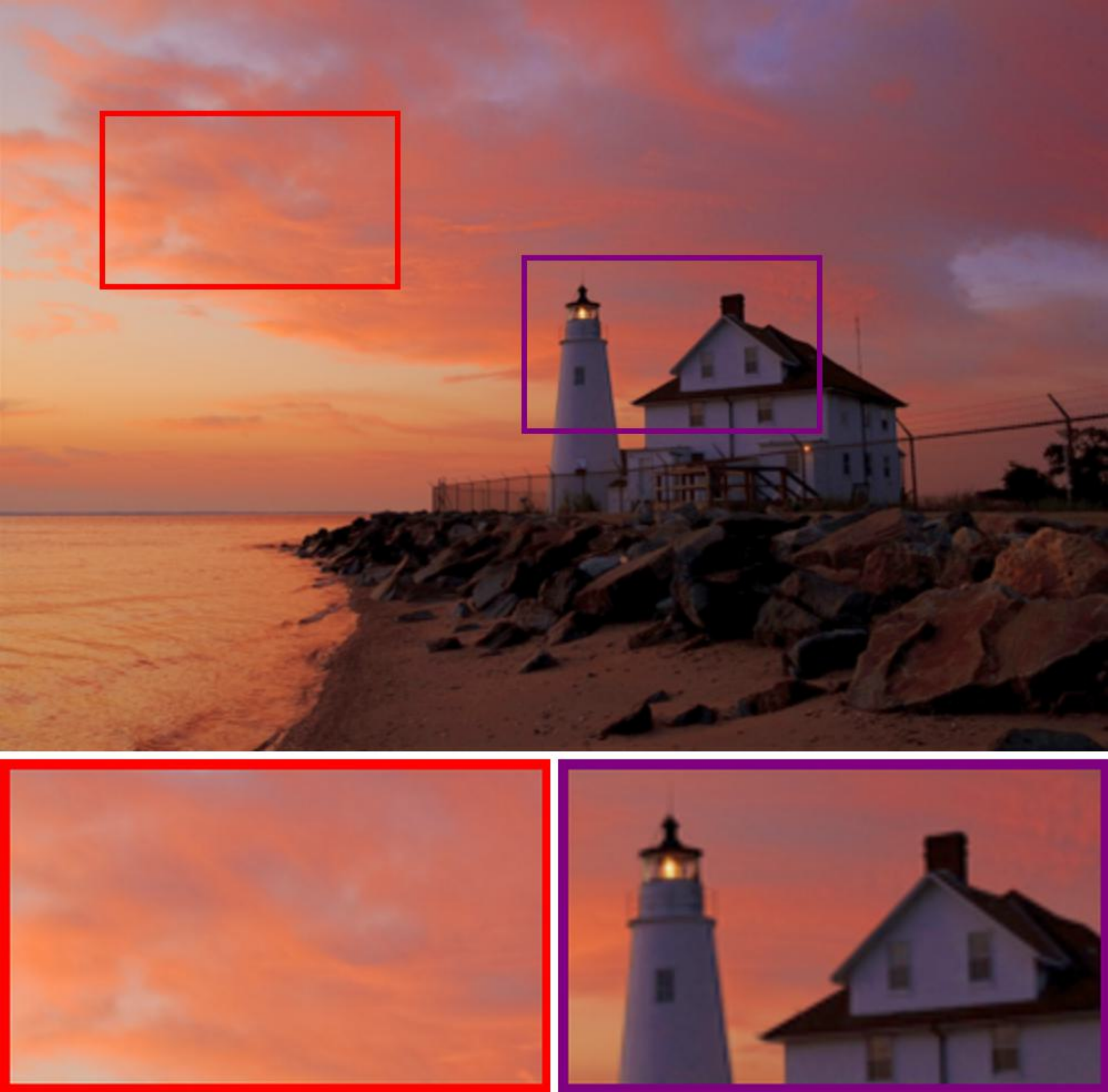}
		\caption{\footnotesize Proposed}
	\end{subfigure}

     \begin{subfigure}{0.16\linewidth}
		\centering
		\includegraphics[width=\linewidth]{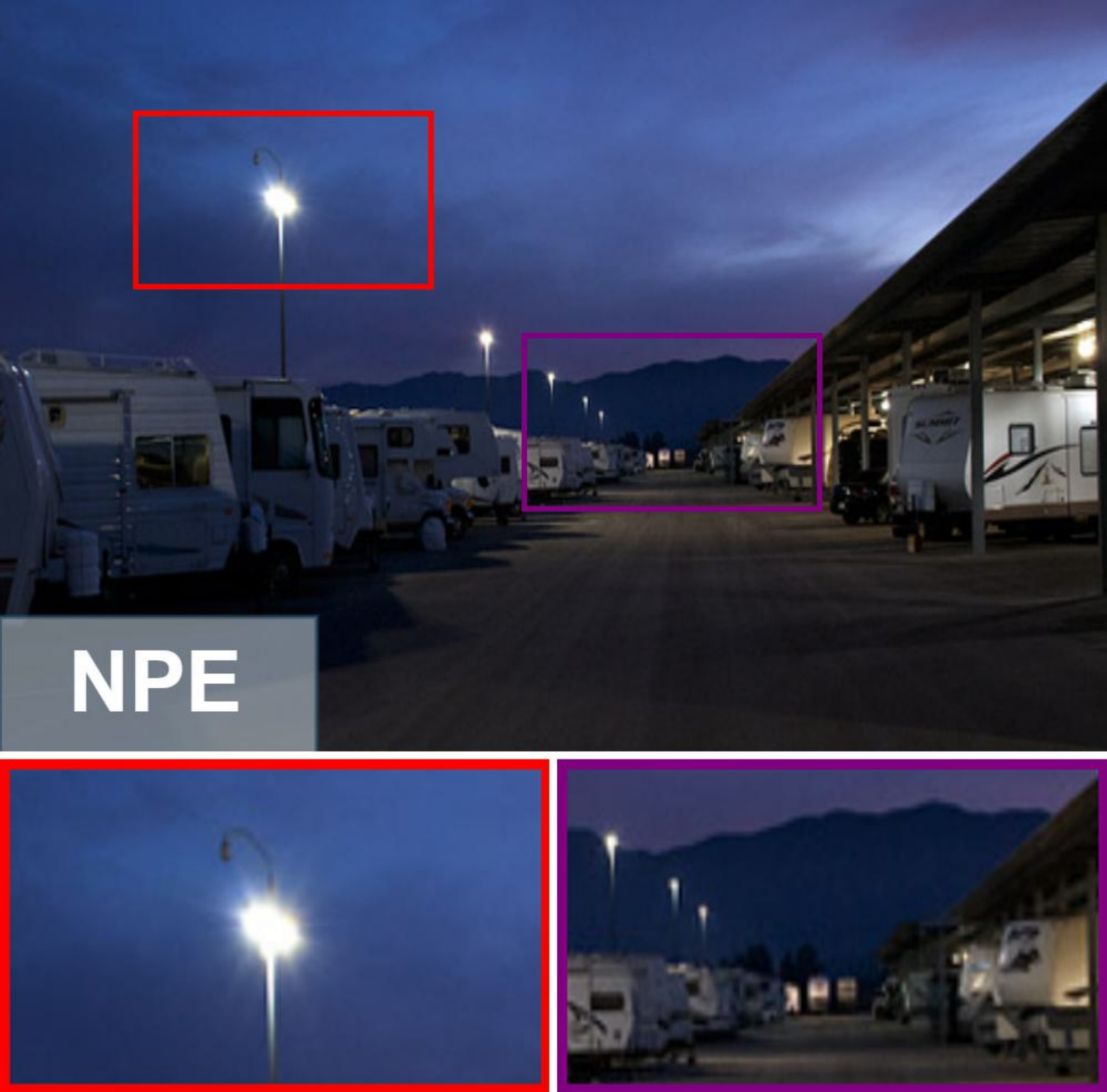} 
		\caption{\footnotesize Low-light}
	\end{subfigure}
	\begin{subfigure}{0.16\linewidth}
		\centering
		\includegraphics[width=\linewidth]{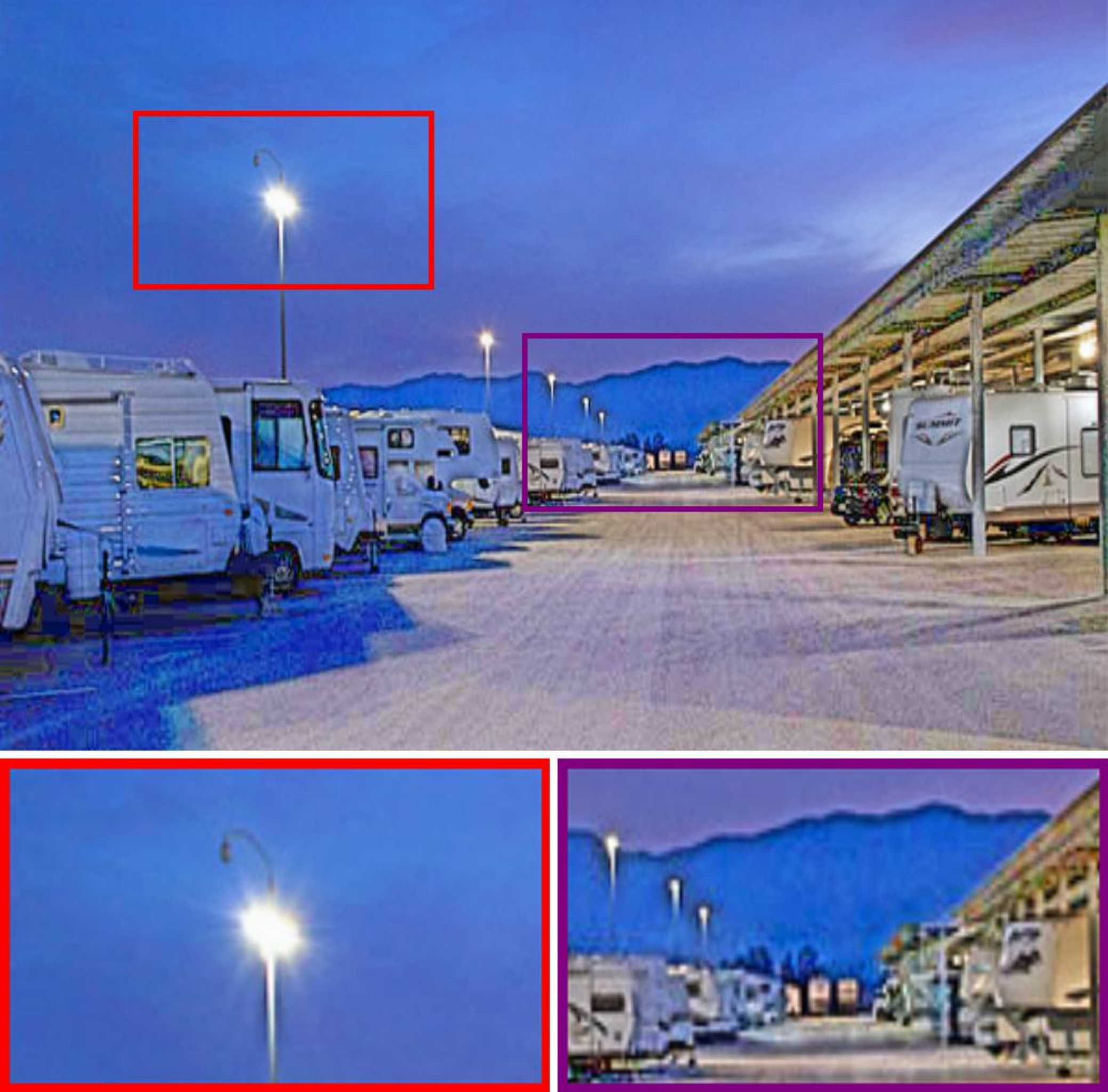} 
		\caption{\footnotesize RetinexNet}
	\end{subfigure}
	\begin{subfigure}{0.16\linewidth}
		\centering
		\includegraphics[width=\linewidth]{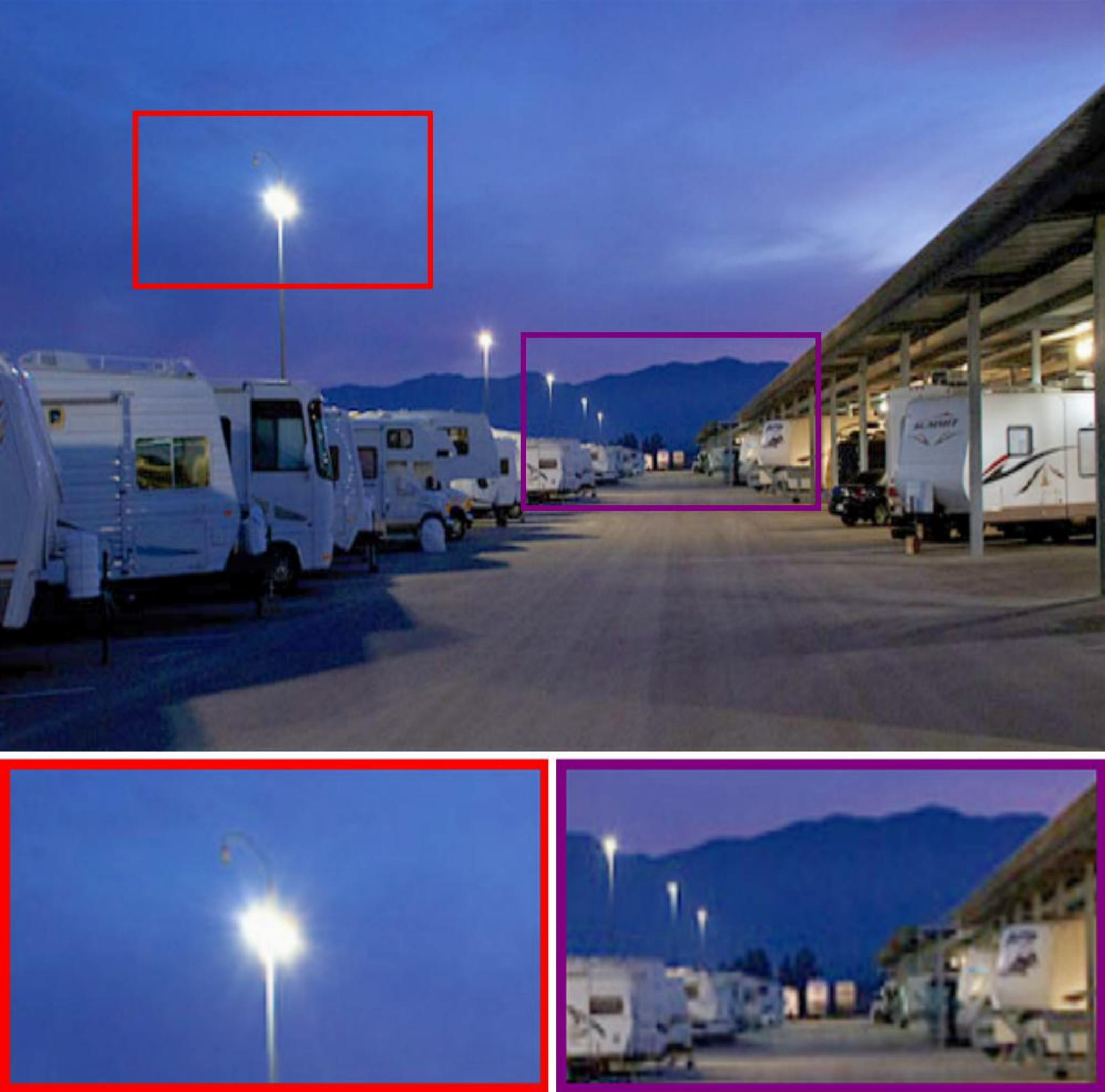}
		\caption{\footnotesize Proposed}
	\end{subfigure}
    \begin{subfigure}{0.16\linewidth}
		\centering
		\includegraphics[width=\linewidth]{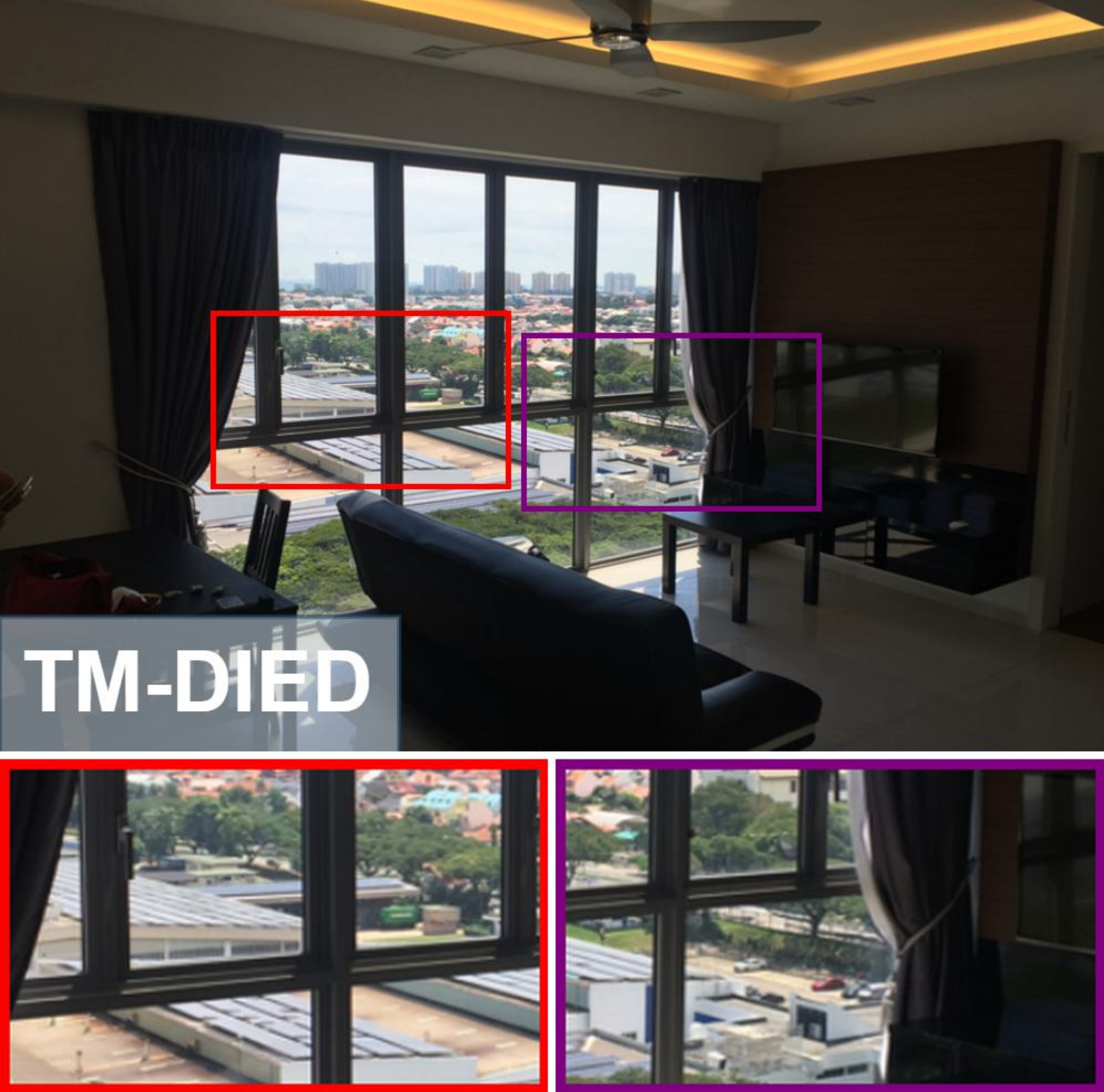} 
		\caption{\footnotesize Low-light}
	\end{subfigure}
	\begin{subfigure}{0.16\linewidth}
		\centering
		\includegraphics[width=\linewidth]{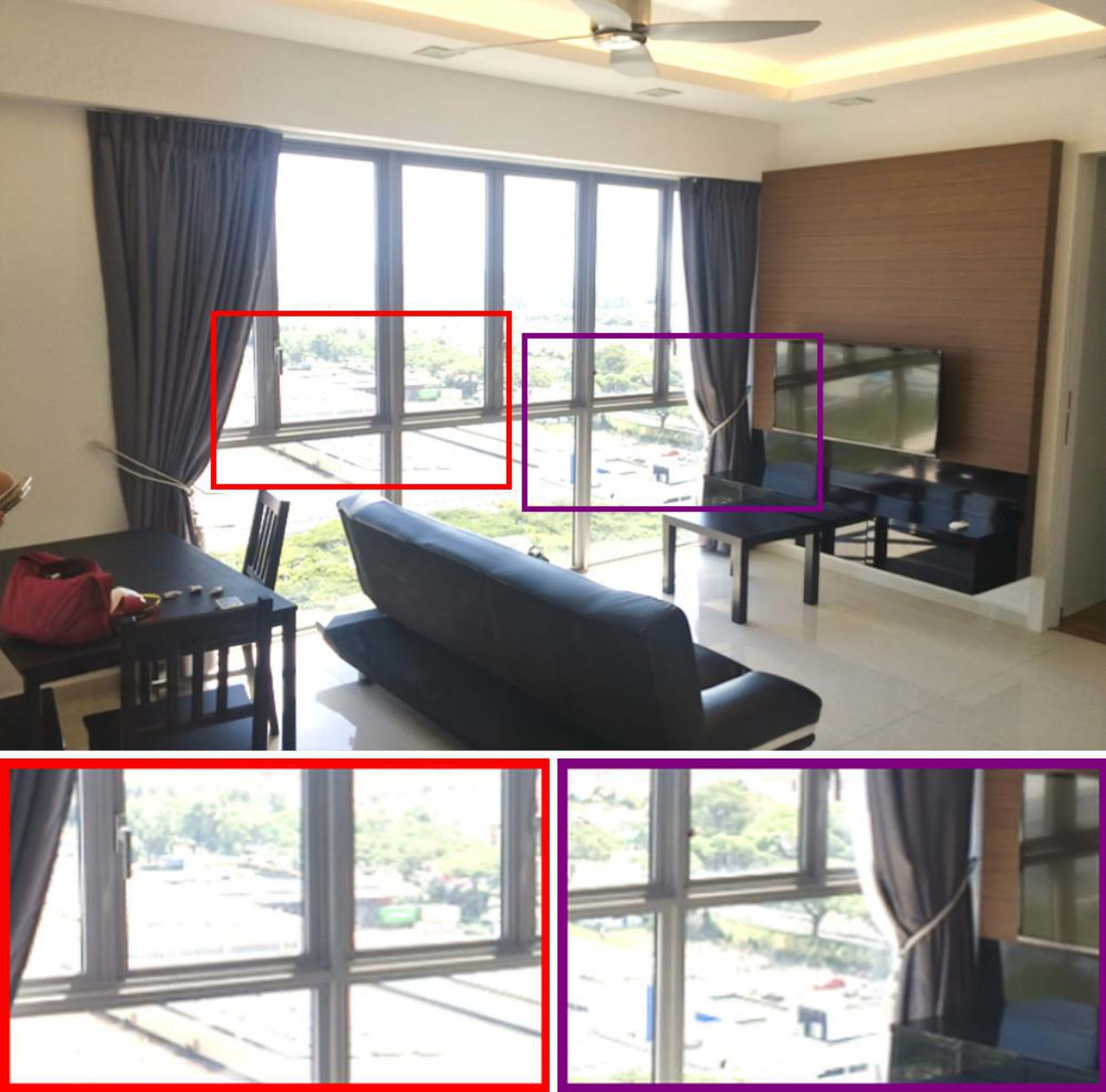} 
		\caption{\footnotesize EMNet}
	\end{subfigure}
	\begin{subfigure}{0.16\linewidth}
		\centering
		\includegraphics[width=\linewidth]{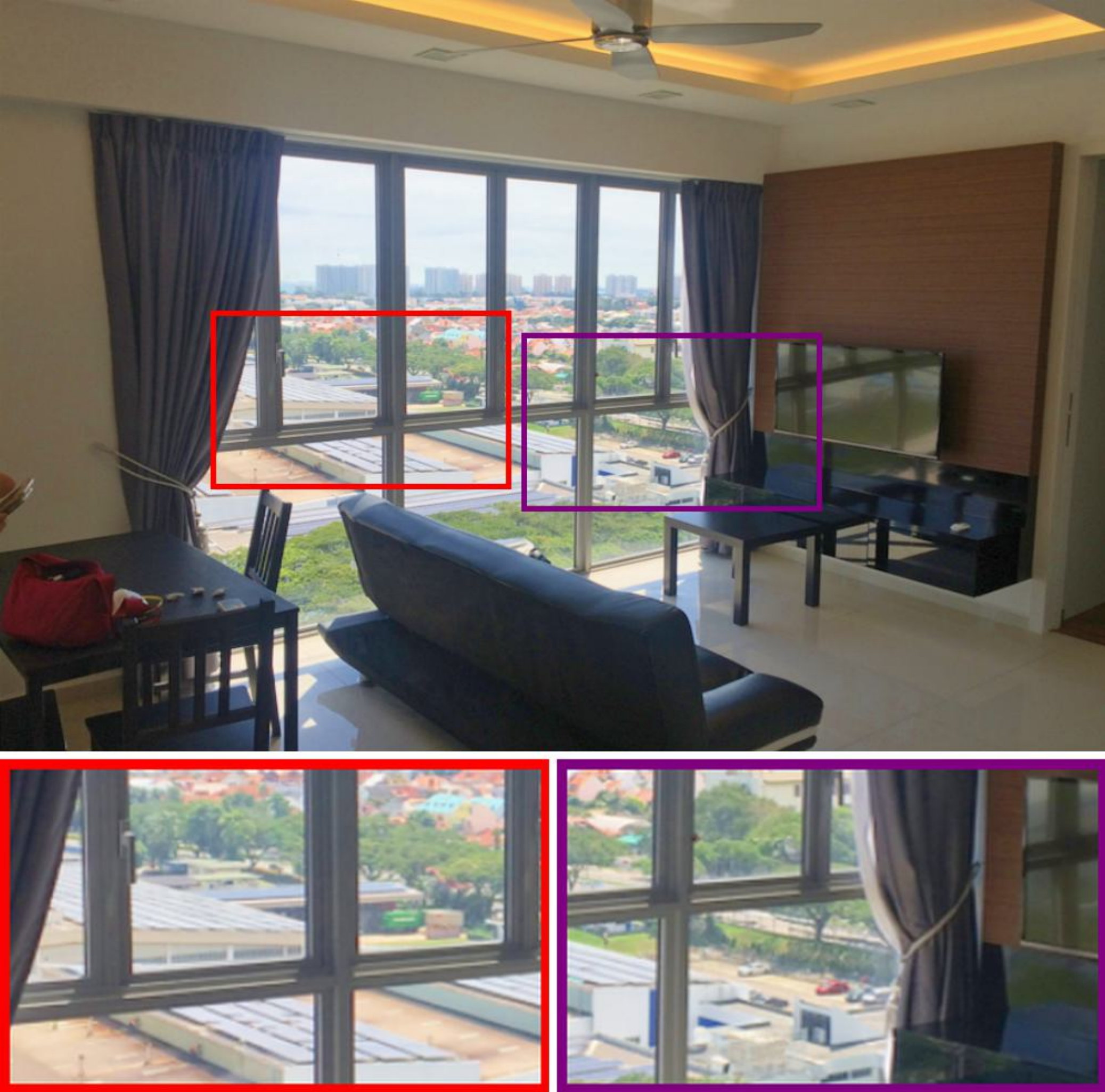}
		\caption{\footnotesize Proposed}
	\end{subfigure}

	\caption{Visual results on the DICM, DIS, LIME, MEF, NPE, and TM-DIED datasets.}
	\label{COLLAGE_SOTA}
\end{figure*}

\subsubsection{Loss functions}
\label{modeldes}
The overall design of the model is based on two fundamental structures encompassing classification and estimation aspects for low-light images. This design is divided into four independent sub-processes. Although the entire architecture focuses on the classifier's decisions, it is essential to consider certain criteria to optimize the performance of each proposed design. A key criterion is the loss function, which measures the difference between the model's predictions and the actual values. During training, the main objective is to minimize the loss function by adjusting the model's parameters. In the proposed architecture, four different loss functions are employed, offering significant potential for improvement in low-light images.

\textbf{Loss function for illuminance classification.} For SLCformer, the Binary Cross-Entropy (BCE) \cite{PyTorchBCELoss} loss function will be used, specifically tailored for binary classification problems. This function measures the difference between two probability distributions: the probability distribution of true labels and the probability distribution predicted by the model. It is mathematically expressed as follows:

\begin{equation}
\begin{aligned}
    \mathcal{L}_{CL}(\hat{y}_i, y_i) = & - \frac{1}{N} \sum_{i=0}^{N} \Big[ y_i \cdot \log(\hat{y}_i) + \\
    & (1 - y_i) \cdot \log(1 - \hat{y}_i) \Big]
\end{aligned}
\end{equation}

\noindent where $N$ is the batch size of low-light images, $y_i$ represents the actual label, indicating whether the image needs global or local illumination enhancement. On the other hand, $\hat{y}_i$ denotes the predicted probability of the category, ranging from 0 to 1. The function $log(\cdot)$ converts probabilities to the natural logarithm scale. The term $y_i \cdot log(\hat{y}_i)$ corresponds to the global category, while $(1-y_i) \cdot log(1-\hat{y}_i)$ refers to the local category.

\textbf{Loss functions for image enhancement.} The estimation part is divided into three sections: $\mathcal{L}_{LE}$ for local illumination enhancement, $\mathcal{L}_{GE}$ for global illumination enhancement, and $\mathcal{L}_{CE}$ for image color enhancement. The first section focused on local illumination enhancement ($\mathcal{L}_{LE}$) and aims to preserve the structure of the image. Therefore, the loss function in this case is:

\begin{equation}
	\mathcal{L}_{LE}=\mathcal{L}_{s}
\end{equation}





The second section, for global illumination enhancement $\mathcal{L}_{GE}$, and the third section, color enhancement $\mathcal{L}_{CE}$, incorporate three shared loss components: contrast ($\mathcal{L}_{c}$), structure ($\mathcal{L}_{s}$), and perceptual quality ($\mathcal{L}_{p}$). Although both  $\mathcal{L}_{GE}$ and $\mathcal{L}_{CE}$ employ the same formulation, they are applied to distinct enhancement objectives. The combined loss function for these tasks is given by:

\begin{equation}
\mathcal{L}_{GE} = \mathcal{L}_{CE} = \mathcal{L}_{c} + \mathcal{L}_{s} + \mathcal{L}_{p}
\end{equation}

Global illumination enhancement $\mathcal{L}_{GE}$ ensures that brightness levels are evenly distributed across the entire scene, whereas color enhancement $\mathcal{L}_{CE}$ focuses on refining color differences to improve perceptual quality. Since these objectives target different aspects of image enhancement, the same loss components  $\mathcal{L}_{c}$, $\mathcal{L}_{s}$ and $\mathcal{L}_{p}$ are optimized separately for each task.

The loss components of $\mathcal{L}_{c}$, $\mathcal{L}_{s}$ and $\mathcal{L}_{p}$ is mathematically expressed as follows:

\begin{equation}
	\mathcal{L}_{c}(\hat{y}, y) = \frac{1}{N} \sum_{i=1}^{n} (y - \hat{y})^2
	\label{contrast}
\end{equation}

\begin{equation}
	\mathcal{L}_{s}(\hat{y}, y) = 1 - \frac{(2\mu_{\hat{y}}\mu_{y} + c_1)(2\sigma_{\hat{y} {y}} + c_2)}{(\mu_{\hat{y}}^2 + \mu_{y}^2 + c_1)(\sigma_{\hat{y}}^2 + \sigma_{y}^2 + c_2)}
	\label{structural}
\end{equation}

\begin{equation}
	\mathcal{L}_{p} = \frac{1}{N}\sum_{i=1}^{N}\frac{1}{C_{j}H_{j}W_{j}}\left\|\phi_j(\hat{y})-\phi_j(y)\right\|^2_2
	\label{perceptual}
\end{equation}

For the loss component $\mathcal{L}_{c}$, it is computed using the Mean Squared Error (MSE) function \cite{pytorch_mseloss}. Here, $\hat{y}$ represents the model's prediction, $y$ is the ground truth of the reference image, and $N$ is the total number of pixels.

On the other hand, the loss component $\mathcal{L}_{s}$ is based on the Structural Similarity Index Measure (SSIM) \cite{wang2004image}. Here, $\hat{y}$ and $y$ represent the two images being compared, $\mu_{\hat{y}}$ and $\mu_{y}$ denote the mean of the images $\hat{y}$ and $y$, respectively. Additionally, $\sigma_{\hat{y}}$ and $\sigma_{y}$ represent the variance of images $\hat{y}$ and $y$, while $\sigma_{{\hat{y}}{y}}$ denotes the covariance between images $\hat{y}$ and $y$. To prevent division by zero, constants $C_1$ and $C_2$ are introduced as small values.

Finally, the loss component $\mathcal{L}_{p}$ utilizes features extracted from the VGG network \cite{johnson2016perceptual,simonyan2014very}. Here, $N$ represents the total number of elements in the dataset, $C_{j}$, $H_{j}$, and $W_{j}$ represent the number of channels, height, and width, respectively, in the $j$-th layer extracted by the VGG network. $\phi_j(\hat{y})$ and $\phi_j(y)$ correspond to the features extracted by the VGG network from the enhanced and reference images, respectively, at the $j$-th layer.

\begin{table*}[ht]
	\centering
	\caption{Comparison results on UHD-LOL4K and LSRW datasets regarding PSNR, SSIM, UQI, LPIPS, DeltaE, MUSIQ, NIQE and LOE. The bold highlights the best result.}
	\label{C3Tab4}
	\resizebox{0.7\textwidth}{!}{
		\begin{tabular}{l c c c c c c c c c}
			\hline
			\multirow{2}{*}{Method} & \multirow{2}{*}{Year} & \multicolumn{5}{c}{FIQA} & \multicolumn{3}{c}{NIQA}\\ 
			\cmidrule(lr){3-7}  \cmidrule(lr){8-10}
			&  & PSNR$\uparrow$ & SSIM$\uparrow$ & UQI$\uparrow$ &  LPIPS$\downarrow$ & DeltaE$\downarrow$ & MUSIQ$\uparrow$ & NIQE$\downarrow$ & LOE$\downarrow$ \\ \hline
			LIME~\cite{guo2016lime} & 2016 & 17.1743	&	0.7229	&	0.8248	&	0.2055	&	16.4256	&	64.3289	&	4.8131	&	836.6412  \\
			RetinexNet~\cite{wei2018deep} & 2018  	& 15.4669	&	0.6430	&	0.7961	&	0.2811	&	20.0278	&	68.2831	&	5.6011	&	870.8079  \\
			Zero-DCE~\cite{guo2020zero} & 2020 	& 18.5357	&	0.7344	&	0.8588	&	0.1515	&	14.0031	&	66.1450	&	4.6980	&	234.1234  \\
			RUAS~\cite{Liu21} & 2021  & 14.7905	&	0.6446	&	0.8171	&	0.2228	&	16.6120	&	64.6593	&	5.2271	&	835.3874 	\\
			UTVNet~\cite{zheng2021adaptive} & 2021 & 16.6559	&	0.6170	&	0.8480	&	0.1802	&	16.2977	&	66.3745	&	4.5571	&	393.0907  \\
			HWMNet~\cite{fan2022half} & 2022  & 18.1116	&	0.7657	&	0.8713	&	0.1500	&	14.4138	&	67.4105	&	4.3973	&	332.1205  \\
			LLFlow~\cite{wang2022lowlight} & 2022  & 18.2048	&	0.7531	&	0.8694	&	0.1565	&	14.2716	&	67.3453	&	4.1433	&	508.7426 	\\
			SCI~\cite{ma2022toward} & 2022   & 15.2598	&	0.6440	&	0.7901	&	0.2057	&	18.9264	&	61.9445	&	4.8367	&	\textbf{193.8519}  \\
			IAT~\cite{cui2022you} & 2022  	& 17.7615	&	0.7311	&	0.8498	&	0.1715	&	15.8407	&	60.0362	&	4.3467	&	382.2881  \\
            URetinex~\cite{wu2022uretinex} & 2022 & 19.8574	&	0.7731	&	0.8788	&	0.1355	&	13.1737	&	68.7522	&	4.4255	&	326.2203  \\
			BL~\cite{ma2023bilevel} & 2023 	& 14.5256	&	0.6096	&	0.6777	&	0.1950	&	24.8426	&	60.9764	&	5.0118	&	511.0316  \\
			PairLIE~\cite{fu2023learning} & 2023  	& 16.9248	&	0.7232	&	0.8305	&	0.1745	&	17.7023	&	66.2740	&	4.4475	&	432.4062 	\\
			CLIP-LIT~\cite{liang2023iterative} & 2023  	& 17.3690	&	0.7473	&	0.8100	&	0.1875	&	17.8064	&	65.8186	&	4.8718	&	732.5396  \\
            EMNet~\cite{ye2023glow} & 2023 & 17.7642	&	0.7465 & 0.8701	&	0.1448	&	14.6159	&	66.8438 & 4.2512	&	386.7146\\
			SHAL-Net~\cite{xu2024degraded} & 2024 	& 17.6318	&	0.6682	&	0.8607	&	0.2031	&	14.5357	&	62.4341	&	4.6036	&	400.4817  \\
			GSAD~\cite{hou2024global} & 2024 & 19.3514	&	0.7478	&	0.8756	&	0.1660	&	13.3485	&	67.1460	&	4.2542	&	258.5360  \\
            PPformer~\cite{dang2024ppformer} & 2024 & 19.4021	&	0.7801 & 0.8722	&	0.1355	&	13.9428	&	65.6636 & 4.0883	&	275.2424\\
            GCP~\cite{jeon2024low} & 2024 & 19.1487	&	0.7091	&	0.8477	&	0.2084	&	12.4680	&	66.5594	&	5.1027	&	504.6734  \\
            NeurBR~\cite{zhao2024non} & 2024 & 17.6668	&	0.7298	&	0.8106	&	0.1719	&	16.5966	&	65.4647	&	4.7649	&	729.5797 \\
            ITRE~\cite{wang2024itre}  & 2024 & 14.7956	&	0.6774	&	0.7062	&	0.1834	&	25.4139	&	61.2213	&	4.5287	&	1040.4291  \\
            PIE~\cite{liang2024pie} & 2024 & 15.6341	&	0.7168	&	0.7630	&	0.1669	&	20.8683	&	65.8341	&	4.6343	&	431.6078  \\
            MSATr~\cite{fang2024non} & 2024 & 16.5095	&	0.6979 & 0.8145 & 0.2544 & 19.1164 & 53.3297 & 3.7131	&	601.7402 \\
			ALEN (Proposed) & 2025  & \textbf{20.1811}	& \textbf{0.8015}	&	\textbf{0.8920}	&	\textbf{0.1215}	&	\textbf{11.8815}	&	\textbf{69.0735}	&	\textbf{3.8941}	&	194.5614\\ \hline
		\end{tabular}
	}
    \end{table*}

\begin{table*}[ht]
	\centering
	\caption{Comparison results on DICM, LIME, MEF, NPE, TM-DIED, and DIS datasets regarding NIQE and LOE. The bold highlights the best result.}
	\label{UPDS}
	\resizebox{2.1\columnwidth}{!}{
		\begin{tabular}{l c c c c c c c c c c c c c | c c}
			\hline
   
                \multirow{2}{*}{Method} & \multirow{2}{*}{Year} & \multicolumn{2}{c}{DICM} & \multicolumn{2}{c}{DIS} & \multicolumn{2}{c}{LIME} & \multicolumn{2}{c}{MEF} & \multicolumn{2}{c}{NPE} & \multicolumn{2}{c}{TM-DIED} & \multicolumn{2}{c}{Average} \\
                \cmidrule(lr){3-4}  \cmidrule(lr){5-6} \cmidrule(lr){7-8} \cmidrule(lr){9-10} \cmidrule(lr){11-12} \cmidrule(lr){13-14} \cmidrule(lr){15-16}
                &  & NIQE$\downarrow$ & LOE$\downarrow$ & NIQE$\downarrow$ & LOE$\downarrow$ & NIQE$\downarrow$ & LOE$\downarrow$ & NIQE$\downarrow$ & LOE$\downarrow$ & NIQE$\downarrow$ & LOE$\downarrow$ & NIQE$\downarrow$ & LOE$\downarrow$ & NIQE$\downarrow$ & LOE$\downarrow$ \\ \hline
                
			LIME~\cite{guo2016lime} & 2016 & 3.5345	&	772.6563	&	3.2577	&	671.6454	&	4.0482	&	698.5153	&	3.4220	&	803.7408	&	3.2599	&	734.3341	&	2.8081	&	690.6513	&	3.3884	&	728.5905\\
			RetinexNet~\cite{wei2018deep} & 2018  & 4.3311	&	624.5084	&	4.1928	&	657.9254	&	5.3635	&	544.4529	&	4.9390	&	690.3006	&	3.5168	&	598.1598	&	3.7575	&	647.5019	&	4.3501	&	627.1415\\
			Zero-DCE~\cite{guo2020zero} & 2020  & 3.4540	&	193.6266	&	3.2197	&	140.9130	&	3.8189	&	138.0490	&	3.2232	&	157.4939	&	\textbf{2.8970}	&	143.4324	&	2.6711	&	164.3887	&	3.2140	&	156.3173\\
			RUAS~\cite{Liu21} & 2021  & 5.0041	&	1414.3637	&	4.1015	&	850.6067	&	4.2925	&	785.9921	&	4.0922	&	767.6648	&	6.4952	&	1290.9727	&	4.9544	&	1114.9357	&	4.8233	&	1037.4230\\
			UTVNet~\cite{zheng2021adaptive} & 2021  & 3.9962	&	519.7970	&	3.6354	&	337.9591	&	3.7031	&	206.4406	&	3.3070	&	210.0038	&	3.7538	&	477.7256	&	3.3238	&	331.2108	&	3.6199	&	347.1895\\
			HWMNet~\cite{fan2022half} & 2022  & 3.2809	&	420.9310	&	3.1222	&	210.0536	&	3.8388	&	171.9222	&	3.4695	&	237.9267	&	3.4595	&	270.6445	&	2.9894	&	290.9019	&	3.3601	&	267.0633\\
			LLFlow~\cite{wang2022lowlight} & 2022  & 3.3801	&	627.6823	&	2.9456	&	437.0992	&	3.9831	&	323.8317	&	3.5434	&	411.6201	&	3.6264	&	468.8866	&	3.0610	&	448.2885	&	3.4233	&	452.9014\\
			SCI~\cite{ma2022toward} & 2022  & 3.6684	&	286.2143	&	3.5215	&	\textbf{129.7757}	&	4.1374	&	79.7487	&	3.4398	&	91.4831	&	3.8805	&	336.6452	&	3.3104	&	236.6095	&	3.6597	&	193.4128\\
			IAT~\cite{cui2022you} & 2022  & 3.6565	&	589.3488	&	3.2308	&	242.0779	&	4.1875	&	151.3161	&	3.6588	&	214.8577	&	3.6106	&	538.4553	&	3.5058	&	378.8967	&	3.6417	&	352.4921\\
            URetinex~\cite{wu2022uretinex} & 2022  & 3.4370	&	374.3998	&	3.1313	&	231.9573	&	4.4272	&	220.4706	&	3.3741	&	229.4568	&	3.9220	&	380.2428	&	3.2063	&	313.3358	&	3.5830	&	291.6439\\
            BL~\cite{ma2023bilevel} & 2023 & 4.0940	&	813.5779	&	3.9838	&	450.3187	&	3.8805	&	187.8713	&	3.3508	&	225.8497	&	4.6247	&	750.0058	&	4.1429	&	562.9484	&	4.0128	&	498.4286\\
			PairLIE~\cite{fu2023learning} & 2023  & 3.8209	&	375.8929	&	3.3186	&	288.5944	&	4.7434	&	259.0895	&	3.9212	&	224.7747	&	3.3467	&	356.3552	&	3.4237	&	281.0192	&	3.7624	&	297.6210\\
			CLIP-LIT~\cite{liang2023iterative} & 2023 & 3.7153	&	614.7664	&	3.3942	&	489.2899	&	4.0989	&	460.4350	&	3.6041	&	557.3280	&	3.1014	&	532.9849	&	2.9500	&	519.6130	&	3.4773	&	529.0695\\
            EMNet~\cite{ye2023glow} & 2023 & 3.6405	&	521.0793	&	3.1560	&	347.4293	&	4.1514	&	198.3746	&	3.5432	&	305.2917	&	3.6855	&	493.2299	&	3.4939	&	449.9543	&	3.6118	&	385.8932\\
			SHAL-Net~\cite{xu2024degraded} & 2024  & 3.6023	&	267.4191	&	3.7415	&	238.2359	&	3.9187	&	223.7167	&	3.8450	&	209.2863	&	3.7725	&	239.3813	&	3.4845	&	218.4001	&	3.7274	&	232.7399\\
            GSAD~\cite{hou2024global} & 2024 & 3.4803	&	304.0283	&	3.2351	&	197.3239	&	5.4784	&	162.6212	&	3.6143	&	211.6887	&	4.0690	&	204.9685	&	3.1760	&	202.6175	&	3.8422	&	213.8747\\
            PPformer~\cite{dang2024ppformer} & 2024 &  \textbf{3.0554}	&	410.8877	&	2.9307	&	379.9775	&	3.7741	&	183.5403	&	\textbf{3.0326}	&	228.0672	&	3.0021	&	340.8865	&	2.8983	&	312.6565	&	3.1155	&	309.3360\\
            GCP~\cite{jeon2024low}  & 2024 & 3.4943	&	324.3585	&	3.3915	&	321.3695	&	4.2018	&	237.9097	&	3.4582	&	267.9230	&	3.0460	&	335.7797	&	2.9202	&	344.8669	&	3.4187	&	305.3679\\
            NeurBR~\cite{zhao2024non} & 2024 & 3.6046	&	696.2333	&	3.2743	&	581.2965	&	3.9914	&	464.9468	&	3.3874	&	562.1682	&	3.0953	&	701.6950	&	2.9839	&	637.9747	&	3.3895	&	607.3858 \\
            ITRE~\cite{wang2024itre} & 2024 & 3.4042	&	882.4511	&	3.1454	&	752.5867	&	3.8092	&	569.9538	&	3.2075	&	511.8575	&	3.0641	&	957.9563	&	2.6791	&	748.5330	&	3.2183	&	737.2231\\
            PIE~\cite{liang2024pie} & 2024 & 3.3417	&	349.2821	&	3.1065	&	219.5983	&	3.7318	&	190.2516	&	3.0819	&	289.1972	&	3.1097	&	322.6611	&	2.7071	&	317.8995	&	3.1798	&	281.4816 \\
            MSATr~\cite{fang2024non} & 2024 & 3.5243 & 546.6340 & 2.9436 & 583.7035 & 3.7512 & 668.4136 & 3.2930 & 807.5738	& 3.6355 & 437.7138 & 3.4295 & 516.8674 & 3.4295 & 593.4844 \\
			ALEN (Proposed) & 2025 & 3.0772	&	\textbf{107.9533} &	\textbf{2.9065}	&	163.9772	&	\textbf{3.6623}	&	\textbf{31.1855}	&	3.0888	&	\textbf{68.3452}	&	2.9734	&	\textbf{58.0590}	&	\textbf{2.5740}	&	\textbf{53.2044}	&	\textbf{3.0470}	&	\textbf{80.4541}\\ \hline
		\end{tabular}
	}
\end{table*}

\subsection{Implementation Details}
Each network dedicated to classification, local and global illumination estimation, and color enhancement underwent separate training processes. PyTorch \cite{Paszke2019Pytorch}, an open-source machine learning library based on Torch known for its efficiency in training complex architectures, facilitated the implementation. The training occurred on a PC equipped with an NVIDIA RTX 3060 GPU, an Intel Core i5-12400F CPU (running at 2.50GHz), and 16 GB of RAM.

For SLCformer, the training involved 75 epochs using batches of 32 images resized to 224$\times$224 pixels. The Adam optimizer \cite{kingma2014adam} was employed with a learning rate set to 1e-4. A total of 2,000 images from the GLI Dataset, proposed in section \ref{GLI_D}, were used for training. The local illumination estimation network underwent 500 training epochs using batches of 8 images at 224$\times$224 pixels resolution. The training also utilized the Adam optimizer with a learning rate of 1e-3, leveraging the SLL~\cite{lv2021attention} dataset comprising 22,472 image pairs.

Similarly, the global illumination estimation network was trained for 40 epochs with batches of 8 images, randomly selecting 128$\times$128 blocks from each image. The Adam optimizer with a learning rate of 1e-3 was applied, reducing the rate by half after 20 epochs. The training utilized the custom HDR+~\cite{zeng2020learning} dataset containing 922 image pairs. Lastly, the color enhancement network underwent 40 epochs of training with batches of 8 images, randomly selecting 128$\times$128 blocks. The Adam optimizer with a learning rate of 1e-4, reduced by half after 20 epochs, was employed. Training data came from the MIT-Adobe FiveK~\cite{bychkovsky2011learning} dataset, comprising 5000 image pairs.

\begin{figure*}[ht]
	\centering
	\begin{subfigure}{0.12\linewidth}
		\centering
		\includegraphics[width=\linewidth]{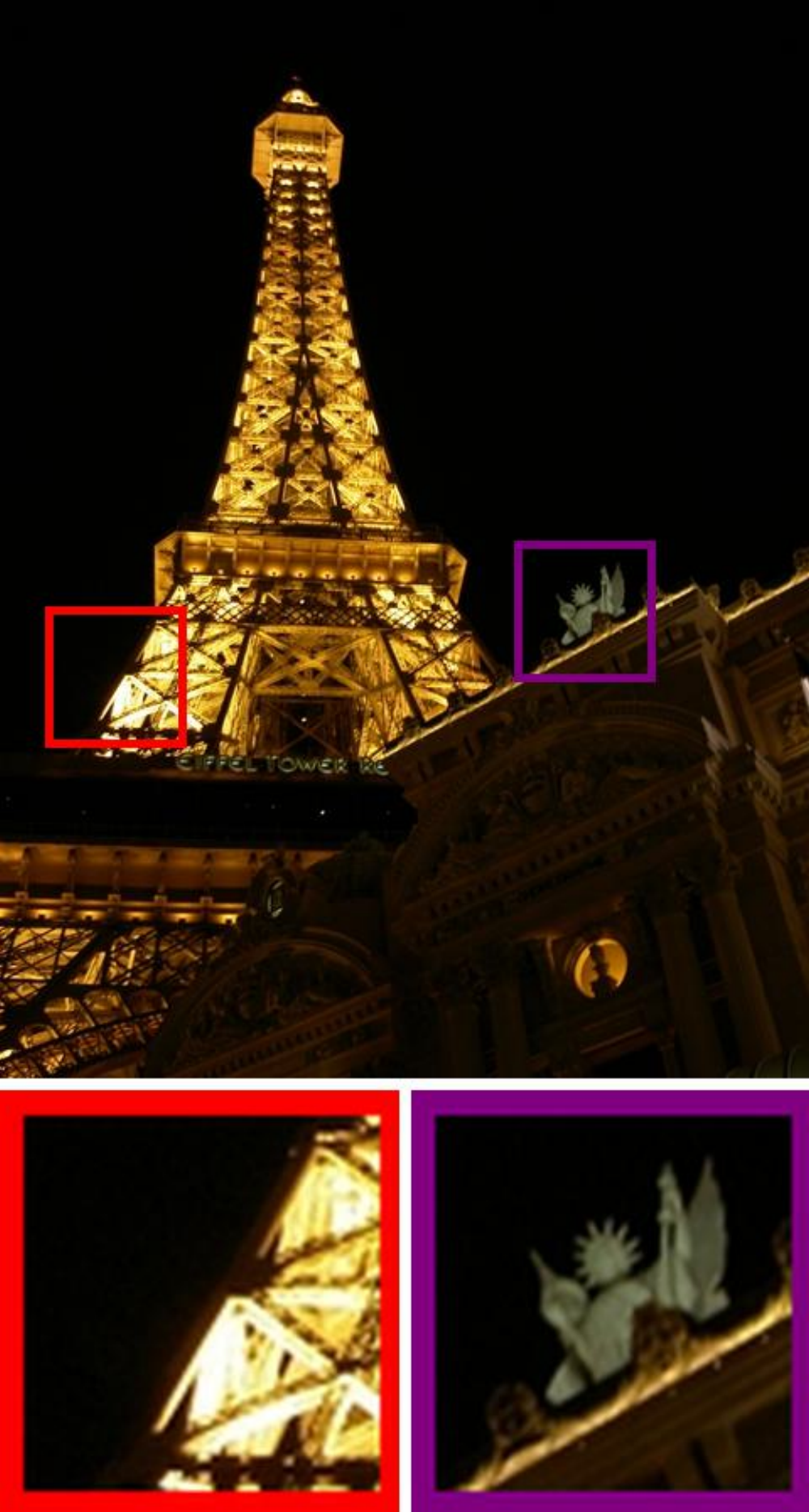} 
        \caption{\footnotesize Low-light}
	\end{subfigure}
	\begin{subfigure}{0.12\linewidth}
		\centering
		\includegraphics[width=\linewidth]{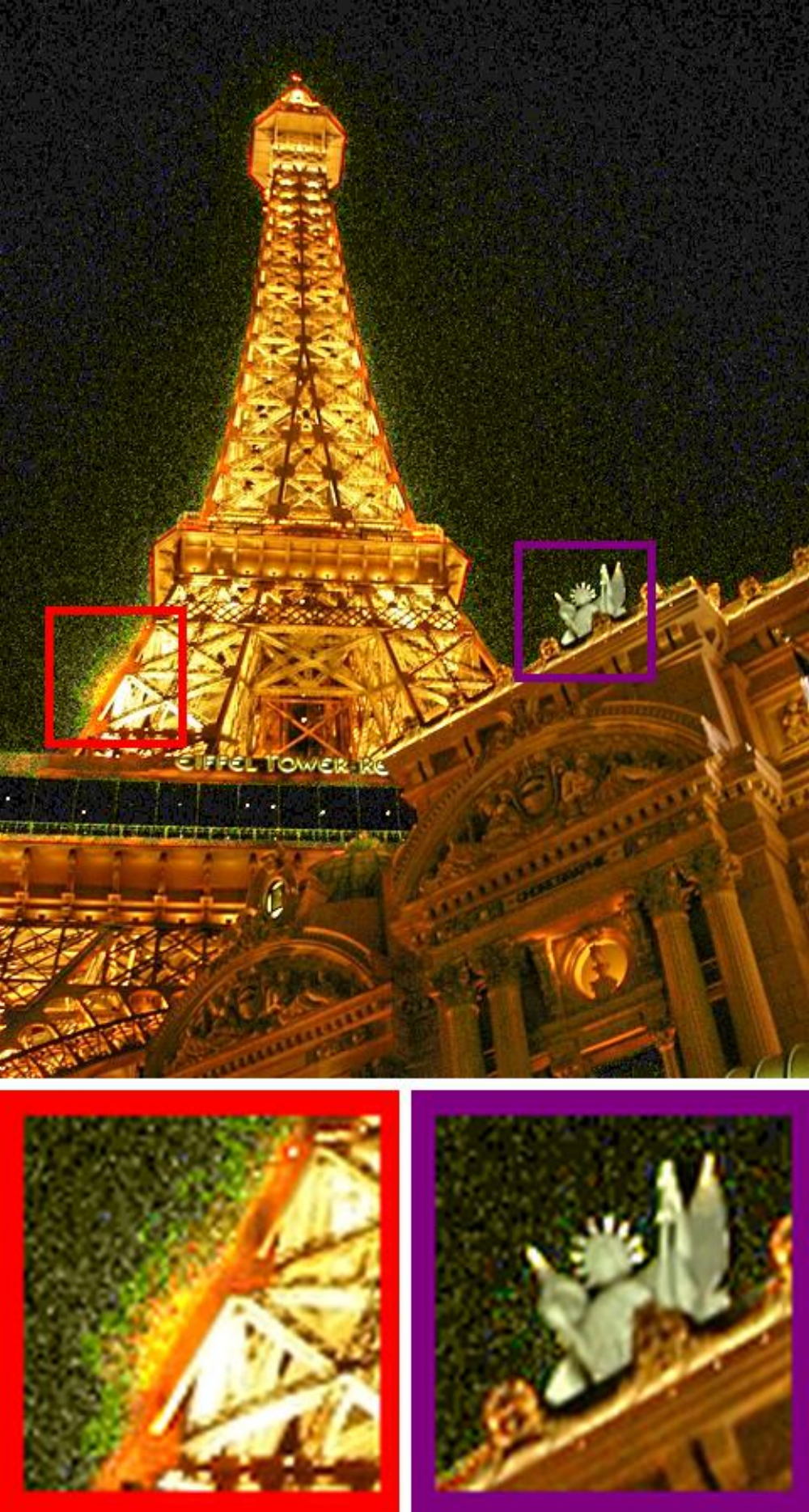} 
        \caption{\footnotesize LIME }
	\end{subfigure}
	\begin{subfigure}{0.12\linewidth}
		\centering
		\includegraphics[width=\linewidth]{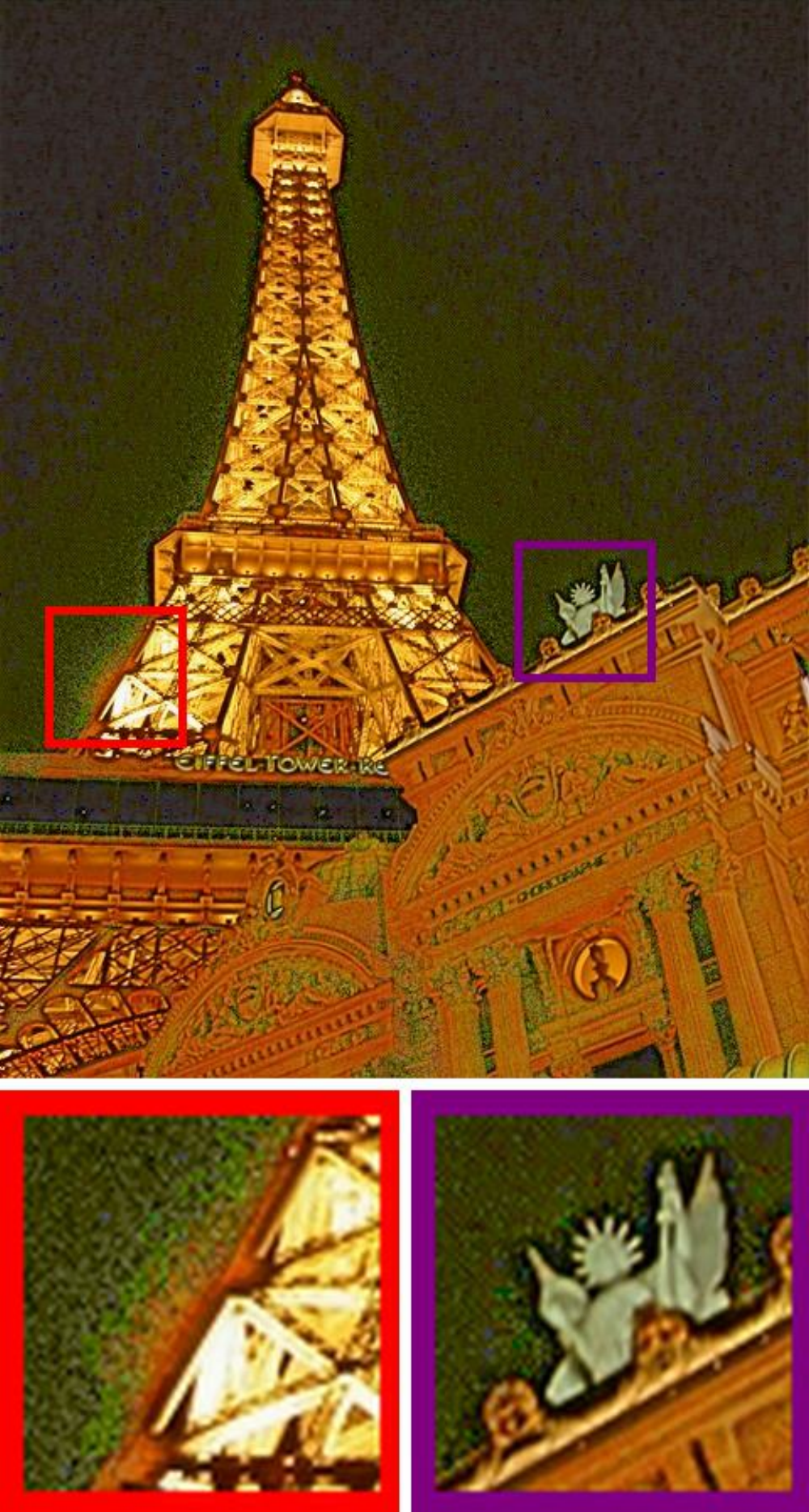}
        \caption{\footnotesize RetinexNet }
	\end{subfigure}
	\begin{subfigure}{0.12\linewidth}
		\centering
		\includegraphics[width=\linewidth]{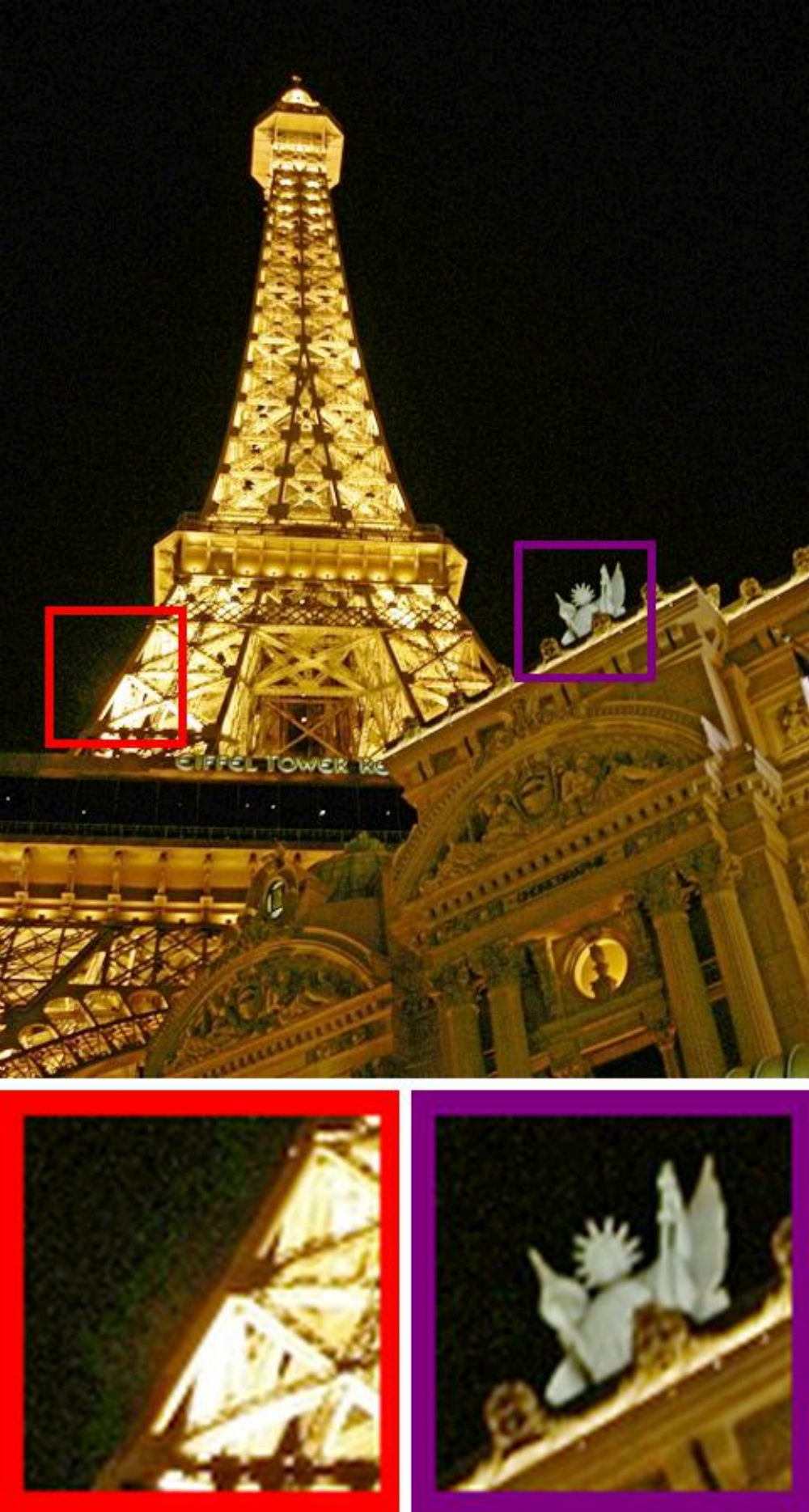} 
		\caption{\footnotesize Zero-DCE }
	\end{subfigure}
	\begin{subfigure}{0.12\linewidth}
		\centering
		\includegraphics[width=\linewidth]{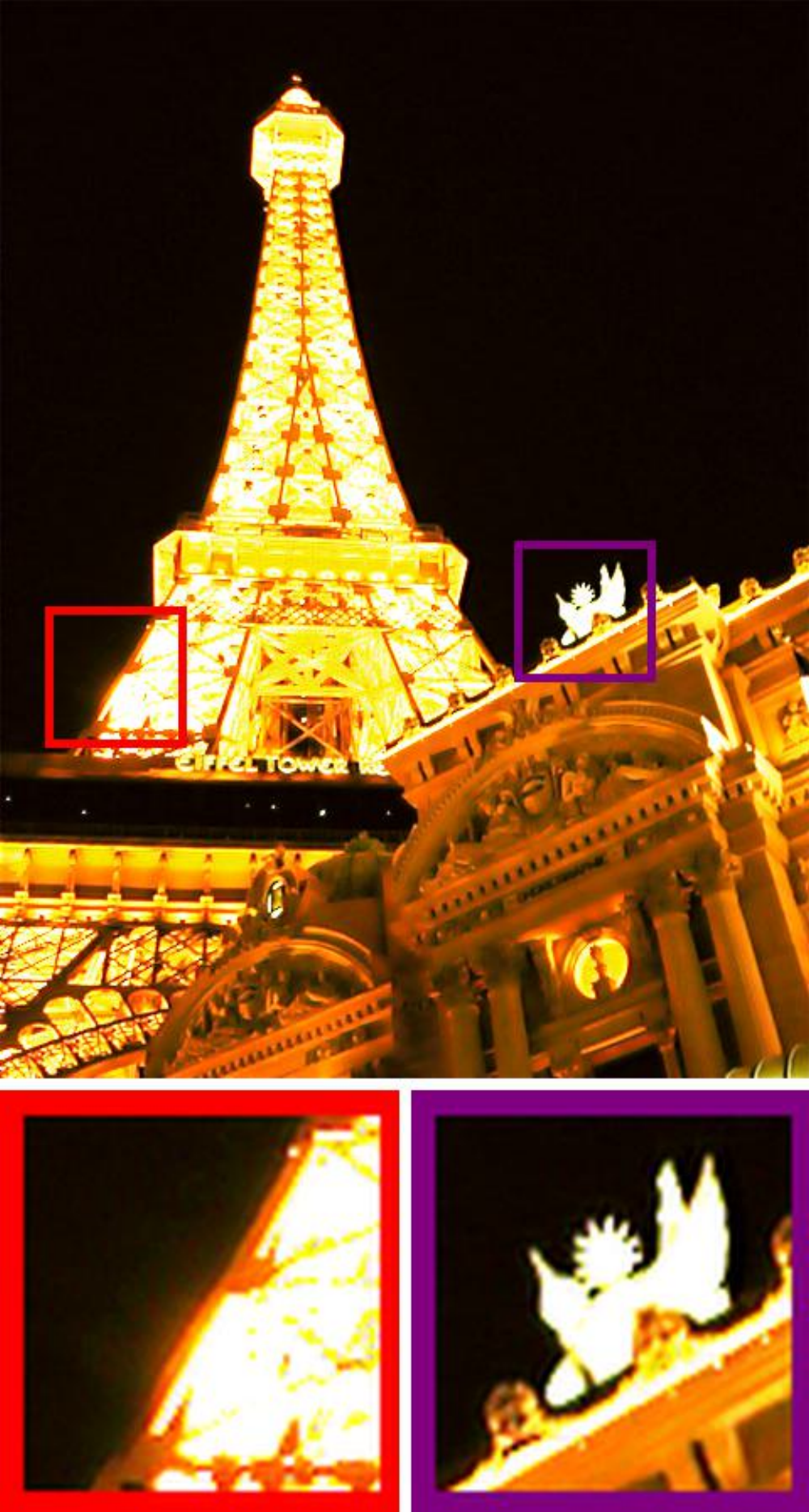} 
		\caption{\footnotesize RUAS }
	\end{subfigure}
	\begin{subfigure}{0.12\linewidth}
		\centering
		\includegraphics[width=\linewidth]{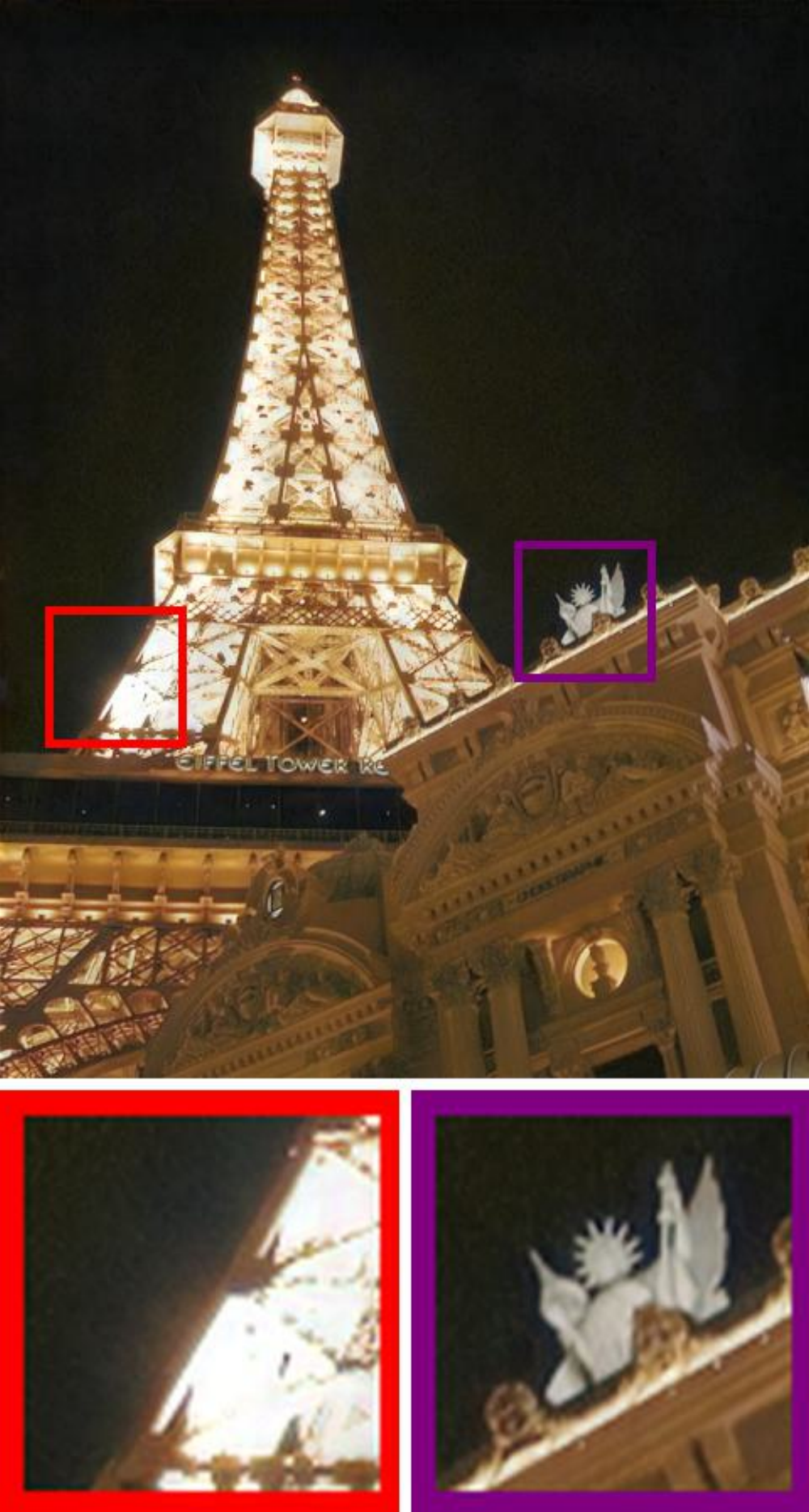}
		\caption{\footnotesize UTVNet }
	\end{subfigure}
    \begin{subfigure}{0.12\linewidth}
		\centering
		\includegraphics[width=\linewidth]{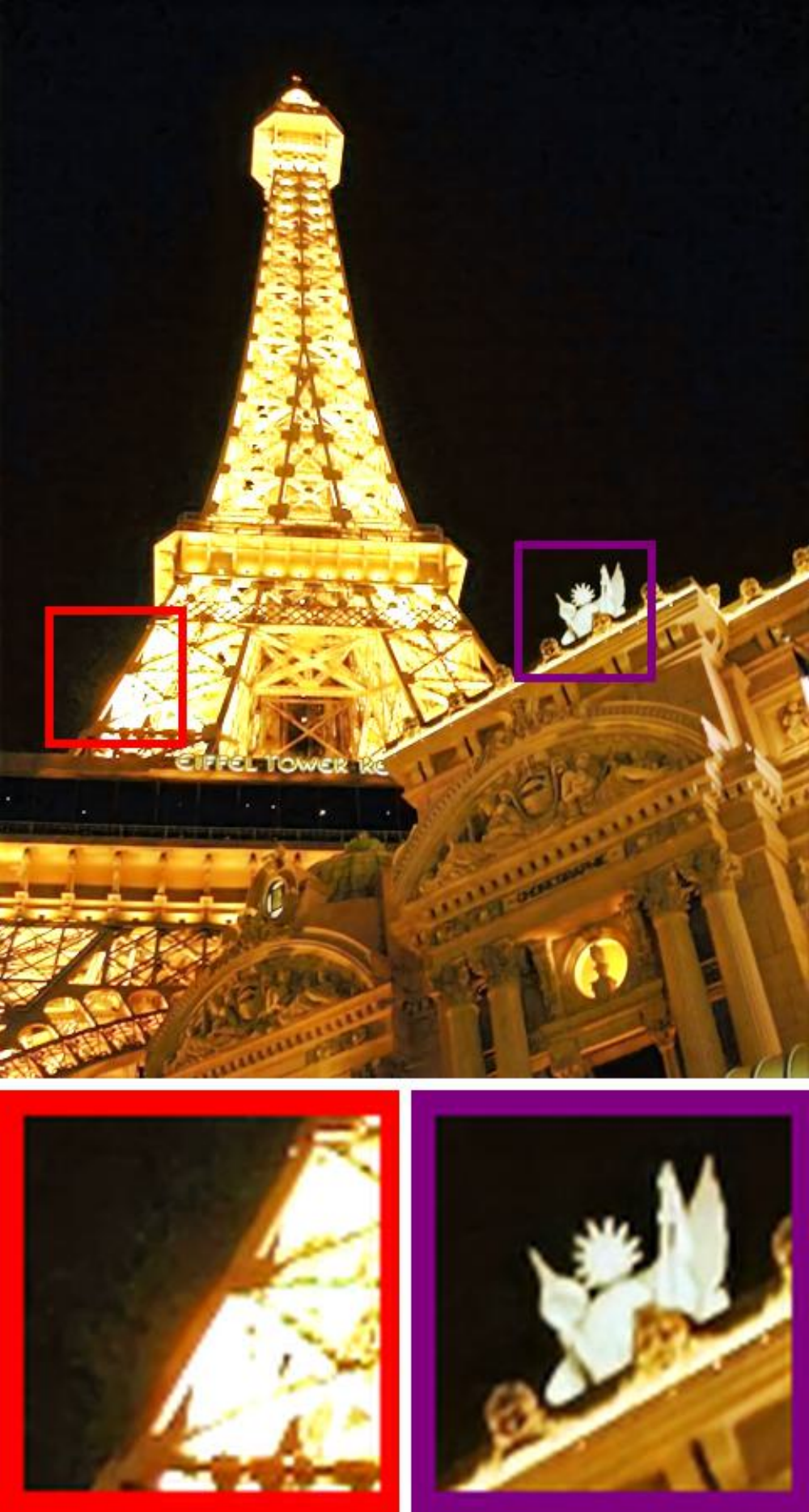} 
		\caption{\footnotesize HWMNet }
	\end{subfigure}
	\begin{subfigure}{0.12\linewidth}
		\centering
		\includegraphics[width=\linewidth]{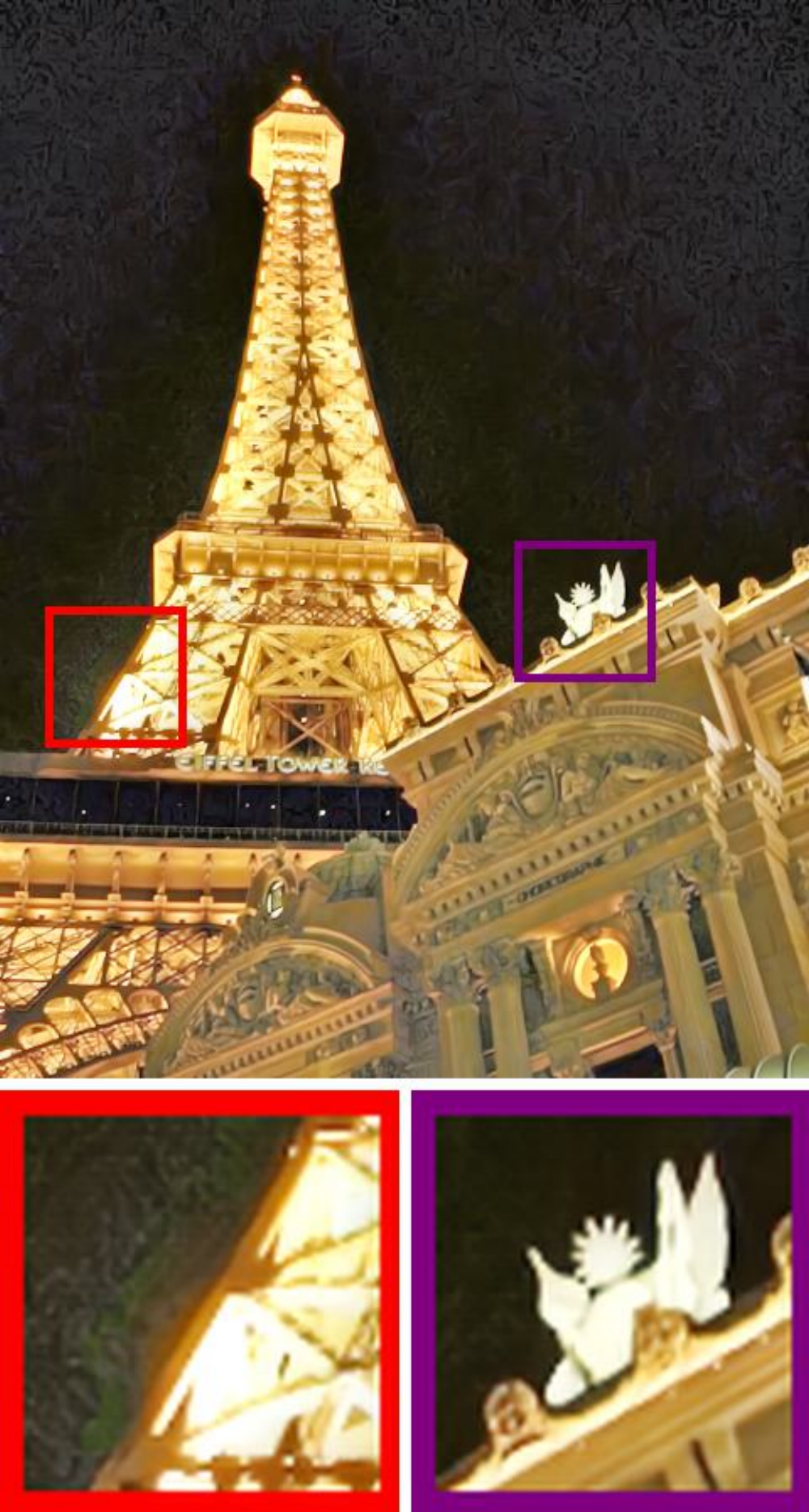} 
		\caption{\footnotesize LLFlow }
	\end{subfigure}


    \begin{subfigure}{0.12\linewidth}
		\centering
		\includegraphics[width=\linewidth]{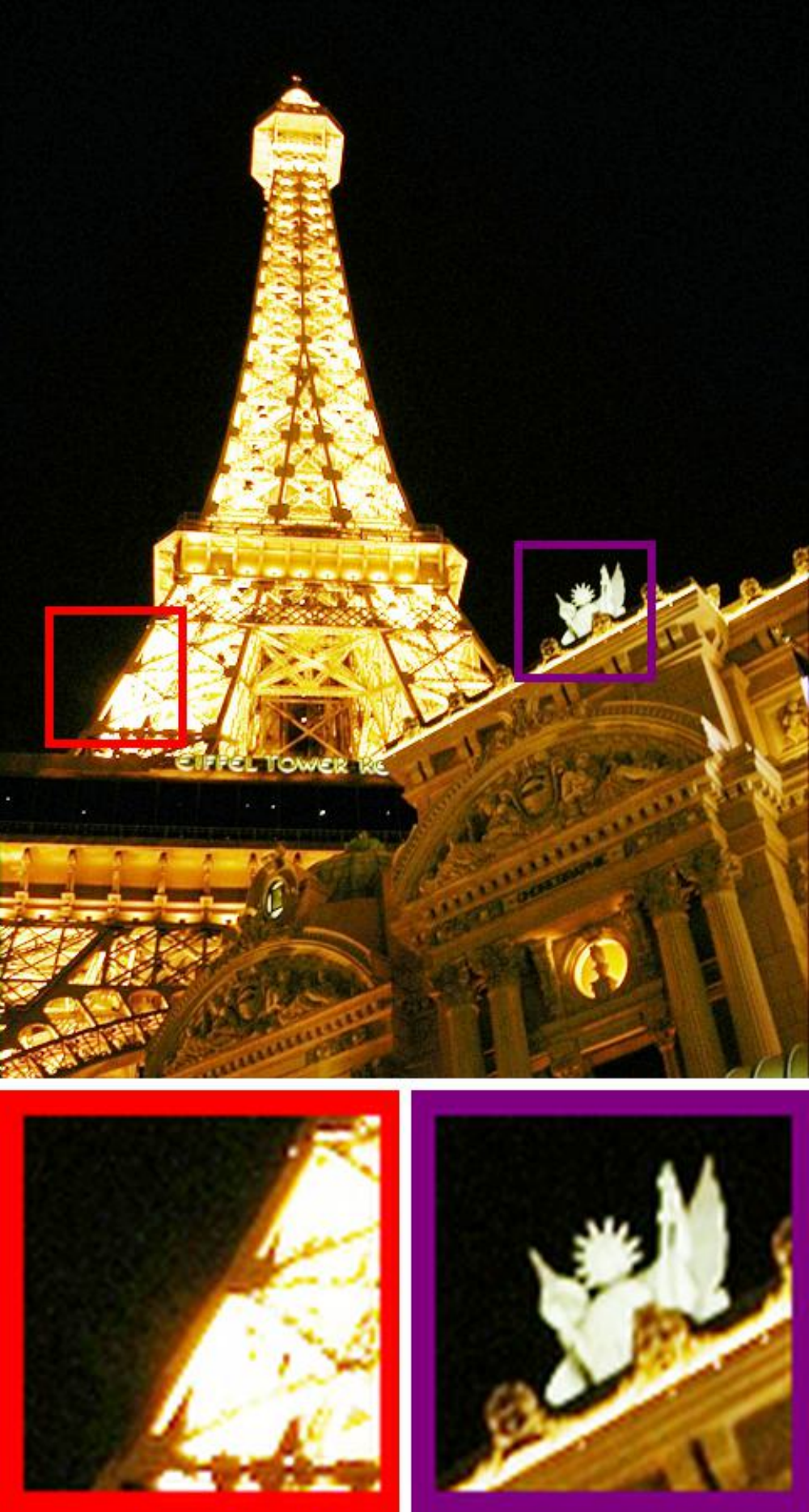} 
        \caption{\footnotesize SCI}
	\end{subfigure}
	\begin{subfigure}{0.12\linewidth}
		\centering
		\includegraphics[width=\linewidth]{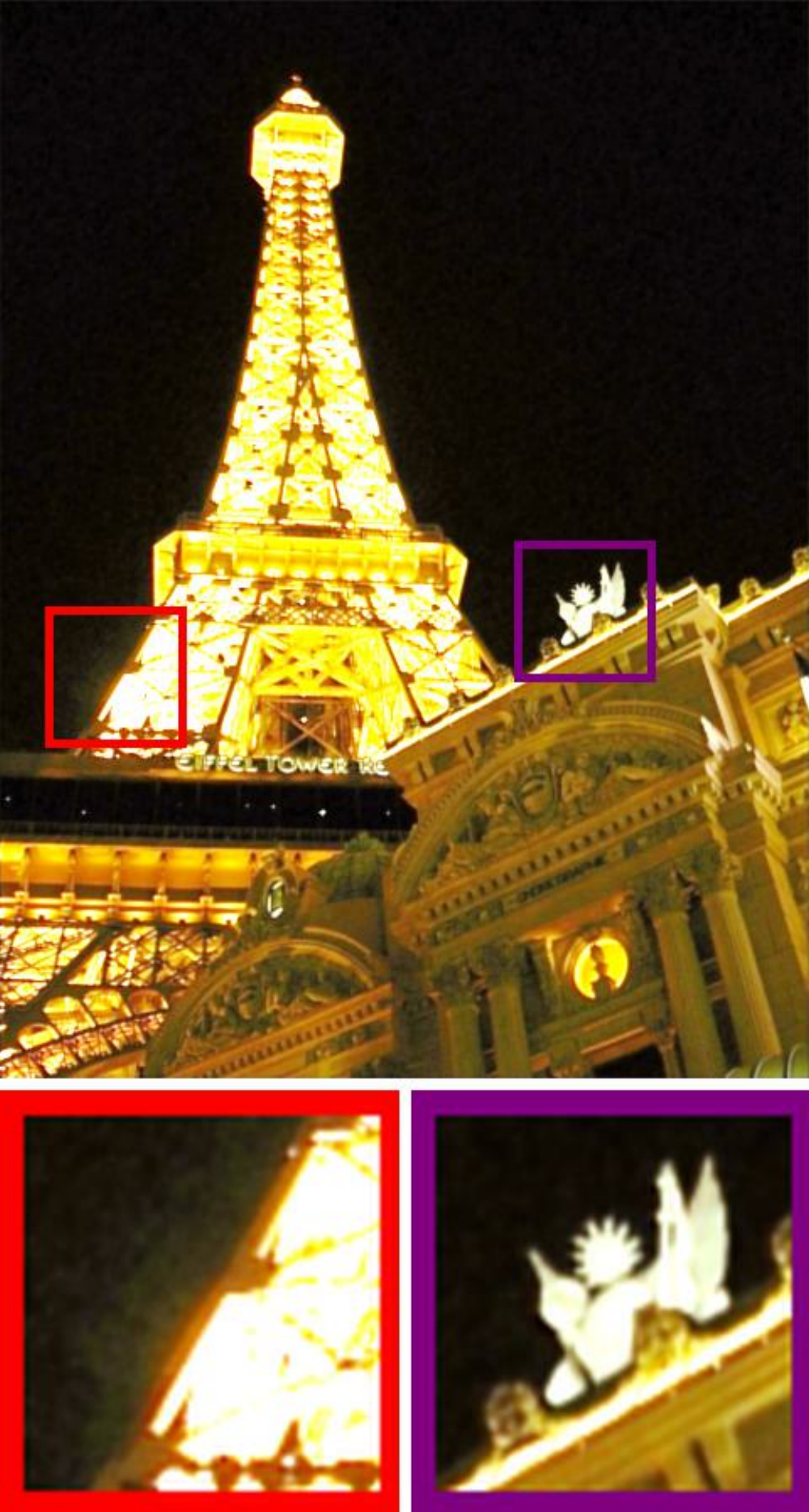} 
        \caption{\footnotesize IAT }
	\end{subfigure}
	\begin{subfigure}{0.12\linewidth}
		\centering
		\includegraphics[width=\linewidth]{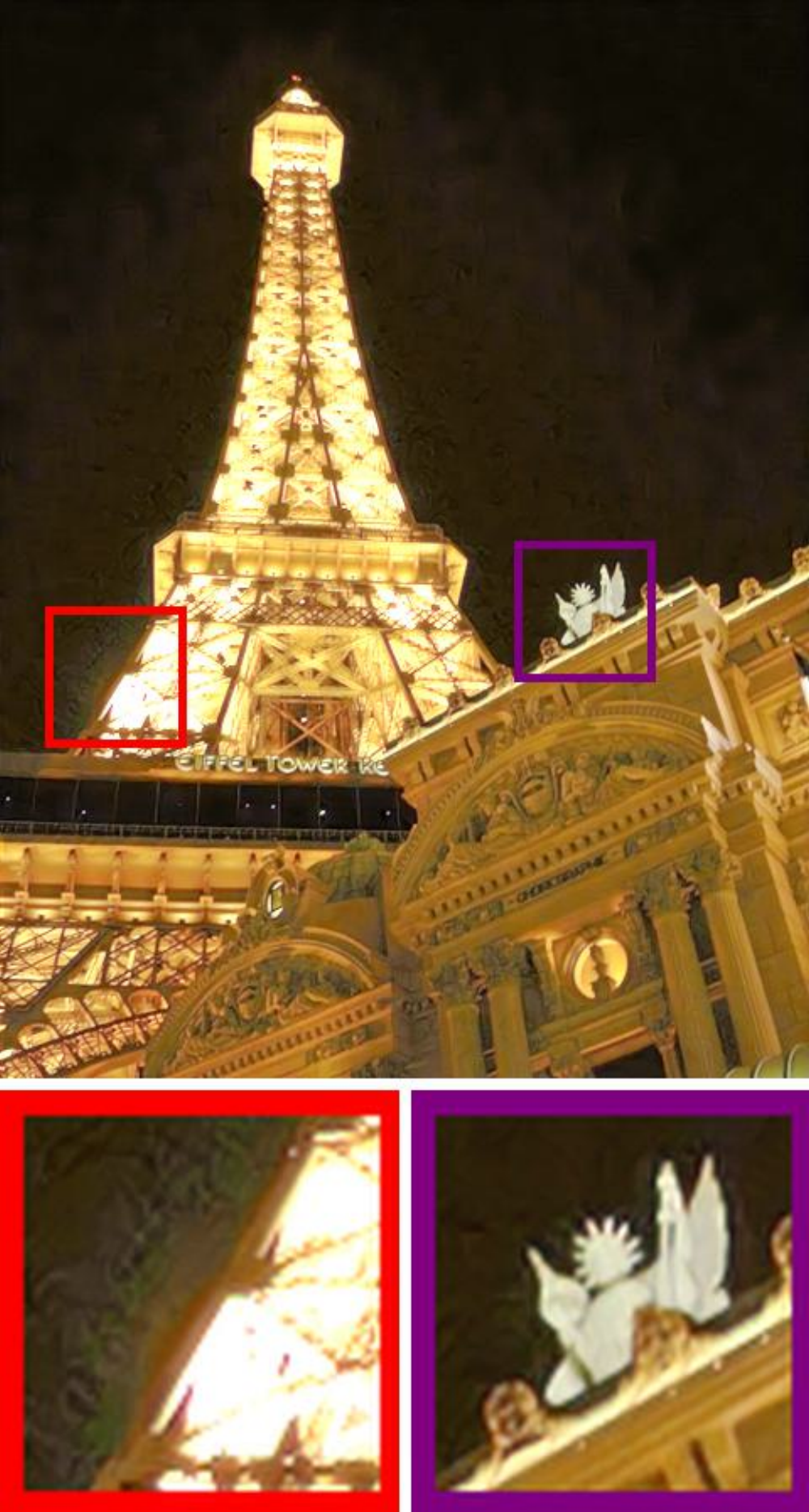}
        \caption{\footnotesize URetinex }
	\end{subfigure}
	\begin{subfigure}{0.12\linewidth}
		\centering
		\includegraphics[width=\linewidth]{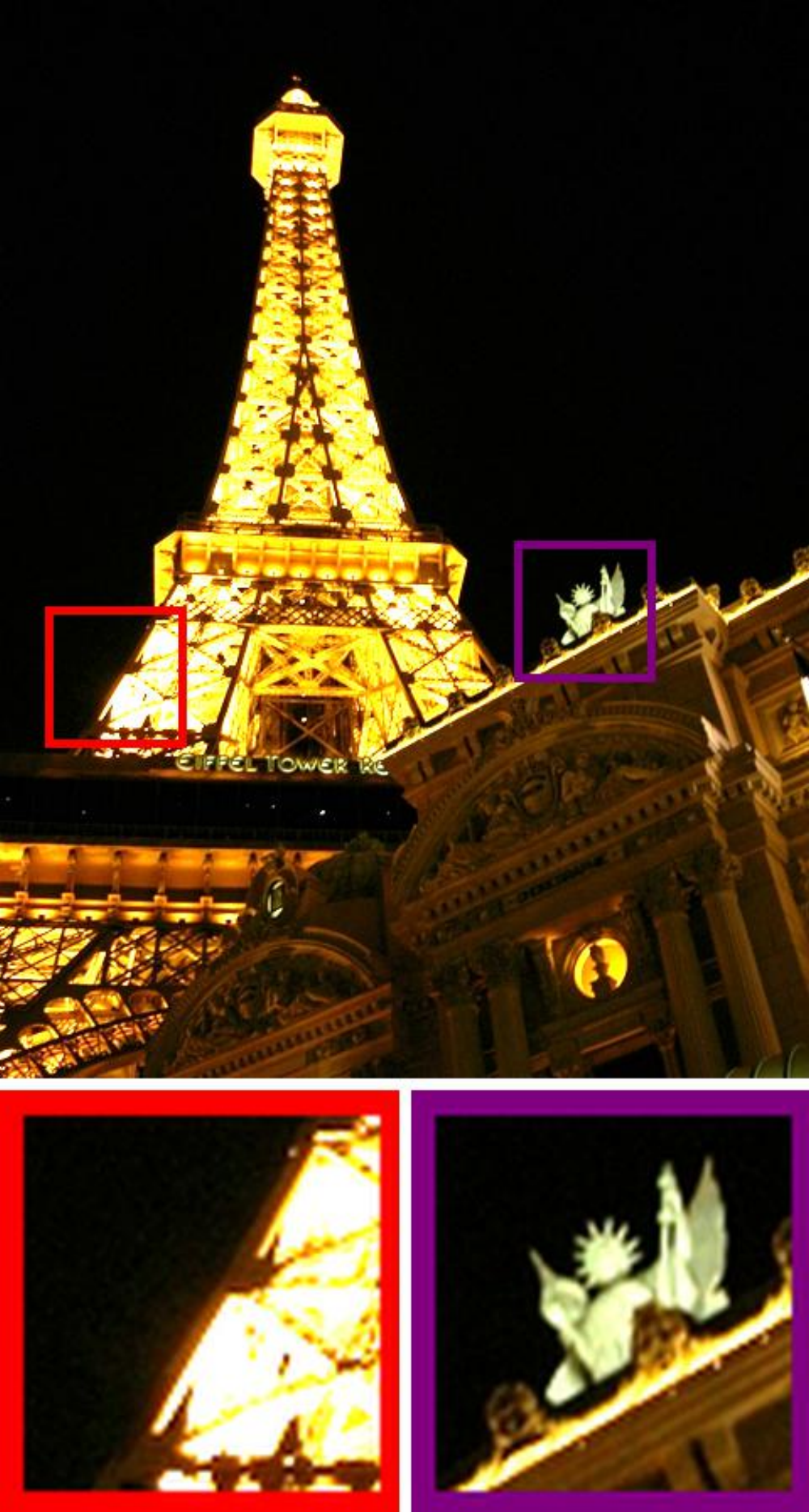} 
		\caption{\footnotesize BL }
	\end{subfigure}
	\begin{subfigure}{0.12\linewidth}
		\centering
		\includegraphics[width=\linewidth]{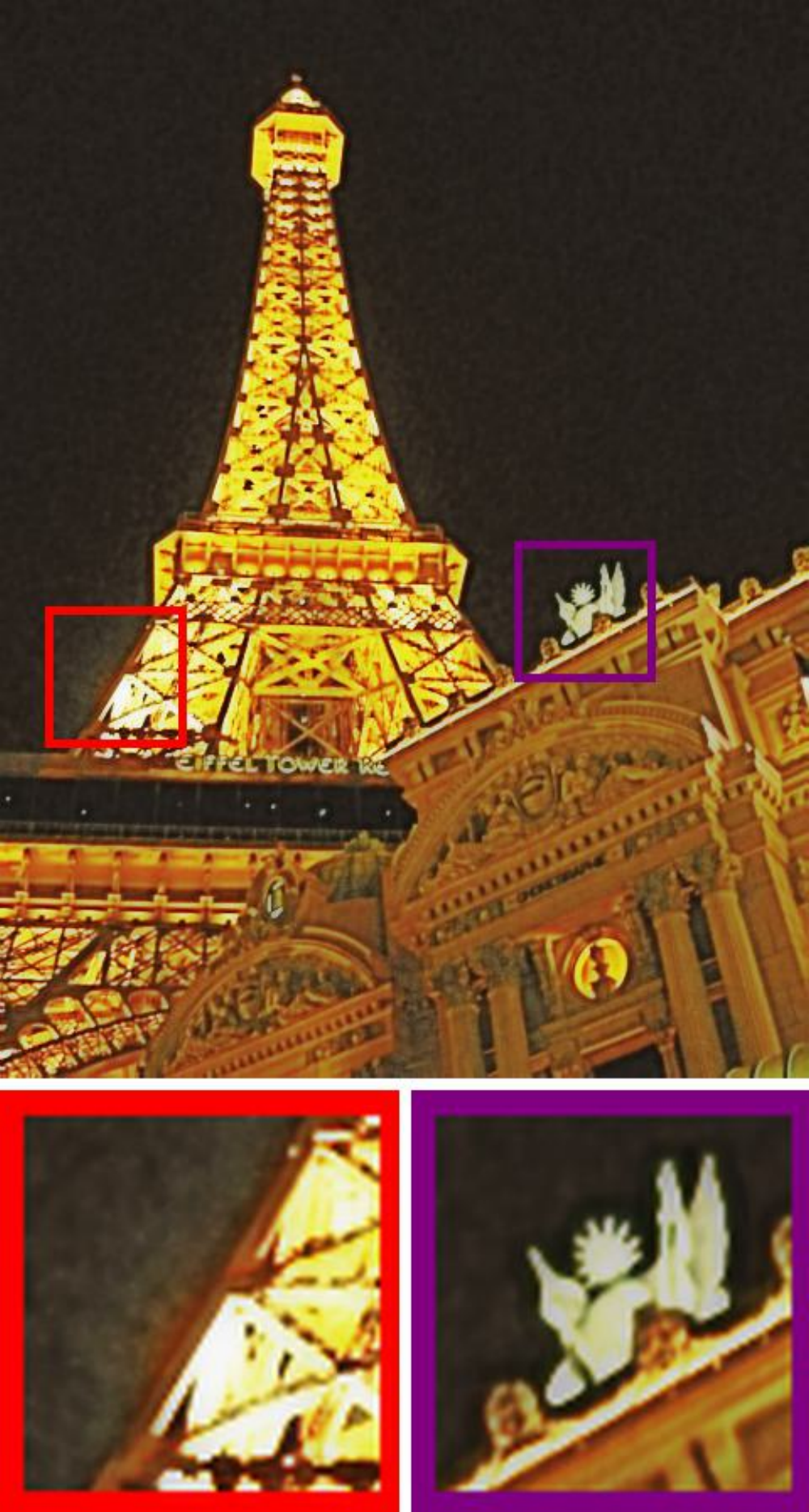} 
		\caption{\footnotesize PairLIE }
	\end{subfigure}
	\begin{subfigure}{0.12\linewidth}
		\centering
		\includegraphics[width=\linewidth]{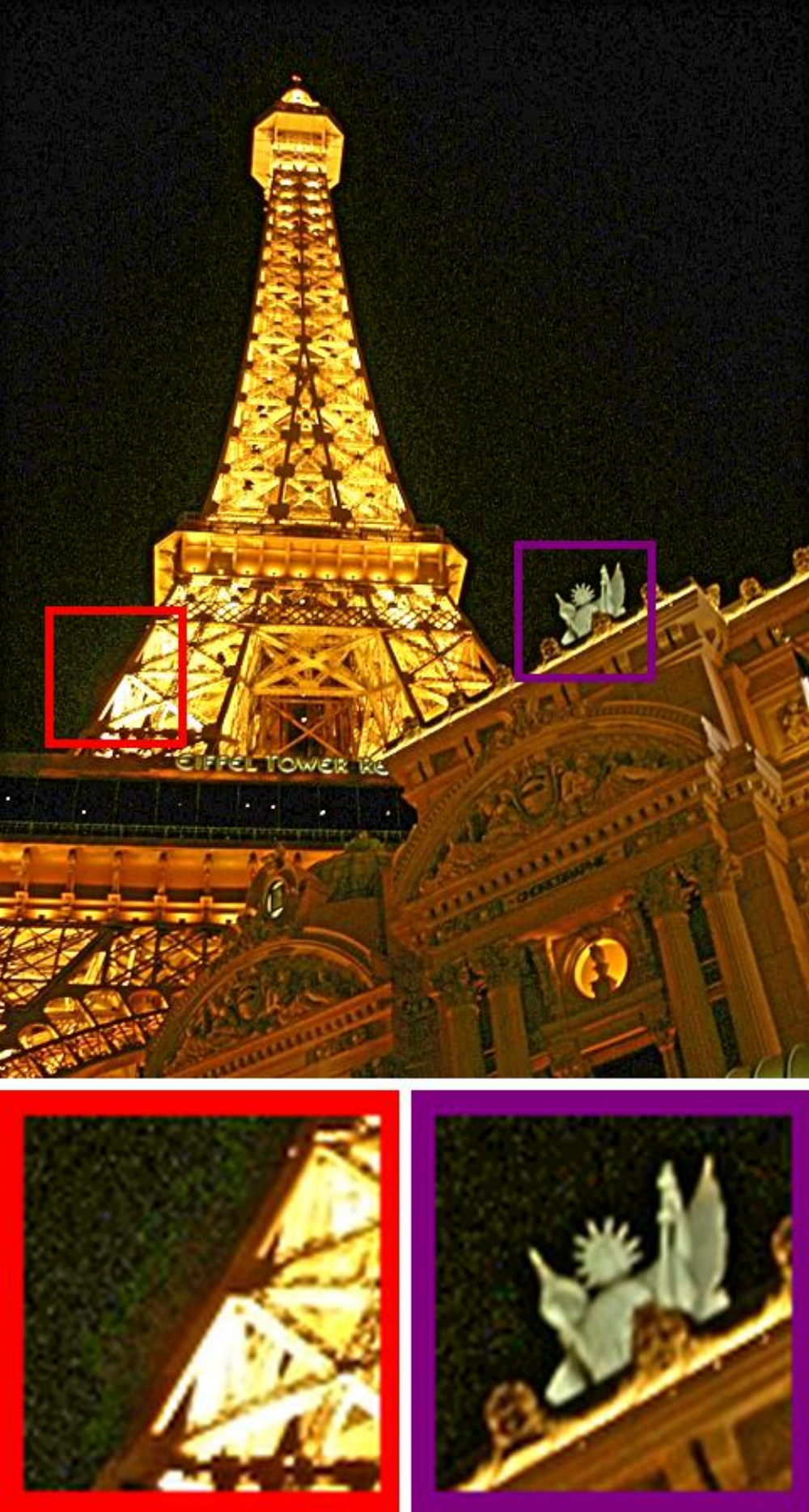}
		\caption{\footnotesize CLIP-LIT }
	\end{subfigure}
    \begin{subfigure}{0.12\linewidth}
		\centering
		\includegraphics[width=\linewidth]{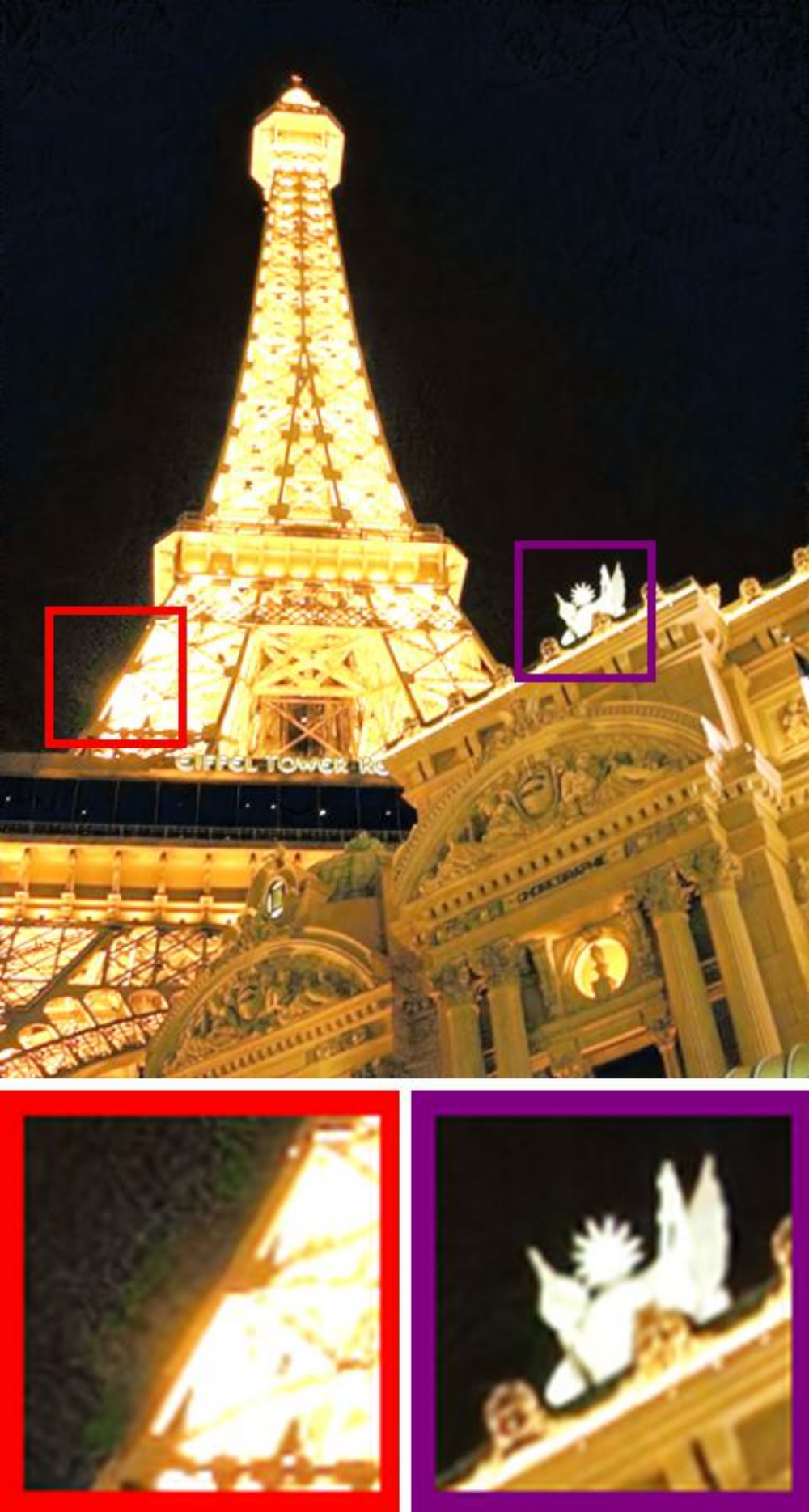} 
		\caption{\footnotesize EMNet }
	\end{subfigure}
	\begin{subfigure}{0.12\linewidth}
		\centering
		\includegraphics[width=\linewidth]{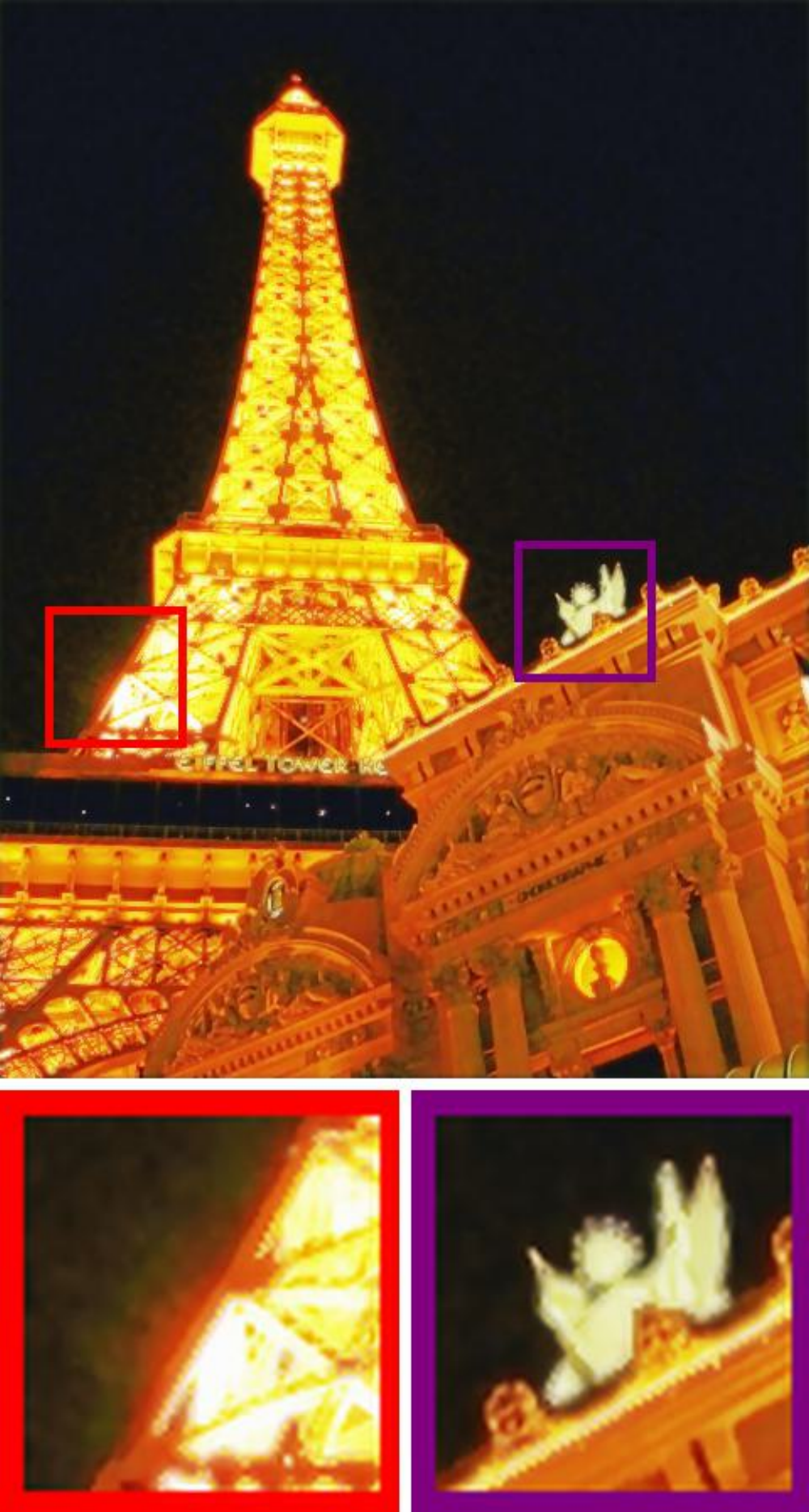} 
		\caption{\footnotesize SHAL-Net }
	\end{subfigure}


    \begin{subfigure}{0.12\linewidth}
		\centering
		\includegraphics[width=\linewidth]{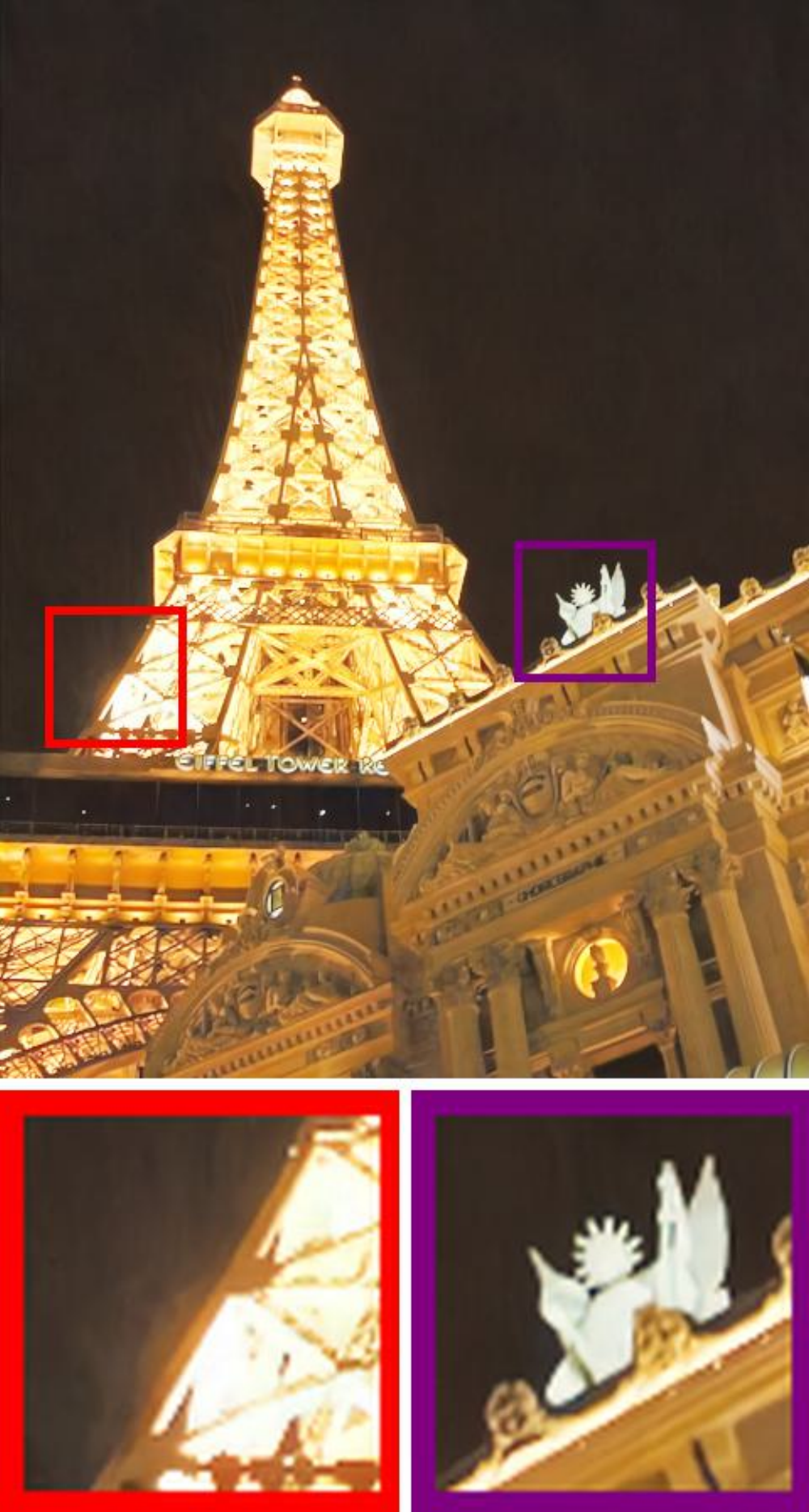} 
        \caption{\footnotesize GSAD}
	\end{subfigure}
	\begin{subfigure}{0.12\linewidth}
		\centering
		\includegraphics[width=\linewidth]{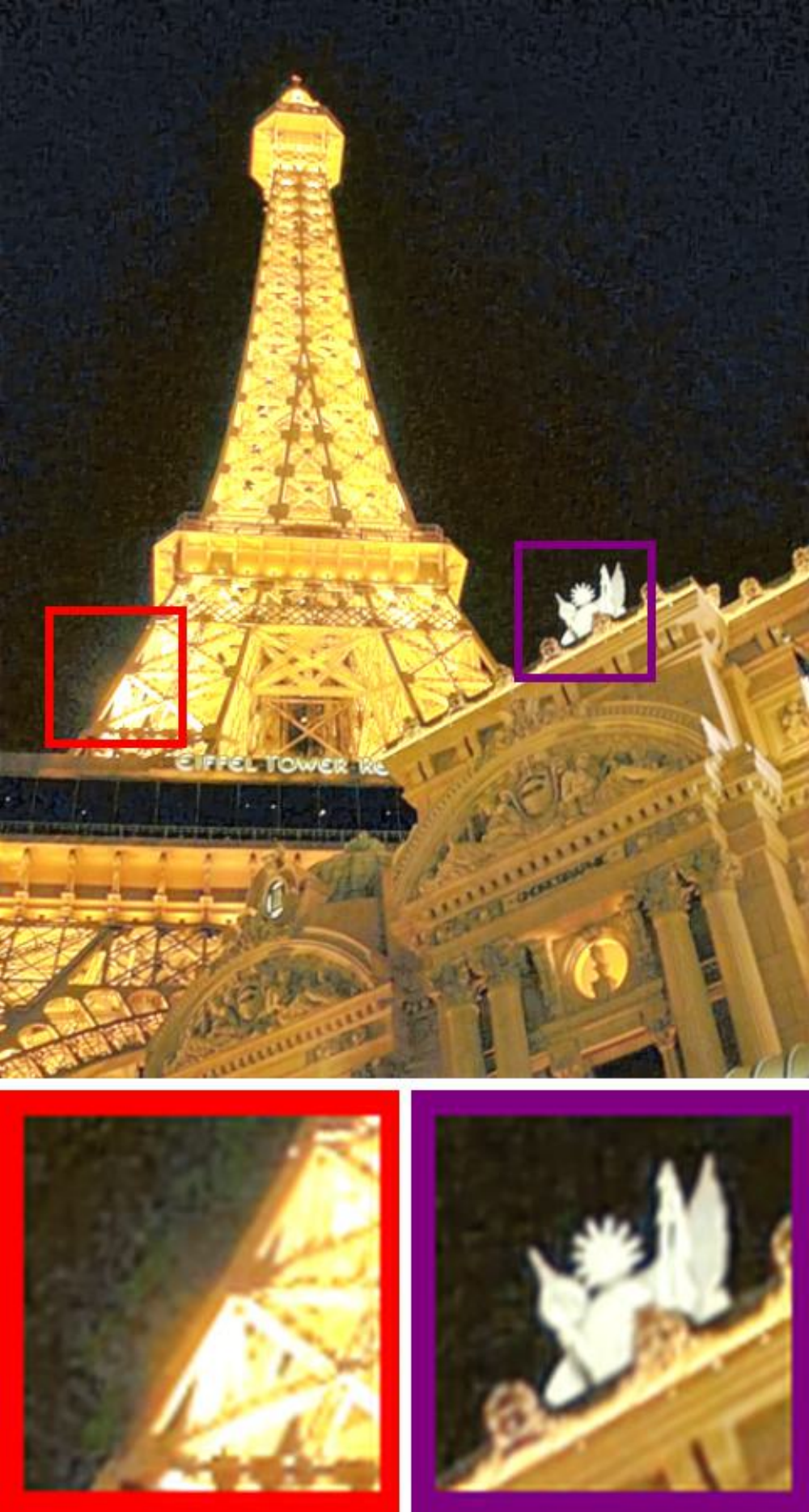} 
        \caption{\footnotesize PPformer }
	\end{subfigure}
	\begin{subfigure}{0.12\linewidth}
		\centering
		\includegraphics[width=\linewidth]{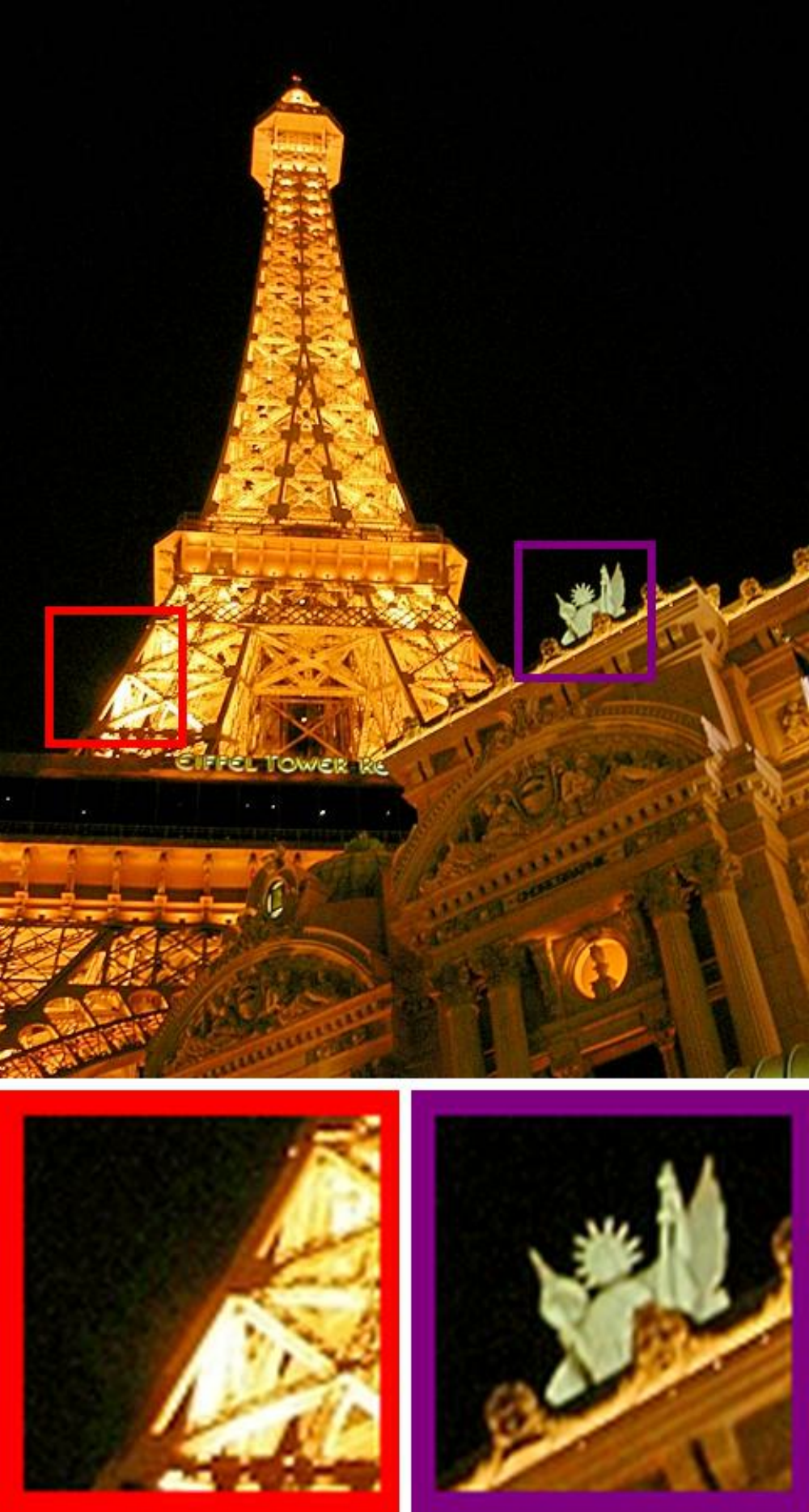}
        \caption{\footnotesize GCP }
	\end{subfigure}
	\begin{subfigure}{0.12\linewidth}
		\centering
		\includegraphics[width=\linewidth]{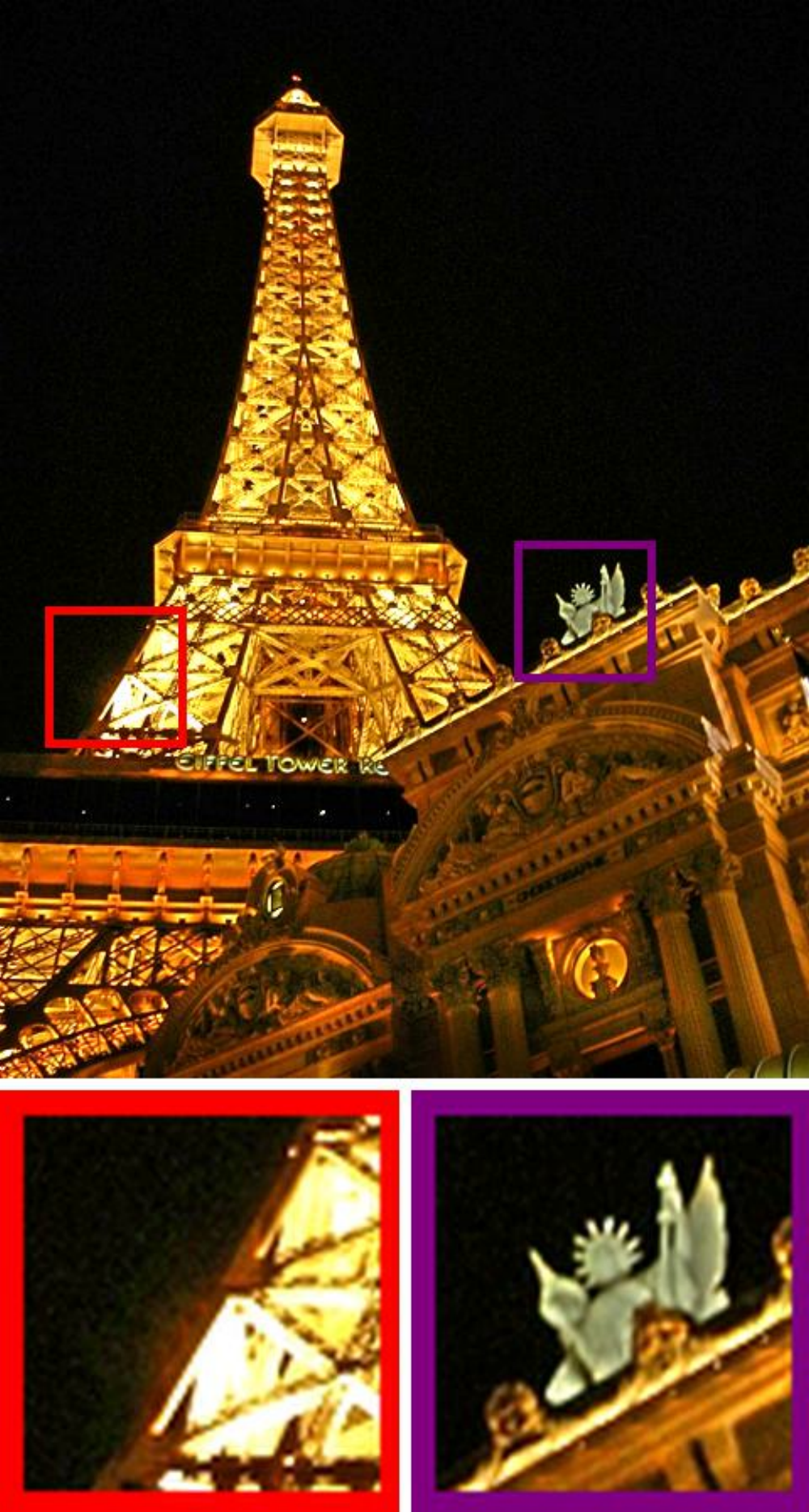} 
		\caption{\footnotesize NeurBR }
	\end{subfigure}
	\begin{subfigure}{0.12\linewidth}
		\centering
		\includegraphics[width=\linewidth]{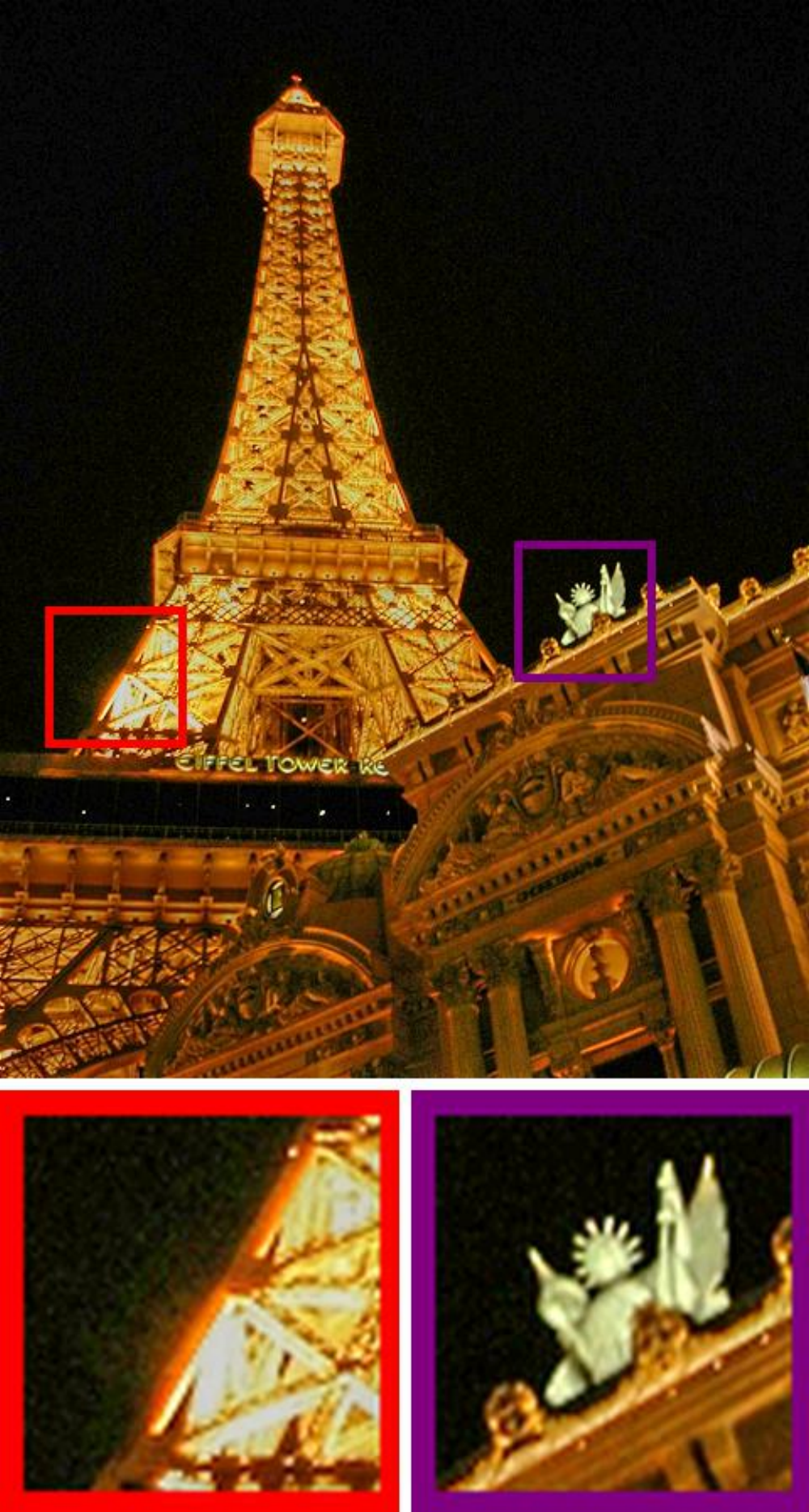} 
		\caption{\footnotesize ITRE }
	\end{subfigure}
	\begin{subfigure}{0.12\linewidth}
		\centering
		\includegraphics[width=\linewidth]{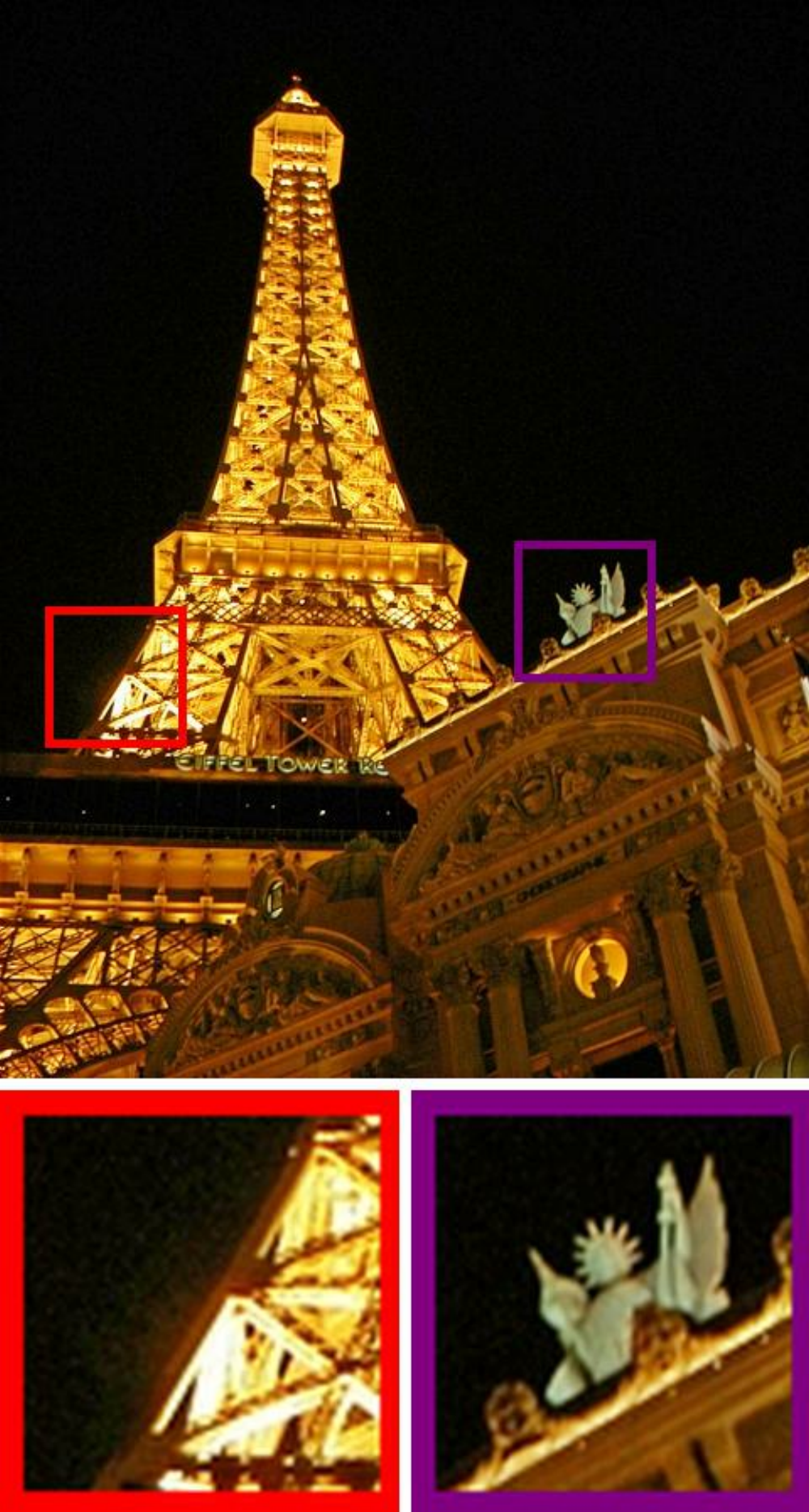}
		\caption{\footnotesize PIE }
	\end{subfigure}
    \begin{subfigure}{0.12\linewidth}
		\centering
		\includegraphics[width=\linewidth]{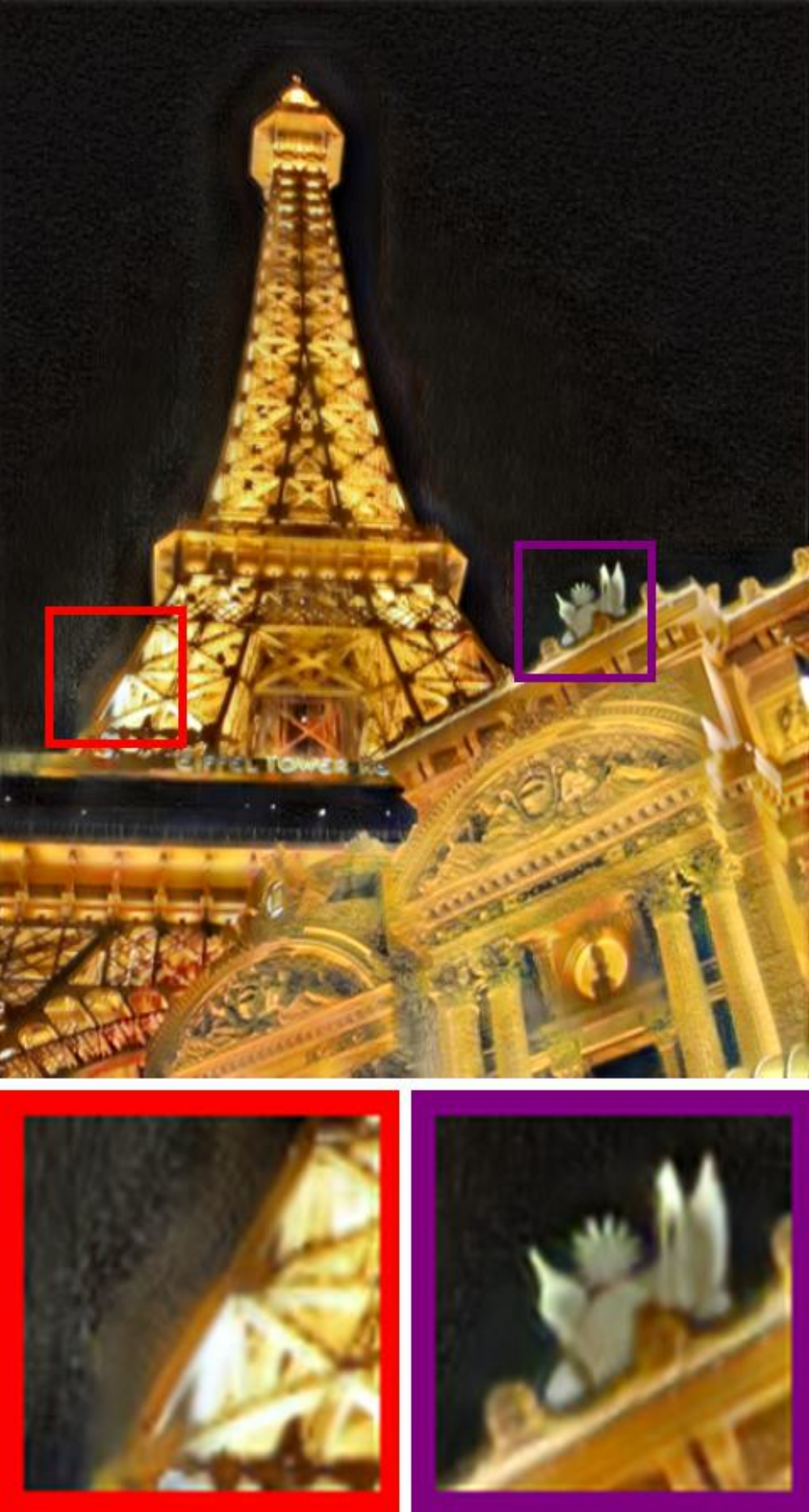} 
		\caption{\footnotesize MSATr }
	\end{subfigure}
	\begin{subfigure}{0.12\linewidth}
		\centering
		\includegraphics[width=\linewidth]{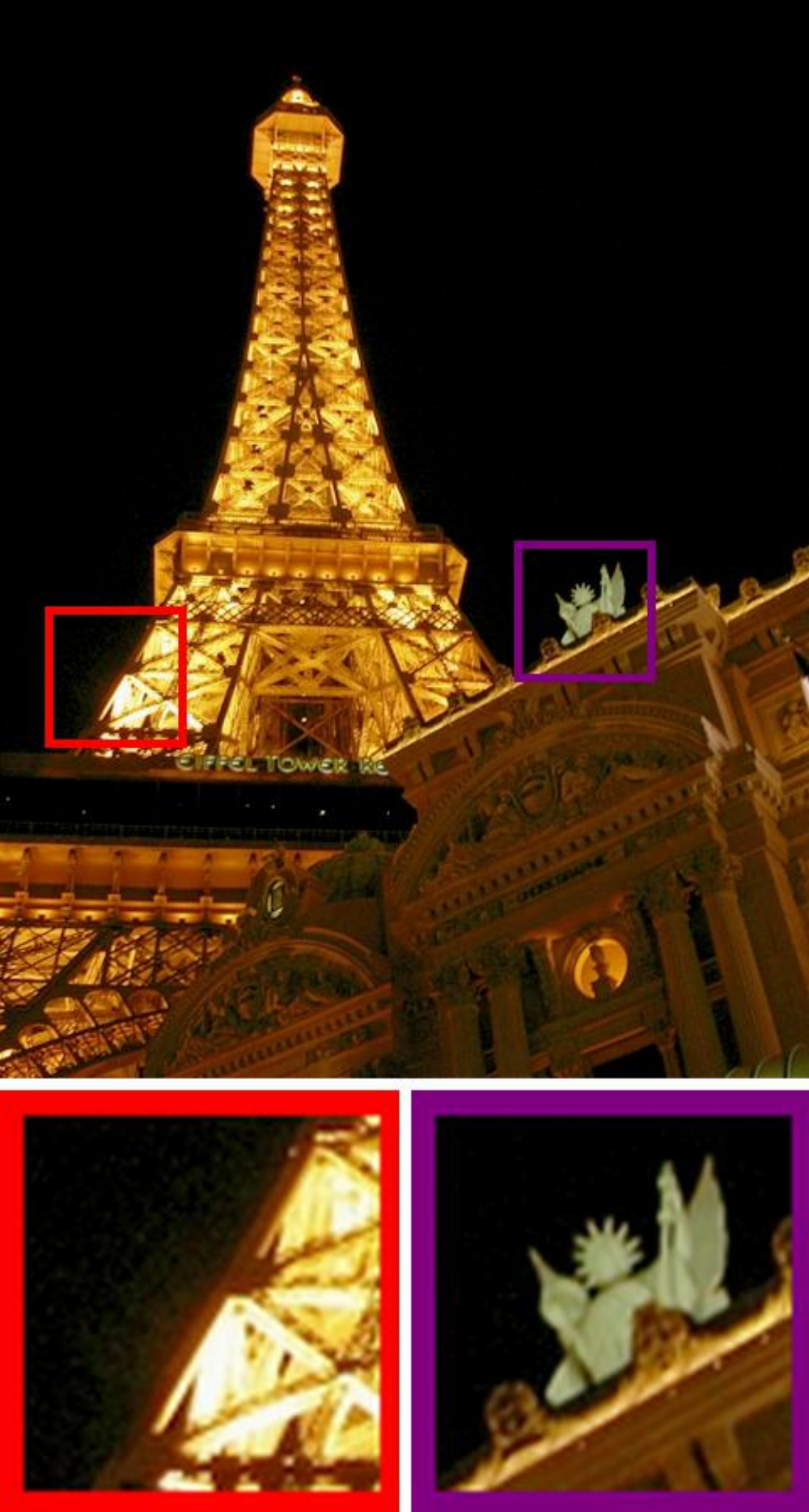} 
		\caption{\footnotesize Proposed }
	\end{subfigure}

	\caption{Visual comparison of methods for enhancing low-light images on DICM dataset.}
	\label{Q1}
\end{figure*}

\section{Results and Discussions}
\label{RD}

To assess the performance of the proposed method ALEN, it was compared against a series of state-of-the-art methods, including LIME (Guo et al., 2016)~\cite{guo2016lime}, RetinexNet (Wei et al., 2018)~\cite{wei2018deep}, Zero-DCE (Guo et al., 2020)~\cite{guo2020zero}, RUAS (Liu et al., 2021)~\cite{Liu21}, UTVNet (Zheng et al., 2021)~\cite{zheng2021adaptive}, HWMNet (Fan et al., 2022)~\cite{fan2022half}, LLFlow (Wang et al., 2022)~\cite{wang2022lowlight}, SCI (Ma et al., 2022)~\cite{ma2022toward}, IAT (Cui et al., 2022)~\cite{cui2022you}, URetinex (Wu et al., 2022)~\cite{wu2022uretinex}, BL (Ma et al., 2023)~\cite{ma2023bilevel}, PairLIE (Fu et al., 2023)~\cite{fu2023learning}, CLIP-LIT (Liang et al., 2023)~\cite{liang2023iterative}, EMNet (Ye et al., 2023)~\cite{ye2023glow}, SHAL-Net (Jiang et al., 2024)~\cite{xu2024degraded}, GSAD (Hou et al., 2024)~\cite{hou2024global}, PPformer (Dang et al., 2024)~\cite{dang2024ppformer}, GCP (Jeon et al., 2024)~\cite{jeon2024low}, NeurBR (Zhao et al., 2024)~\cite{zhao2024non}, ITRE (Wang et al., 2024)~\cite{wang2024itre}, PIE (Liang et al., 2024)~\cite{liang2024pie} and MSATr(Fang et al.,2024)~\cite{fang2024non}. The selected methods, publicly available for research purposes, were evaluated using their official pre-trained models provided by the authors, along with the default configurations specified for each method.

\subsection{Quantitative results}

To evaluate the efficiency of the proposed ALEN method alongside other state-of-the-art methods, Full-Reference IQA (FIQA) and No-Reference IQA (NIQA) metrics were employed. FIQA metrics require reference images for measurements, including Peak Signal-to-Noise Ratio (PSNR)~\cite{wang2004image}, Structural Similarity Index Measure (SSIM)~\cite{wang2004image}, Universal Quality Index (UQI)~\cite{wang2002universal}, Learned Perceptual Image Patch Similarity (LPIPS)~\cite{zhang2018unreasonable}, and DeltaE~\cite{xu2024degraded}. In contrast, NIQA metrics such as Naturalness Image Quality Evaluator (NIQE)~\cite{mittal2012making}, MUlti-Scale Image Quality Transformer (MUSIQ)~\cite{ke2021musiq} and Lightness Order Error (LOE)~\cite{wang2013naturalness} do not rely on a reference image.

\begin{figure*}[ht]
	\centering
	\begin{subfigure}{0.12\linewidth}
		\centering
		\includegraphics[width=\linewidth]{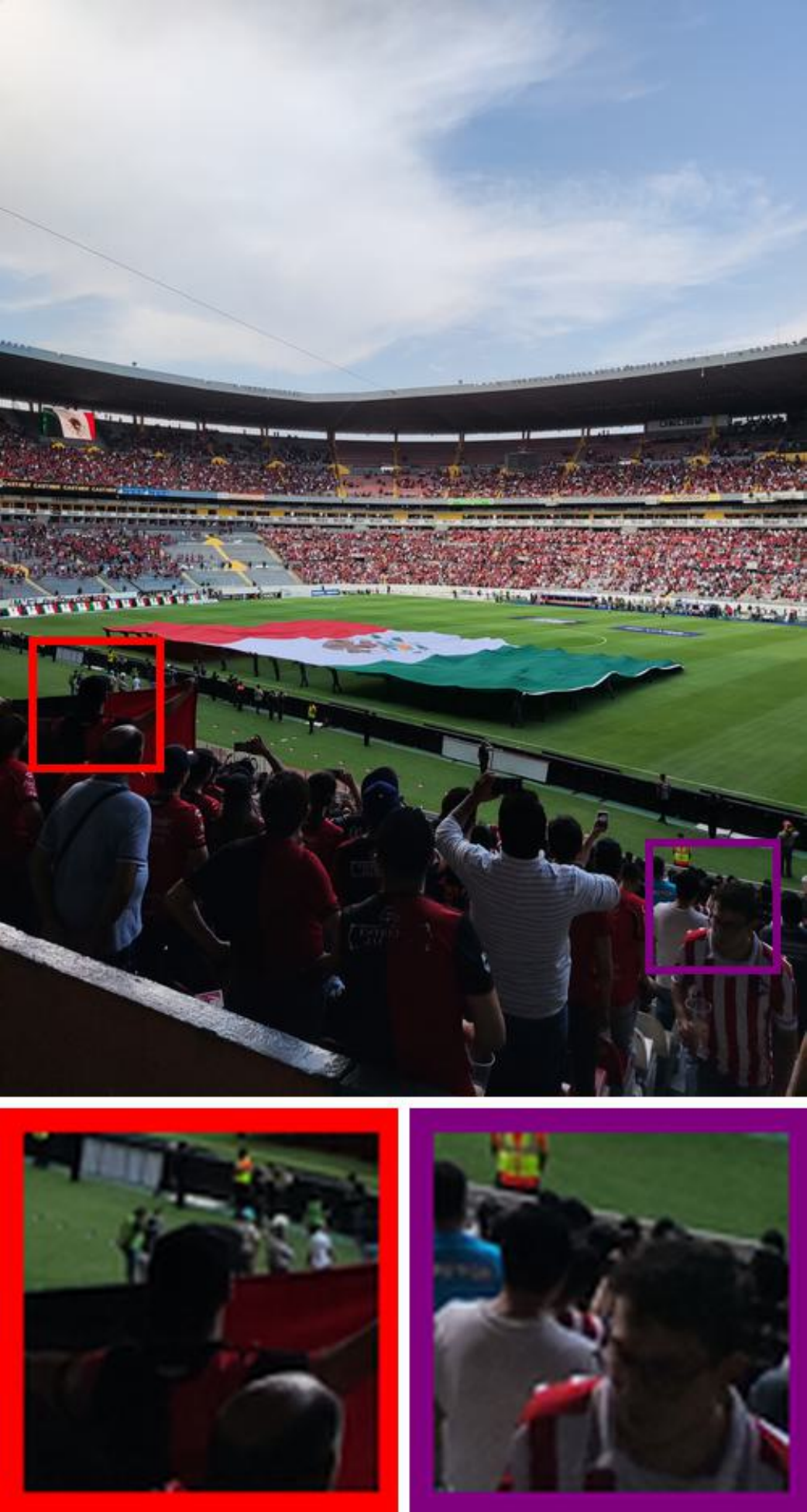} 
        \caption{\footnotesize Low-light}
	\end{subfigure}
	\begin{subfigure}{0.12\linewidth}
		\centering
		\includegraphics[width=\linewidth]{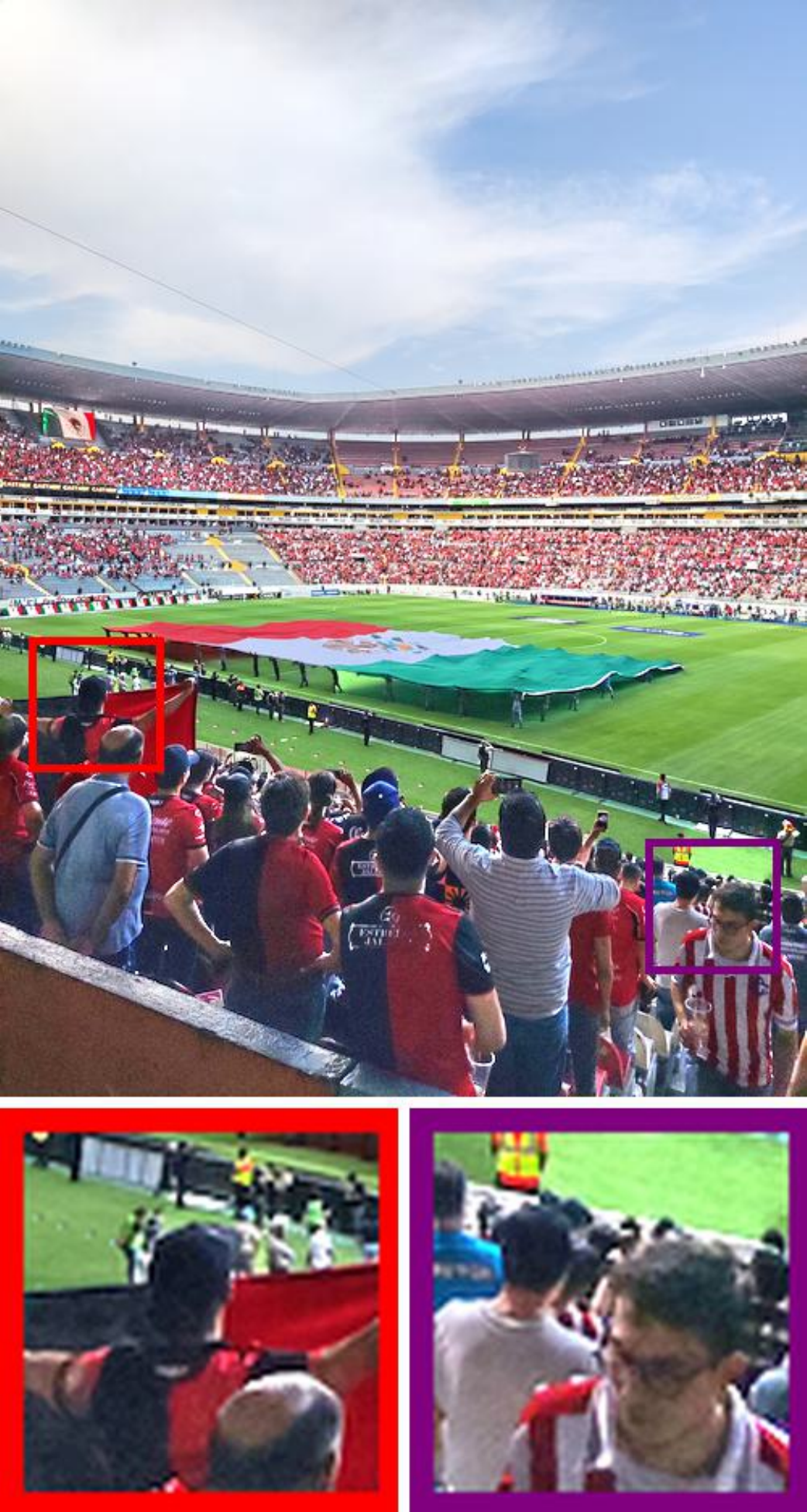} 
        \caption{\footnotesize LIME }
	\end{subfigure}
	\begin{subfigure}{0.12\linewidth}
		\centering
		\includegraphics[width=\linewidth]{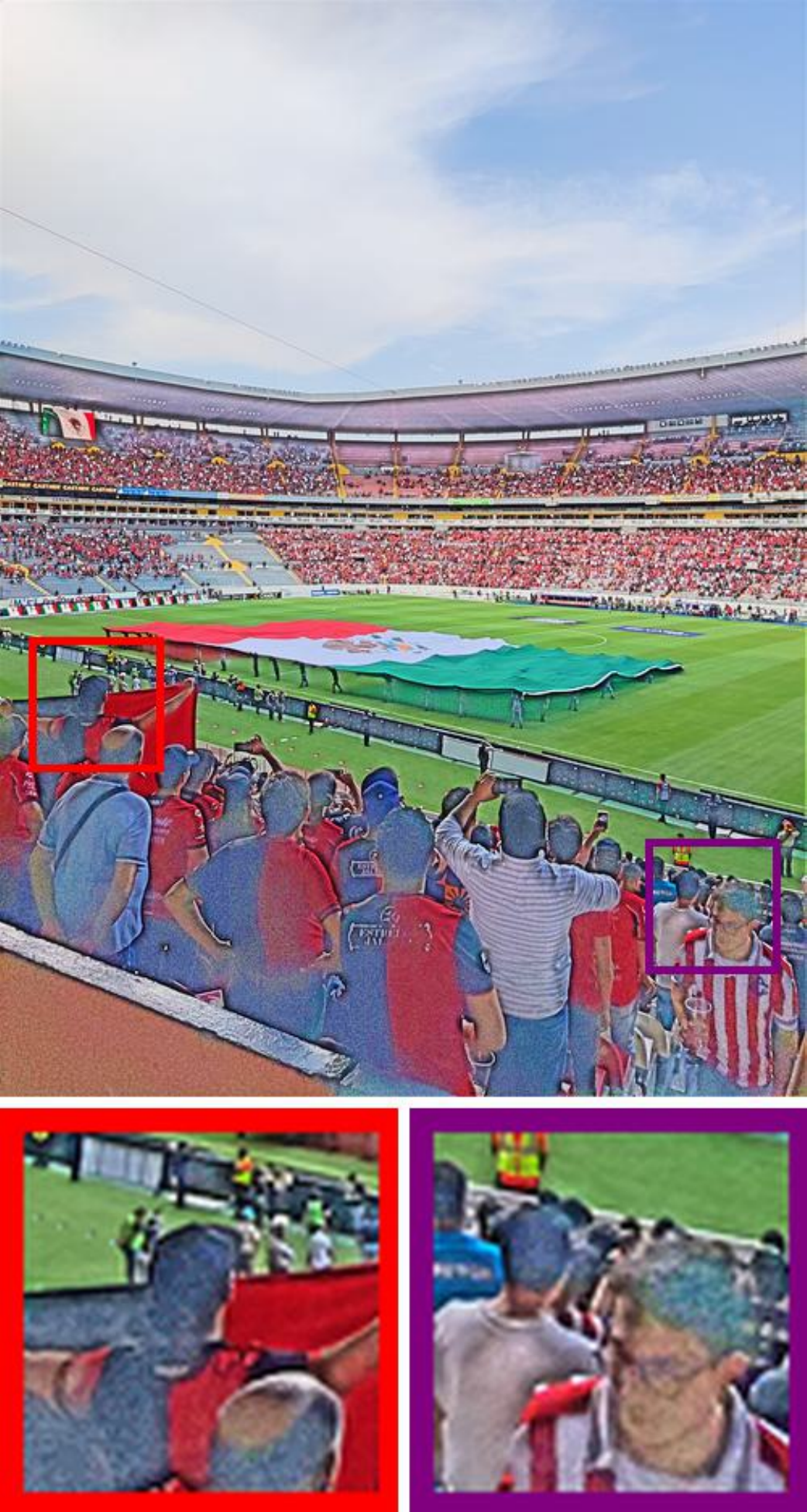}
        \caption{\footnotesize RetinexNet }
	\end{subfigure}
	\begin{subfigure}{0.12\linewidth}
		\centering
		\includegraphics[width=\linewidth]{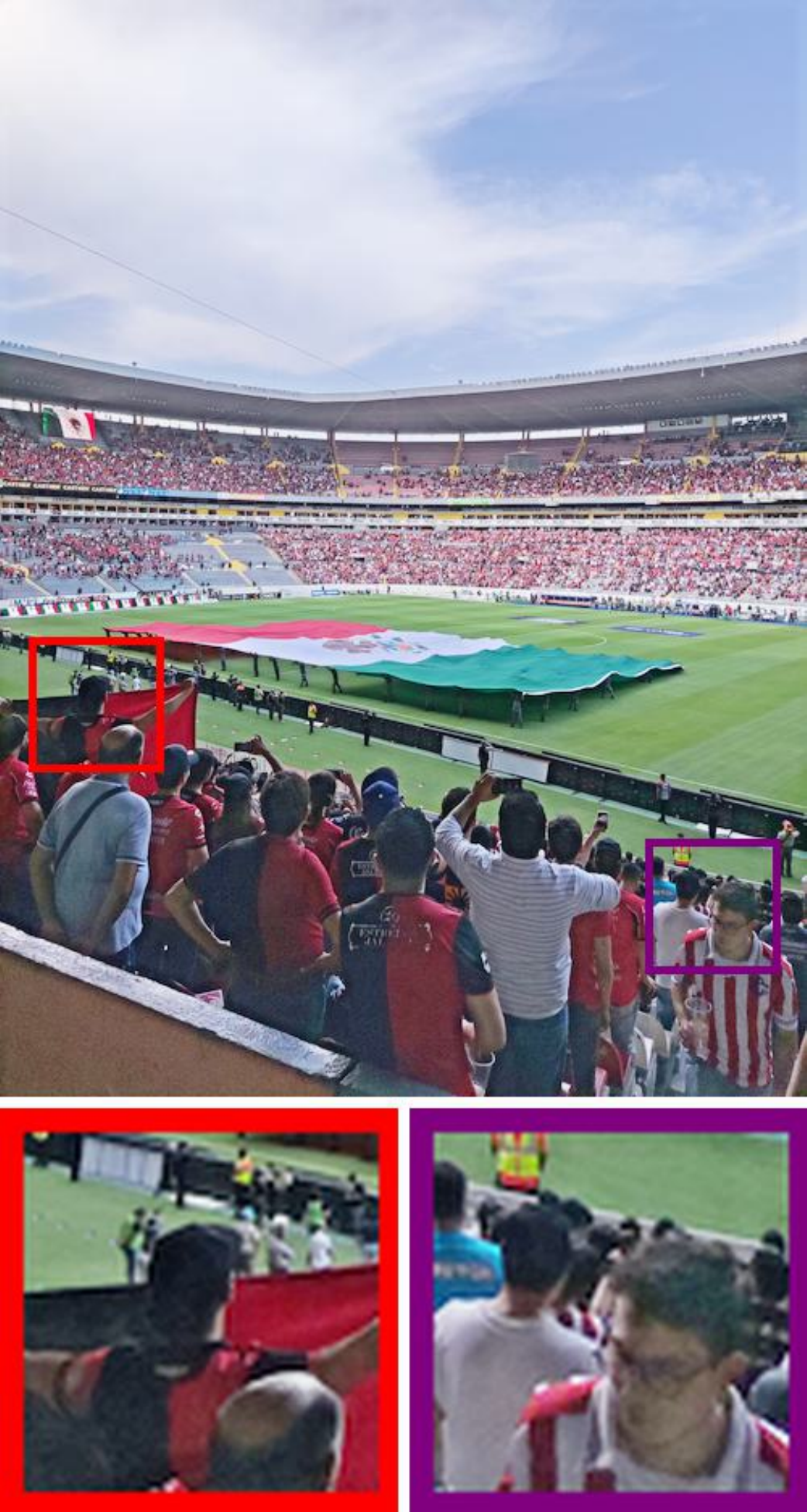} 
		\caption{\footnotesize Zero-DCE }
	\end{subfigure}
	\begin{subfigure}{0.12\linewidth}
		\centering
		\includegraphics[width=\linewidth]{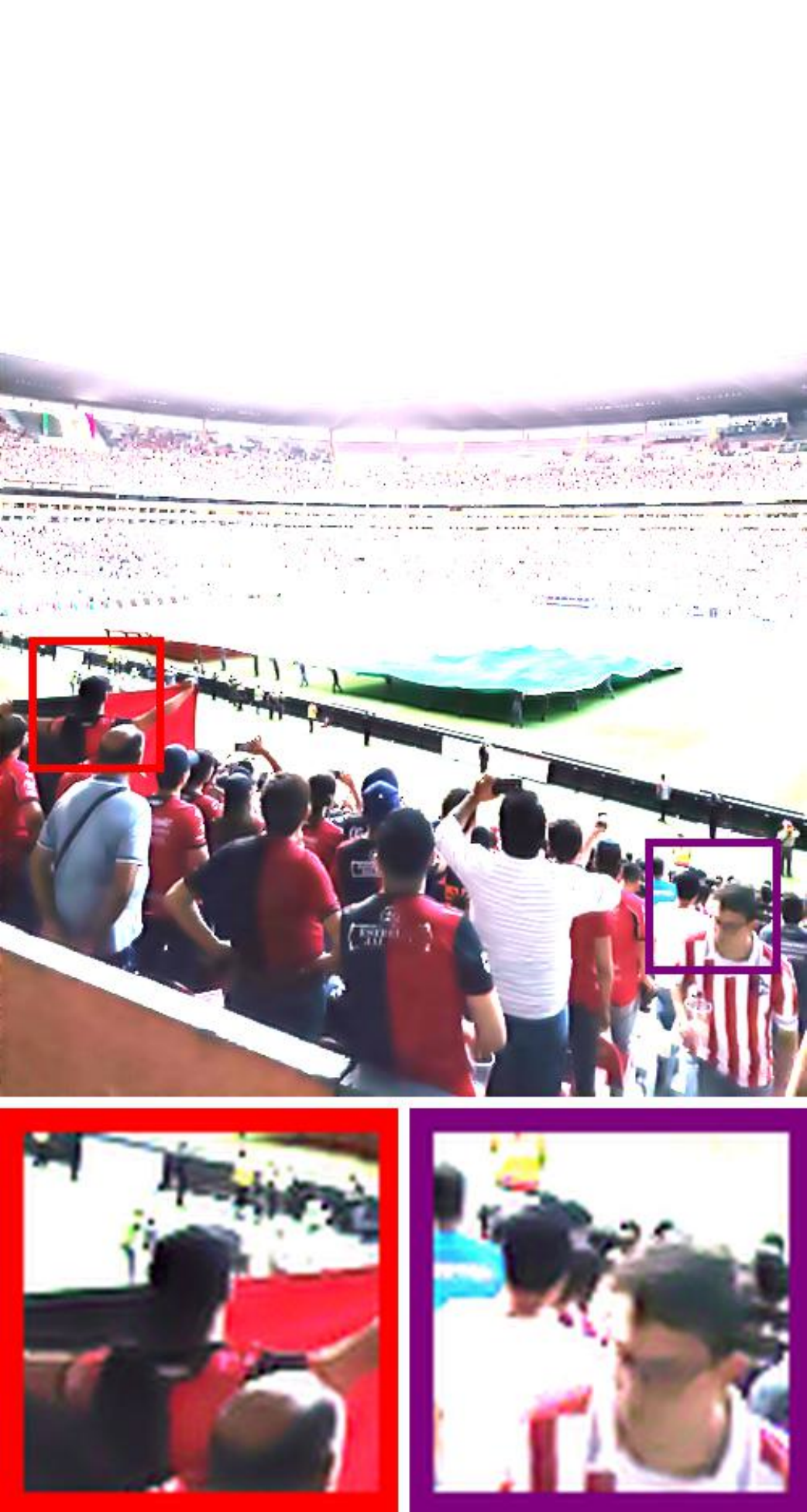} 
		\caption{\footnotesize RUAS }
	\end{subfigure}
	\begin{subfigure}{0.12\linewidth}
		\centering
		\includegraphics[width=\linewidth]{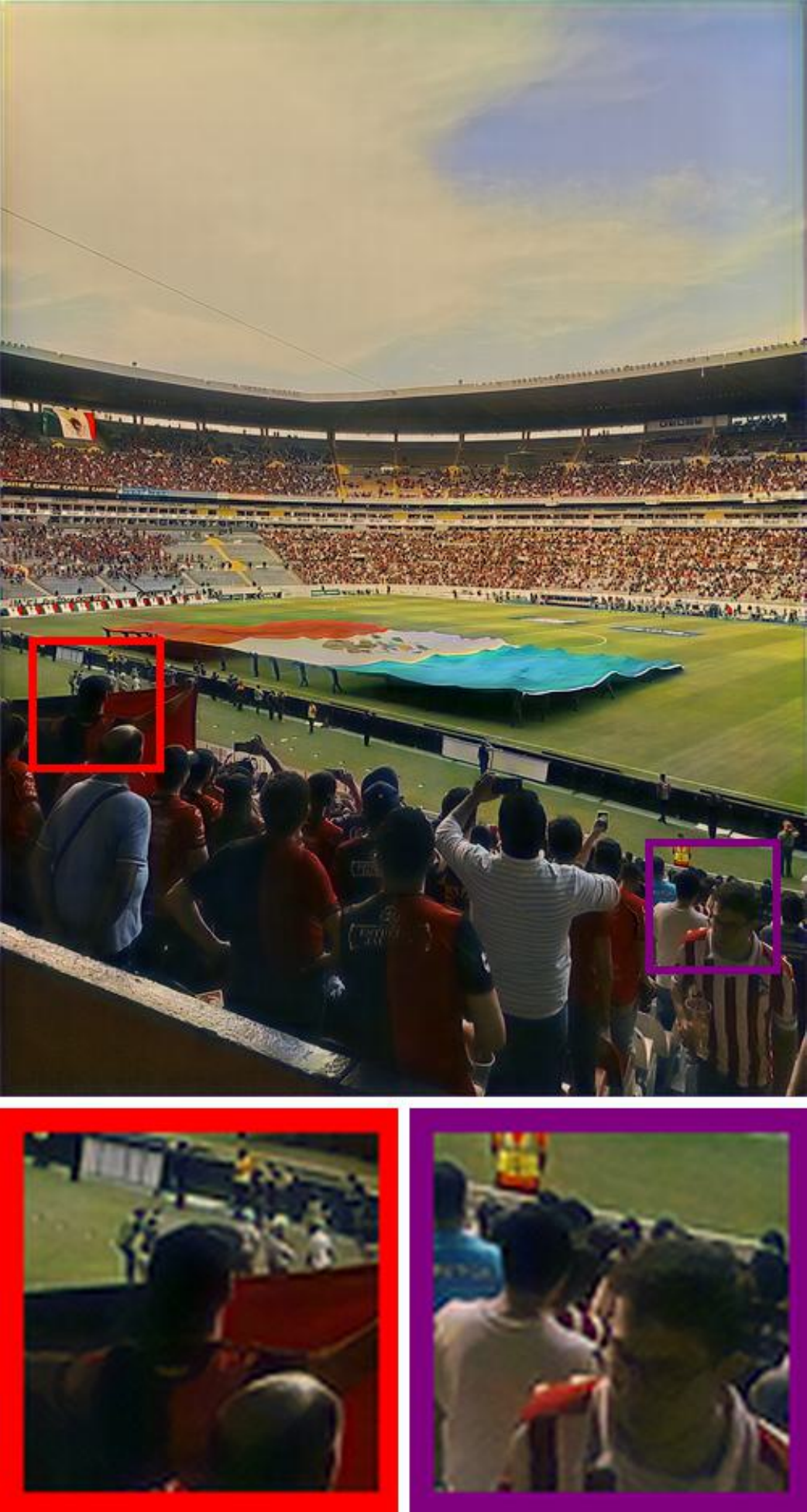}
		\caption{\footnotesize UTVNet }
	\end{subfigure}
    \begin{subfigure}{0.12\linewidth}
		\centering
		\includegraphics[width=\linewidth]{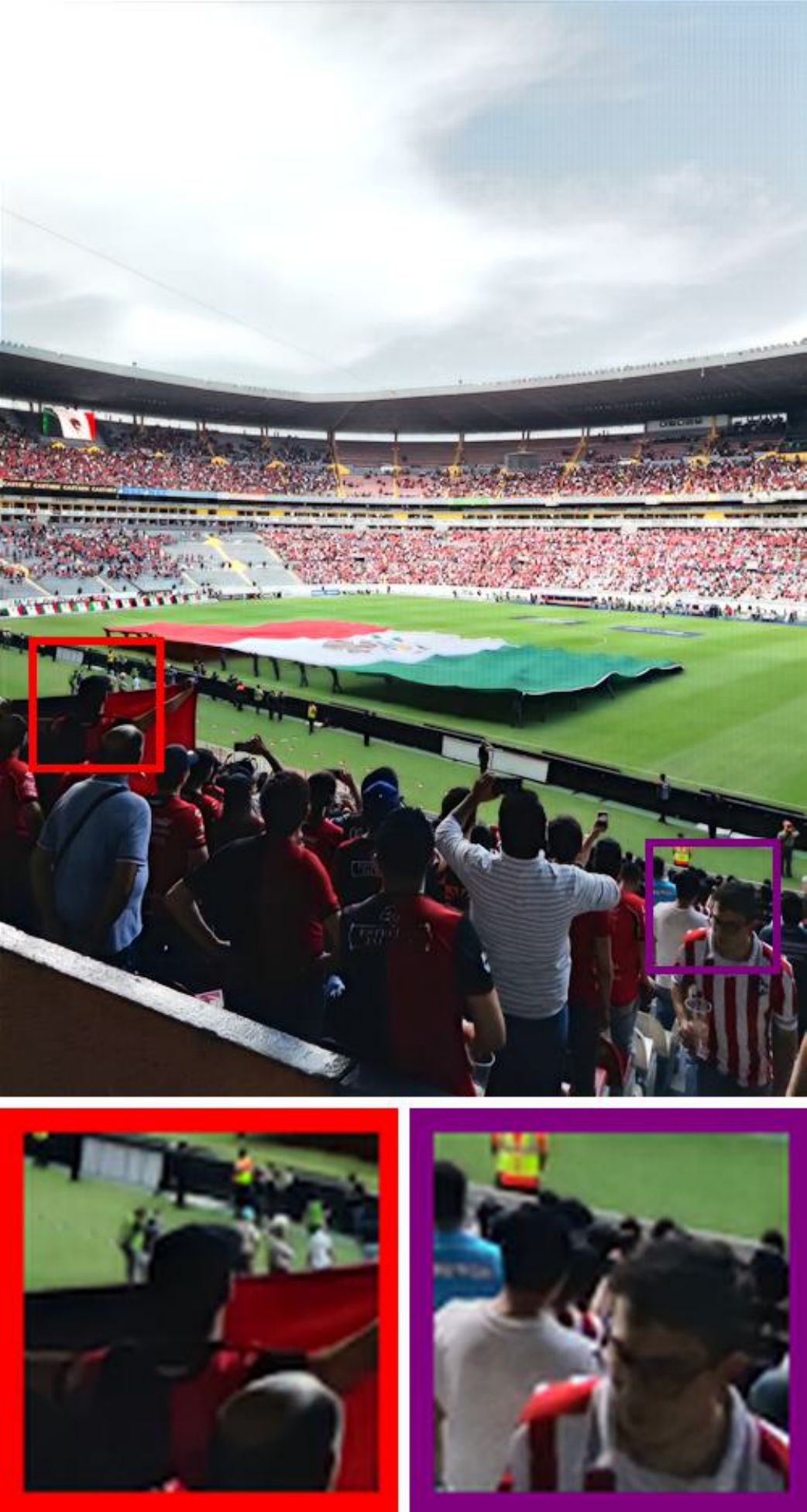} 
		\caption{\footnotesize HWMNet }
	\end{subfigure}
	\begin{subfigure}{0.12\linewidth}
		\centering
		\includegraphics[width=\linewidth]{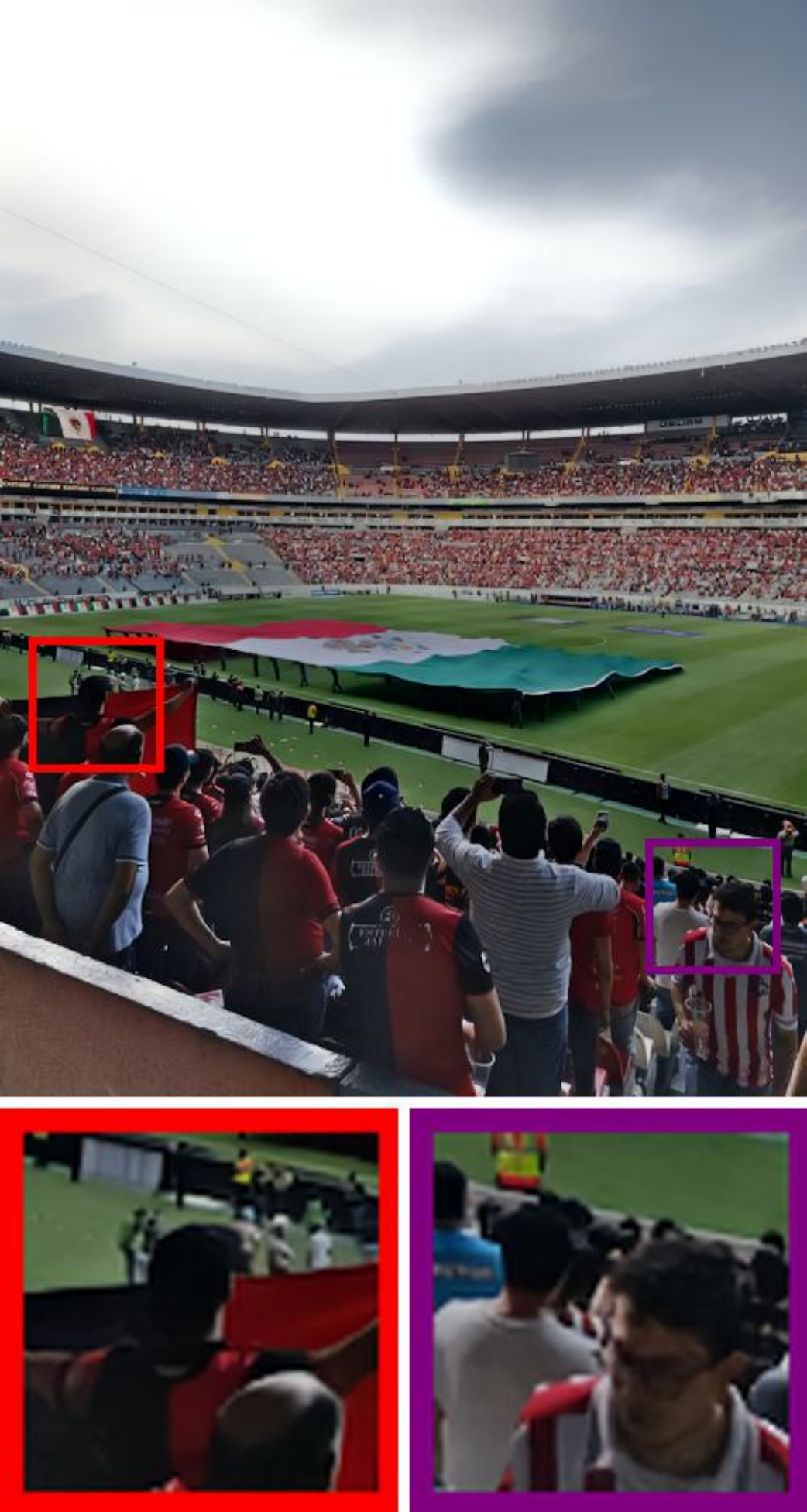} 
		\caption{\footnotesize LLFlow }
	\end{subfigure}


    \begin{subfigure}{0.12\linewidth}
		\centering
		\includegraphics[width=\linewidth]{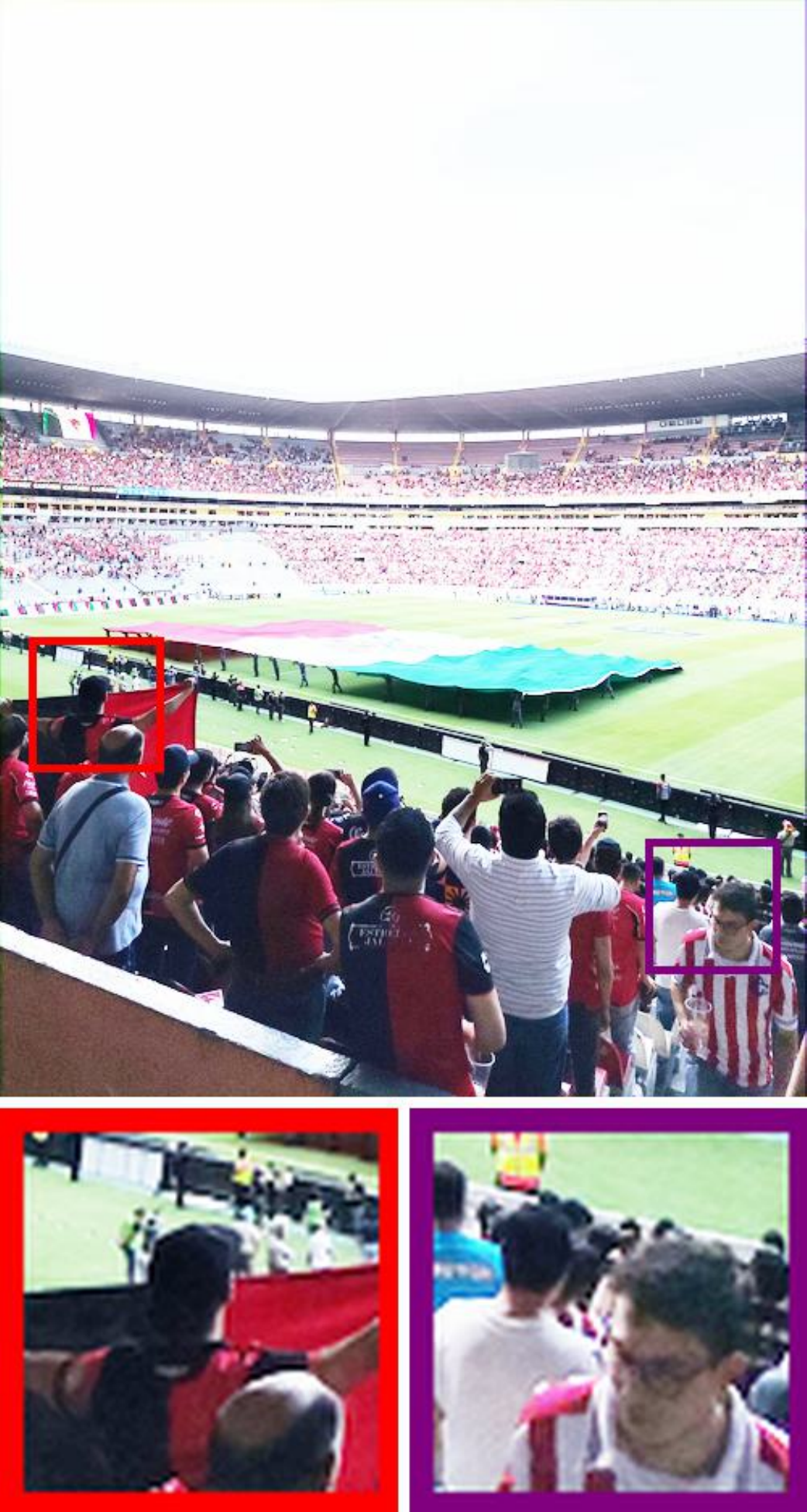} 
        \caption{\footnotesize SCI}
	\end{subfigure}
	\begin{subfigure}{0.12\linewidth}
		\centering
		\includegraphics[width=\linewidth]{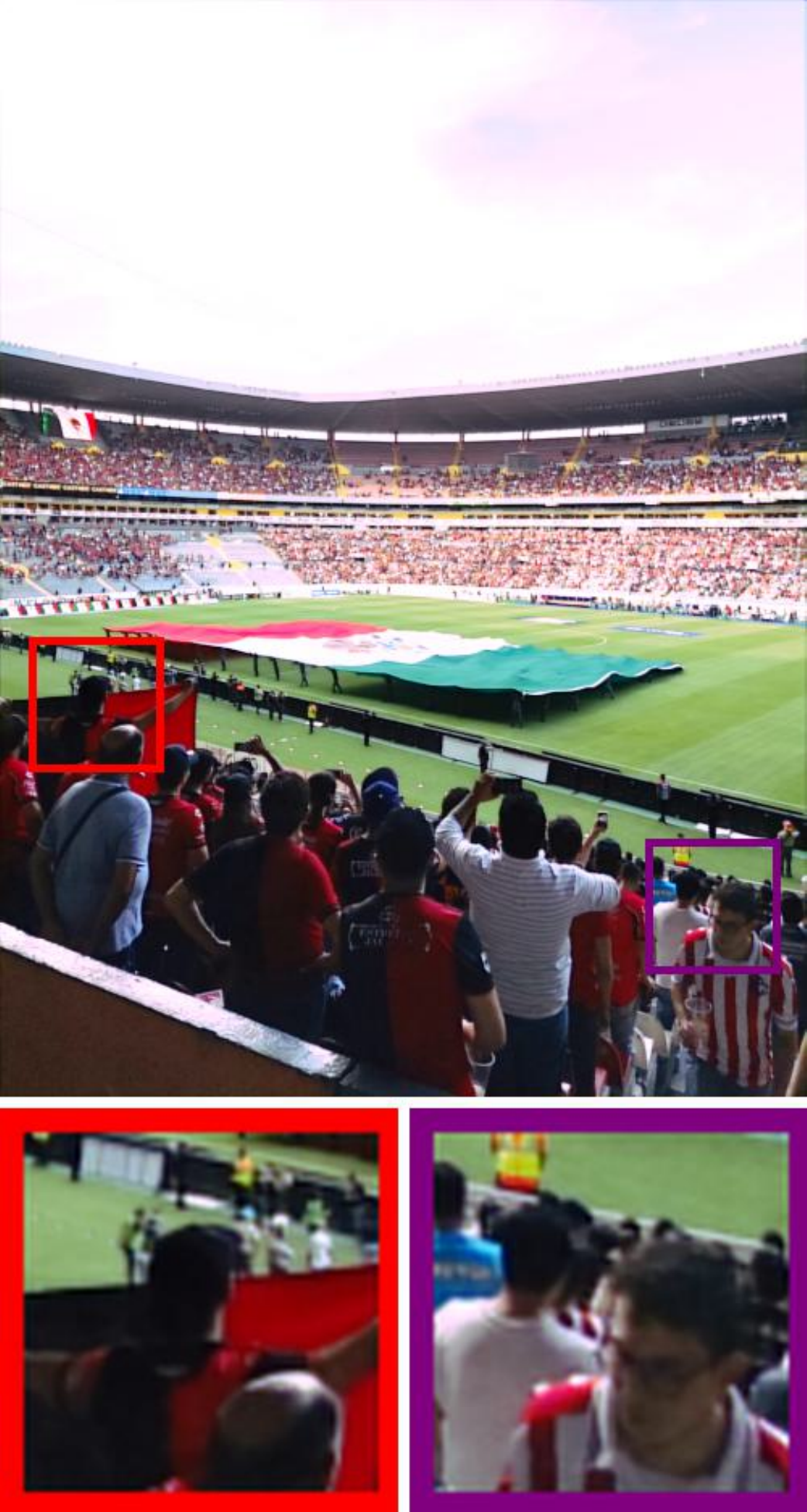} 
        \caption{\footnotesize IAT }
	\end{subfigure}
	\begin{subfigure}{0.12\linewidth}
		\centering
		\includegraphics[width=\linewidth]{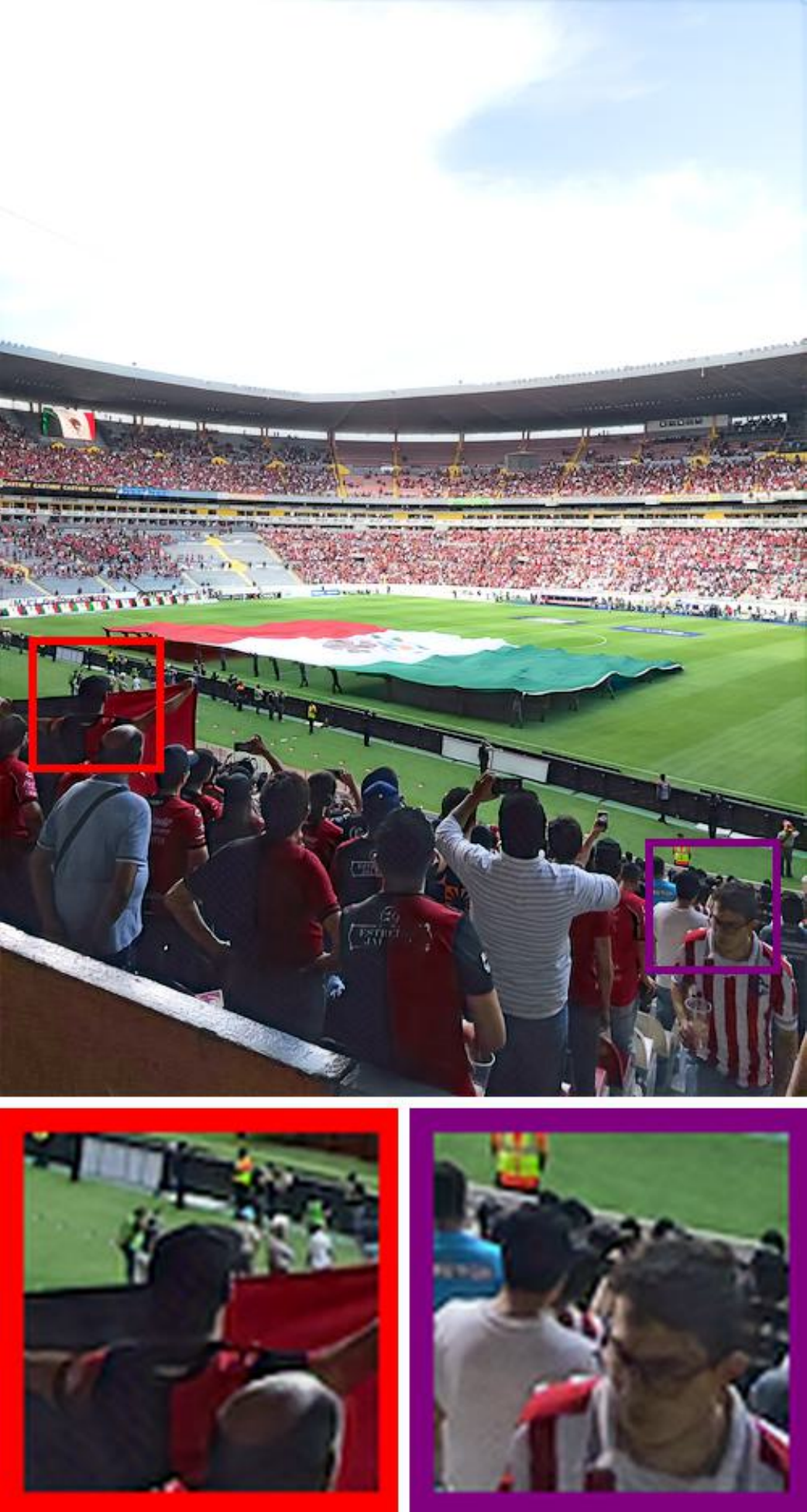}
        \caption{\footnotesize URetinex }
	\end{subfigure}
	\begin{subfigure}{0.12\linewidth}
		\centering
		\includegraphics[width=\linewidth]{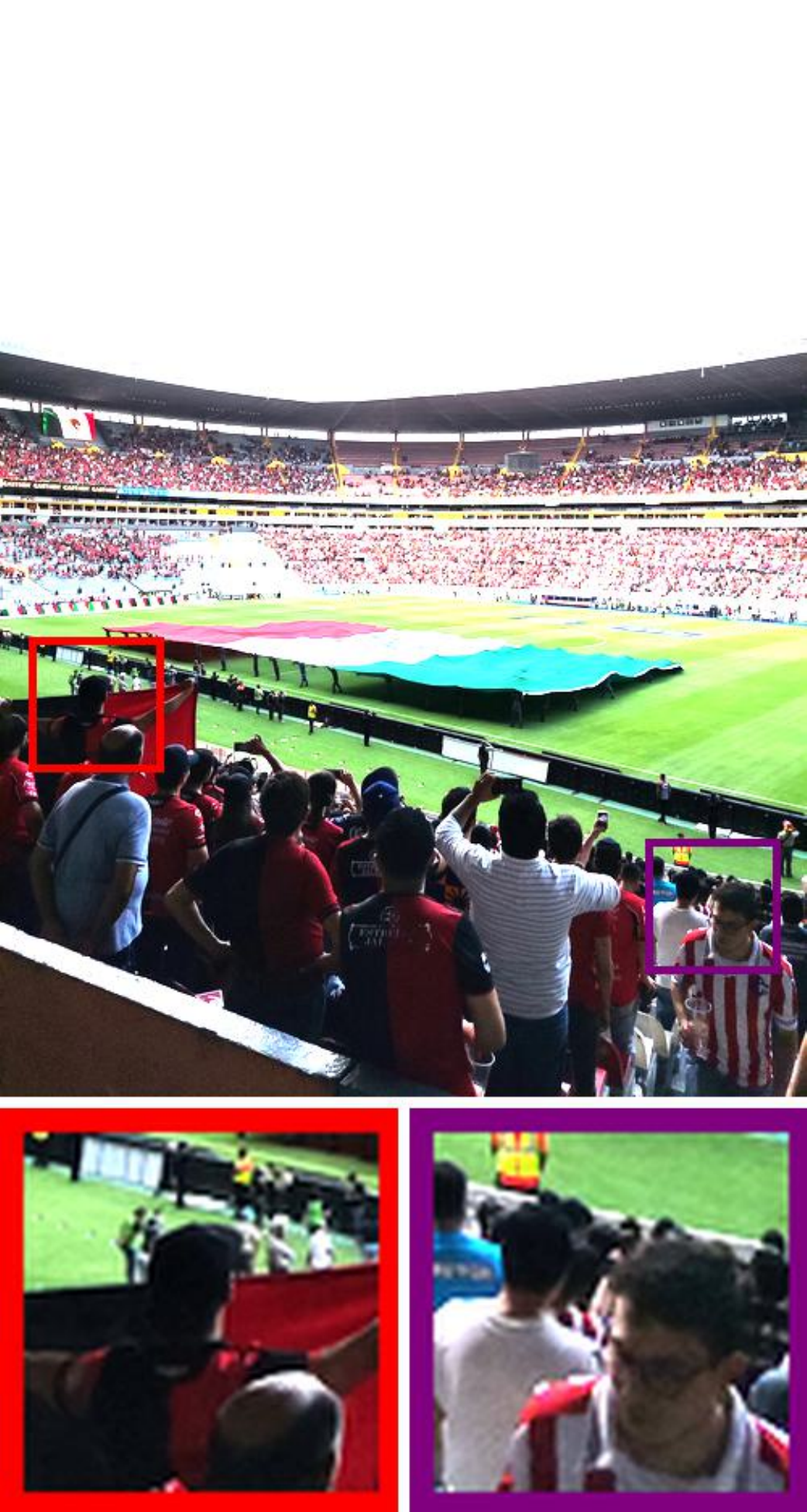} 
		\caption{\footnotesize BL }
	\end{subfigure}
	\begin{subfigure}{0.12\linewidth}
		\centering
		\includegraphics[width=\linewidth]{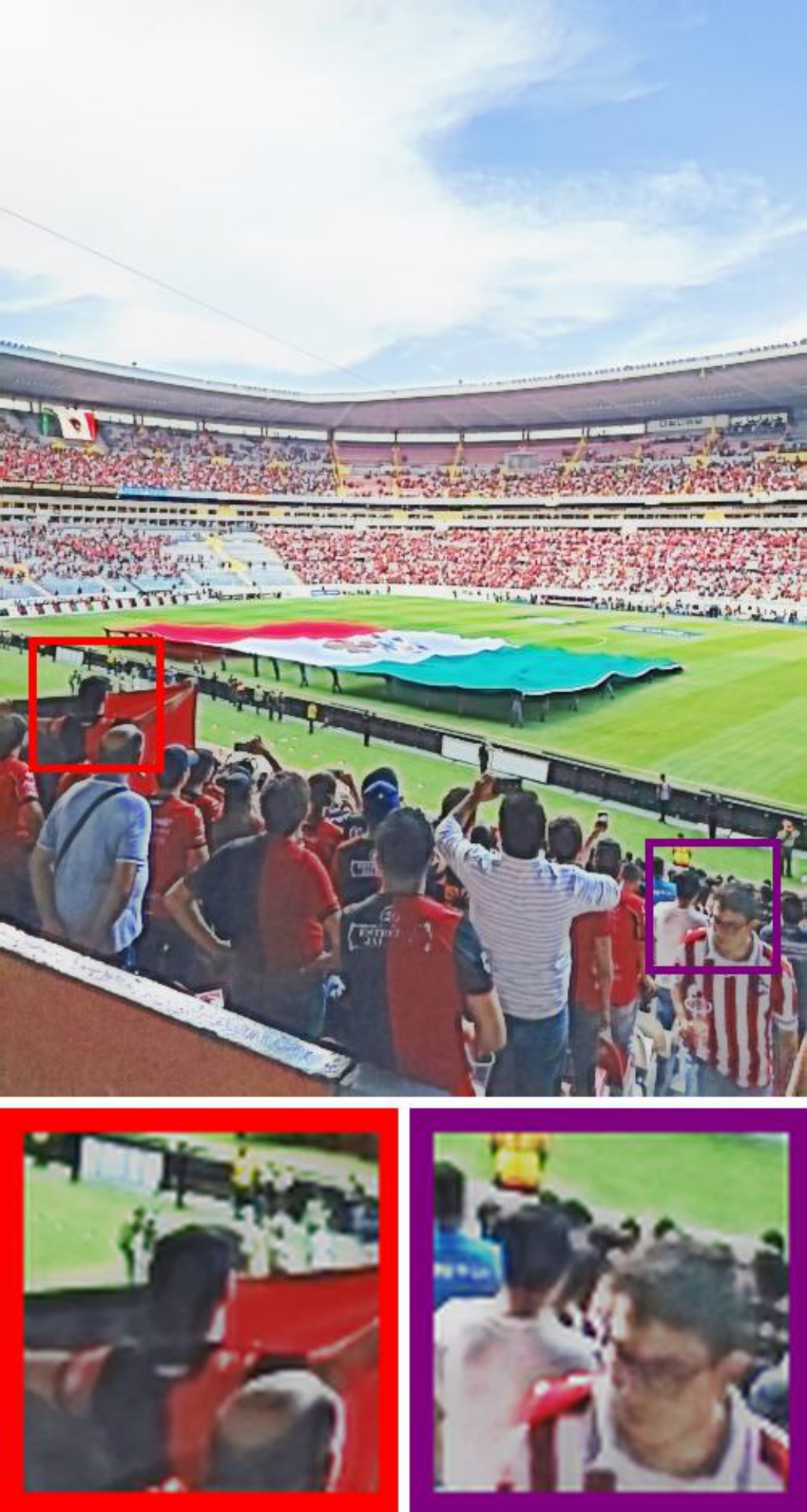} 
		\caption{\footnotesize PairLIE }
	\end{subfigure}
	\begin{subfigure}{0.12\linewidth}
		\centering
		\includegraphics[width=\linewidth]{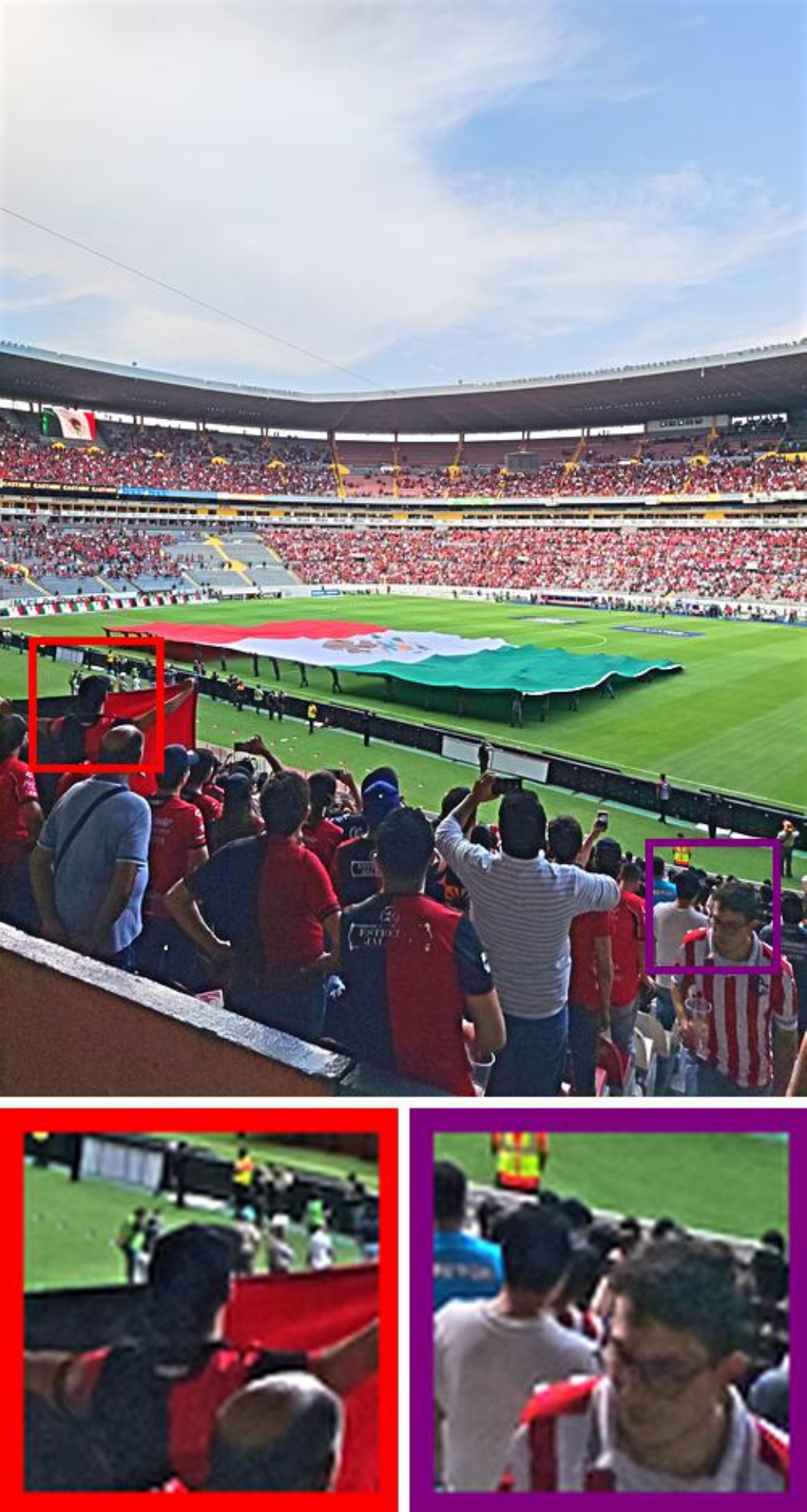}
		\caption{\footnotesize CLIP-LIT }
	\end{subfigure}
    \begin{subfigure}{0.12\linewidth}
		\centering
		\includegraphics[width=\linewidth]{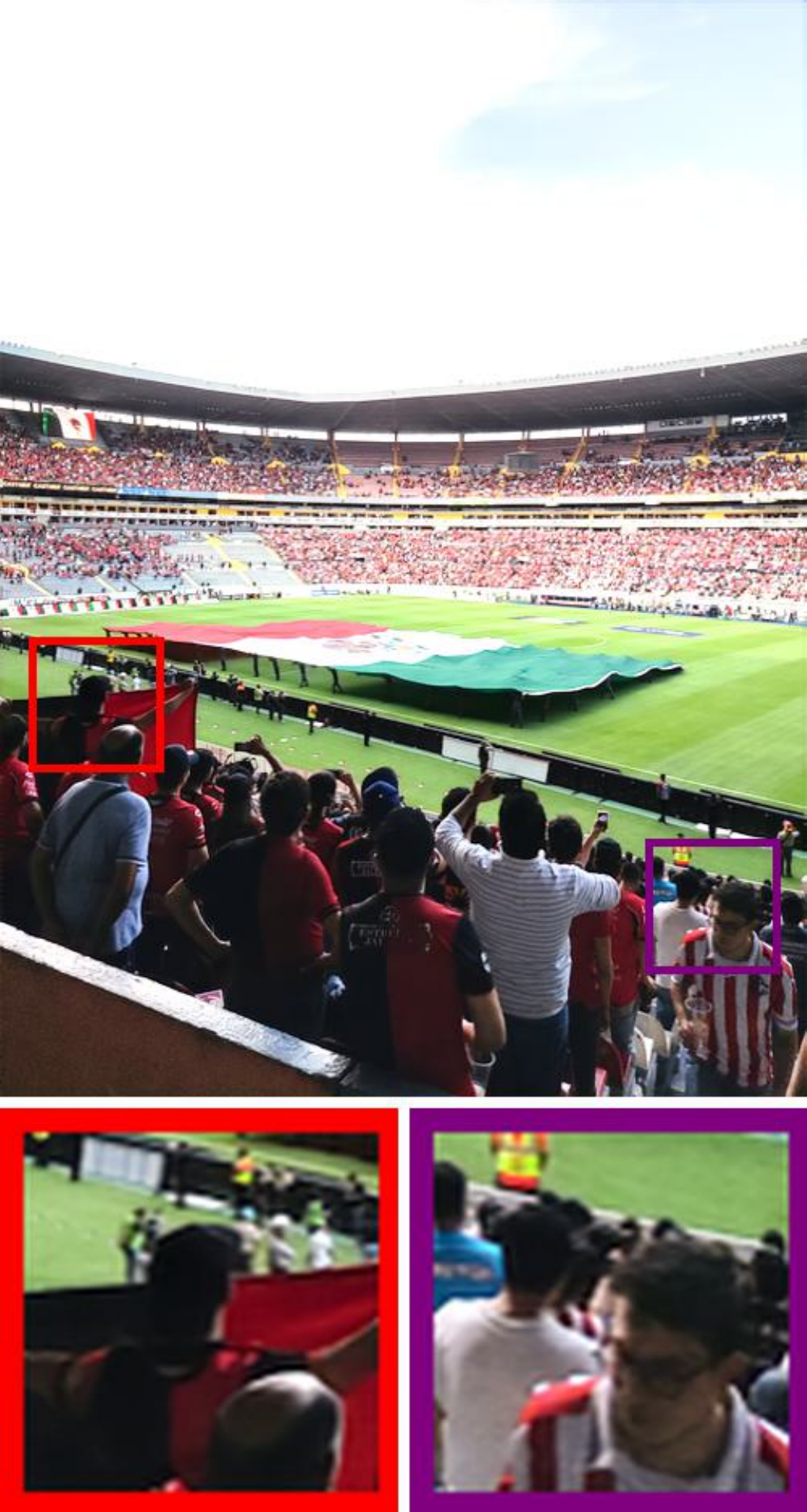} 
		\caption{\footnotesize EMNet }
	\end{subfigure}
	\begin{subfigure}{0.12\linewidth}
		\centering
		\includegraphics[width=\linewidth]{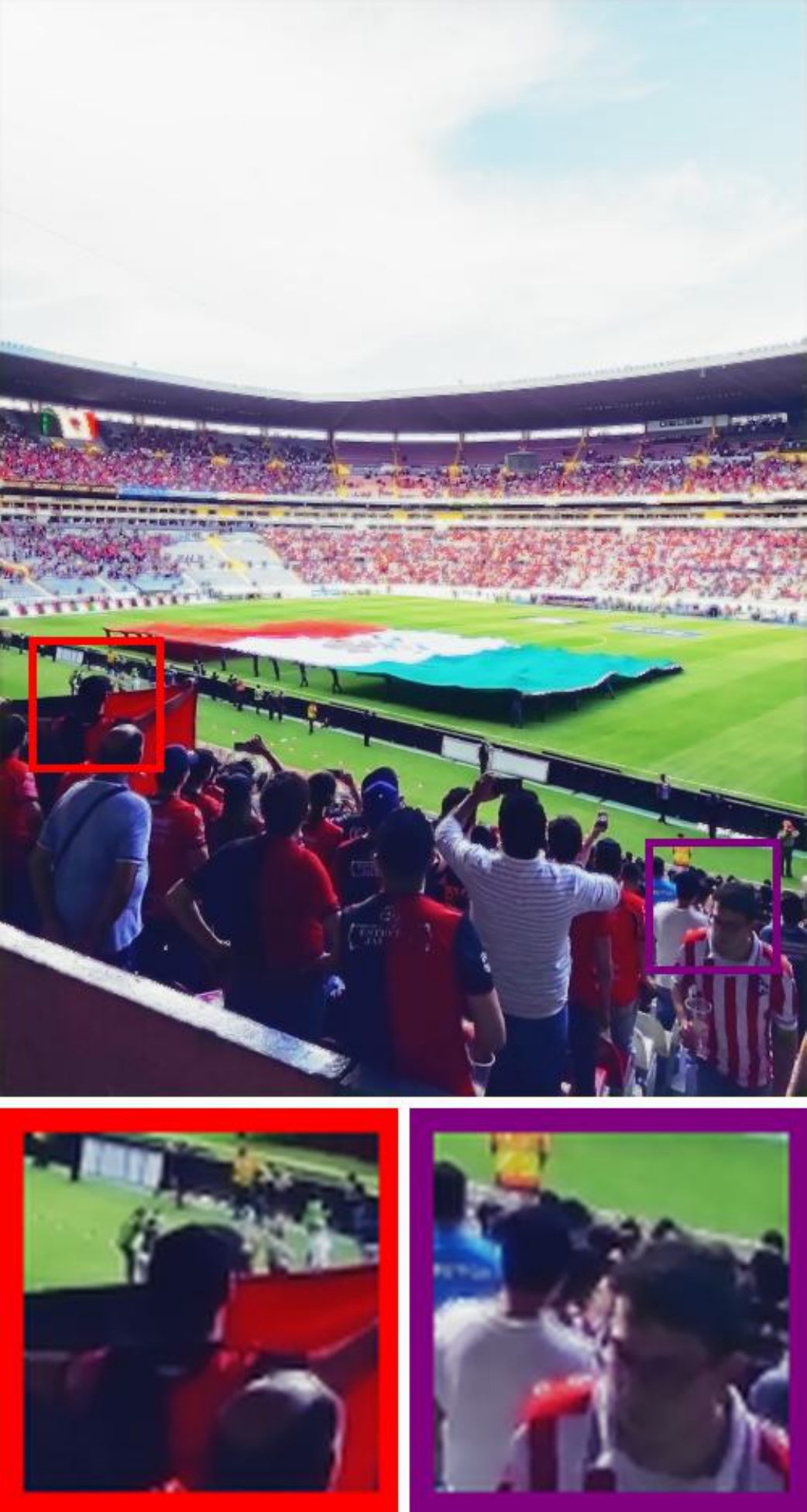} 
		\caption{\footnotesize SHAL-Net }
	\end{subfigure}


    \begin{subfigure}{0.12\linewidth}
		\centering
		\includegraphics[width=\linewidth]{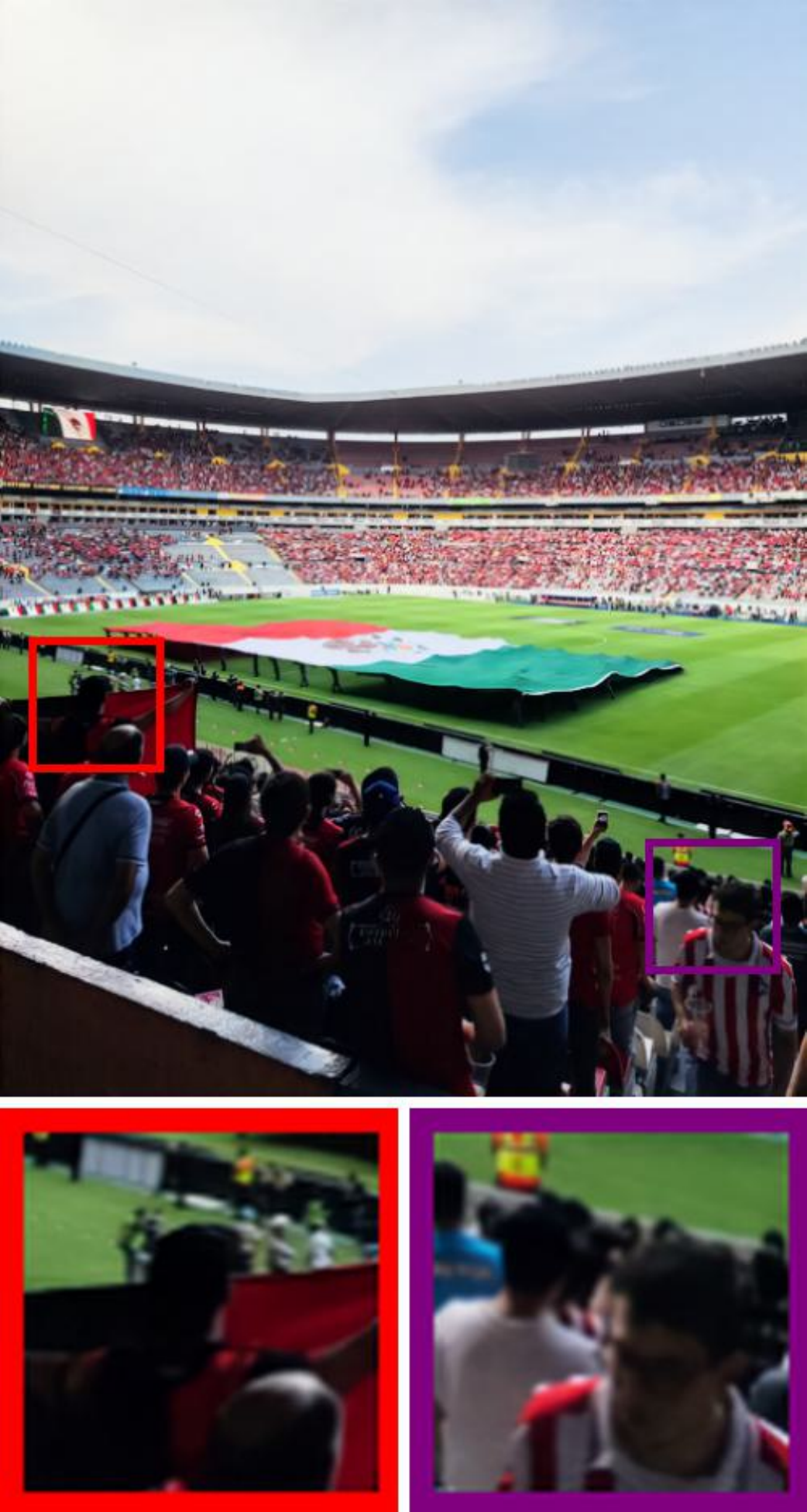} 
        \caption{\footnotesize GSAD}
	\end{subfigure}
	\begin{subfigure}{0.12\linewidth}
		\centering
		\includegraphics[width=\linewidth]{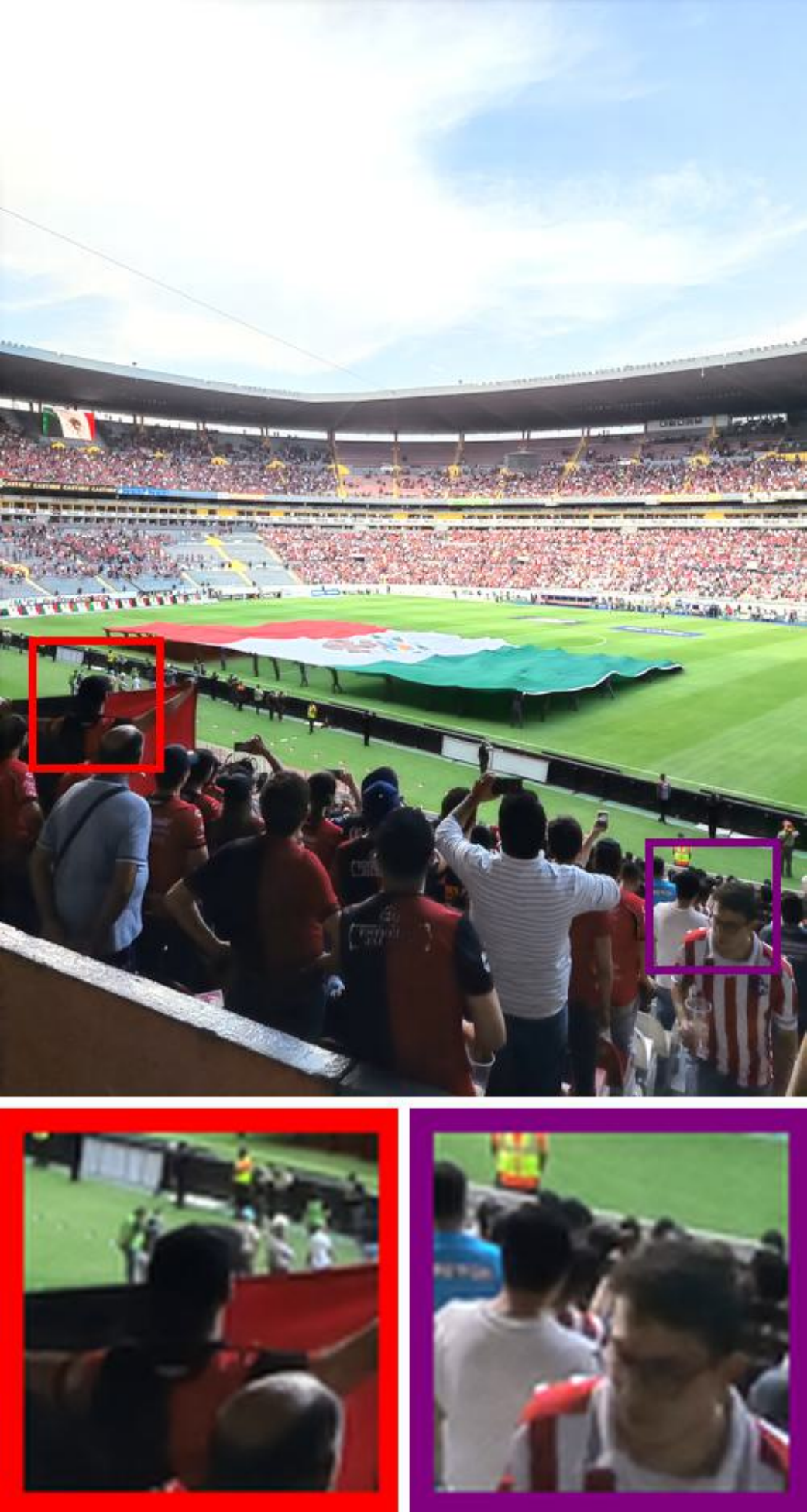} 
        \caption{\footnotesize PPformer }
	\end{subfigure}
	\begin{subfigure}{0.12\linewidth}
		\centering
		\includegraphics[width=\linewidth]{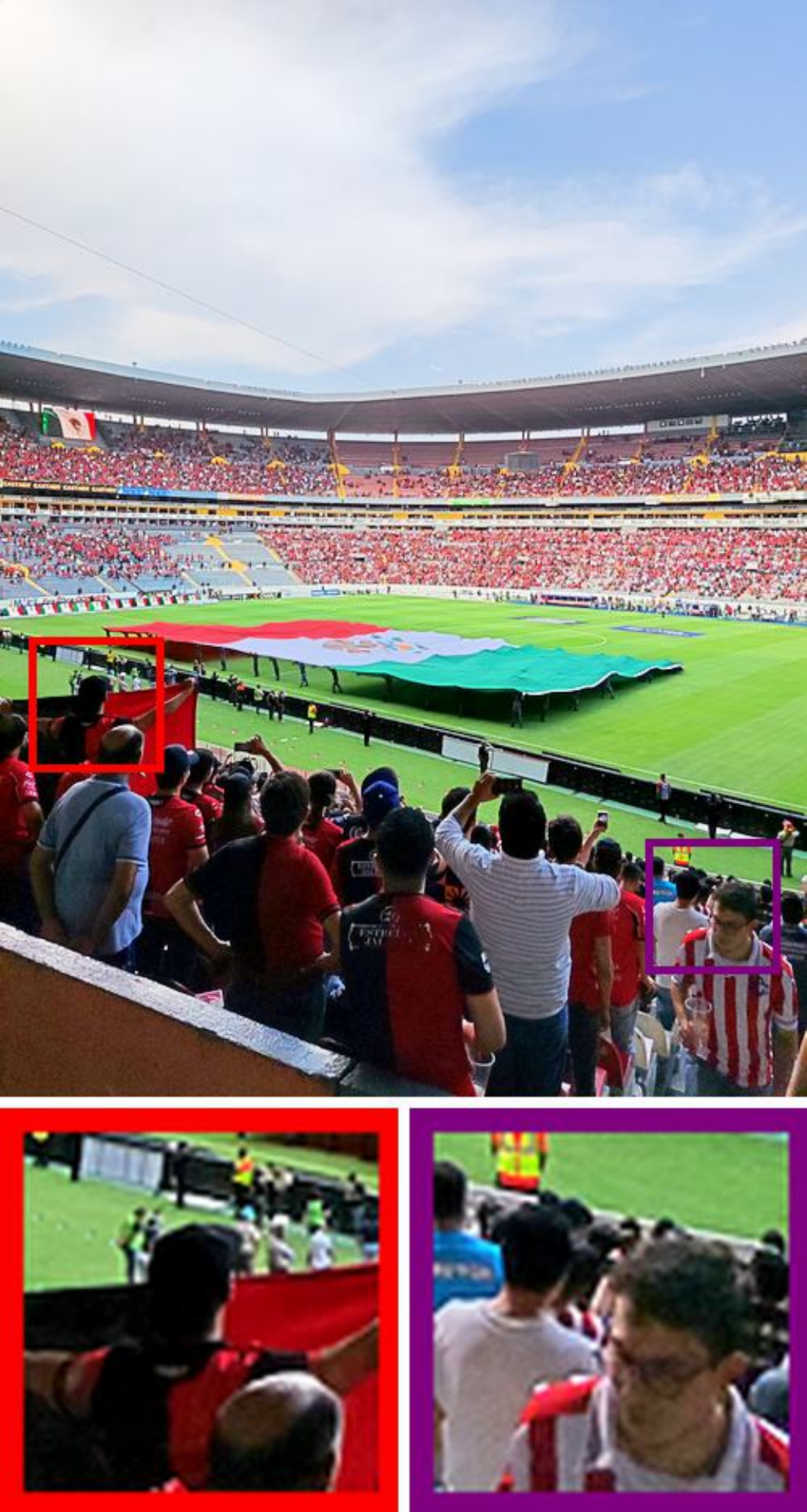}
        \caption{\footnotesize GCP }
	\end{subfigure}
	\begin{subfigure}{0.12\linewidth}
		\centering
		\includegraphics[width=\linewidth]{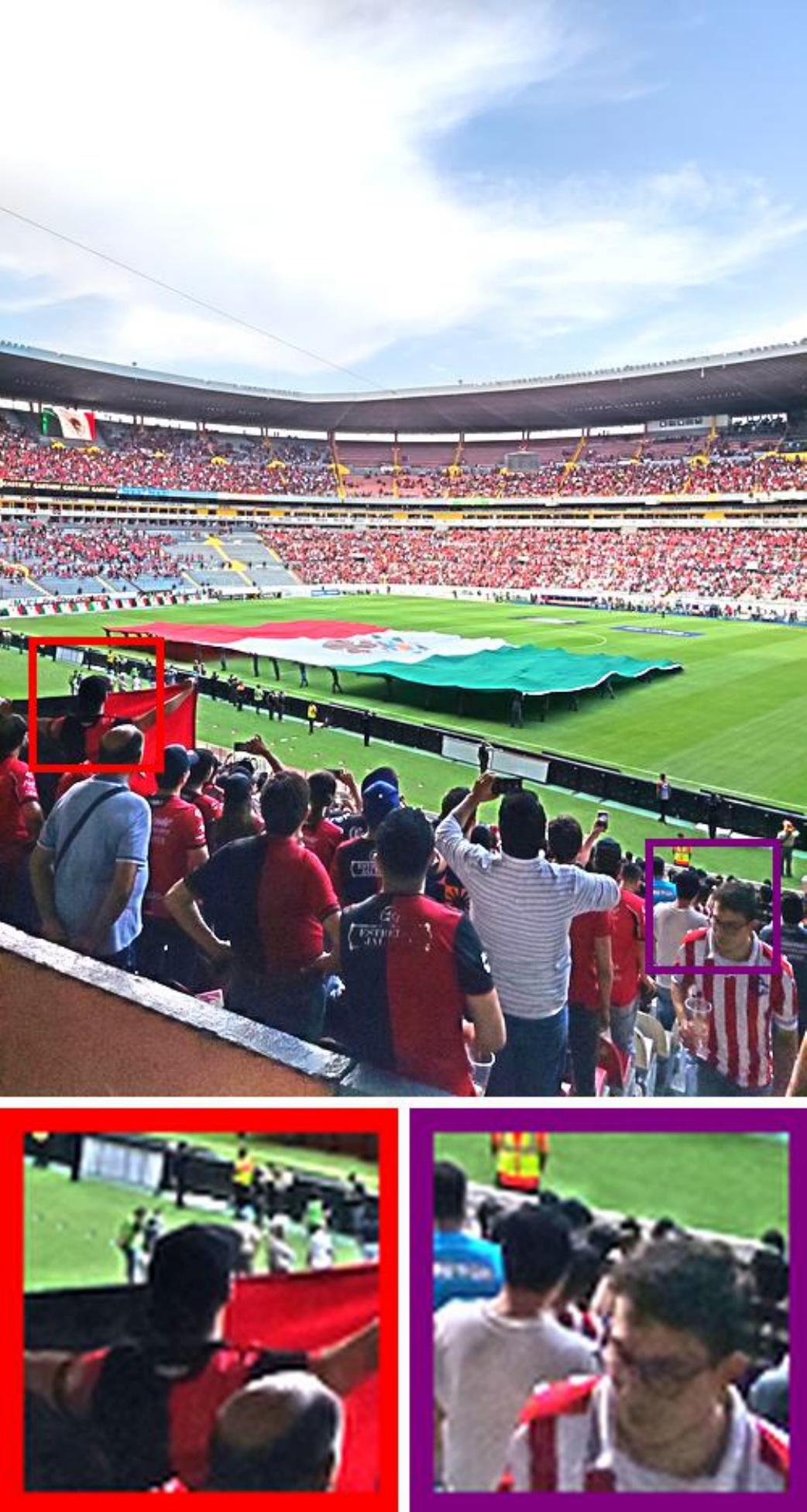} 
		\caption{\footnotesize NeurBR }
	\end{subfigure}
	\begin{subfigure}{0.12\linewidth}
		\centering
		\includegraphics[width=\linewidth]{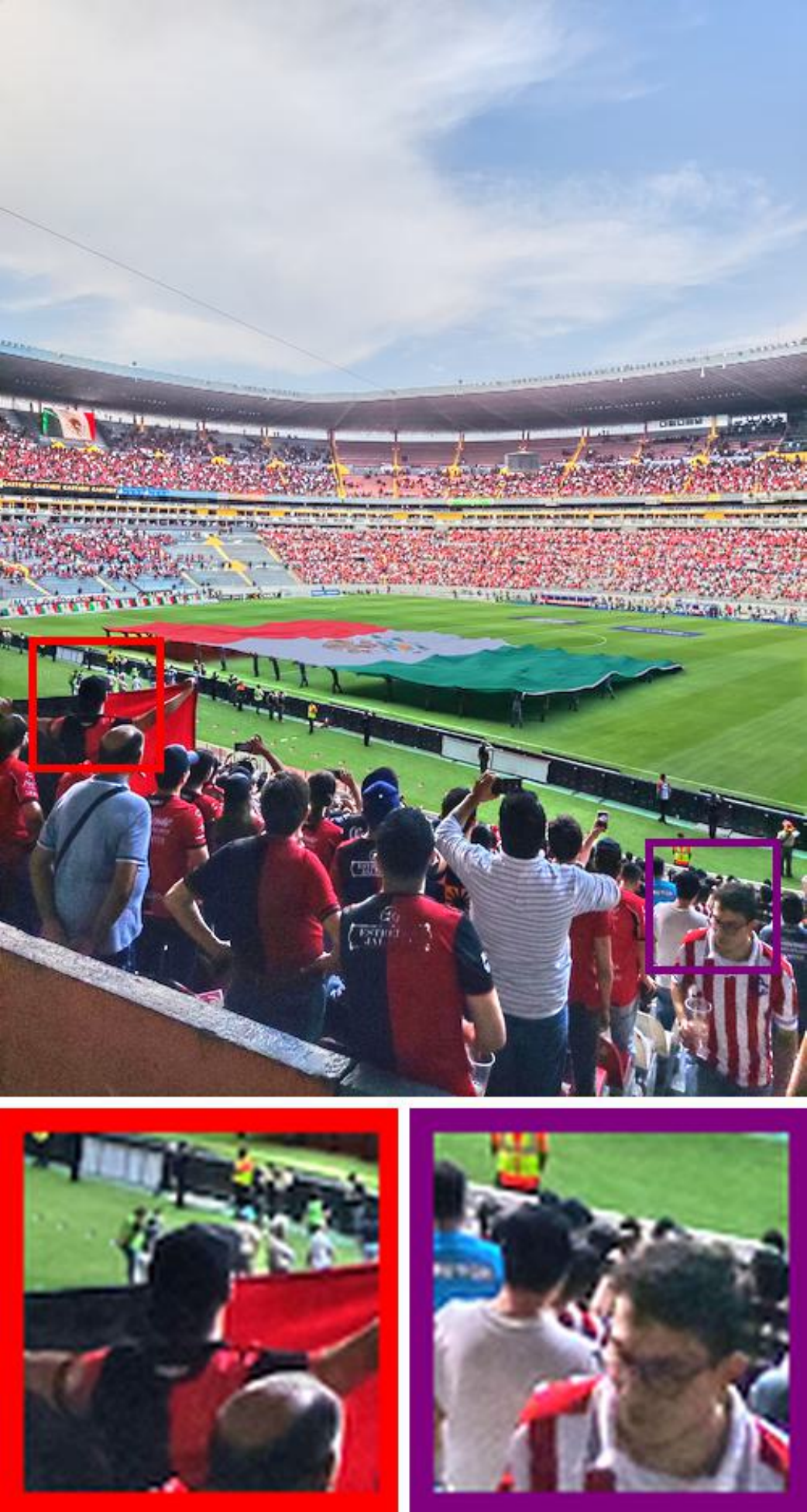} 
		\caption{\footnotesize ITRE }
	\end{subfigure}
	\begin{subfigure}{0.12\linewidth}
		\centering
		\includegraphics[width=\linewidth]{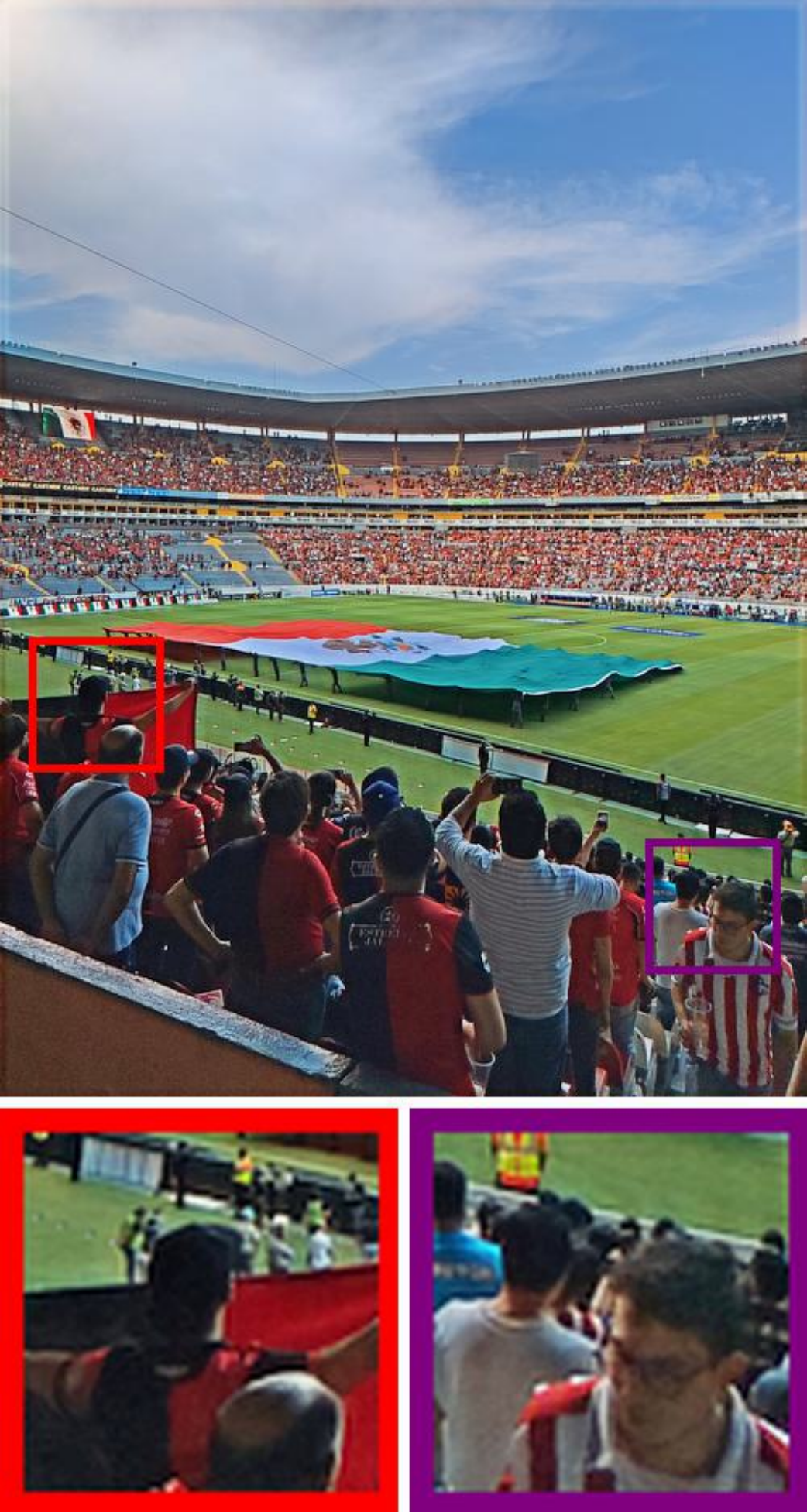}
		\caption{\footnotesize PIE }
	\end{subfigure}
    \begin{subfigure}{0.12\linewidth}
		\centering
		\includegraphics[width=\linewidth]{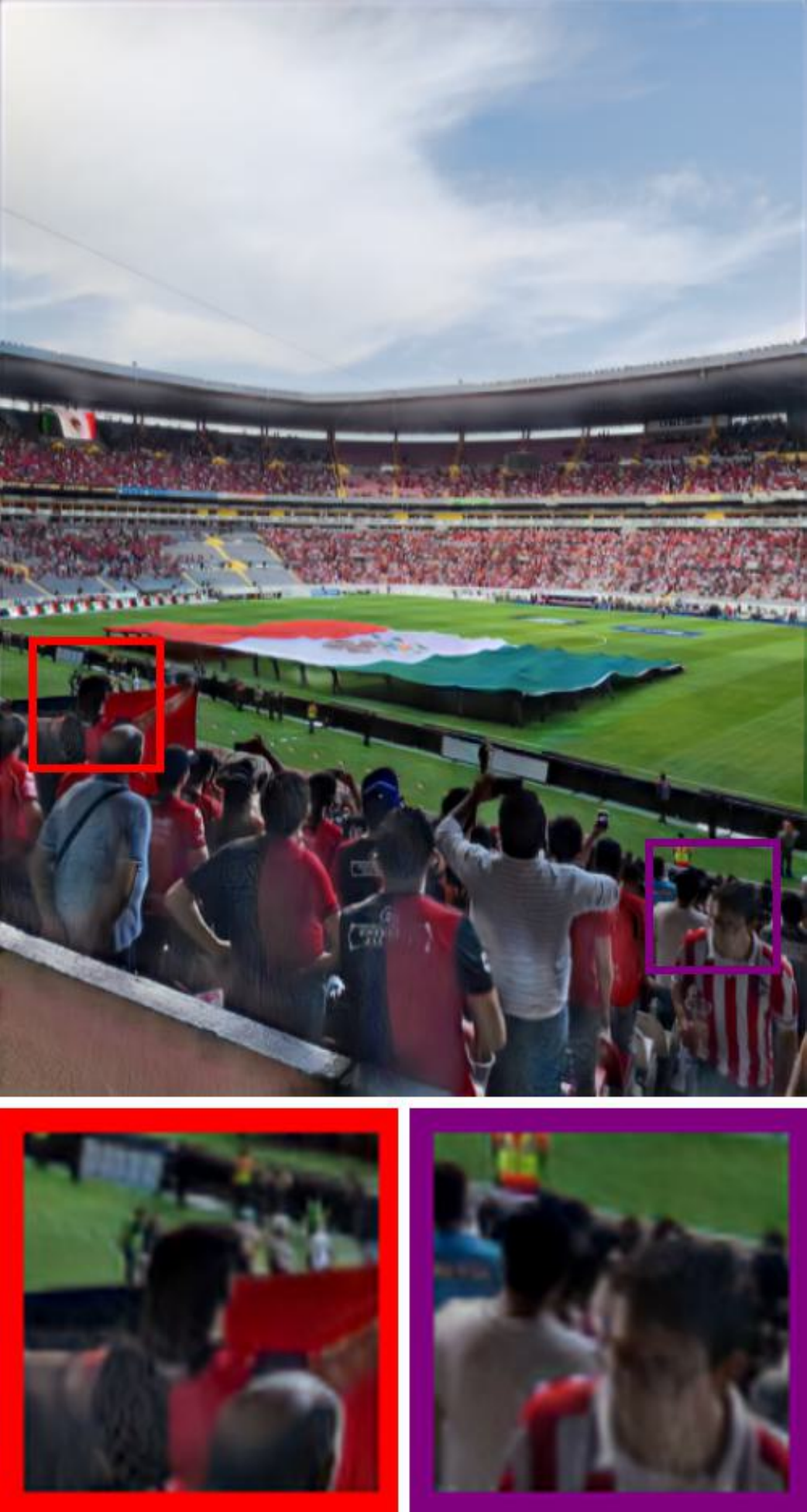} 
		\caption{\footnotesize MSATr }
	\end{subfigure}
	\begin{subfigure}{0.12\linewidth}
		\centering
		\includegraphics[width=\linewidth]{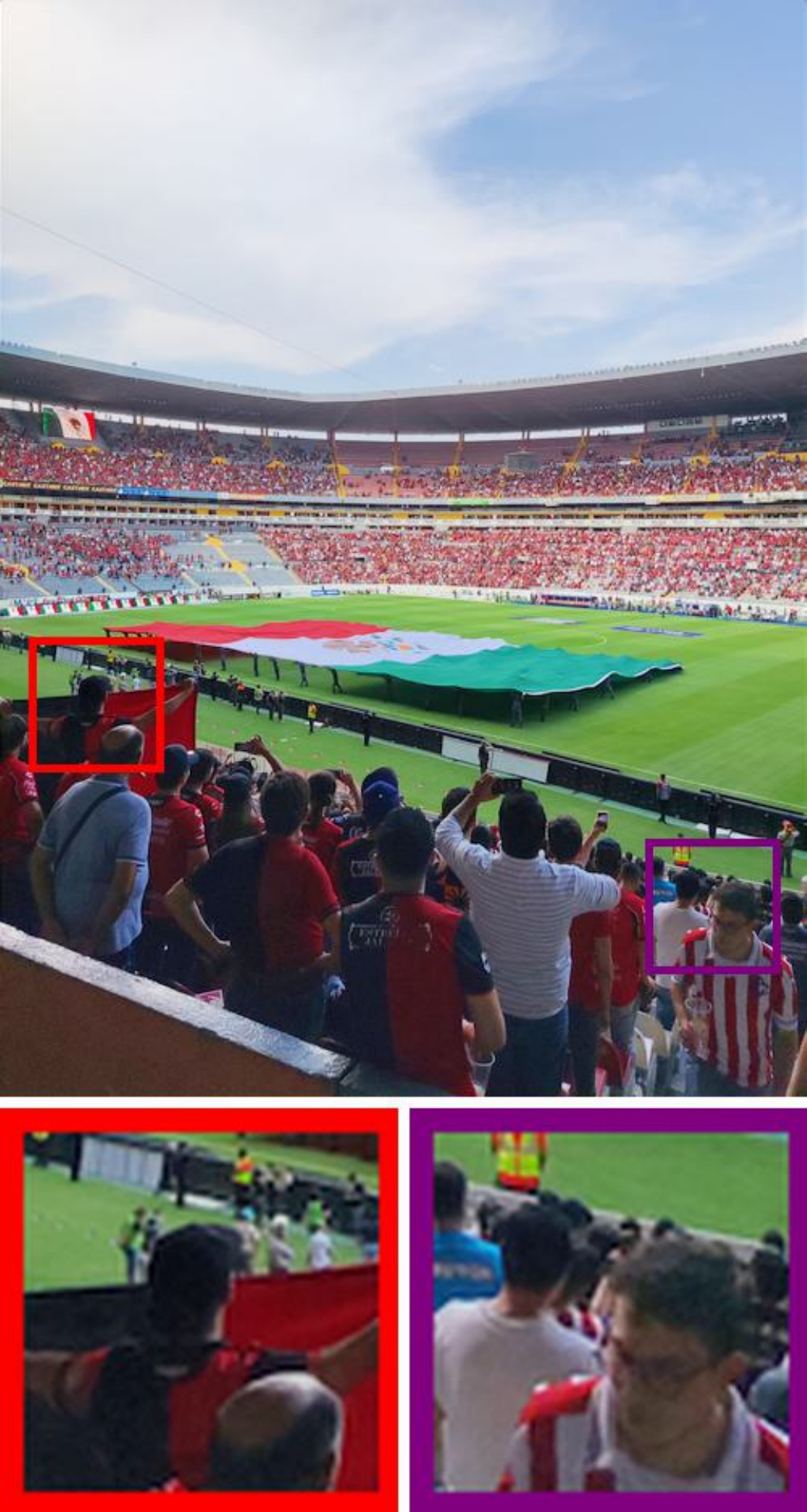} 
		\caption{\footnotesize Proposed }
	\end{subfigure}

	\caption{Visual comparison of methods for enhancing low-light images on the DIS dataset.}
	\label{Q2}
\end{figure*}

To ensure a fair comparison, a performance evaluation was conducted on independent datasets not used in the training of any of the methodologies, including the proposed approach ALEN. The assessment encompassed 1,470 images: 735 from the UHD-LOL4K~\cite{wang2023ultra} dataset, designed to test local enhancement capabilities, and 735 from the Large-Scale Real-World (LSRW)~\cite{hai2023r2rnet} dataset, focused on global enhancement performance. This hybrid dual-dataset strategy enables comprehensive validation of the proposed approach, which uniquely addresses global and local enhancement through an adaptive classification mechanism. Before processing, the system classifies input images to determine whether global or local enhancement strategies should be applied, ensuring that each image receives the most appropriate treatment.  

The LOL dataset~\cite{wei2018deep} was excluded due to its prevalent use in model training across existing methods. The UHD-LOL4K/LSRW combination, being a paired dataset, offers a more generalized assessment framework by better reflecting real-world scenarios, thus avoiding potential evaluation biases from training data overlap. Table \ref{C3Tab4} presents the results of these assessments using various metrics, where $\uparrow$ indicates higher values are better and $\downarrow$ indicates lower values are better.

As shown in Table \ref{C3Tab4}, ALEN outperforms other methods, achieving the highest PSNR of 20.1811, surpassing URetinex at 19.8574. ALEN also leads in SSIM with 0.8015, which indicates better image structure preservation, and in UQI with 0.8920, maintaining global image fidelity. In LPIPS, ALEN demonstrates the best performance with a value of 0.1215, reflecting high perceptual similarity. For DeltaE, ALEN achieves the lowest color difference at 11.8815. Additionally, ALEN excels in NIQA metrics with a MUSIQ of 69.0735 and a NIQE of 3.8941, ranking second in LOE with a value of 194.5614, while achieving first place in SCI with 193.8519, demonstrating effective illumination correction.

In another complementary quantitative evaluation, real-world scene datasets were considered, including DICM~\cite{lee2013contrast} (69 images), LIME~\cite{guo2016lime} (10 images), MEF~\cite{ma2015perceptual} (17 images), NPE~\cite{wang2013naturalness} (8 images), TM-DIED~\cite{VV_TM-DIED} (222 images), and the proposed DIS dataset (10 images). The DIS dataset, designed to evaluate both local and global illumination characteristics, is inspired by smaller datasets like NPE and LIME, allowing for faster and more focused assessments. All these datasets are unpaired, meaning they lack a reference image for comparison. As a result, they were assessed solely using the NIQE and LOE metrics. Table \ref{UPDS} presents the comprehensive evaluation results, where $\downarrow$ indicates that a lower value is better.

From the data provided in Table \ref{UPDS}, ALEN demonstrates competitive performance, particularly in DIS, LIME, and TM-DIED datasets, with NIQE values of 2.9065, 3.6623, and 2.5740, respectively. ALEN leads most datasets in the LOE metric, except for DIS, where SCI ranks first at 129.7757. ALEN ranks best overall, with average NIQE and LOE values of 3.0470 and 80.4541. While PPformer excels in specific datasets like DICM and MEF, it ranks second in overall NIQE. ALEN’s balanced average LOE outperforms Zero-DCE with 156.3173 and surpasses older methods like LIME with an average NIQE of 3.3884 and LOE of 728.5905, showcasing its advancements in luminance and quality preservation.

\subsection{Qualitative results}
The qualitative evaluation includes images of real-world scenes captured in low-light conditions. Figures \ref{COLLAGE_SOTA}, \ref{Q1}, and \ref{Q2} show images taken at different times of the day: Figure \ref{COLLAGE_SOTA} compares various methods in day-time, night-time, indoor, and outdoor scenarios, with images taken from the DICM~\cite{lee2013contrast}, LIME~\cite{guo2016lime}, NPE~\cite{wang2013naturalness}, MEF~\cite{ma2015perceptual}, TM-DIED~\cite{VV_TM-DIED} and DIS datasets; Figure \ref{Q1} presents a night-time image from the DICM dataset, and Figure \ref{Q2} shows a day-time image from the DIS dataset.

In \ref{COLLAGE_SOTA}, the PPformer method exhibits opacity and overexposes the sky in day-time environments, while in night-time scenes, RetinexNet and SHAL-Net introduce noise and distort colors. In indoor settings, BL and EMNet show overexposure in some areas, and SCI excessively overexposes during sunset. Overall, the ALEN method provides the best results, adapting well to various conditions and enhancing the image appropriately.

\begin{figure*}[ht]
    \newlength{\asheight}
	\setlength{\asheight}{2.1cm}
	\centering
    \begin{subfigure}{0.19\linewidth}
		\centering
		\includegraphics[width=\linewidth,  height=\asheight]{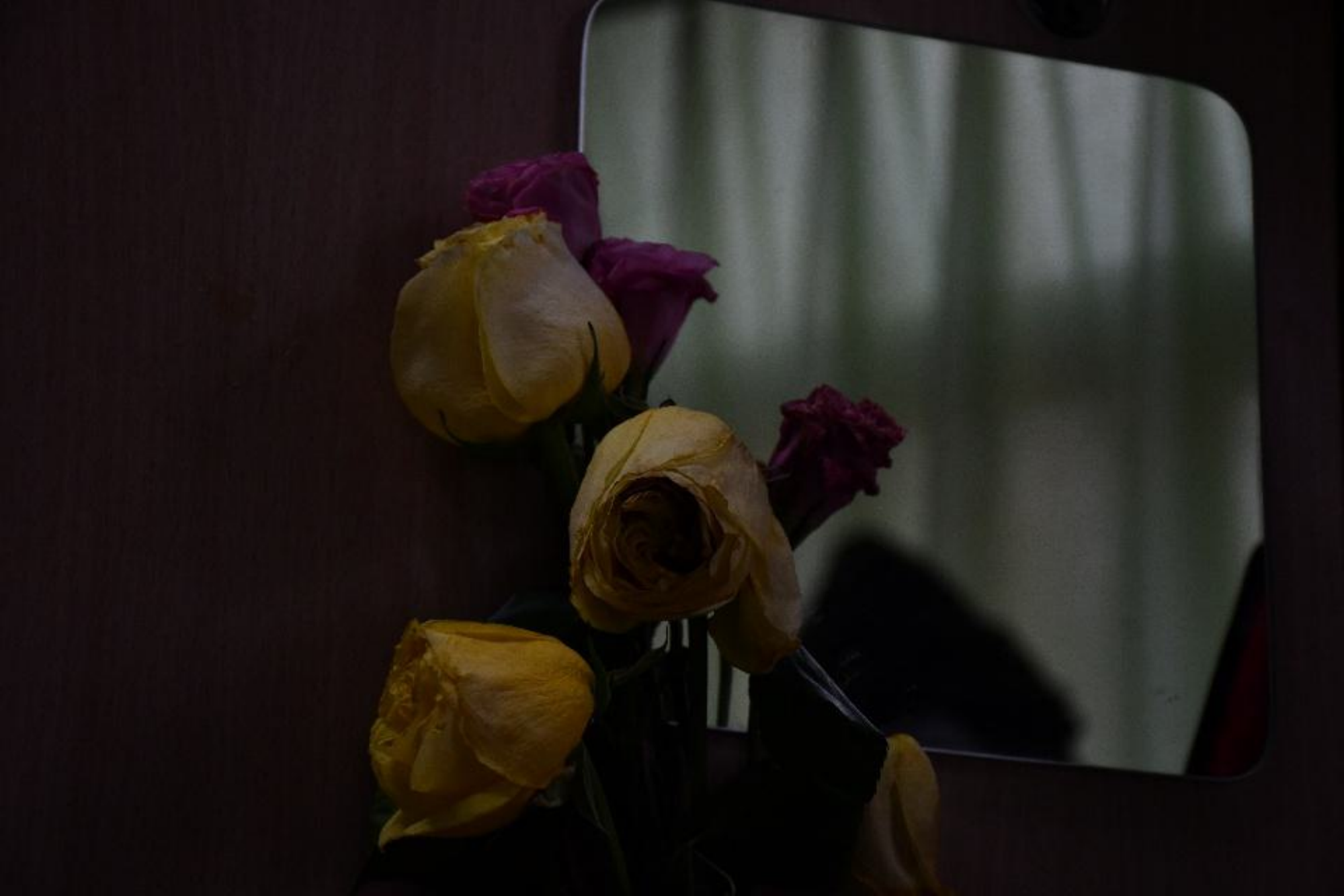} 
        \includegraphics[width=\linewidth,  height=\asheight]{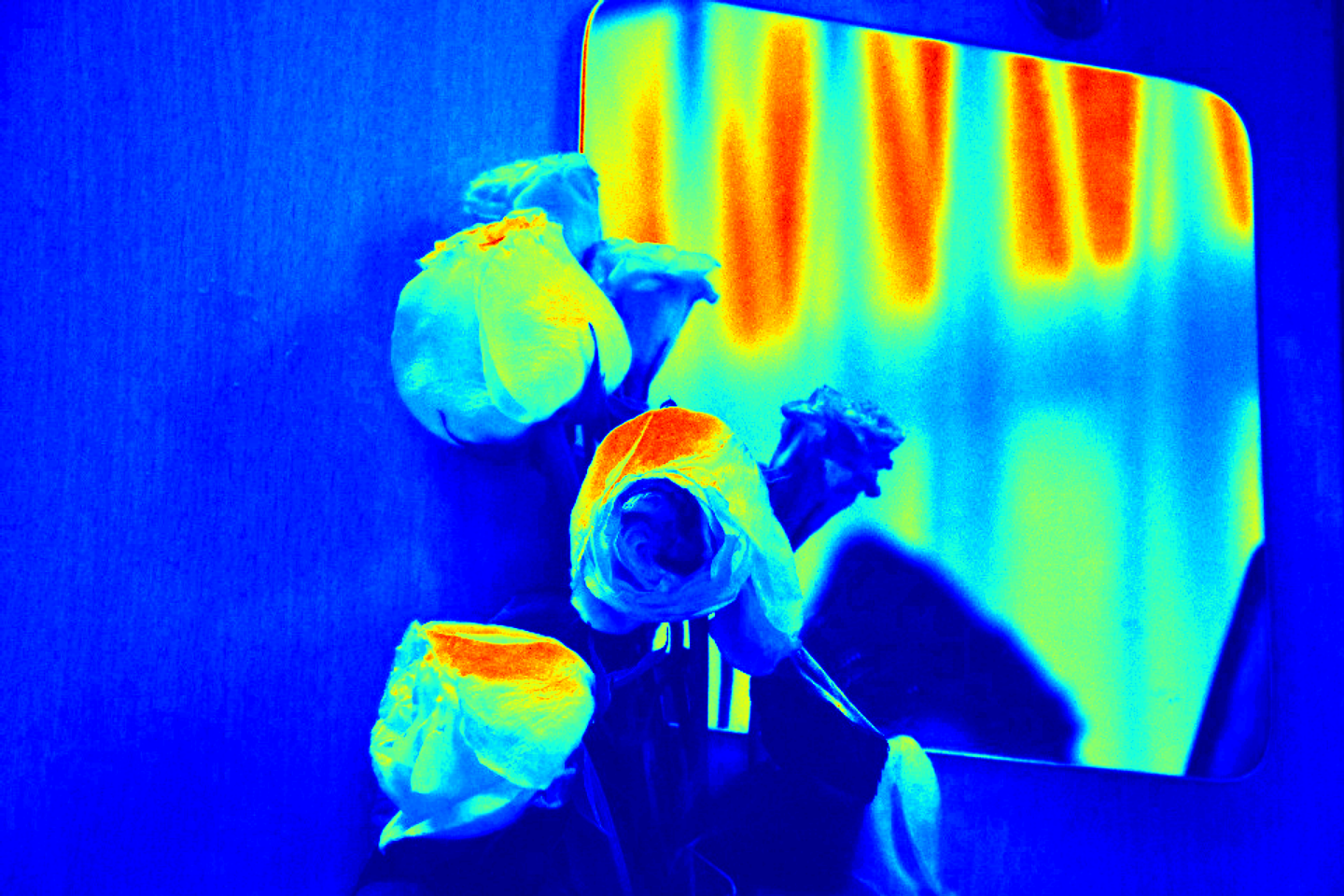} 
		\caption{\scriptsize Global Input}
		\label{ASRW1_LL}
	\end{subfigure}
	\begin{subfigure}{0.19\linewidth}
		\centering
		\includegraphics[width=\linewidth,  height=\asheight]{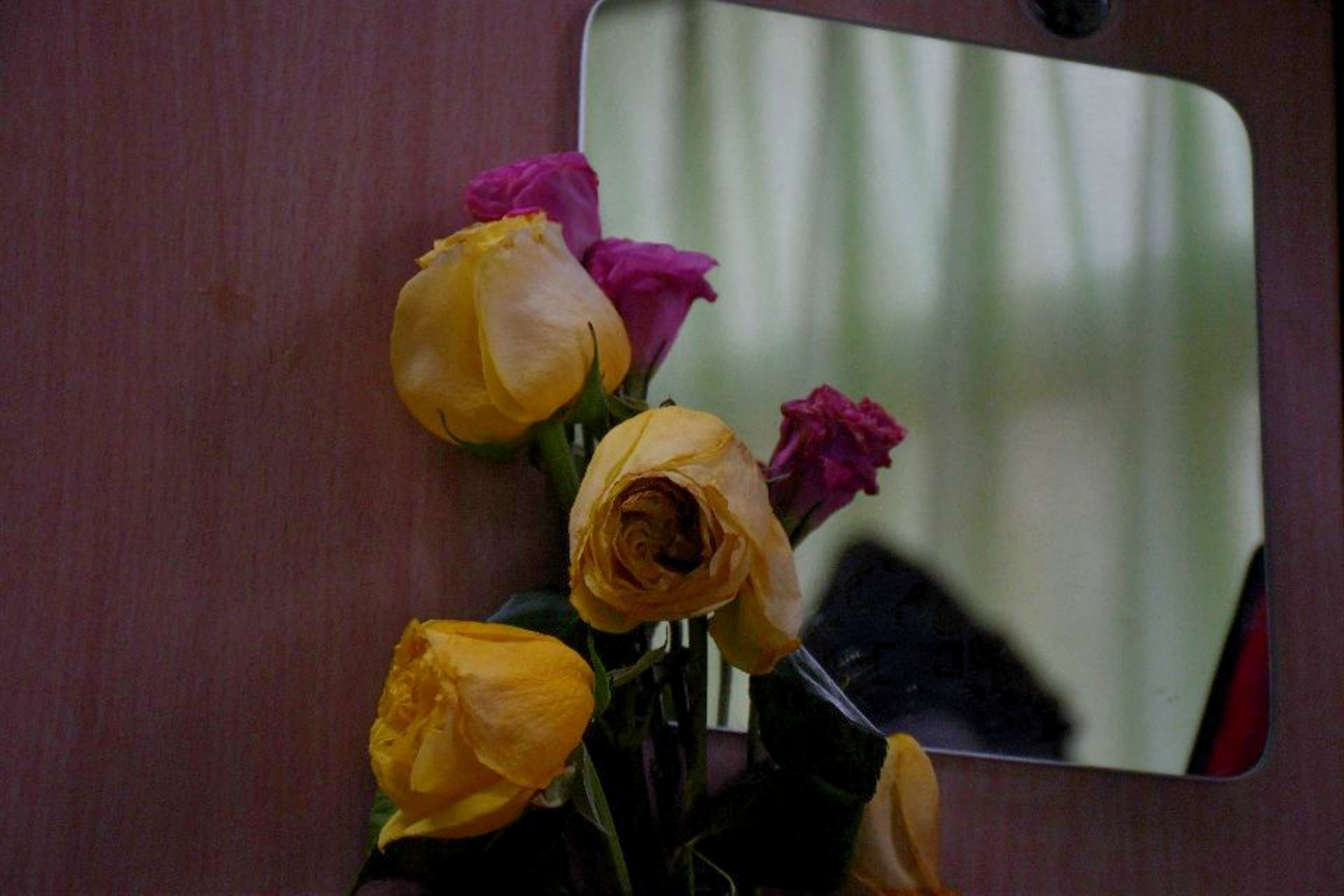} 
        \includegraphics[width=\linewidth,  height=\asheight]{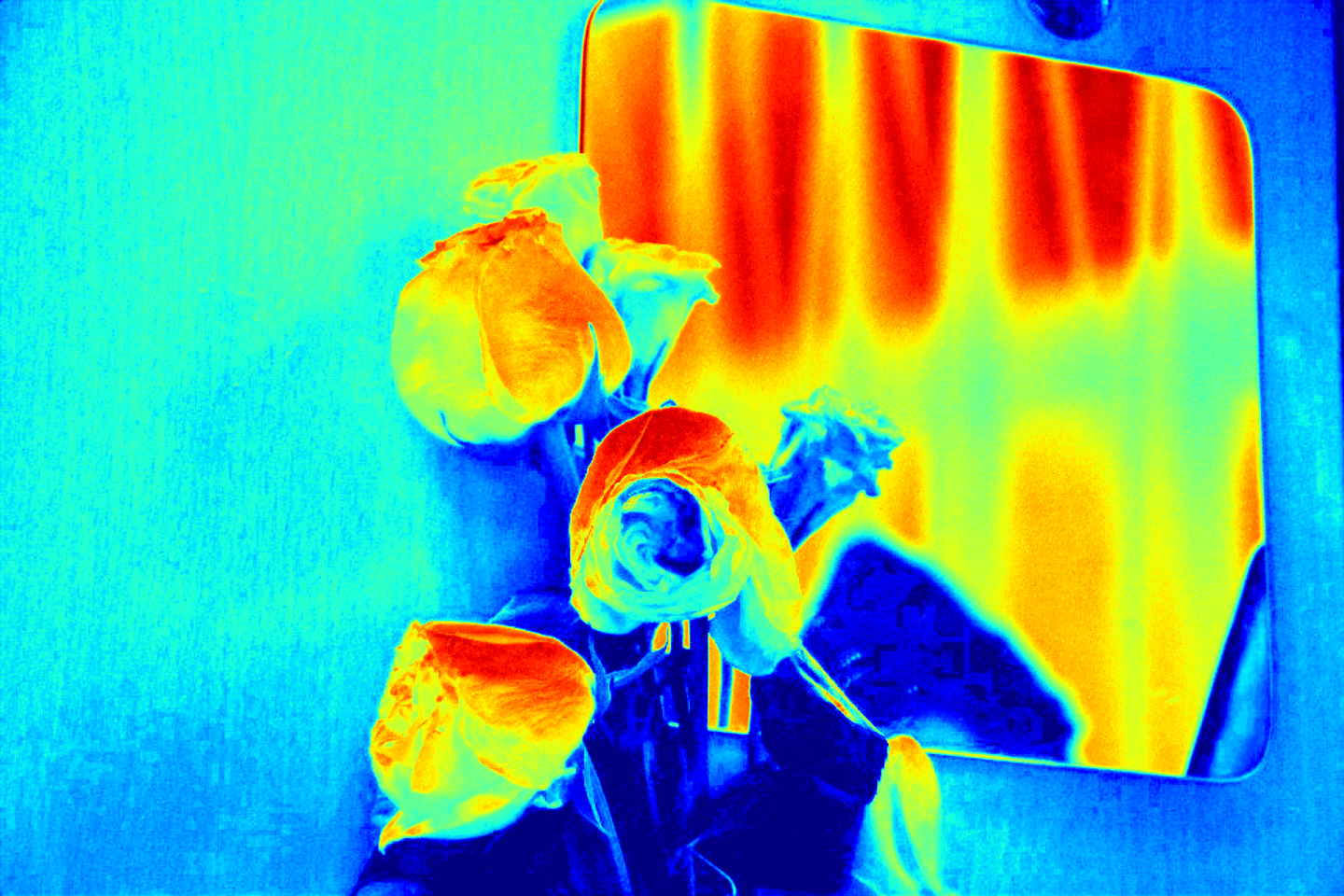} 
		\caption{\scriptsize w/o CL and GE}
		\label{ASRW1_GECE}
	\end{subfigure}
	\begin{subfigure}{0.19\linewidth}
		\centering
		\includegraphics[width=\linewidth,  height=\asheight]{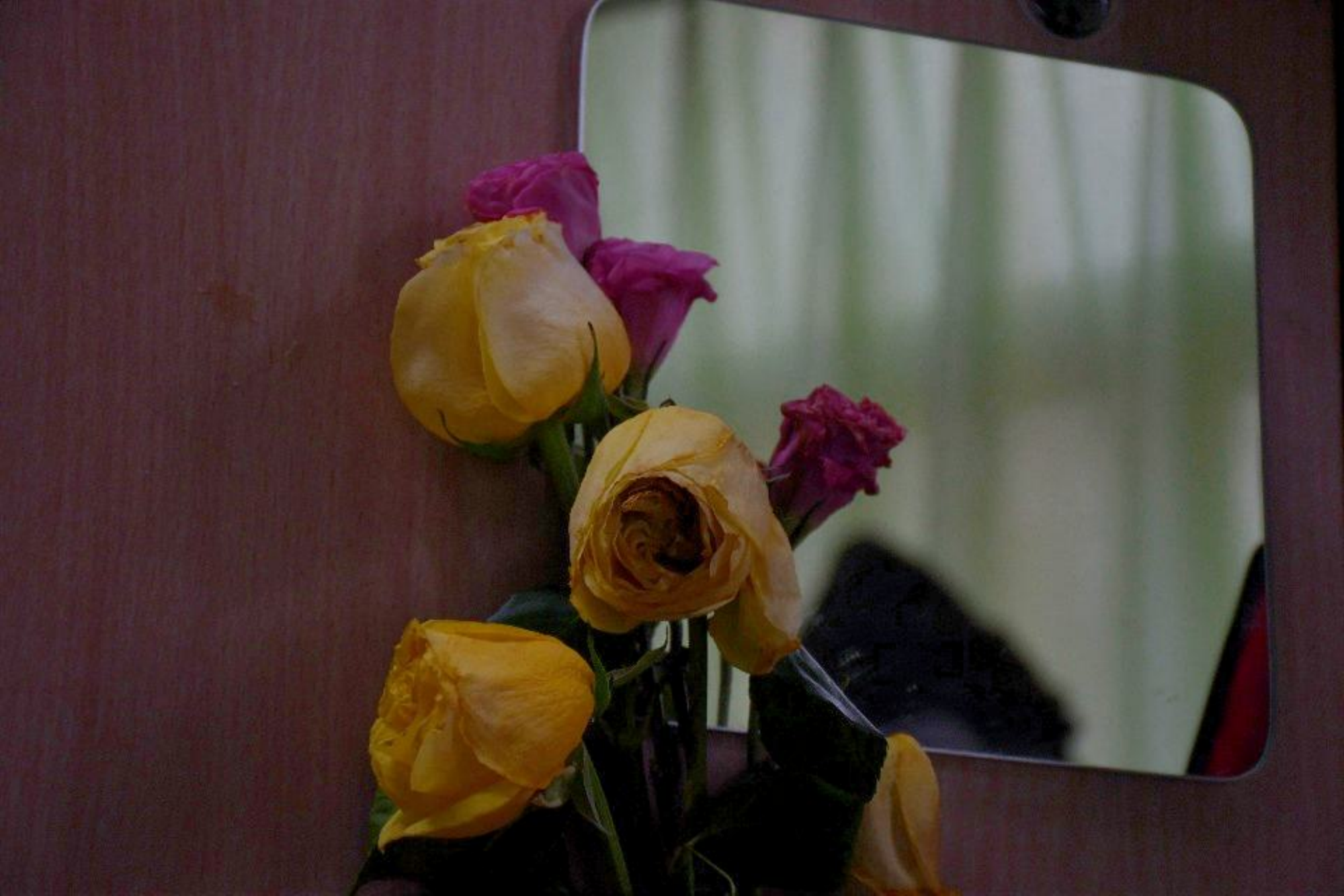}
        \includegraphics[width=\linewidth,  height=\asheight]{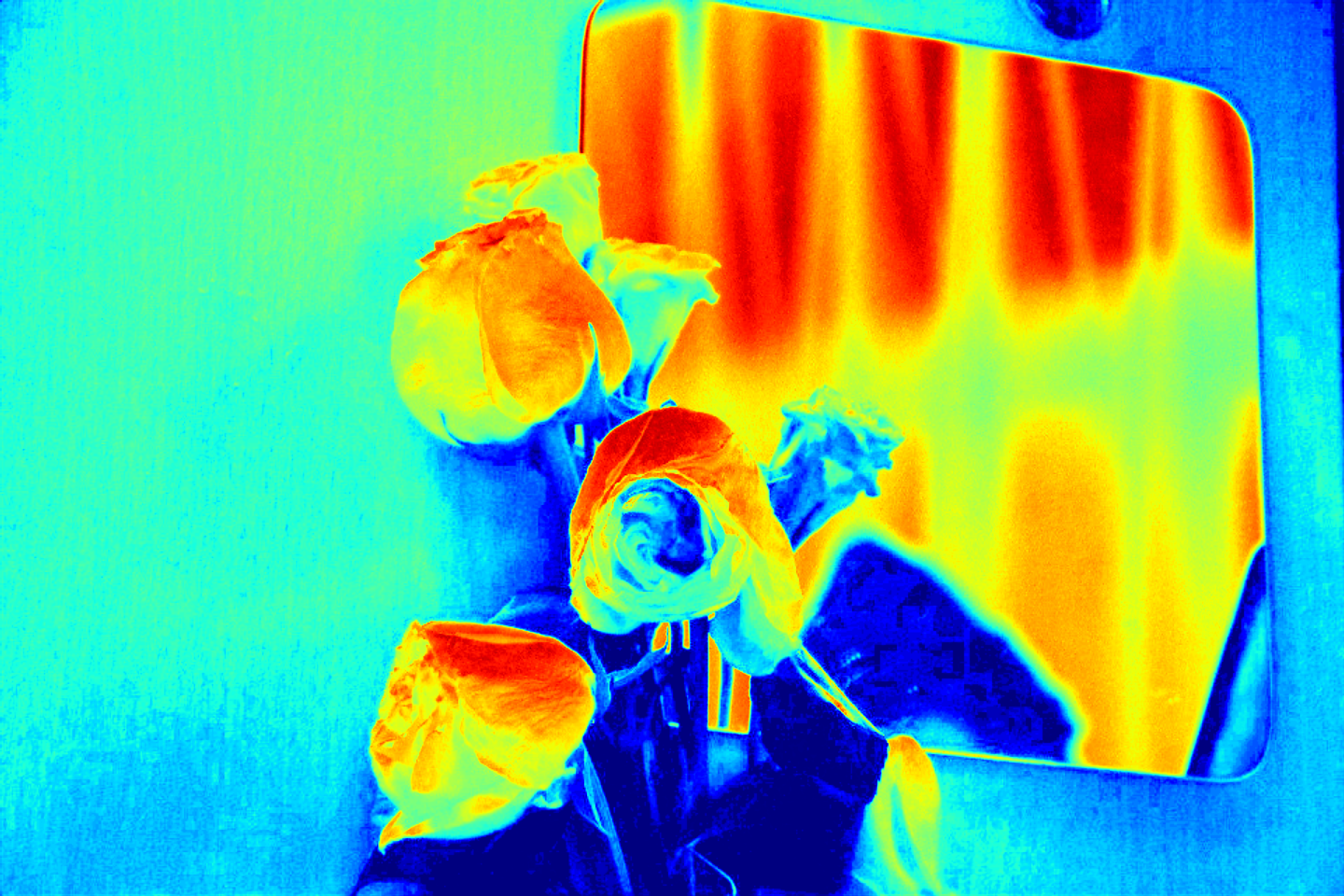}
		\caption{\scriptsize w/o CL, GE and CE}
		\label{ASRW1_GE}
	\end{subfigure}
	\begin{subfigure}{0.19\linewidth}
		\centering
		\includegraphics[width=\linewidth,  height=\asheight]{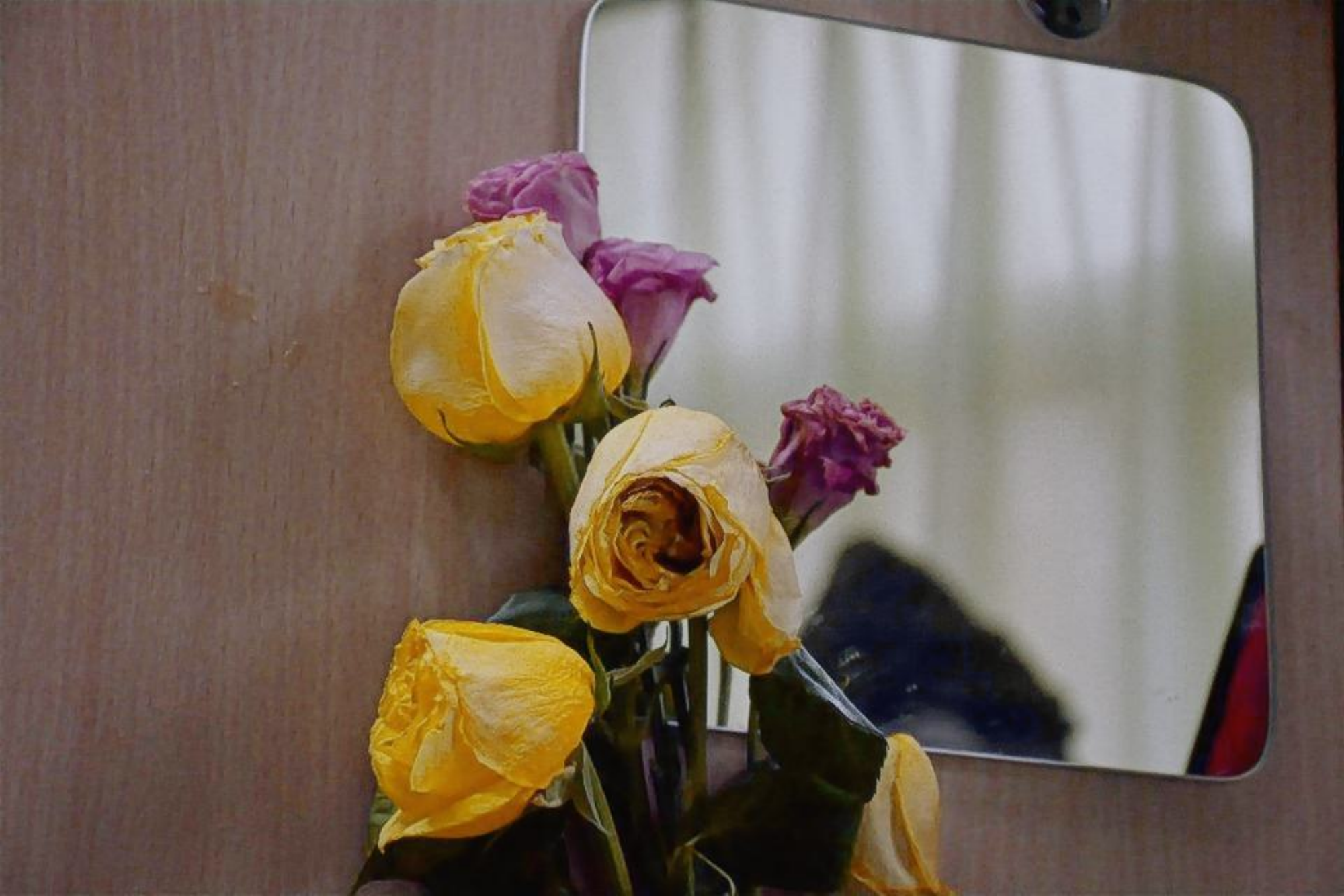}
        \includegraphics[width=\linewidth,  height=\asheight]{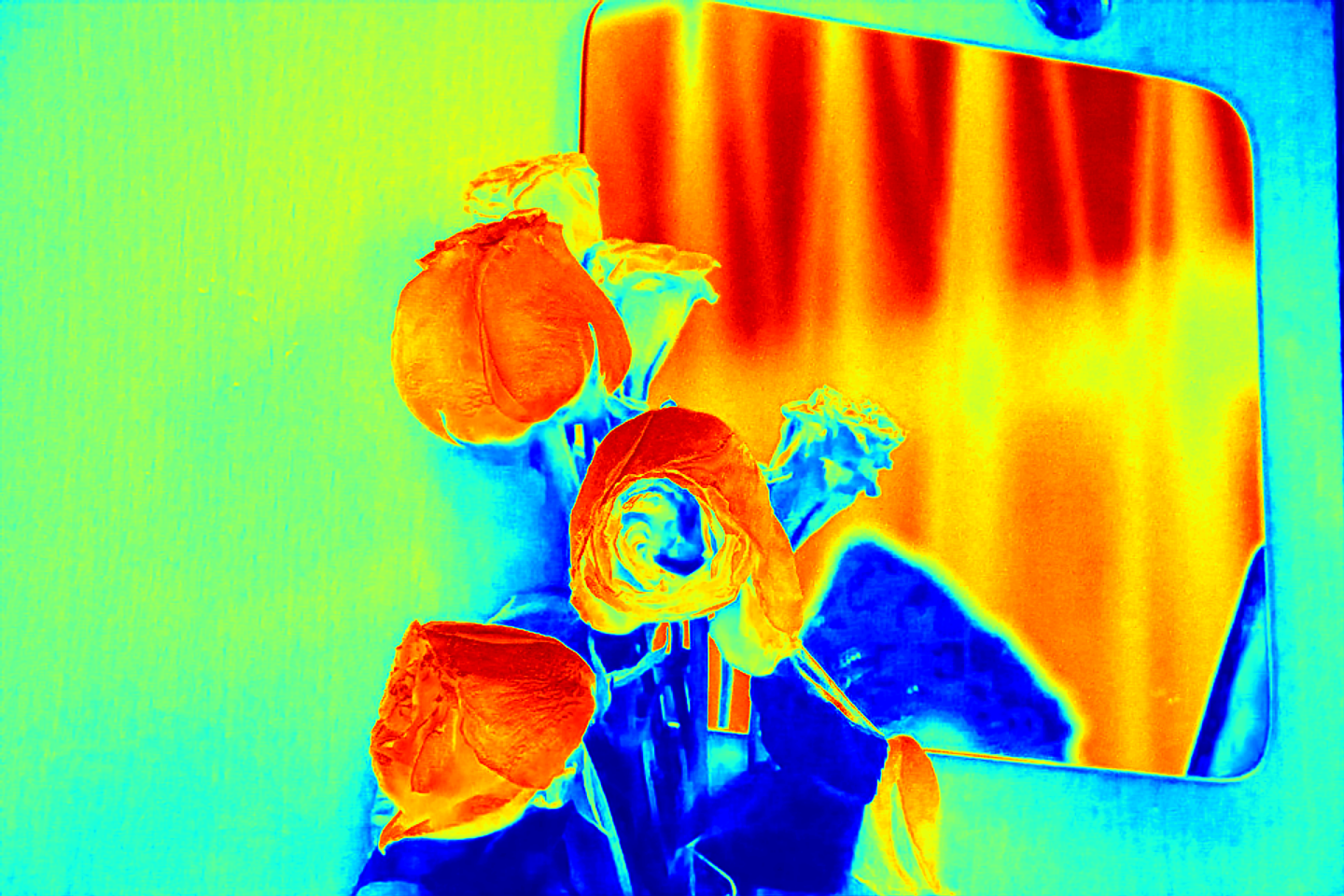}
		\caption{\scriptsize w/o CE}
		\label{ASRW1_CLLEGE}
	\end{subfigure}
	\begin{subfigure}{0.19\linewidth}
		\centering
		\includegraphics[width=\linewidth,  height=\asheight]{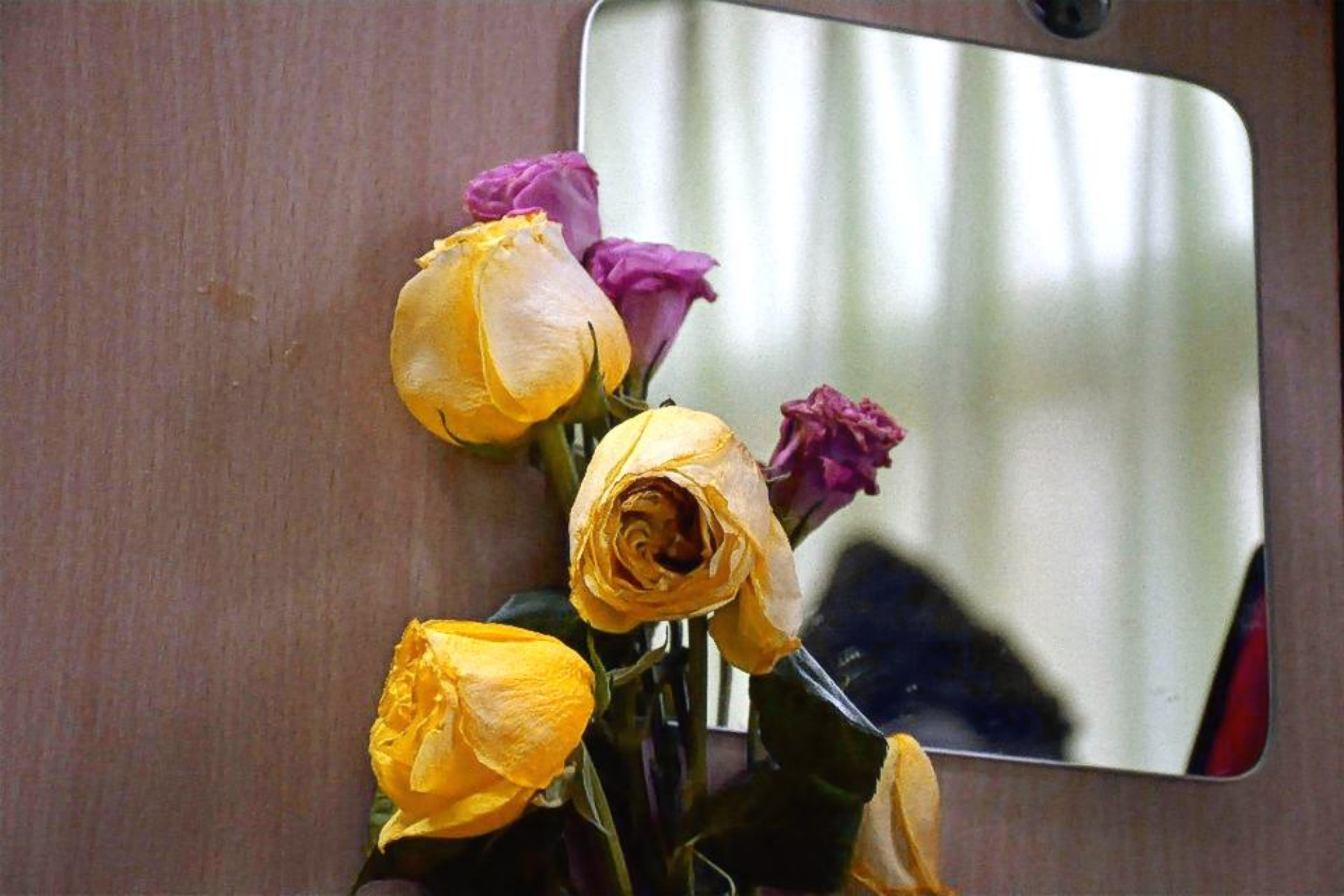}
        \includegraphics[width=\linewidth,  height=\asheight]{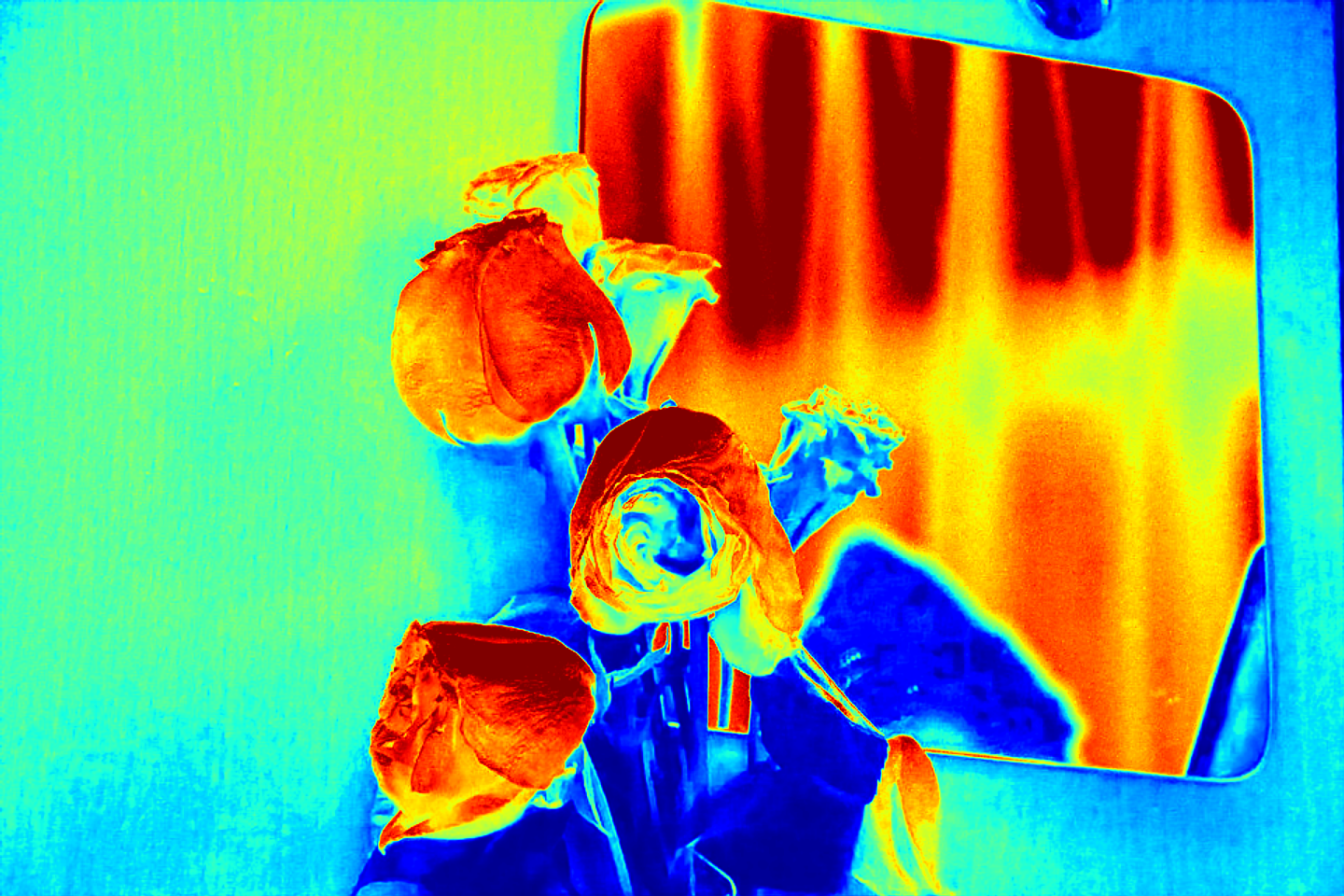}
		\caption{\scriptsize Global Enhancement}
		\label{ASRW1_FM}
	\end{subfigure}

    \vspace{2pt} 
    
	\begin{subfigure}{0.19\linewidth}
		\centering
		\includegraphics[width=\linewidth,  height=\asheight]{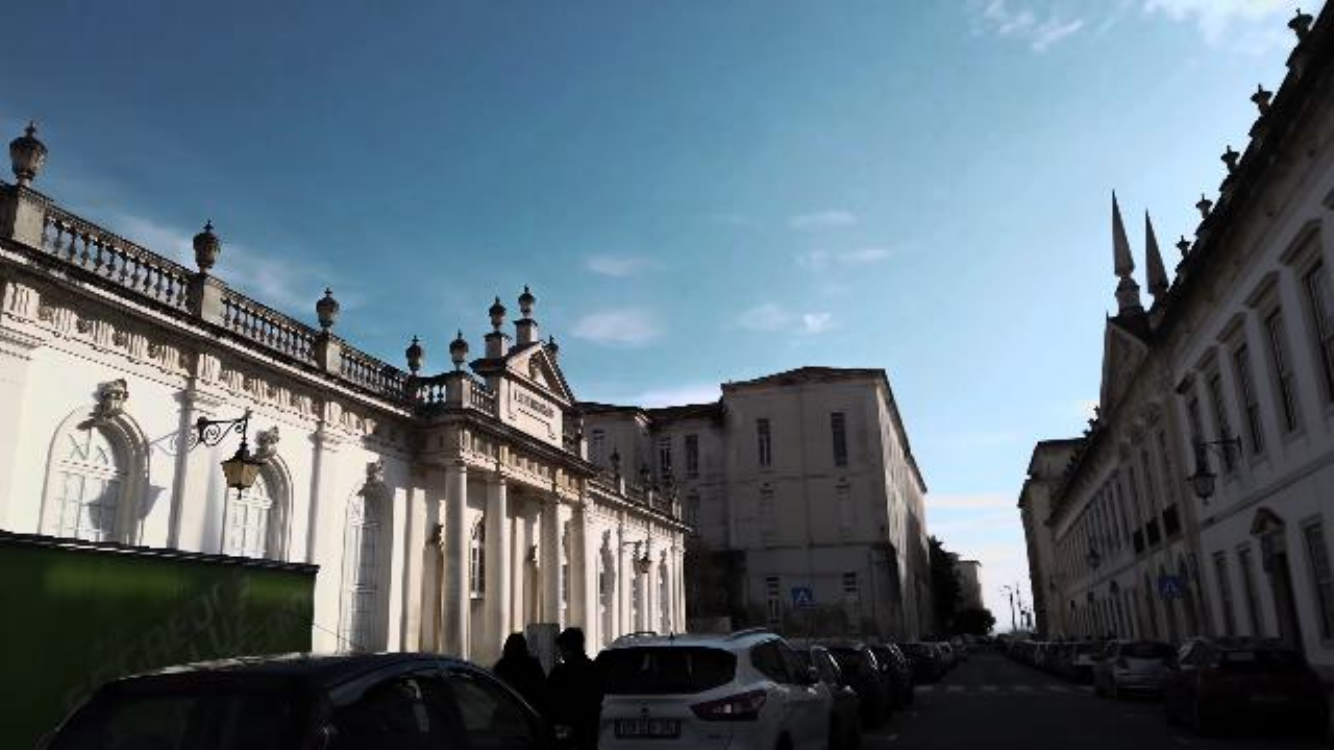} 
        \includegraphics[width=\linewidth,  height=\asheight]{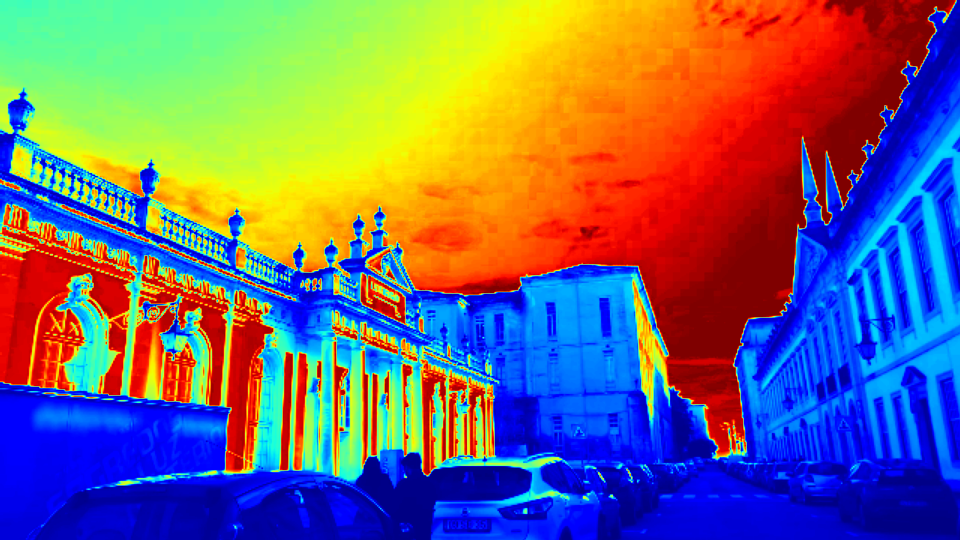} 
            \caption{\scriptsize Local Input}
	\end{subfigure}
	\begin{subfigure}{0.19\linewidth}
		\centering
		\includegraphics[width=\linewidth,  height=\asheight]{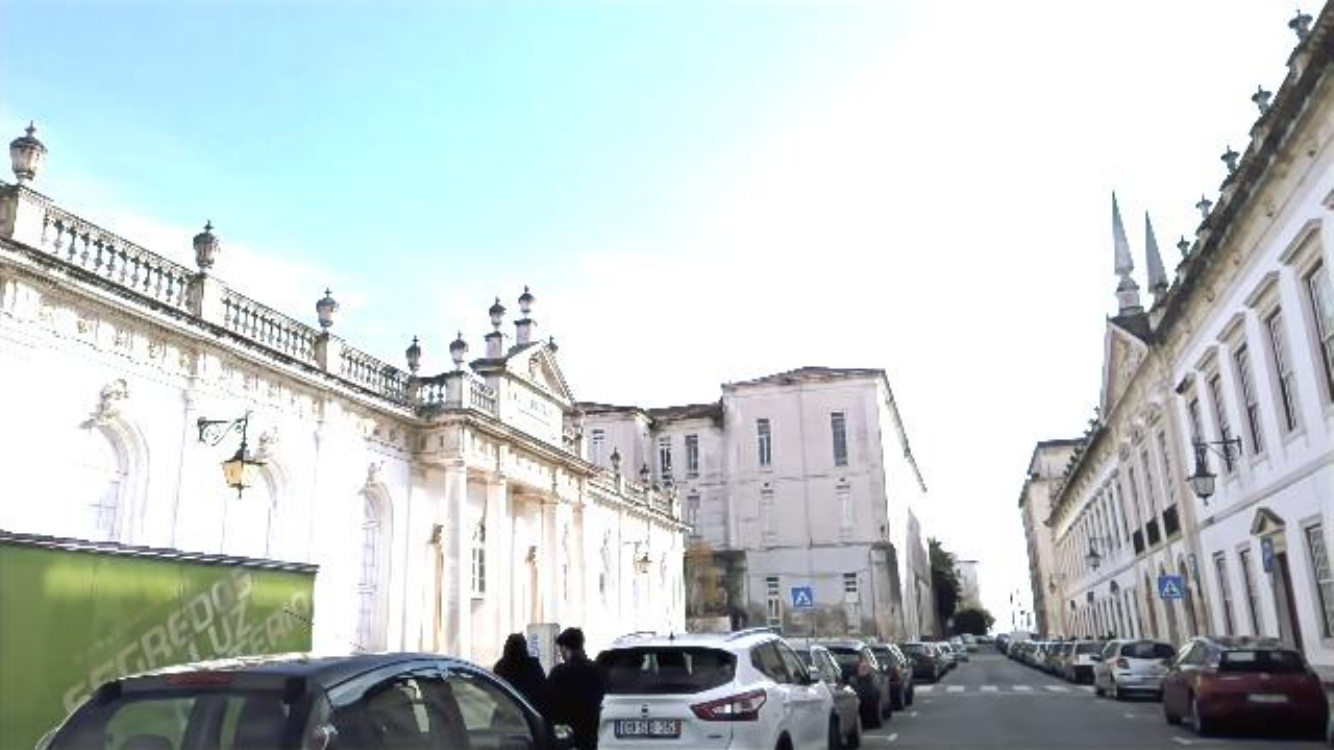} 
        \includegraphics[width=\linewidth,  height=\asheight]{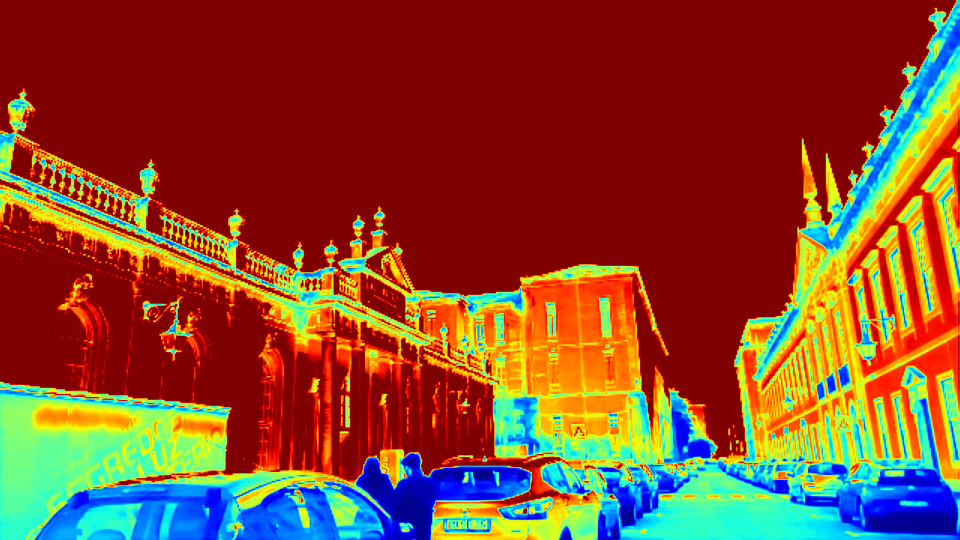} 
            \caption{\scriptsize w/o CL and LE}
	\end{subfigure}
	\begin{subfigure}{0.19\linewidth}
		\centering
		\includegraphics[width=\linewidth,  height=\asheight]{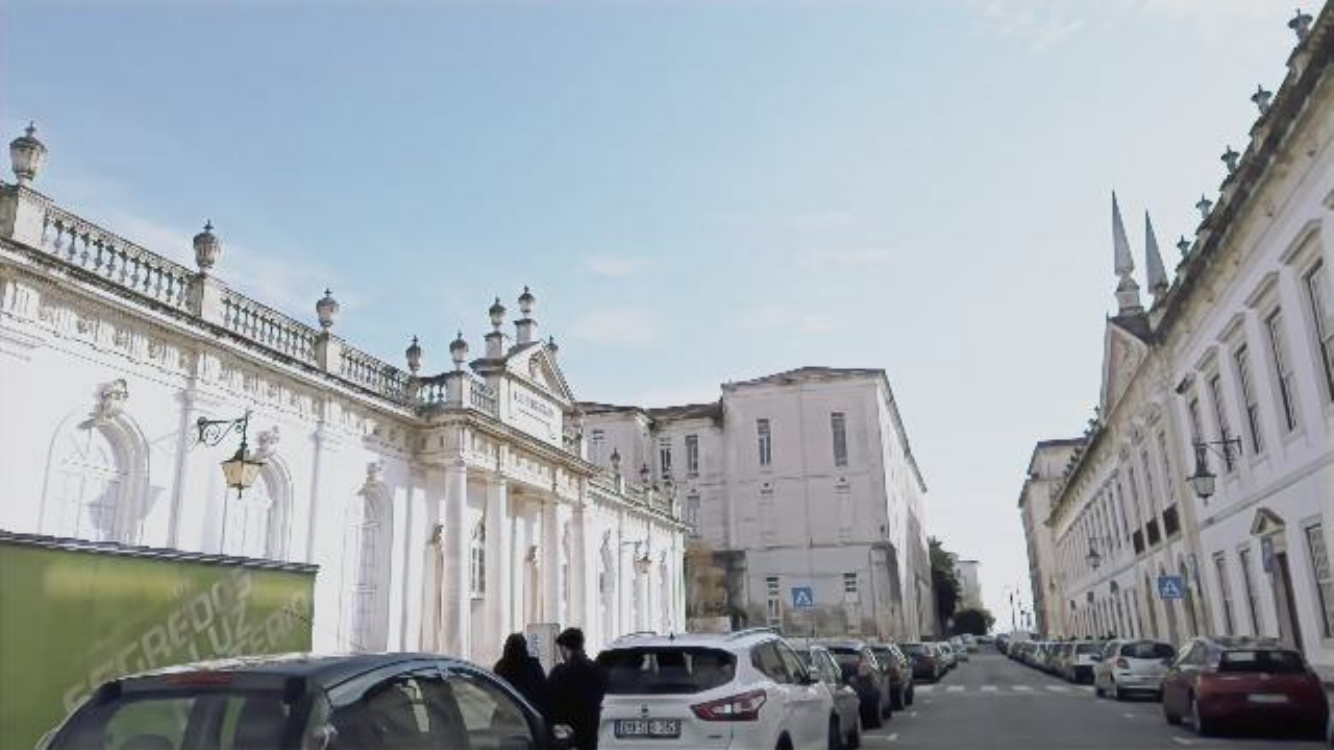}
        \includegraphics[width=\linewidth,  height=\asheight]{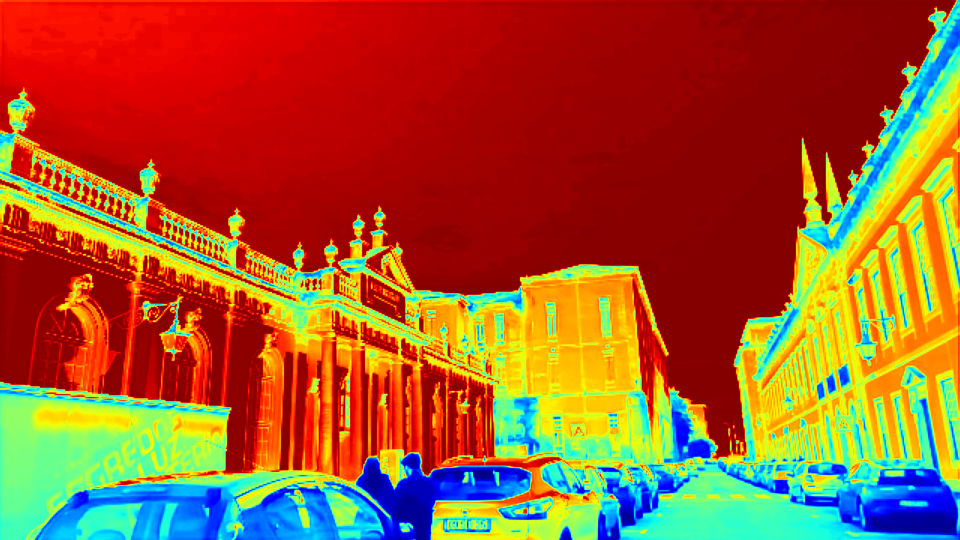}
            \caption{\scriptsize w/o CL, LE and CE}
	\end{subfigure}
	\begin{subfigure}{0.19\linewidth}
		\centering
		\includegraphics[width=\linewidth,  height=\asheight]{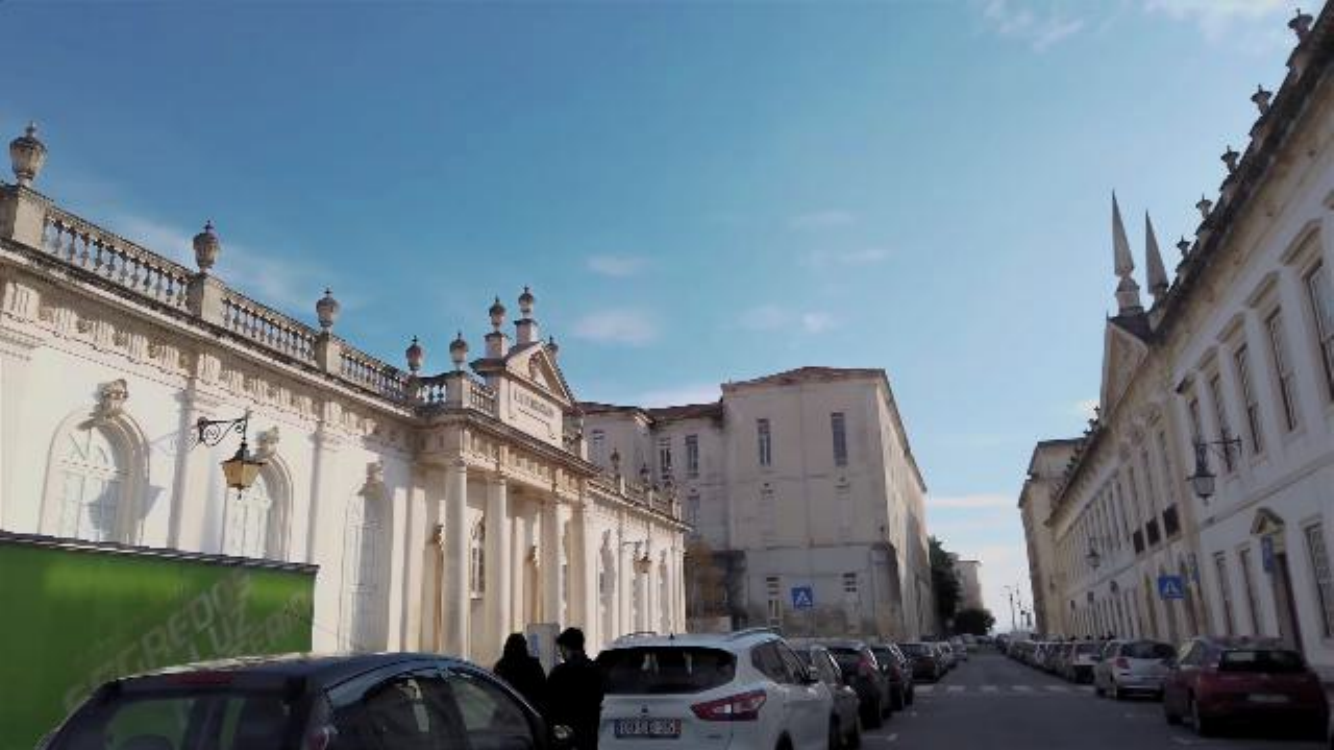}
        \includegraphics[width=\linewidth,  height=\asheight]{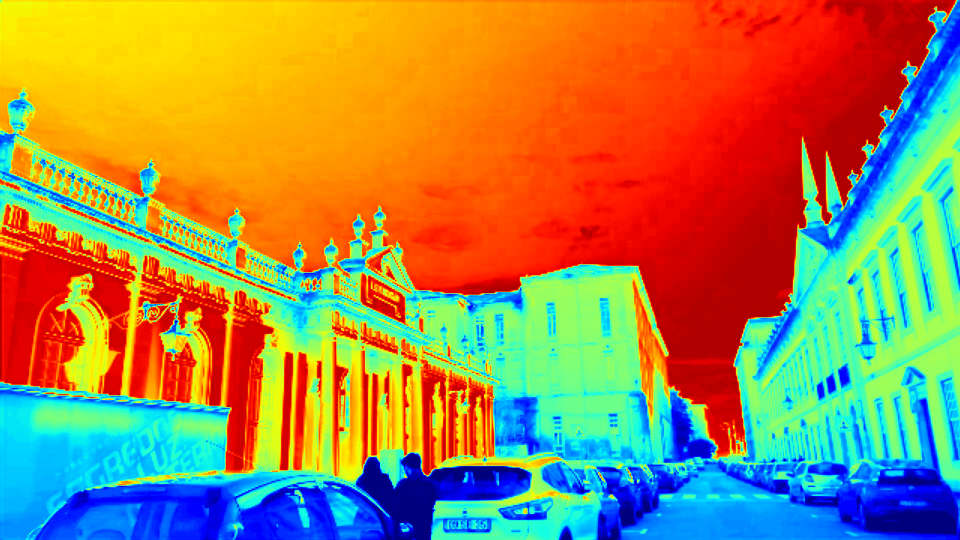}
            \caption{\scriptsize w/o CE}
	\end{subfigure}
	\begin{subfigure}{0.19\linewidth}
		\centering
		\includegraphics[width=\linewidth,  height=\asheight]{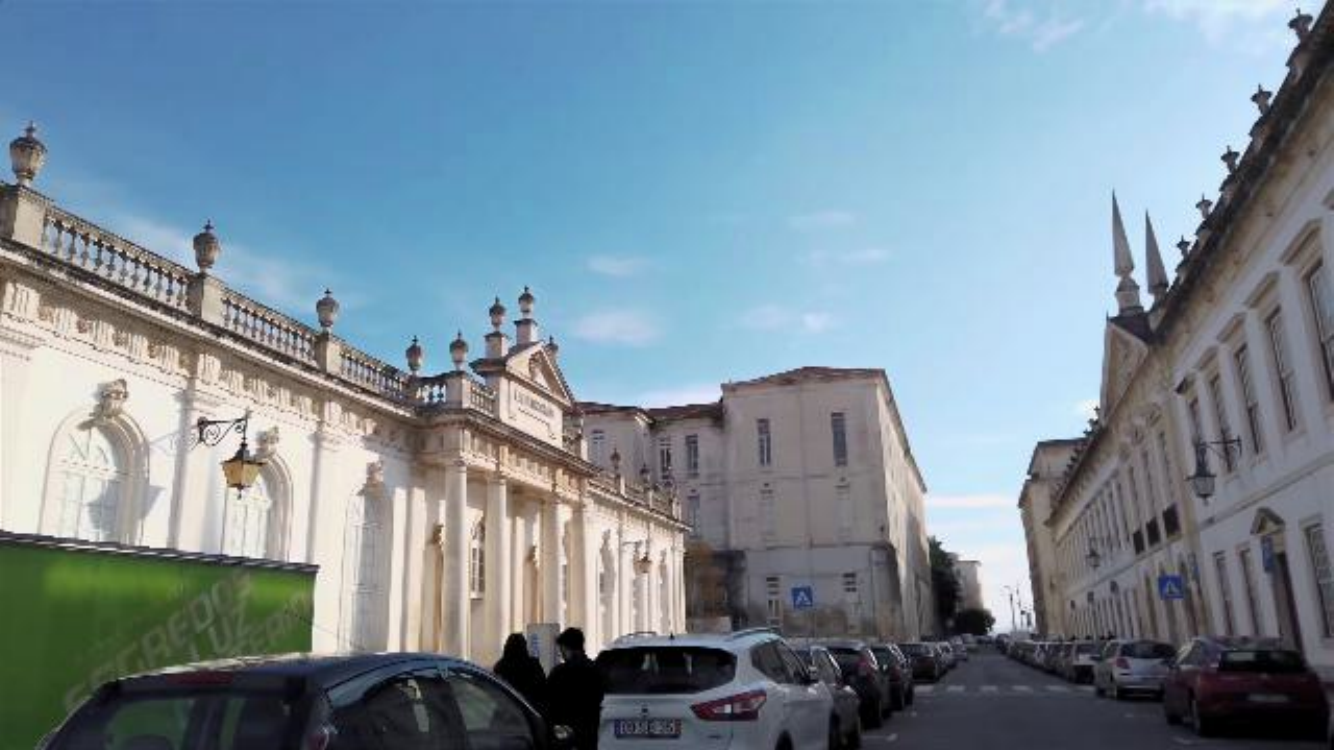}
        \includegraphics[width=\linewidth,  height=\asheight]{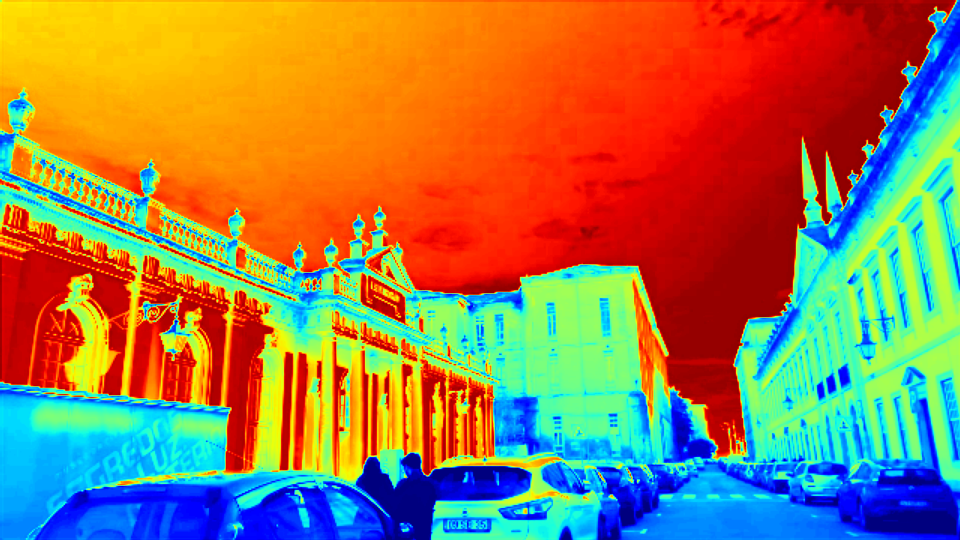}
            \caption{\scriptsize  Local Enhancement}
	\end{subfigure}
	
	\caption{Visual comparisons from the ablation study of global and local enhancement components}
	\label{AS1}
\end{figure*}

\begin{table*}[ht]
	\centering
	\caption{Quantitative evaluation results of the ablation study on the performance of $C_L$, $L_E$, $G_E$, and $C_E$ modules, evaluated using PSNR and SSIM metrics on UHD-LOL4K and LSRW dataset.}
	\label{RW1Ablation}
    \resizebox{1.5\columnwidth}{!}{
    \begin{tabular}{lcccccccc}
		\toprule
		\multicolumn{1}{c}{\multirow{2}{*}{Model}} & \multicolumn{4}{c}{Components}  & \multicolumn{2}{c}{Model Complexity} & \multicolumn{2}{c}{Metrics}\\
		\cmidrule(lr){2-5} \cmidrule(lr){6-7} \cmidrule(lr){8-9}
		& $C_L$ & $L_E$ & $G_E$ & $C_E$  & \# Parameters $\downarrow$ & FLOPs (G) $\downarrow$ & PSNR $\uparrow$ & SSIM $\uparrow$\\
		\midrule
		Model w/o $C_L$ and $G_E$ & & $\checkmark$ &  & $\checkmark$  & 1.30 (M) & 799.87 & 17.8703 & 0.7352\\
		
		Model w/o $C_L$ and $L_E$ &  &  & $\checkmark$ & $\checkmark$  & 2.17 (M) & 1,328.99 & 16.0125 & 0.7015\\
		
		Model w/o $C_L$, $G_E$ and $C_E$ &  & $\checkmark$& &  & \textbf{0.44 (M)} & \textbf{270.75} & 16.6347 & 0.7229\\
		
		Model w/o $C_L$, $L_E$ and $C_E$ &  &  & $\checkmark$ &  & 1.30 (M) & 799.87 & 18.0890 & 0.7488\\
		
		Model w/o $C_E$ & $\checkmark$ & $\checkmark$ & $\checkmark$ &  &   4.81 (M) & 1,071.11 & 19.0049 & 0.7991 \\
		
		Full Model & $\checkmark$ & $\checkmark$ & $\checkmark$ & $\checkmark$  & 5.67 (M) & 1,600.23 & \textbf{20.1811} & \textbf{0.8015} \\
		\bottomrule
	\end{tabular}
 }
\end{table*}

In the comparison of night-time images in \ref{Q1}, methods such as RUAS, HWMNet, SCI, IAT, and EMNet tend to overexpose the scene, leading to a significant loss of detail. LLFlow, URetinex, GSAD, PPformer, and MSATr generate noise, especially in the sky, and reduce color saturation. RetinexNet, UTVNet, PairLIE, and SHAL-Net distort colors and introduce noise, with overexposure in some cases, such as with SHAL-Net. LIME and CLIP-LIT present noise despite adequate lighting, while BL reduces noise well but offers limited lighting improvement. Methods like Zero-DCE, GCP, NeurBR, ITRE, PIE, and ALEN strike the best balance between lighting, color, and noise reduction.

In the day-time scene shown in \ref{Q2}, RUAS and SCI overexpose the entire image, while UTVNet, RetinexNet, and SHAL-Net distort colors, reducing visual quality. IAT, URetinex, BL, and EMNet eliminate details in bright areas such as the sky, whereas HWMNet, LLFlow, GSAD, and MSATr fail to illuminate certain regions properly. Zero-DCE, PairLIE, and PPformer maintain a good lighting balance but produce dull colors. Finally, LIME, CLIP-LIT, GCP, NeurBR, ITRE, PIE, and ALEN achieve the best compromise between lighting, detail, and color.

\begin{table*}[ht]
	\centering
	\caption{Comparison of computational cost for GPU-based methods.}
	\label{GPU}
	\resizebox{2.05\columnwidth}{!}{
		\begin{tabular}{l c c c c c c}
			\hline
			Method & Year & Inference Speed (s) $\downarrow$ & \#Parameters  $\downarrow$ & FLOPs (G) $\downarrow$ & Memory Usage (MB) $\downarrow$ & Platform \\ \hline
			RetinexNet~\cite{wei2018deep} & 2018  & 1.6624	&	0.44 (M)	 & \textbf{0.0011} &	\textbf{0.4240}	&	TensorFlow	\\
			Zero-DCE~\cite{guo2020zero} & 2020 	& 0.2595 &	0.08 (M)	& 41.06 &	818.64	&	Pytorch	\\
			RUAS~\cite{Liu21} & 2021 &  0.2856	&	3,438	& 2.01 &	830.40	&	Pytorch	\\
			UTVNet~\cite{zheng2021adaptive} & 2021 	& 0.4605	&	7.75 (M)	& 212.86 &	882.21 	&	Pytorch	\\
			HWMNet~\cite{fan2022half} & 2022  	&  0.9067	&	66.56 (M)	& 2,085.38 &	4,831.06	&	Pytorch	\\
			LLFlow~\cite{wang2022lowlight} & 2022 	& 0.2314 	&	5.43 (M)	& (unavailable) &	325.19	&	Pytorch	\\
			SCI~\cite{ma2022toward} & 2022 	& \textbf{0.1907}	&	\textbf{258} 	&  0.28 &	42.19	&	Pytorch	\\
			IAT~\cite{cui2022you} & 2022   & 0.3952	 &	0.09 (M)	& 11.35 &	9,015.92	&	Pytorch	\\
            URetinex~\cite{wu2022uretinex} & 2022  & 0.5657 &	0.34 (M)	& 533.75 &	1,412.43	&	Pytorch	\\
            BL~\cite{ma2023bilevel} & 2023 & 0.5991 &	38.81 (M)	& 979.82 &	5,538.39	&	Pytorch	\\
			PairLIE~\cite{fu2023learning} & 2023   & 0.6670 &	0.34 (M)	& 176.77 &	476.45	&	Pytorch	\\
			CLIP-LIT~\cite{liang2023iterative} & 2023  & 0.2628	&	0.28 (M)	& 144.01 &	1,062.87	&	Pytorch	\\
            EMNet~\cite{ye2023glow} & 2023 & 1.1672 & 12.54 (M) &  (unavailable) & 3,601.44 &	Pytorch\\
			SHAL-Net~\cite{xu2024degraded} & 2024  &  0.5505	&	 0.27 (M)	& 163.43 &	1,565.92	&	Pytorch	\\
            GSAD~\cite{hou2024global} & 2024  & 3.6568	&	17.43 (M)	&	(unavailable)	& 1,969.35 &	Pytorch	\\
            PPformer~\cite{dang2024ppformer} & 2024 & 0.8016 & 0.12 (M) &  (unavailable) & 1,244.70 & Pytorch \\
            NeurBr~\cite{zhao2024non} & 2024  & 0.9151 &	0.17 (M)	& (unavailable) &	5.89	&	Pytorch	\\
            PIE~\cite{liang2024pie} & 2024  &  0.2152	&  0.08 (M)	& 48.66 &	818.64	&	Pytorch	\\
            MSATr~\cite{fang2024non} & 2024 & 0.4272 & 13.43 (M) & (unavailable) & 1,156.56 & Pytorch \\ 
			ALEN (Proposed) & 2025  & 0.7969 &	5.67 (M) & 1,600.23  &	2,776.67	&	Pytorch	\\ \hline
		\end{tabular}
	}
\end{table*}

 \begin{table}[ht]
    \centering
    \caption{Comparison of computational cost for CPU-based methods.}
    \label{CPU}
    \begin{tabular}{lccc} 
        \toprule
        Method & Year & Inference Speed (s) $\downarrow$ & Platform \\ 
        \midrule
        LIME~\cite{guo2016lime} & 2016 & 6.8984 & Python \\
        GCP~\cite{jeon2024low} & 2024 & \textbf{0.1704} & Python \\
        ITRE~\cite{wang2024itre} & 2024 & 4.2325 & Matlab \\ 
        ALEN (Prop.) & 2025 & 20.9679 & Pytorch \\ 
        \bottomrule
    \end{tabular}
\end{table}

\subsection{Ablation study}
The ablation study results in Table \ref{RW1Ablation} highlight the impact of each ALEN component on image enhancement, using PSNR, SSIM, and model complexity (parameters and FLOPs) as metrics. The complete model, which integrates all components $C_L$, $L_E$, $G_E$, and $C_E$, achieved the highest performance with a PSNR of 20.1811, SSIM of 0.8015, 5.67 million parameters, and 1,600.23 GFLOPs, showing the benefits of including all modules despite the increased complexity.

Excluding certain components led to significant drops in performance. When $C_L$ and $G_E$ were removed, the PSNR dropped to 17.8703 and SSIM to 0.7352, while the complexity decreased to 1.30 million parameters and 799.87 GFLOPs. The worst performance was observed when $C_L$ and $L_E$ were omitted, resulting in a PSNR of 16.0125 and SSIM of 0.7015, underlining the importance of $L_E$. Excluding $C_E$ led to a PSNR of 19.0049 and SSIM of 0.7991, which still fell short of the full model's performance. The model with only $L_E$ had the lowest complexity, with 0.44 million parameters and 270.75 GFLOPs, but its PSNR was 16.6347 and SSIM 0.7229, highlighting the tradeoff between complexity and performance.

Figure \ref{AS1} provides visual comparisons showing the input image for global or local enhancements, the impact of omitting certain modules, and the complete model results. The full configuration yields the best visual quality.

\begin{figure*}[ht]
	\centering
	\begin{subfigure}{0.16\linewidth}
		\centering
		\includegraphics[width=\linewidth]{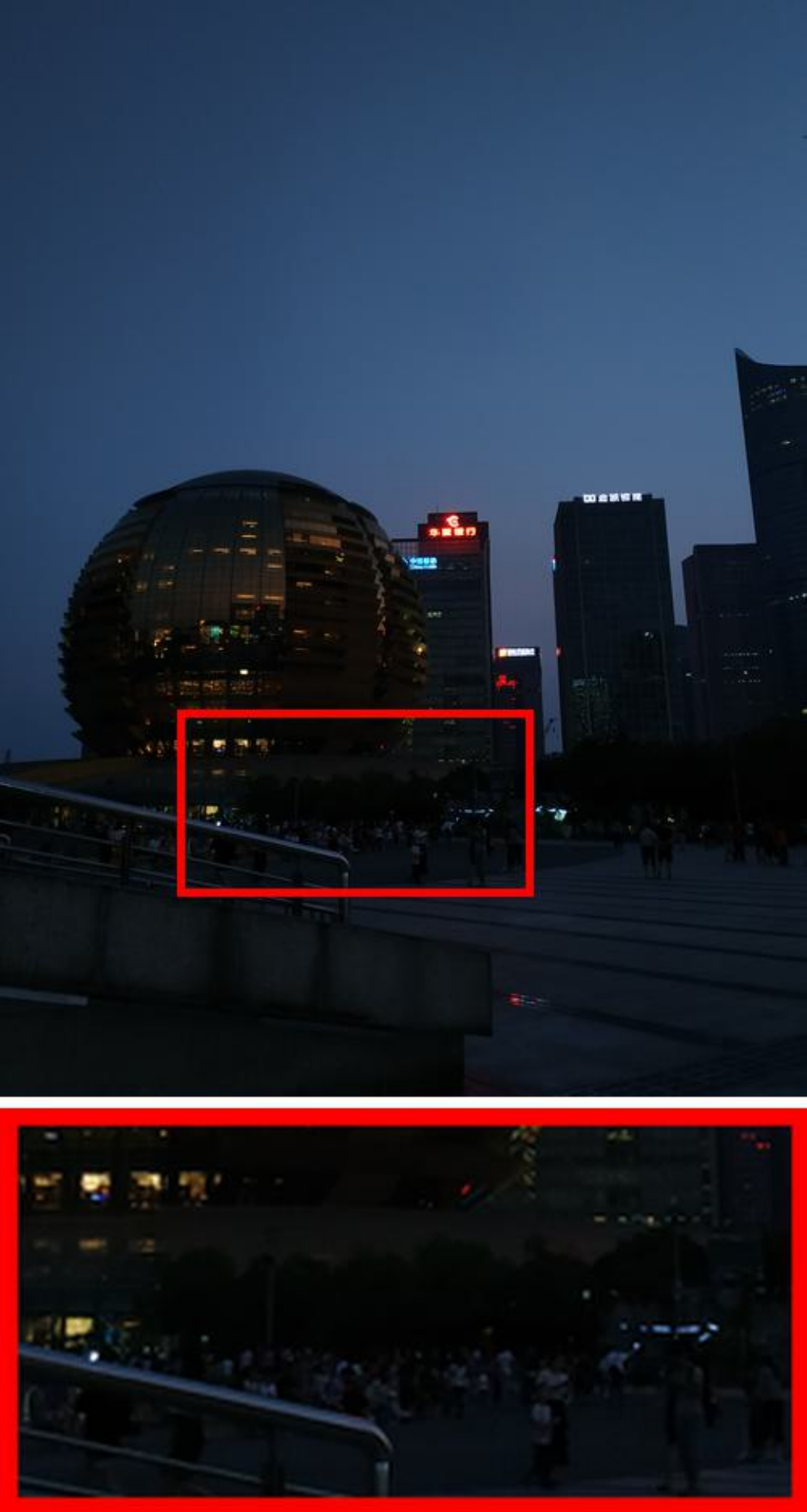} 
	\end{subfigure}
	\begin{subfigure}{0.16\linewidth}
		\centering
		\includegraphics[width=\linewidth]{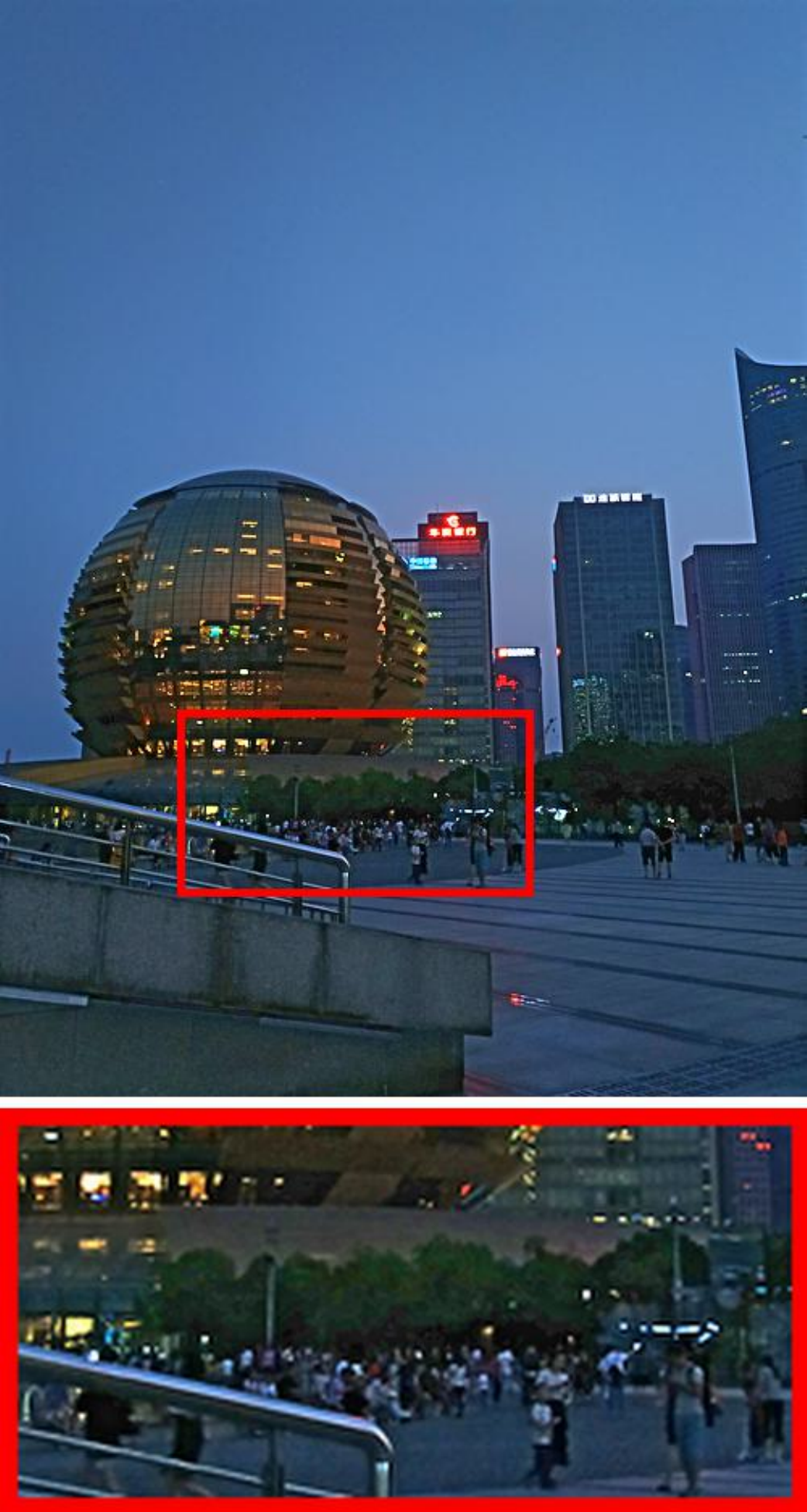} 
	\end{subfigure}
	\begin{subfigure}{0.16\linewidth}
		\centering
		\includegraphics[width=\linewidth]{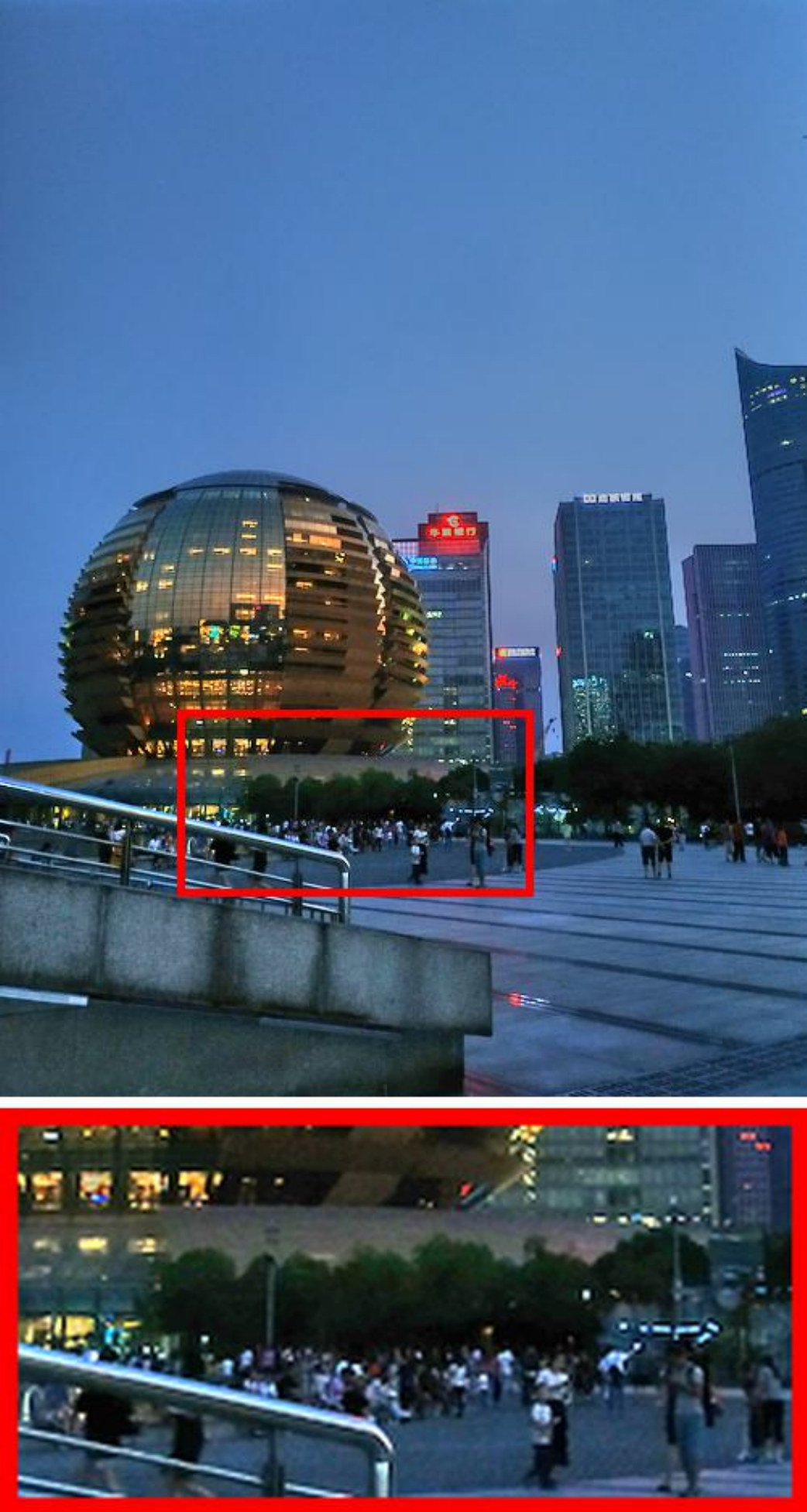}
	\end{subfigure}
	\begin{subfigure}{0.16\linewidth}
		\centering
		\includegraphics[width=\linewidth]{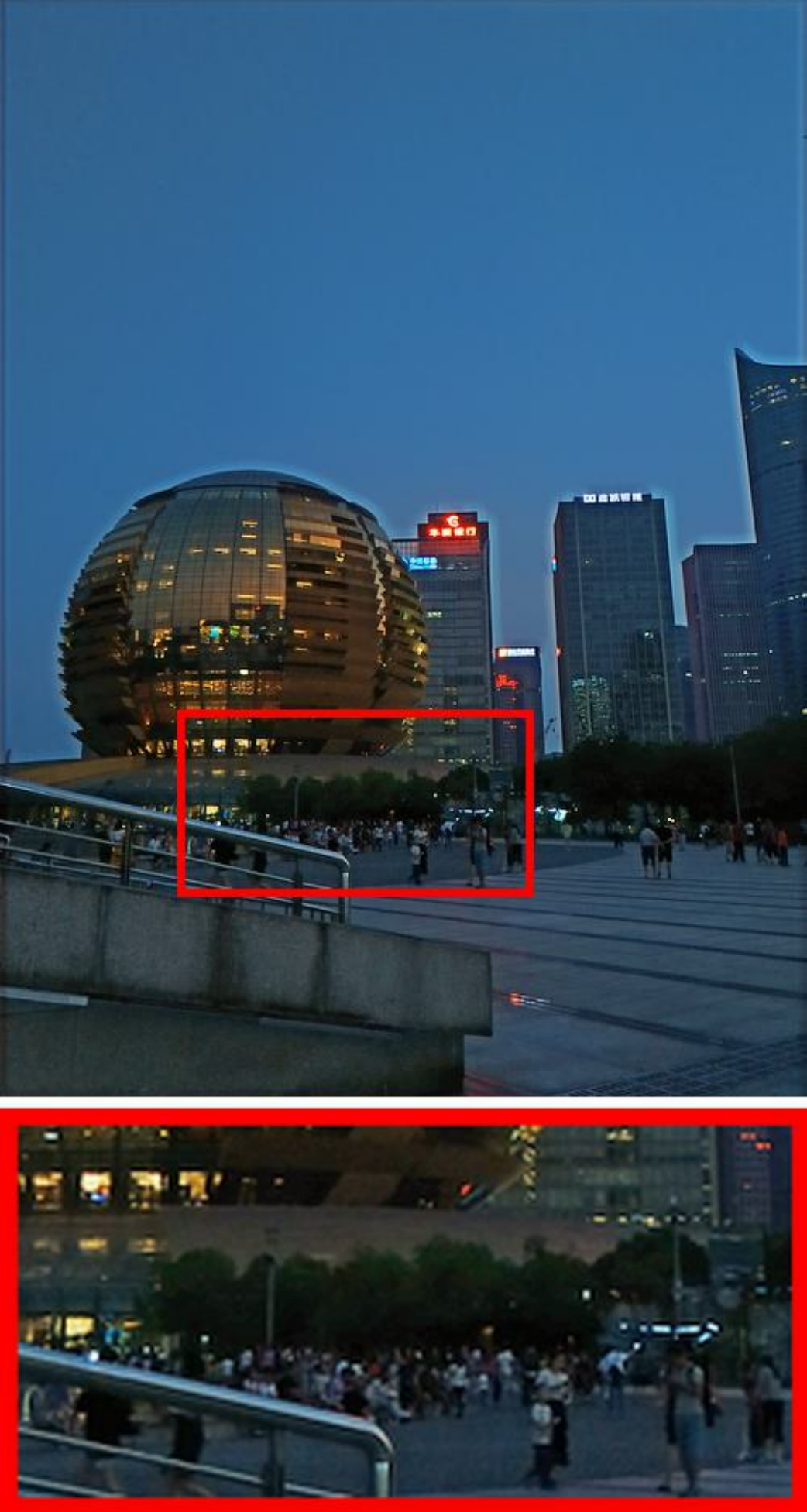} 
	\end{subfigure}
	\begin{subfigure}{0.16\linewidth}
		\centering
		\includegraphics[width=\linewidth]{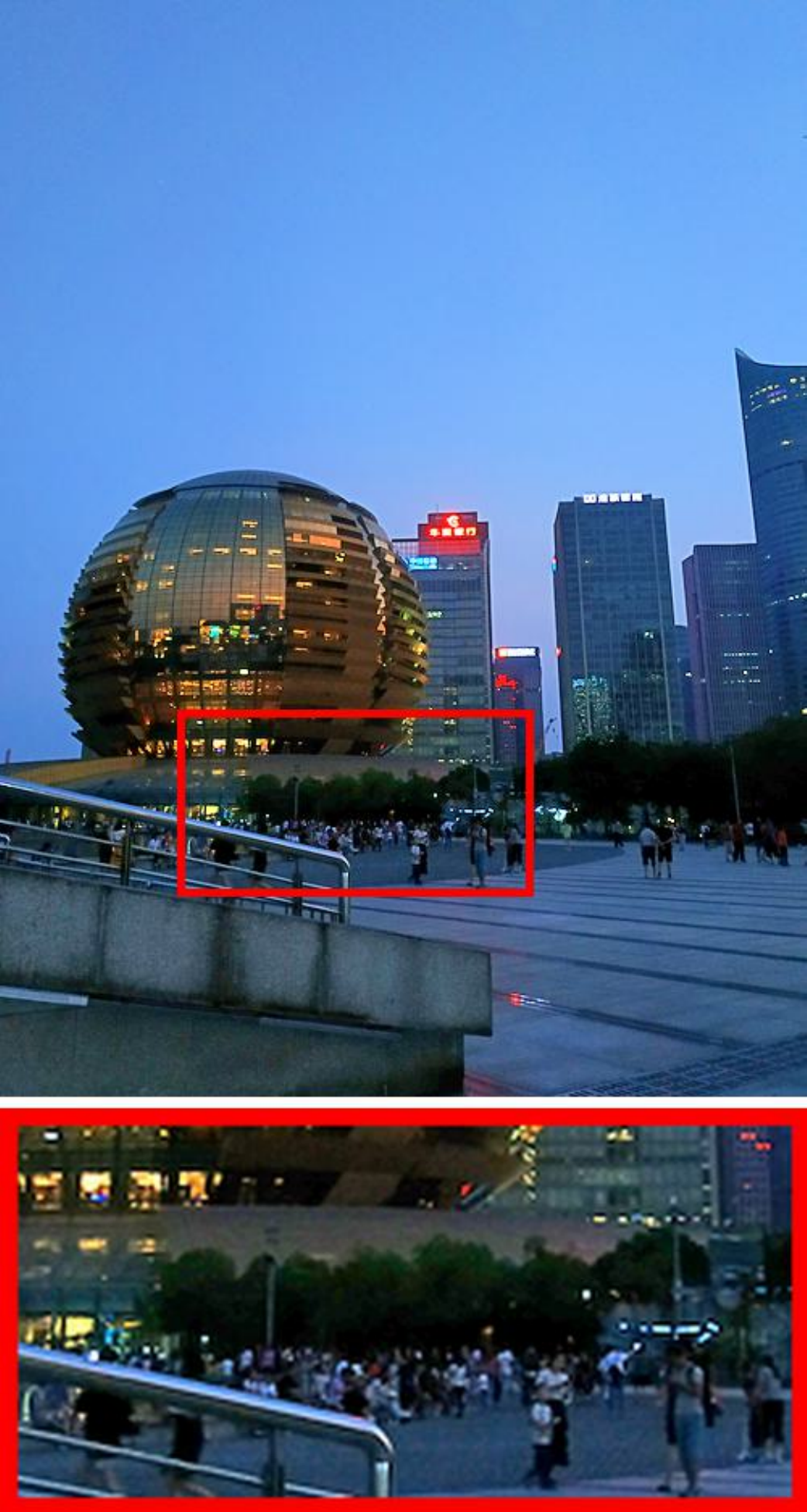} 
	\end{subfigure}
	\begin{subfigure}{0.16\linewidth}
		\centering
		\includegraphics[width=\linewidth]{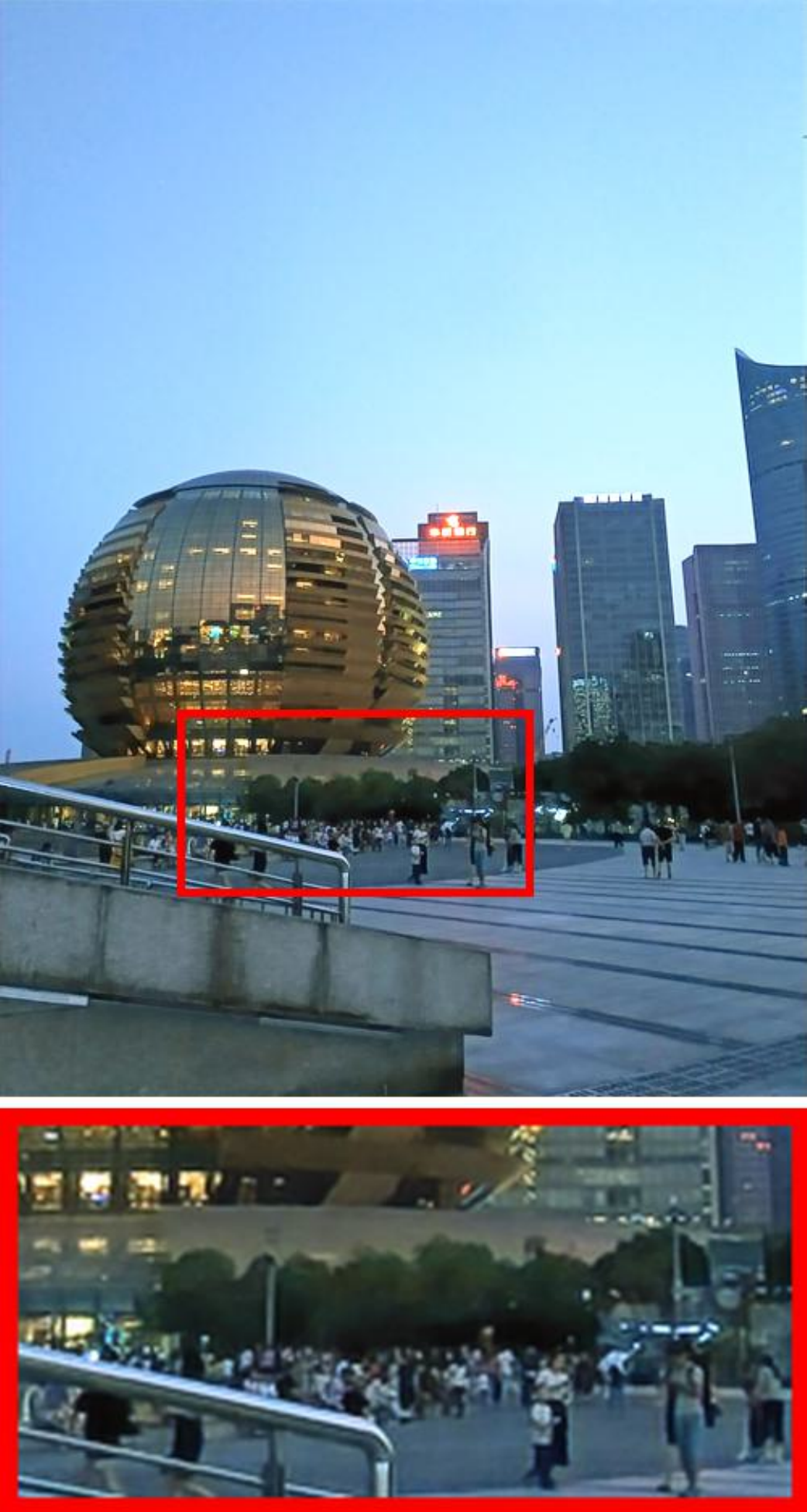}
	\end{subfigure}

    \begin{subfigure}{0.16\linewidth}
		\centering
		\includegraphics[width=\linewidth]{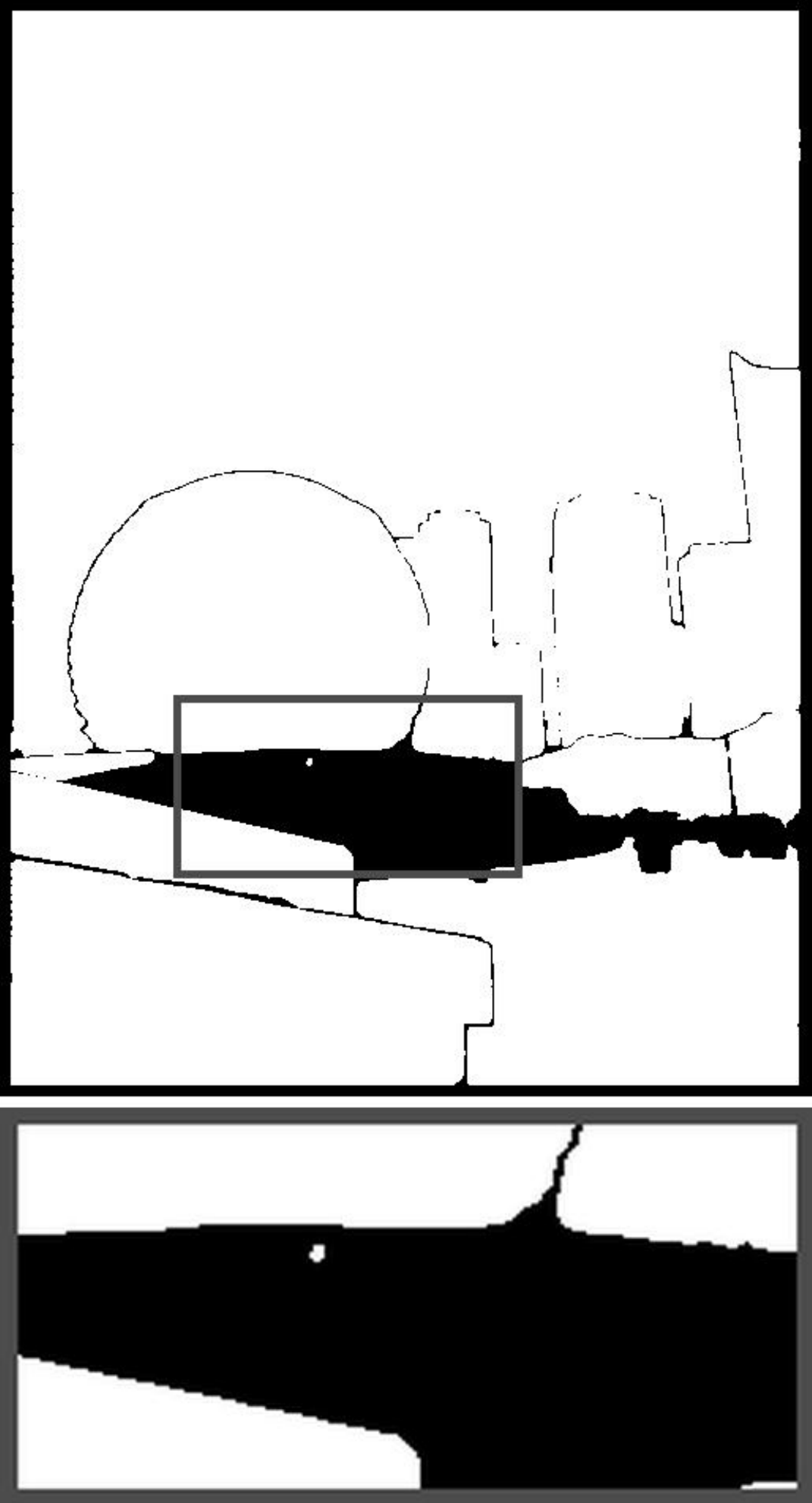} 
		\caption{\footnotesize Low-light}
	\end{subfigure}
	\begin{subfigure}{0.16\linewidth}
		\centering
		\includegraphics[width=\linewidth]{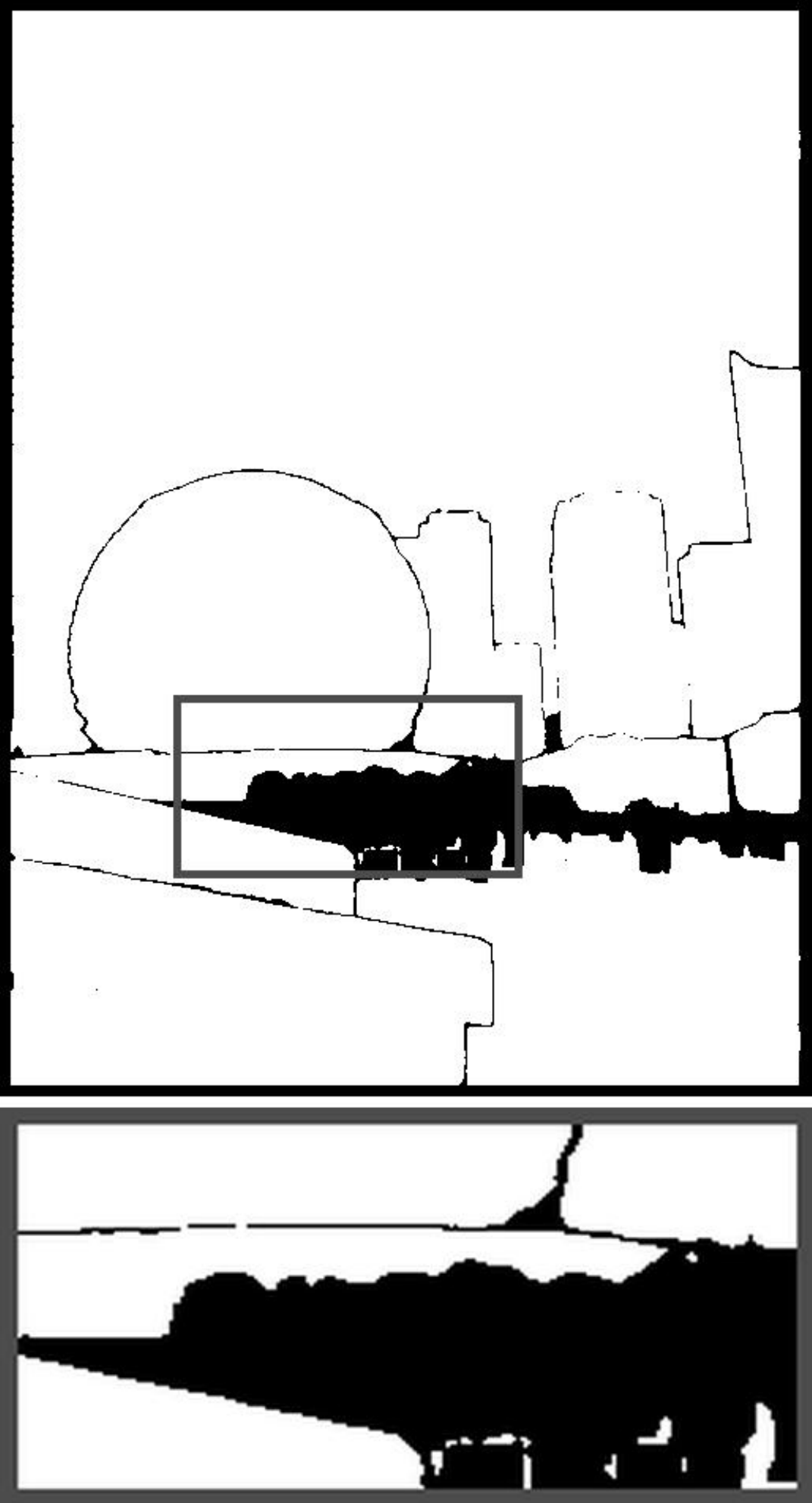} 
		\caption{\footnotesize CLIP-LIT}
	\end{subfigure}
	\begin{subfigure}{0.16\linewidth}
		\centering
		\includegraphics[width=\linewidth]{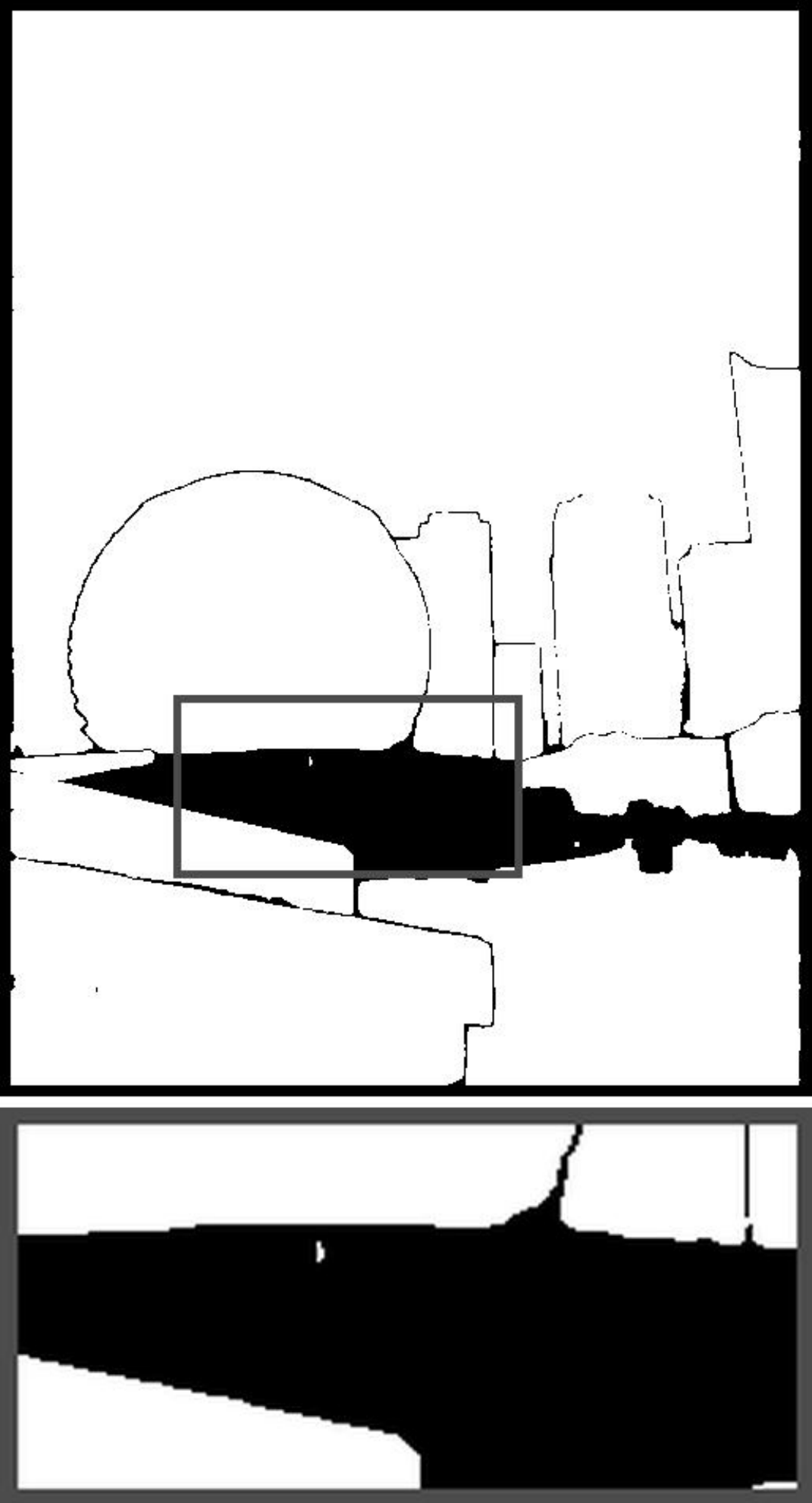}
		\caption{\footnotesize ITRE}
	\end{subfigure}
     \begin{subfigure}{0.16\linewidth}
		\centering
		\includegraphics[width=\linewidth]{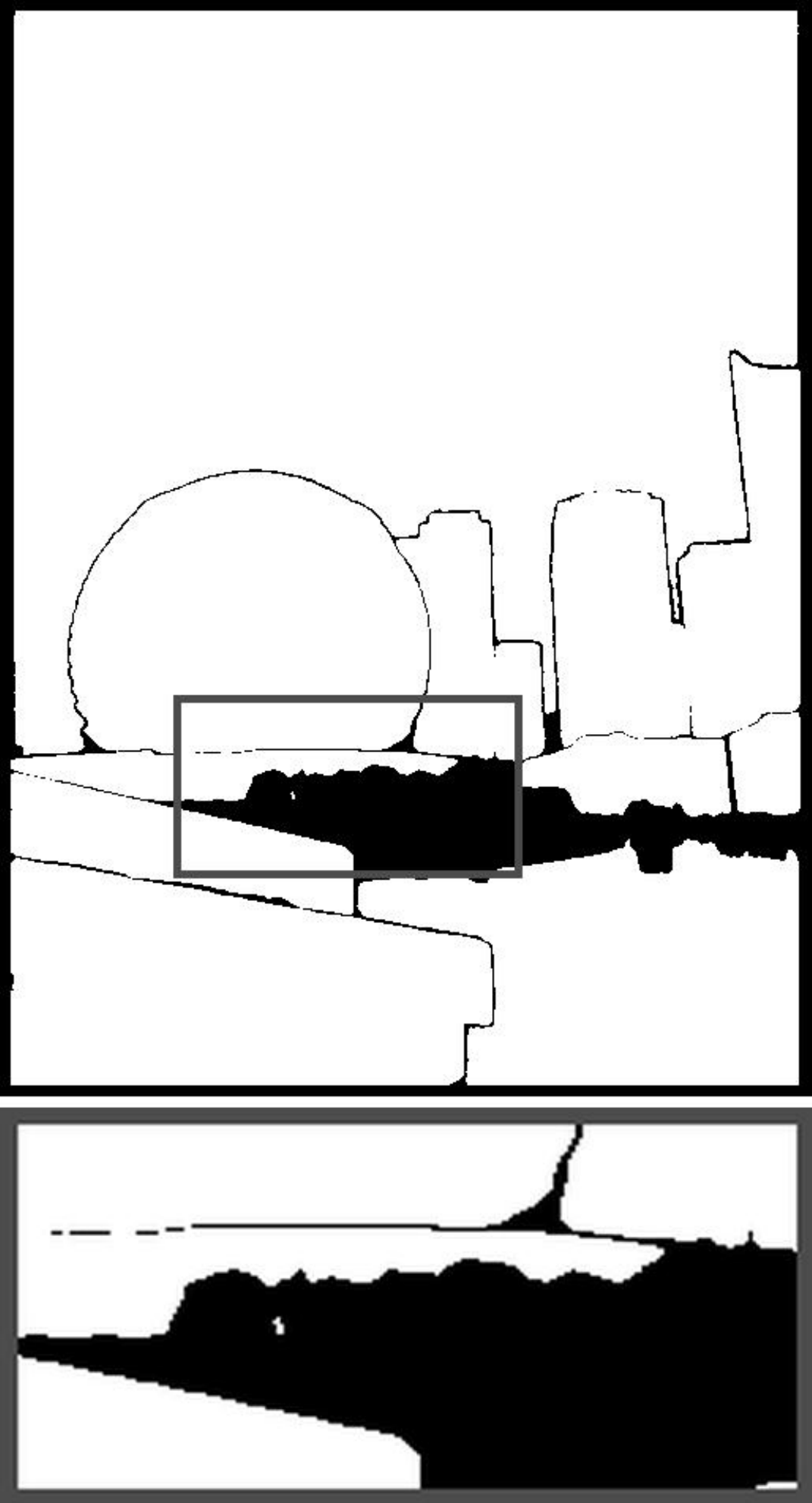} 
		\caption{\footnotesize PIE}
	\end{subfigure}
	\begin{subfigure}{0.16\linewidth}
		\centering
		\includegraphics[width=\linewidth]{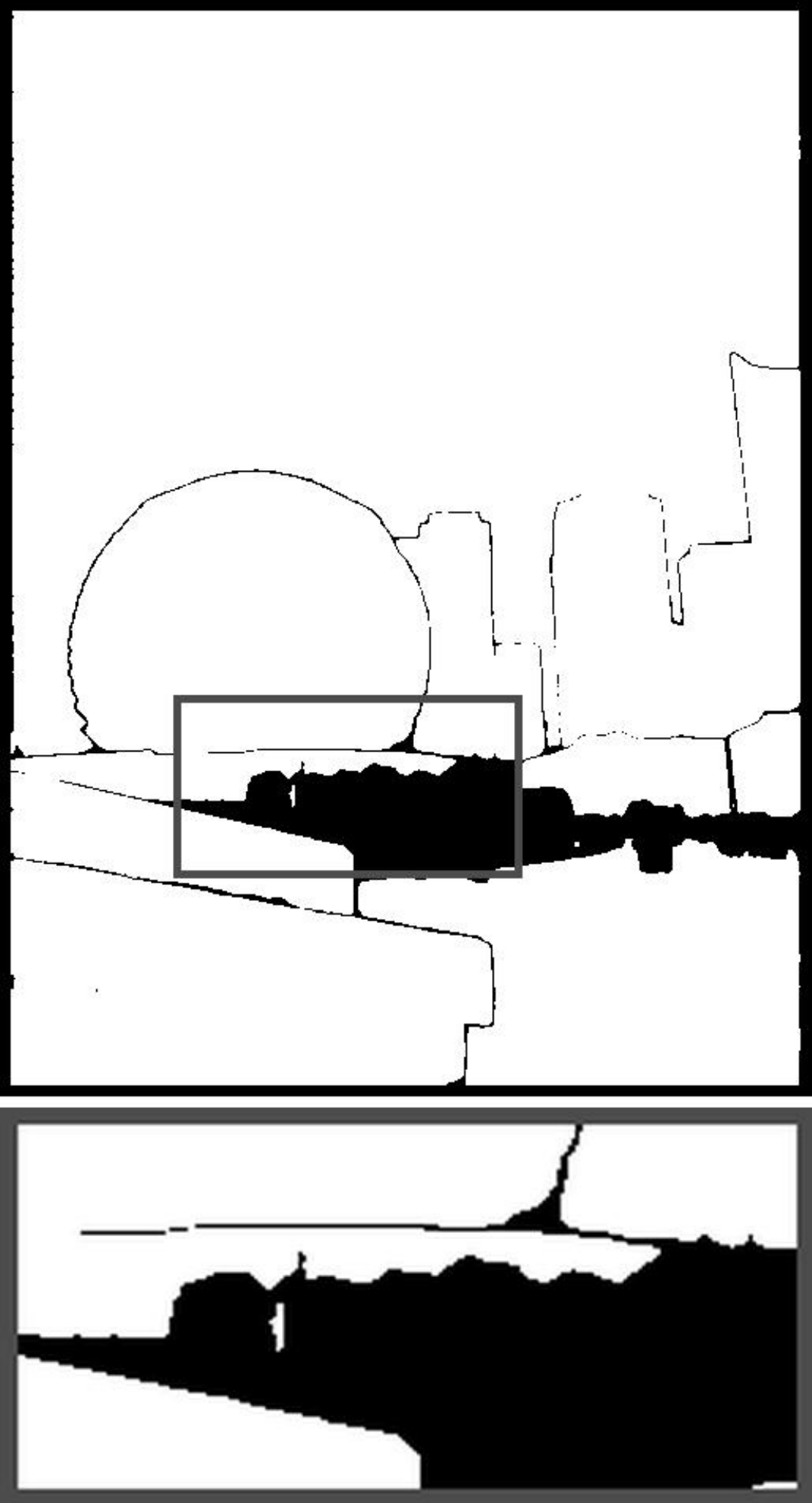} 
		\caption{\footnotesize GCP}
	\end{subfigure}
	\begin{subfigure}{0.16\linewidth}
		\centering
		\includegraphics[width=\linewidth]{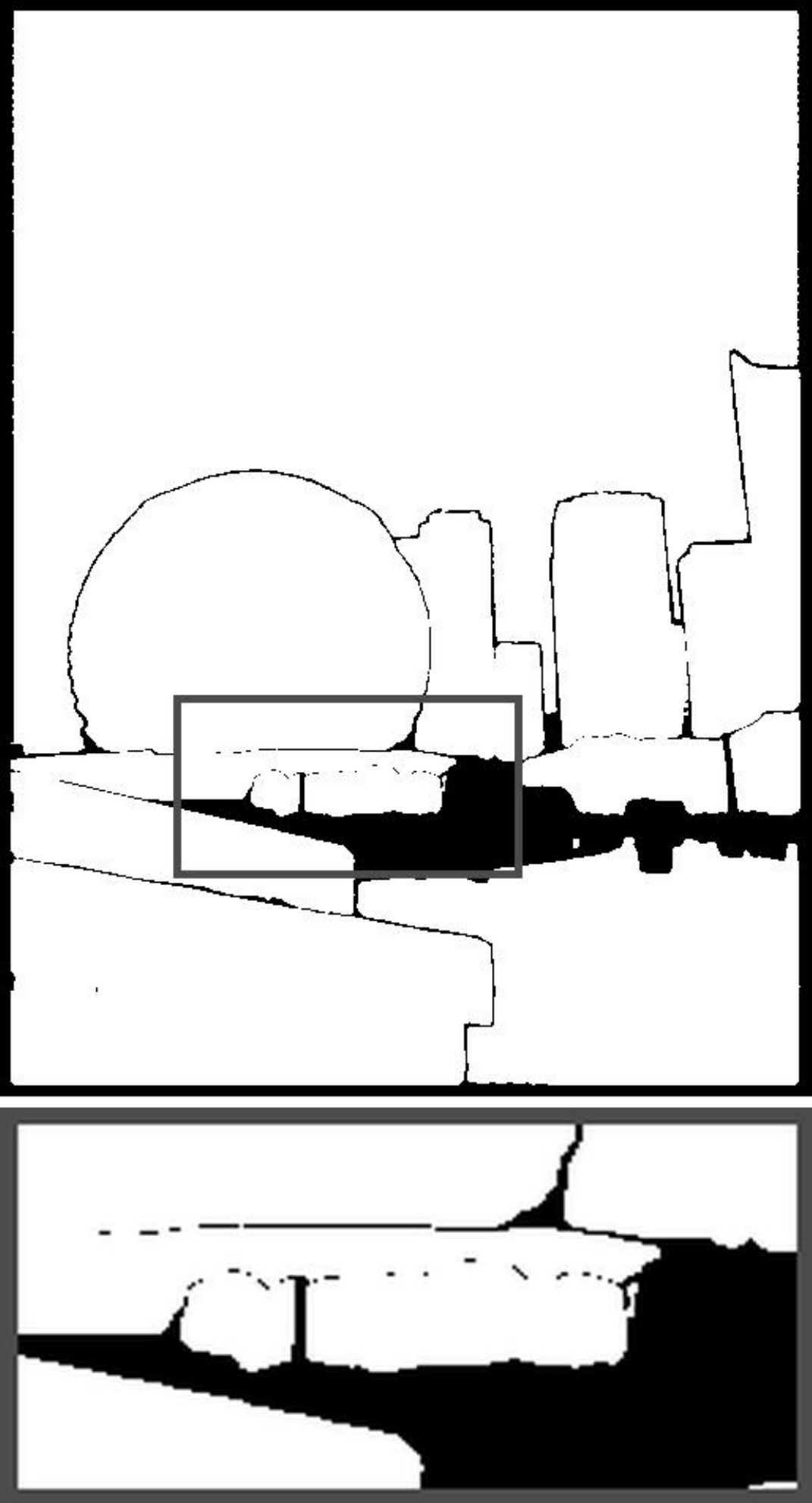}
		\caption{\footnotesize Proposed}
	\end{subfigure}

	\caption{Comparisons of different enhancement methods results for later segmentation in low-light images.}
	\label{segment}
\end{figure*}

\subsection{Computational Cost Comparison}
The analysis of Tables \ref{GPU} and \ref{CPU} highlights the performance and computational cost of various image enhancement methods under low-light conditions. All metrics are recorded using full processing on a set of 30 images, each with dimensions 960 × 640 × 3

In the comparison of GPU-based methods presented in Table \ref{GPU}, SCI stands out as the most efficient, with the fastest inference speed (0.1907 s), the fewest parameters (258), and one of the lowest numbers of FLOPs (0.28 G). Regarding memory usage, RetinexNet consumes the least (0.4240 MB), while IAT consumes the most (9,015.92 MB), a significant amount considering its low number of parameters (0.09 million). On the other hand, more advanced methods such as HWMNet and BL require more parameters and higher memory usage, increasing their computational cost. Although ALEN is not the fastest, with an inference speed of 0.7969 s and 5.67 million parameters, it demonstrates competitive performance compared to other state-of-the-art approaches.

In the evaluation of CPU-based methods presented in Table \ref{CPU}, GCP stands out with the best inference speed (0.1704 s), outperforming LIME (6.8984 s) and ITRE (4.2325 s). On the other hand, ALEN has an inference speed of 20.9679 s on CPU, indicating a significantly higher computational cost than other CPU-based methods.

Overall, SCI is considered the method with the lowest computational cost. However, when generalizing enhancement across different datasets, its results are not the most optimal, as ALEN significantly outperforms it, despite its higher computational cost. On the other hand, although ALEN provides good performance in low-light image enhancement, its application should be evaluated based on the usage environment and the availability of computational resources.

\subsection{Improving high-level vision tasks}
The proposed method, ALEN, demonstrates promising potential as a preprocessing technique to enhance semantic segmentation in low-light conditions. To evaluate this application, the DIS dataset served as the basis for comparison with four state-of-the-art methods: CLIP-LIT (2023), ITRE (2024), PIE (2024), and GCP (2024). The impact of these methods and ALEN was evaluated by implementing them as preprocessing steps before segmentation. The segmentation was conducted using the Segment Anything Model (SAM)~\cite{kirillov2023segment}, which features an image encoder based on a pre-trained Vision Transformer, a prompt encoder that supports multiple inputs, and a mask decoder for precise segmentation. Binary masks were generated to evaluate the segmentation of each image enhanced by the different methods.

Figure \ref{segment} presents an example from the DIS dataset. The first row shows the low-light input image followed by the versions enhanced by each method. The second row displays the corresponding segmentations. It can be observed that in the original low-light image, the areas containing trees are not segmented adequately, an issue that is also seen with the CLIP-LIT, ITRE, PIE, and GCP methods. However, the ALEN method achieves a notable improvement in segmentation, accurately capturing the trees in the image.

These results highlight the significant advantages of employing ALEN as a preprocessing step before applying SAM, demonstrating a substantial enhancement in the segmentation quality of images captured under challenging low-light conditions. By effectively addressing the limitations posed by poor illumination, ALEN enables SAM to achieve more precise and reliable segmentation results. This improvement surpasses the performance of several state-of-the-art methods introduced in 2023 and 2024, highlighting the robustness and superiority of the proposed approach for processing images under challenging lighting conditions.

\section{Conclusions and future work}
\label{CF}

This article introduces ALEN, a novel method specifically designed to enhance the visual quality of images captured under low-light conditions. By integrating global and local enhancements through a lighting classification network and convolutional estimator networks, ALEN has demonstrated its capability to enhance both illumination and color in dark environments significantly. This approach optimizes the perceptual quality of images and reduces noise and facilitates advanced tasks in high-level vision, such as semantic segmentation, by serving as a powerful preprocessing step.

Both quantitative and qualitative evaluations confirm the effectiveness of ALEN, outperforming recent state-of-the-art methods. This approach enhances image sharpness and fidelity and demonstrates strong adaptability to different illumination conditions. Specifically, this study focuses on global and local enhancement adaptability. However, future work will expand this scope by exploring a broader range of lighting conditions, refining the labeling of the GLI dataset to improve adaptability further, and extending it into a large-scale classification dataset. Additionally, developing specific NIQA metrics for evaluating unpaired images will be considered. These improvements aim to enhance the method’s robustness across diverse capture conditions and broaden its applicability to practical scenarios, such as night-time surveillance and imaging in environments with variable and challenging lighting conditions.

\begin{acknowledgements}
This work is supported by the Zhejiang Provincial Natural Science Foundation of China (No.LY24F020004, No.LZ23F020004), and the Zhejiang Gongshang University "Digital+" Disciplinary Construction Management Project (No.SZJ2022B016) and the University of Guadalajara.
\end{acknowledgements}








%

\vspace{0.5 cm}

\textbf{Data availability}
The data used to support the findings are cited within
the article. Also, the datasets generated during and/or analyzed during the current study are available from the corresponding author upon reasonable request.


\textbf{Author contributions}
All authors contributed equally to the study's conception and design.

\section*{Compliance with ethical standards statements}
\textbf{I. Ethical approval}
This article does not contain any studies with human
participants or animals performed by any of the authors.\\

\textbf{II. Funding}
The authors declare that no funds, grants, or other support were received during the preparation of this manuscript. \\

\textbf{III. Conflict of interest}
All the authors declare that there is no conflict of interest. \\

\textbf{IV. Informed consent}
Informed consent was obtained from all individual participants included in the study.

\bibliographystyle{spmpsci}      
\bibliography{sn-bibliography}

\begin{thebibliography}{10}
\providecommand{\url}[1]{{#1}}
\providecommand{\urlprefix}{URL }
\expandafter\ifx\csname urlstyle\endcsname\relax
  \providecommand{\doi}[1]{DOI~\discretionary{}{}{}#1}\else
  \providecommand{\doi}{DOI~\discretionary{}{}{}\begingroup \urlstyle{rm}\Url}\fi

\bibitem{abdullah2007dynamic}
Abdullah-Al-Wadud, M., Kabir, M.H., Dewan, M.A.A., Chae, O.: {A Dynamic Histogram Equalization for Image Contrast Enhancement}.
\newblock IEEE Transactions on Consumer Electronics \textbf{53}(2), 593--600 (2007)

\bibitem{bai2024retinexmamba}
Bai, J., Yin, Y., He, Q.: {Retinexmamba: Retinex-Based Mamba for Low-Light Image Enhancement}.
\newblock arXiv preprint arXiv:2405.03349  (2024)

\bibitem{bychkovsky2011learning}
Bychkovsky, V., Paris, S., Chan, E., Durand, F.: {Learning Photographic Global Tonal Adjustment with a Database of Input/Output Image Pairs}.
\newblock In: Proceedings of the IEEE Conference on Computer Vision and Pattern Recognition, pp. 97--104. IEEE, Colorado Springs, USA (2011)

\bibitem{cai2023retinexformer}
Cai, Y., Bian, H., Lin, J., Wang, H., Timofte, R., Zhang, Y.: {Retinexformer: One-stage Retinex-based Transformer for Low-light Image Enhancement}.
\newblock In: Proceedings of the IEEE/CVF International Conference on Computer Vision, pp. 12504--12513. IEEE/CVF, Paris, France (2023)

\bibitem{cao2023physics}
Cao, Y., Liu, M., Liu, S., Wang, X., Lei, L., Zuo, W.: {Physics-Guided ISO-Dependent Sensor Noise Modeling for Extreme Low-Light Photography}.
\newblock In: Proceedings of the IEEE/CVF Conference on Computer Vision and Pattern Recognition, pp. 5744--5753. IEEE/CVF, Vancouver, Canada (2023)

\bibitem{choi2008color}
Choi, D.H., Jang, I.H., Kim, M.H., Kim, N.C.: {Color Image Enhancement Using Single-Scale Retinex Based on an Improved Image Formation Model}.
\newblock In: Proceedings of the 16th European Signal Processing Conference, pp. 1--5. IEEE, LLausanne, Switzerland (2008)

\bibitem{cui2022you}
Cui, Z., Li, K., Gu, L., Su, S., Gao, P., Jiang, Z., Qiao, Y., Harada, T.: {You Only Need 90K Parameters to Adapt Light: A Light Weight Transformer for Image Enhancement and Exposure Correction}.
\newblock arXiv preprint arXiv:2205.14871  (2022)

\bibitem{cui2021multitask}
Cui, Z., Qi, G.J., Gu, L., You, S., Zhang, Z., Harada, T.: {Multitask AET with Orthogonal Tangent Regularity for Dark Object Detection}.
\newblock In: Proceedings of the IEEE/CVF International Conference on Computer Vision, pp. 2553--2562. IEEE/CVF, Virtual (2021)

\bibitem{dale1993study}
Dale-Jones, R., Tjahjadi, T.: {A Study and Modification of the Local Histogram Equalization Algorithm}.
\newblock Pattern Recognition \textbf{26}(9), 1373--1381 (1993)

\bibitem{dang2024ppformer}
Dang, J., Zhong, Y., Qin, X.: {PPformer: Using Pixel-Wise and Patch-Wise Cross-Attention for Low-Light Image Enhancement}.
\newblock Computer Vision and Image Understanding \textbf{241}, 103930 (2024)

\bibitem{deng2009imagenet}
Deng, J., Dong, W., Socher, R., Li, L.J., Li, K., Li, F.F.: {ImageNet: A Large-Scale Hierarchical Image Database}.
\newblock In: Proceedings of the IEEE Conference on Computer Vision and Pattern Recognition, pp. 248--255. IEEE, Miami, USA (2009)

\bibitem{fan2022half}
Fan, C.M., Liu, T.J., Liu, K.H.: {Half Wavelet Attention on M-Net+ for Low-Light Image Enhancement}.
\newblock In: Proceedings of the 29th IEEE International Conference on Image Processing, pp. 3878--3882. IEEE, Bordeaux, France (2022)

\bibitem{fang2024non}
Fang, X., Gao, X., Li, B., Zhai, F., Qin, Y., Meng, Z., Lu, J., Xiao, C.: {A Non-Uniform Low-Light Image Enhancement Method with Multi-Scale Attention Transformer and Luminance Consistency Loss}.
\newblock The Visual Computer pp. 1--18 (2024)

\bibitem{fu2023learning}
Fu, Z., Yang, Y., Tu, X., et~al.: {Learning a Simple Low-Light Image Enhancer From Paired Low-Light Instances}.
\newblock In: Proceedings of the IEEE/CVF Conference on Computer Vision and Pattern Recognition, pp. 22252--22261. IEEE/CVF, Vancouver, Canada (2023)

\bibitem{guo2020zero}
Guo, C., Li, C., Guo, J., et~al.: {Zero-Reference Deep Curve Estimation for Low-Light Image Enhancement}.
\newblock In: Proceedings of the IEEE/CVF Conference on Computer Vision and Pattern Recognition, pp. 1780--1789. IEEE/CVF, Virtual (2020)

\bibitem{guo2016lime}
Guo, X., Li, Y., Ling, H.: {LIME: Low-Light Image Enhancement via Illumination Map Estimation}.
\newblock IEEE Transactions on Image Processing \textbf{26}(2), 982--993 (2016)

\bibitem{hai2023r2rnet}
Hai, J., Xuan, Z., Yang, R., Hao, Y., Zou, F., Lin, F., Han, S.: {R2RNet: Low-Light Image Enhancement via Real-Low to Real-Normal Network}.
\newblock Journal of Visual Communication and Image Representation \textbf{90}, 103712 (2023)

\bibitem{han2022survey}
Han, K., Wang, Y., Chen, H., Chen, X., Guo, J., Liu, Z., Tang, Y., Xiao, A., Xu, C., Xu, Y., et~al.: {A Survey on Vision Transformer}.
\newblock IEEE Transactions on Pattern Analysis and Machine Intelligence \textbf{45}(1), 87--110 (2022)

\bibitem{han2023low}
Han, Y., Chen, X., Zhong, Y., Huang, Y., Li, Z., Han, P., Li, Q., Yuan, Z.: {Low-Illumination Road Image Enhancement by Fusing Retinex Theory and Histogram Equalization}.
\newblock Electronics \textbf{12}(4), 990 (2023)

\bibitem{hasinoff2016burst}
Hasinoff, S.W., Sharlet, D., Geiss, R., Adams, A., Barron, J.T., Kainz, F., Chen, J., Levoy, M.: {Burst Photography for High Dynamic Range and Low-Light Imaging on Mobile Cameras}.
\newblock ACM Transactions on Graphics \textbf{35}(6), 1--12 (2016)

\bibitem{hitam2013mixture}
Hitam, M.S., Awalludin, E.A., Yussof, W.N.J.H.W., Bachok, Z.: {Mixture Contrast Limited Adaptive Histogram Equalization for Underwater Image Enhancement}.
\newblock In: {2013 International Conference on Computer Applications Technology (ICCAT)}, pp. 1--5. IEEE (2013)

\bibitem{hou2024global}
Hou, J., Zhu, Z., Hou, J., Liu, H., Zeng, H., Yuan, H.: {Global Structure-Aware Diffusion Process for Low-light Image Enhancement}.
\newblock Advances in Neural Information Processing Systems \textbf{36}, 79734–79747 (2024)

\bibitem{hussein2019retinex}
Hussein, R.R., Hamodi, Y.I., Rooa, A.S.: {Retinex Theory for Color Image Enhancement: A Systematic Review}.
\newblock International Journal of Electrical and Computer Engineering \textbf{9}(6), 5560 (2019)

\bibitem{ibrahim2007brightness}
Ibrahim, H., Kong, N.S.P.: {Brightness Preserving Dynamic Histogram Equalization for Image Contrast Enhancement}.
\newblock IEEE Transactions on Consumer Electronics \textbf{53}(4), 1752--1758 (2007)

\bibitem{jeon2024low}
Jeon, J.J., Park, J.Y., Eom, I.K.: {Low-Light Image Enhancement Using Gamma Correction Prior in Mixed Color Spaces}.
\newblock Pattern Recognition \textbf{146}, 110001 (2024)

\bibitem{jin2022unsupervised}
Jin, Y., Yang, W., Tan, R.T.: {Unsupervised Night Image Enhancement: When Layer Decomposition Meets Light-Effects Suppression}.
\newblock In: Proceedings of the 17th European Conference on Computer Vision, pp. 404--421. Springer, Tel Aviv, Israel (2022)

\bibitem{jobson1997multiscale}
Jobson, D.J., Rahman, Z.u., Woodell, G.A.: {A Multiscale Retinex for Bridging the Gap Between Color Images and Human Observation of Scenes}.
\newblock IEEE Transactions on Image processing \textbf{6}(7), 965--976 (1997)

\bibitem{johnson2016perceptual}
Johnson, J., Alahi, A., Li, F.: {Perceptual Losses for Real-Time Style Transfer and Super-Resolution}.
\newblock In: Proceedings of the 14th European Conference on Computer Vision, pp. 694--711. Springer, Amsterdam, The Netherlands (2016)

\bibitem{kaur2011survey}
Kaur, M., Kaur, J., Kaur, J.: {Survey of Contrast Enhancement Techniques Based on Histogram Equalization}.
\newblock International Journal of Advanced Computer Science and Applications \textbf{2}(7), 137–141 (2011)

\bibitem{ke2021musiq}
Ke, J., Wang, Q., Wang, Y., Milanfar, P., Yang, F.: {MUSIQ: Multi-Scale Image Quality Transformer}.
\newblock In: Proceedings of the IEEE/CVF International Conference on Computer Vision, pp. 5148--5157. IEEE/CVF, Nashville, USA (2021)

\bibitem{khan2020survey}
Khan, A., Sohail, A., Zahoora, U., Qureshi, A.S.: {A Survey of the Recent Architectures of Deep Convolutional Neural Networks}.
\newblock Artificial Intelligence Review \textbf{53}(8), 5455--5516 (2020)

\bibitem{khan2014segment}
Khan, M.F., Khan, E., Abbasi, Z.A.: {Segment-Dependent Dynamic Multi-Histogram Equalization for Image Contrast Enhancement}.
\newblock Digital Signal Processing \textbf{25}, 198--223 (2014)

\bibitem{khan2022transformers}
Khan, S., Naseer, M., Hayat, M., Zamir, S.W., Khan, F.S., Shah, M.: {Transformers in Vision: A Survey}.
\newblock ACM Computing Surveys \textbf{54}(10), 1--41 (2022)

\bibitem{kingma2014adam}
Kingma, D.P., Ba, J.: {Adam: A Method for Stochastic Optimization}.
\newblock arXiv preprint arXiv:1412.6980  (2014)

\bibitem{kirillov2023segment}
Kirillov, A., Mintun, E., Ravi, N., Mao, H., Rolland, C., Gustafson, L., Xiao, T., Whitehead, S., Berg, A.C., Lo, W.Y., et~al.: {Segment Anything}.
\newblock In: Proceedings of the IEEE/CVF International Conference on Computer Vision, pp. 4015--4026. IEEE/CVF, Paris, France (2023)

\bibitem{land1971lightness}
Land, E.H., McCann, J.J.: {Lightness and Retinex Theory}.
\newblock Journal of the Optical Society of America \textbf{61}(1), 1--11 (1971)

\bibitem{lee2013contrast}
Lee, C., Lee, C., Kim, C.S.: {Contrast Enhancement Based on Layered Difference Representation of 2D Histograms}.
\newblock IEEE Transactions on Image Processing \textbf{22}(12), 5372--5384 (2013)

\bibitem{li2021deep}
Li, G., Yang, Y., Qu, X., Cao, D., Li, K.: {A Deep Learning-Based Image Enhancement Approach for Autonomous Driving at Night}.
\newblock Knowledge-Based Systems \textbf{213}, 106617 (2021)

\bibitem{liang2024pie}
Liang, D., Xu, Z., Li, L., Wei, M., Chen, S.: {PIE: Physics-Inspired Low-Light Enhancement}.
\newblock International Journal of Computer Vision \textbf{132}(9), 1--22 (2024)

\bibitem{liang2023iterative}
Liang, Z., Li, C., Zhou, S., Feng, R., Loy, C.C.: {Iterative Prompt Learning for Unsupervised Backlit Image Enhancement}.
\newblock In: Proceedings of the IEEE/CVF International Conference on Computer Vision, pp. 8094--8103. IEEE/CVF, Paris, France (2023)

\bibitem{lin2014microsoft}
Lin, T.Y., Maire, M., Belongie, S., Hays, J., Perona, P., Ramanan, D., Doll{\'a}r, P., Zitnick, C.L.: {Microsoft COCO: Common Objects in Context}.
\newblock In: Proceedings of the 13th European Conference on Computer Vision, pp. 740--755. Springer, Zurich, Switzerland (2014)

\bibitem{Liu21}
Liu, R., Ma, L., Zhang, J., et~al.: {Retinex-Inspired Unrolling with Cooperative Prior Architecture Search for Low-Light Image Enhancement}.
\newblock In: Proceedings of the IEEE/CVF Conference on Computer Vision and Pattern Recognition, pp. 10561--10570. IEEE/CVF, Virtual (2021)

\bibitem{liu2021swin}
Liu, Z., Lin, Y., Cao, Y., Hu, H., Wei, Y., Zhang, Z., Lin, S., Guo, B.: {Swin Transformer: Hierarchical Vision Transformer Using Shifted Windows}.
\newblock In: Proceedings of the IEEE/CVF International Conference on Computer Vision, pp. 10012--10022. Nashville, USA (2021)

\bibitem{lv2021attention}
Lv, F., Li, Y., Lu, F.: {Attention Guided Low-Light Image Enhancement with a Large Scale Low-Light Simulation Dataset}.
\newblock International Journal of Computer Vision \textbf{129}(7), 2175--2193 (2021)

\bibitem{ma2015perceptual}
Ma, K., Zeng, K., Wang, Z.: {Perceptual Quality Assessment for Multi-Exposure Image Fusion}.
\newblock IEEE Transactions on Image Processing \textbf{24}(11), 3345--3356 (2015)

\bibitem{ma2023bilevel}
Ma, L., Jin, D., An, N., Liu, J., Fan, X., Luo, Z., Liu, R.: {Bilevel Fast Scene Adaptation for Low-Light Image Enhancement}.
\newblock International Journal of Computer Vision pp. 1--19 (2023)

\bibitem{ma2022toward}
Ma, L., Ma, T., Liu, R., Fan, X., Luo, Z.: {Toward Fast, Flexible, and Robust Low-Light Image Enhancement}.
\newblock In: Proceedings of the IEEE/CVF Conference on Computer Vision and Pattern Recognition, pp. 5637--5646. IEEE/CVF, New Orleans, USA (2022)

\bibitem{mittal2012making}
Mittal, A., Soundararajan, R., Bovik, A.C.: {Making a “Completely Blind” Image Quality Analyzer}.
\newblock IEEE Signal Processing Letters \textbf{20}(3), 209--212 (2012)

\bibitem{pan2019recent}
Pan, Z., Yu, W., Yi, X., Khan, A., Yuan, F., Zheng, Y.: {Recent Progress on Generative Adversarial Networks (GANs): A Survey}.
\newblock IEEE Access \textbf{7}, 36322--36333 (2019)

\bibitem{Paszke2019Pytorch}
Paszke, A., Gross, S., Massa, F., Lerer, A., Bradbury, J., Chanan, G.: {PyTorch: An Imperative Style, High-Performance Deep Learning Library}.
\newblock Advances in Neural Information Processing Systems \textbf{32}, 8026--8037 (2019)

\bibitem{perez2023loli}
Perez-Zarate, E., Ramos-Soto, O., Rodr{\'\i}guez-Esparza, E., Aguilar, G.: {LoLi-IEA: Low-Light Image Enhancement Algorithm}.
\newblock In: Proceedings of the SPIE Optical Engineering + Applications, pp. 230--245. SPIE, San Diego, USA (2023)

\bibitem{pizer1987adaptive}
Pizer, S.M., Amburn, E.P., Austin, J.D., Cromartie, R., Geselowitz, A., Greer, T., ter Haar~Romeny, B., Zimmerman, J.B., Zuiderveld, K.: {Adaptive Histogram Equalization and Its Variations}.
\newblock Computer Vision, Graphics, and Image Processing \textbf{39}(3), 355--368 (1987)

\bibitem{PyTorchBCELoss}
{PyTorch}: {BCELoss}.
\newblock \url{https://pytorch.org/docs/stable/generated/torch.nn.BCELoss.html} (2019).
\newblock Accessed: June 1, 2023

\bibitem{pytorch_mseloss}
{PyTorch}: {MSELoss}.
\newblock \url{https://pytorch.org/docs/stable/generated/torch.nn.MSELoss.html} (2019).
\newblock Accessed: June 1, 2023

\bibitem{qu2024double}
Qu, J., Liu, R.W., Gao, Y., Guo, Y., Zhu, F., Wang, F.Y.: {Double Domain Guided Real-Time Low-Light Image Enhancement for Ultra-High-Definition Transportation Surveillance}.
\newblock IEEE Transactions on Intelligent Transportation Systems \textbf{25}(8), 9550 -- 9562 (2024)

\bibitem{rahman1996multi}
Rahman, Z.u., Jobson, D.J., Woodell, G.A.: {Multi-Scale Retinex for Color Image Enhancement}.
\newblock In: Proceedings of the 3rd International Conference on Image Processing, pp. 1003--1006. IEEE, Lausanne, Switzerland (1996)

\bibitem{rahman2004retinex}
Rahman, Z.u., Jobson, D.J., Woodell, G.A.: {Retinex Processing for Automatic Image Enhancement}.
\newblock Journal of Electronic Imaging \textbf{13}(1), 100--110 (2004)

\bibitem{reflectance2011retinex}
Reflectance-Illuminance, B.: {Retinex Image Processing: Improving the Visual Realism of Color Images}.
\newblock International Journal of Information Technology and Knowledge Management \textbf{4}(2), 371--377 (2011)

\bibitem{shi2024maco}
Shi, Y., Liu, D., Zhang, L., Xia, X., Sun, J.: {MaCo: Efficient Unsupervised Low-Light Image Enhancement via Illumination-Based Magnitude Control}.
\newblock The Visual Computer \textbf{40}(12), 1--19 (2024)

\bibitem{simonyan2014very}
Simonyan, K., Zisserman, A.: {Very Deep Convolutional Networks for Large-Scale Image Recognition}.
\newblock arXiv preprint arXiv:1409.1556  (2014)

\bibitem{singh2015enhancement}
Singh, K., Kapoor, R., Sinha, S.K.: {Enhancement of Low Exposure Images via Recursive Histogram Equalization Algorithms}.
\newblock Optik \textbf{126}(20), 2619--2625 (2015)

\bibitem{thepade2022contrast}
Thepade, S.D., Pardhi, P.M.: {Contrast Enhancement with Brightness Preservation of Low-Light Images Using a Blending of CLAHE and BPDHE Histogram Equalization Methods}.
\newblock International Journal of Information Technology \textbf{14}(6), 3047--3056 (2022)

\bibitem{VV_TM-DIED}
{Vasileios Vonikakis}: {TM-DIED: The Most Difficult Image Enhancement Dataset}.
\newblock \url{https://sites.google.com/site/vonikakis/datasets/tm-died} (2019).
\newblock Accessed: October 15, 2024

\bibitem{wang2007fast}
Wang, Q., Ward, R.K.: {Fast Image/Video Contrast Enhancement Based on Weighted Thresholded Histogram Equalization}.
\newblock IEEE Transactions on Consumer Electronics \textbf{53}(2), 757--764 (2007)

\bibitem{wang2013naturalness}
Wang, S., Zheng, J., Hu, H.M., et~al.: {Naturalness Preserved Enhancement Algorithm for Non-Uniform Illumination Images}.
\newblock IEEE Transactions on Image Processing \textbf{22}(9), 3538--3548 (2013)

\bibitem{wang2023ultra}
Wang, T., Zhang, K., Shen, T., Luo, W., Stenger, B., Lu, T.: {Ultra-High-Definition Low-Light Image Enhancement: A Benchmark and Transformer-Based Method}.
\newblock In: Proceedings of the 37th AAAI Conference on Artificial Intelligence, pp. 2654--2662. AAAI, Washington DC, USA (2023)

\bibitem{wang2022lowlight}
Wang, Y., Wan, R., Yang, W., et~al.: {Low-Light Image Enhancement with Normalizing Flow}.
\newblock In: Proceedings of the 36th AAAI Conference on Artificial Intelligence, pp. 2604--2612. AAAI, Vancouver, Canada (2022)

\bibitem{wang2024itre}
Wang, Y., Wang, Y., Liu, T., Li, J., Sui, X., Chen, Q.: {ITRE: Low-Light Image Enhancement Based on Illumination Transmission Ratio Estimation}.
\newblock Knowledge-Based Systems \textbf{303}, 112427 (2024)

\bibitem{wang2002universal}
Wang, Z., Bovik, A.C.: {A Universal Image Quality Index}.
\newblock IEEE Signal Processing Letters \textbf{9}(3), 81--84 (2002)

\bibitem{wang2004image}
Wang, Z., Bovik, A.C., Sheikh, H.R., et~al.: {Image Quality Assessment: From Error Visibility to Structural Similarity}.
\newblock IEEE Transactions on Image Processing \textbf{13}(4), 600--612 (2004)

\bibitem{wei2018deep}
Wei, C., Wang, W., Yang, W., Liu, J.: {Deep Retinex Decomposition for Low-Light Enhancement}.
\newblock arXiv preprint arXiv:1808.04560  (2018)

\bibitem{wu2022uretinex}
Wu, W., Weng, J., Zhang, P., et~al.: {URetinex-Net: Retinex-Based Deep Unfolding Network for Low-Light Image Enhancement}.
\newblock In: Proceedings of the IEEE/CVF Conference on Computer Vision and Pattern Recognition, pp. 5901--5910. IEEE/CVF, New Orleans, USA (2022)

\bibitem{xia2023cmda}
Xia, R., Zhao, C., Zheng, M., Wu, Z., Sun, Q., Tang, Y.: {CMDA: Cross-Modality Domain Adaptation for Nighttime Semantic Segmentation}.
\newblock In: Proceedings of the IEEE/CVF International Conference on Computer Vision, pp. 21572--21581. IEEE/CVF, Paris, France (2023)

\bibitem{xu2024degraded}
Xu, H., Liu, X., Zhang, H., Wu, X., Zuo, W.: {Degraded Structure and Hue Guided Auxiliary Learning for Low-Light Image Enhancement}.
\newblock Knowledge-Based Systems \textbf{295}, 111779 (2024)

\bibitem{ye2023glow}
Ye, D., Ni, Z., Yang, W., Wang, H., Wang, S., Kwong, S.: {Glow in the Dark: Low-Light Image Enhancement With External Memory}.
\newblock IEEE Transactions on Multimedia \textbf{26} (2023)

\bibitem{yi2023diff}
Yi, X., Xu, H., Zhang, H., Tang, L., Ma, J.: {Diff-Retinex: Rethinking Low-light Image Enhancement with A Generative Diffusion Model}.
\newblock In: Proceedings of the IEEE/CVF International Conference on Computer Vision, pp. 12302--12311. IEEE/CVF, Paris, France (2023)

\bibitem{zeng2020learning}
Zeng, H., Cai, J., Li, L., et~al.: {Learning Image-Adaptive 3D Lookup Tables for High Performance Photo Enhancement in Real-Time}.
\newblock IEEE Transactions on Pattern Analysis and Machine Intelligence \textbf{14}(8), 2058--2073 (2020)

\bibitem{zhang2021unsupervised}
Zhang, F., Shao, Y., Sun, Y., Zhu, K., Gao, C., Sang, N.: {Unsupervised Low-Light Image Enhancement via Histogram Equalization Prior}.
\newblock arXiv preprint arXiv:2112.01766  (2021)

\bibitem{zhang2019dual}
Zhang, Q., Nie, Y., Zheng, W.S.: {Dual Illumination Estimation for Robust Exposure Correction}.
\newblock Computer Graphics Forum \textbf{38}, 243--252 (2019)

\bibitem{zhang2018unreasonable}
Zhang, R., Isola, P., Efros, A.A., Shechtman, E., Wang, O.: {The Unreasonable Effectiveness of Deep Features as a Perceptual Metric}.
\newblock In: Proceedings of the IEEE Conference on Computer Vision and Pattern Recognition, pp. 586--595. IEEE, Salt Lake City, USA (2018)

\bibitem{zhao2024non}
Zhao, Z., Lin, H., Shi, D., Zhou, G.: {A Non-Regularization Self-Supervised Retinex Approach to Low-Light Image Enhancement with Parameterized Illumination Estimation}.
\newblock Pattern Recognition \textbf{146}, 110025 (2024)

\bibitem{zheng2021adaptive}
Zheng, C., Shi, D., Shi, W.: {Adaptive Unfolding Total Variation Network for Low-Light Image Enhancement}.
\newblock In: Proceedings of the IEEE/CVF International Conference on Computer Vision, pp. 4439--4448. IEEE/CVF, Virtual (2021)

\end{thebibliography}

\vspace{9\baselineskip}

\setlength\intextsep{0pt} 
    \begin{wrapfigure}{l}{25mm}
        \centering
        \includegraphics[width=0.17\textwidth]{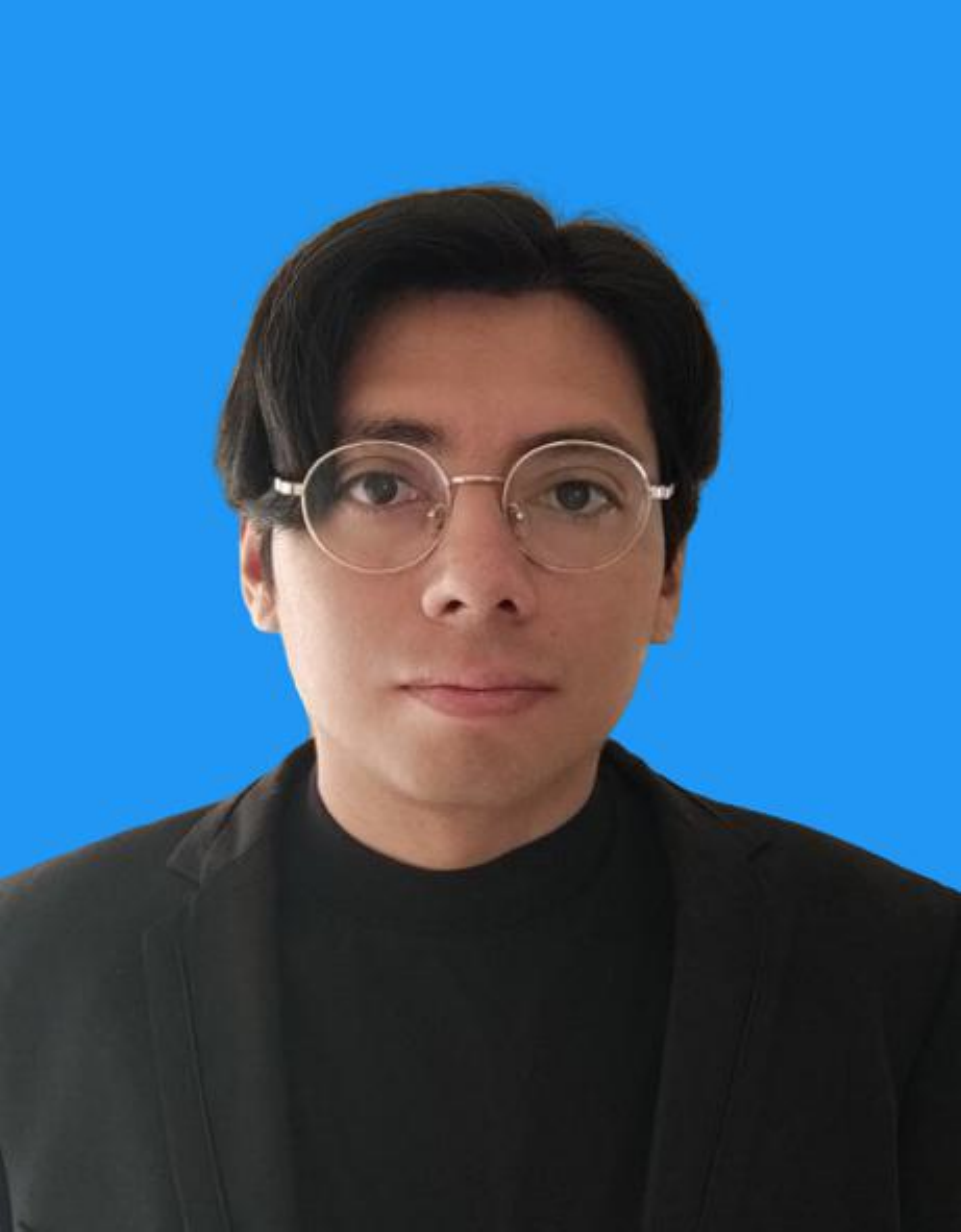}
    \end{wrapfigure}
    \noindent \textbf{Ezequiel Perez-Zarate} received the B.S. degree in Communications and Electronics Engineering from the University of Guadalajara, Mexico, in 2019. He is currently pursuing the M.E. degree in Computer Science and Technology at the Zhejiang Gongshang University, China. His research interests include computer vision, computer graphics, machine and deep learning, digital image processing, and real-time image processing.

\vspace{5\baselineskip}

\setlength\intextsep{0pt} 
    \begin{wrapfigure}{l}{25mm}
        \centering
        \includegraphics[width=0.17\textwidth]{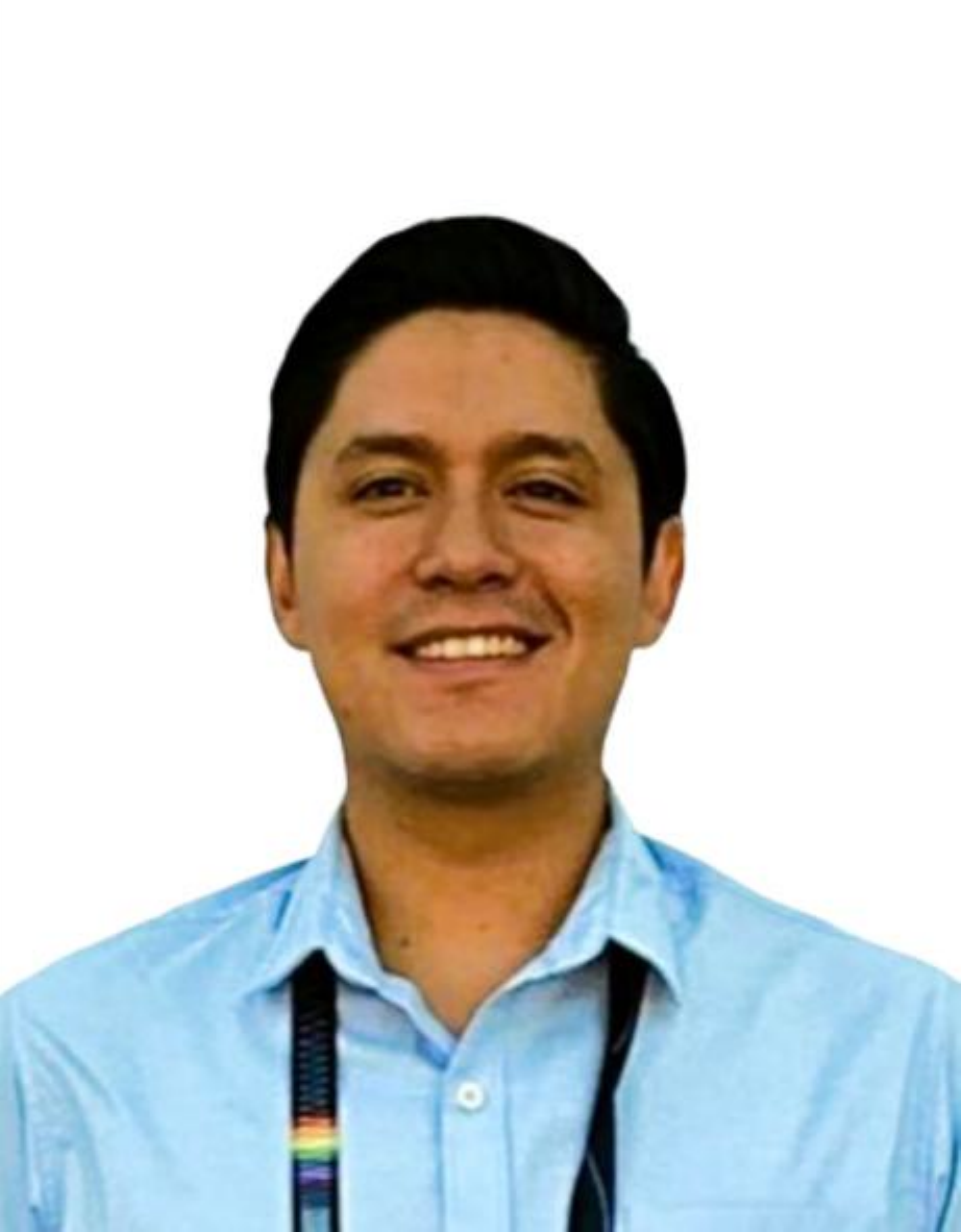}
    \end{wrapfigure}
    \noindent \textbf{Oscar Ramos-Soto} received the B.S. degree in Communications and Electronics Engineering from the National Polytechnic Institute, Mexico, in 2019, and the M.Sc. degree in Electronic Engineering and Computer Sciences from University of Guadalajara, Mexico, in 2021. Currently, he is pursuing a Ph.D. degree in Electronics and Computer Sciences at the University of Guadalajara, Mexico. His research interests primarily involve areas such as computer vision, digital image processing, machine and deep learning, biomedical engineering, and vision science.
    
\vspace{5\baselineskip}

\setlength\intextsep{0pt} 
    \begin{wrapfigure}{l}{25mm}
        \centering
        \includegraphics[width=0.17\textwidth]{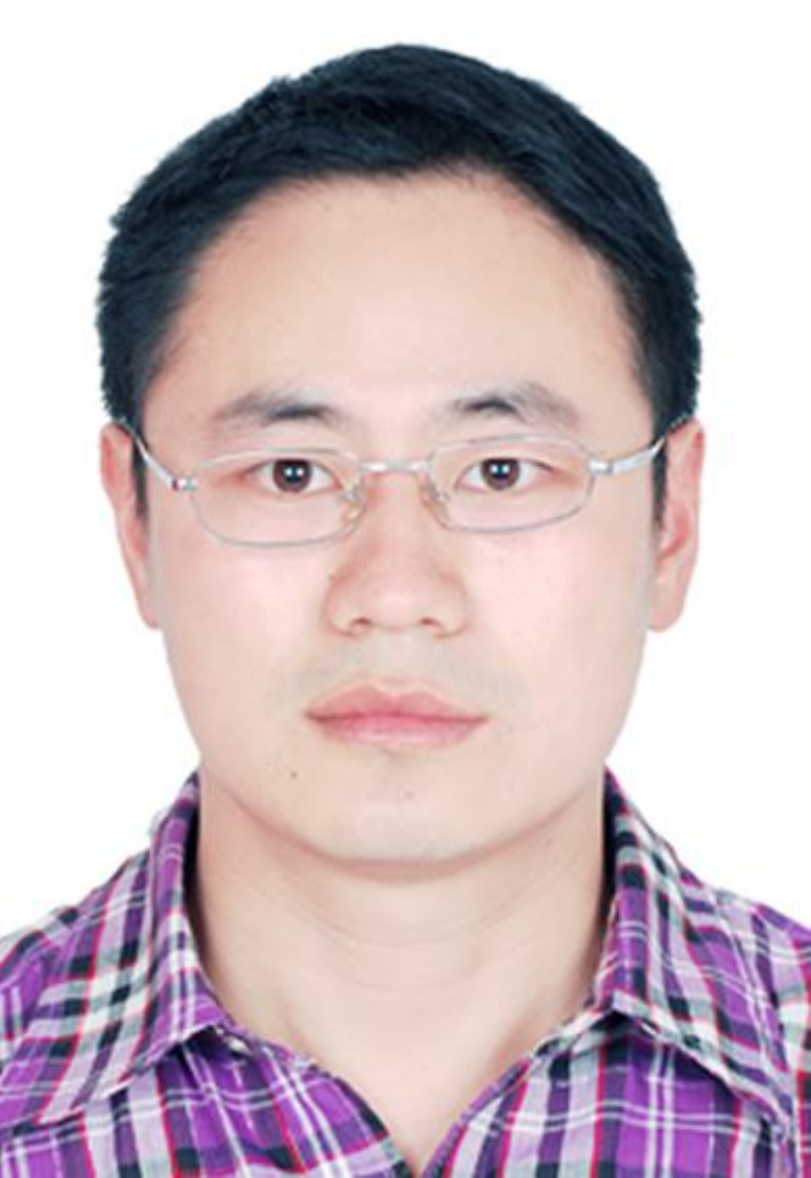}
    \end{wrapfigure}
    \noindent \textbf{Chunxiao Liu} is an associate professor and master supervisor in Computer Science and Technology at the School of Computer Science and Technology, Zhejiang Gongshang University. He got his Ph.D in Mathematics from the State Key Lab of CAD\&CG, Zhejiang University, Hangzhou, China, in 2019. His current research interests include image and video processing, computer vision, computer graphics, machine learning, intelligent systems.
    
\vspace{6\baselineskip}

\setlength\intextsep{0pt} 
    \begin{wrapfigure}{l}{25mm}
        \centering
        \includegraphics[width=0.17\textwidth]{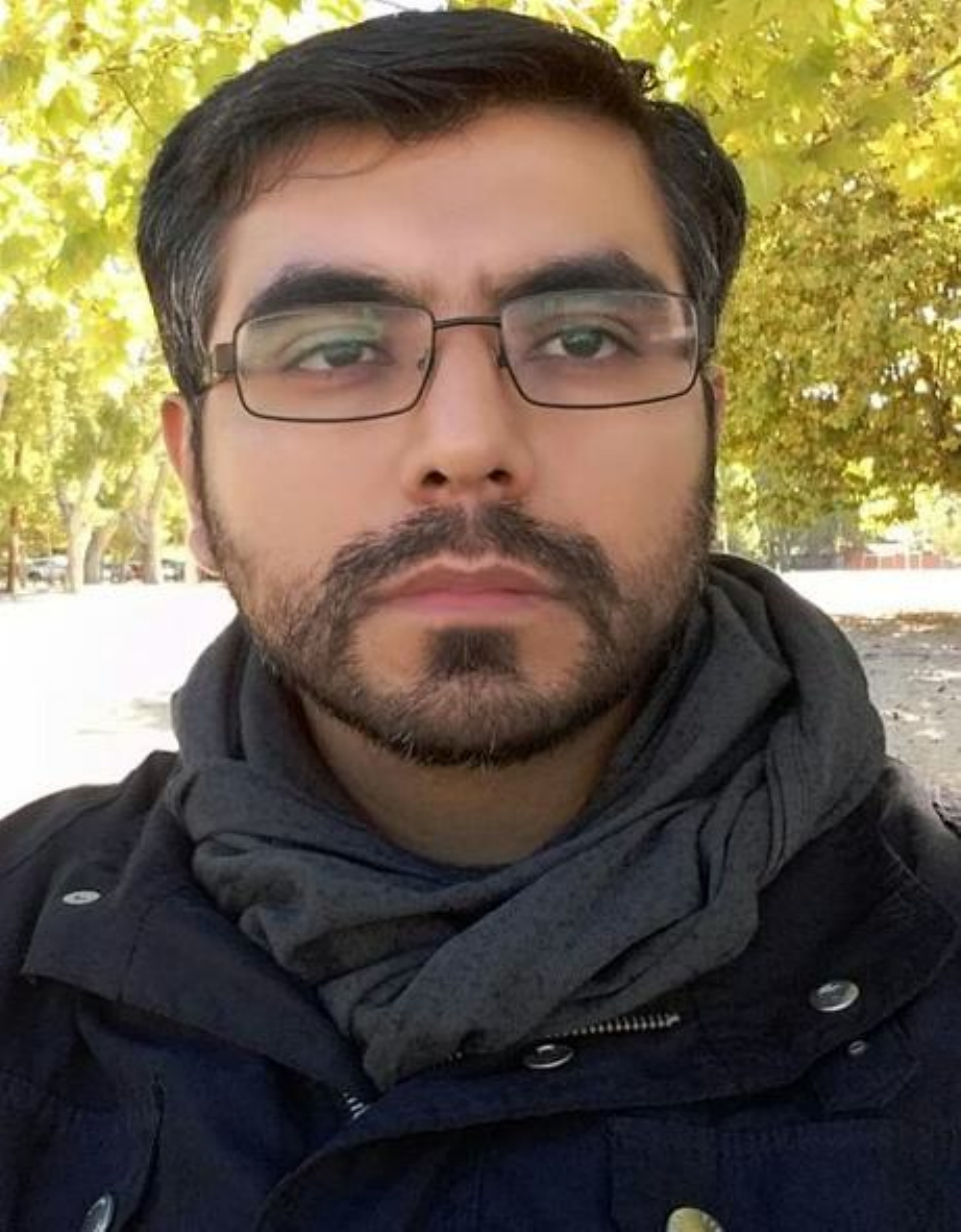}
    \end{wrapfigure}
    \noindent \textbf{Prof. Diego Oliva} received a B.S. degree in Electronics and Computer Engineering from the Industrial Technical Education Center (CETI) of Guadalajara, Mexico, in 2007 and an M.Sc. degree in Electronic Engineering and Computer Sciences from the University of Guadalajara, Mexico, in 2010. He obtained a Ph. D. in Informatics in 2015 from the Universidad Complutense de Madrid. Currently, he is an Associate Professor at the University of Guadalajara in Mexico. He is a member of the Mexican National Research System (SNII), a Senior member of the IEEE, a member of the Association for Computing Machinery (ACM), and a member of the Mexican Academy of Computer Sciences (AMEXCOMP). His research interests include evolutionary and swarm algorithms, hybridization of evolutionary and swarm algorithms, and computational intelligence.

\vspace{3\baselineskip}

\setlength\intextsep{0pt} 
    \begin{wrapfigure}{l}{25mm}
        \centering
        \includegraphics[width=0.17\textwidth]{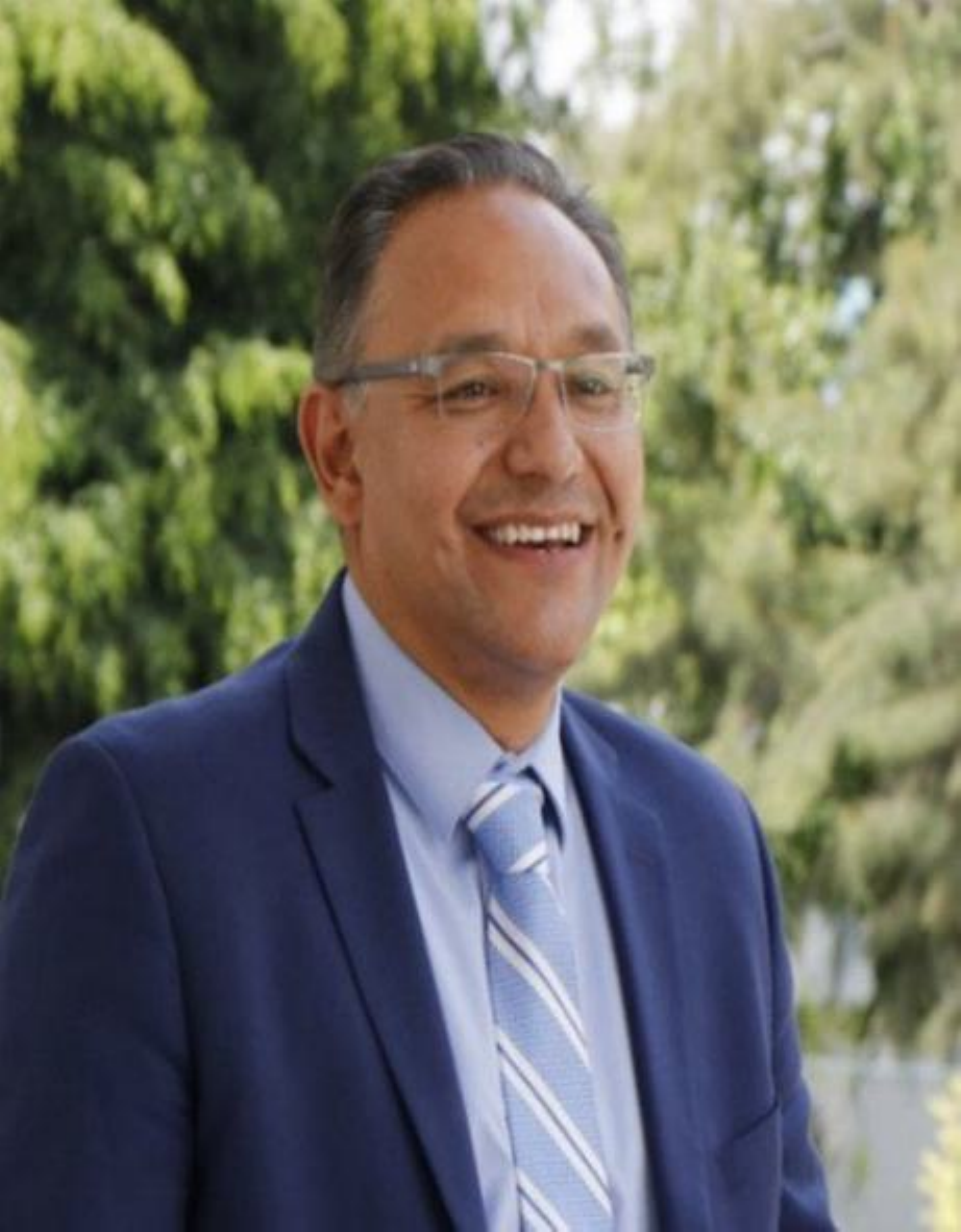}
    \end{wrapfigure}
    \noindent \textbf{Marco Pérez-Cisneros} received the B.S. degree (Hons.) in electronics and communications engineering from Universidad de Guadalajara, Mexico, in 1995, the M.Sc. degree in industrial electronics from ITESO University, Guadalajara, in 2000, and the Ph.D. degree from the Institute of Science and Technology, UMIST, The University of Manchester, U.K., in 2004. Since 2005, he has been with CUCEI, Universidad de Guadalajara, where he is currently a Professor and the Rector of CUCEI. He has published over 75 indexed papers, currently holding an H-index of 17. He is the author of seven textbooks about his research interests. His current research interests include computational intelligence and evolutionary algorithms and their applications to robotics, computational vision, and automatic control. He is a member of Mexican Science Academy, Mexican National Research System (SNI), and a Senior Member of IET.

\end{document}